\documentclass[12pt,a4paper]{Thesis} 

\graphicspath{%
	{./Pictures/}%
	{./Figures/}%
}

\let\degree\relax

\usepackage{adjustbox}
\usepackage{amsmath}
\usepackage[amssymb]{SIunits}
\usepackage{amssymb}
\usepackage{multirow}
\usepackage{wrapfig}
\usepackage{enumitem}
\usepackage{subcaption}
\usepackage{mathtools}
\usepackage[ruled,vlined,noresetcount]{algorithm2e}
\usepackage[square, numbers, comma, sort&compress]{natbib} 

\title{\ttitle} 

\usepackage{float}
\usepackage{color,soul}               
\usepackage{xcolor}
\usepackage{colortbl}
\usepackage{enumerate}
\usepackage{makecell}
\usepackage{boldline}
\usepackage{Styles/mydefs}
\usepackage{arydshln}
\usepackage{hyperref}
\usepackage{enumerate}

\def\eg{\emph{e.g.}\xspace}
\def\ie{\emph{i.e.}\xspace}

\def\wrt{\emph{w.r.t.}\xspace}
\def\etal{\emph{et al.}\xspace}

\begin{document}

\frontmatter 

\setstretch{1.3} 

\pagestyle{fancy} 
\fancyhead{}   
\fancyhead[LO]{\sl{\leftmark}}
\fancyhead[RE]{\sl{\rightmark}}
\fancyhead[LE,RO]{\thepage}



\maketitle




\thesisdeclareAuthorship{This thesis contains material from Six paper(s) published in the following peer-reviewed journal(s) / from papers accepted at conferences in which I am listed as the first author.}
{
\noindent The work in Chapter \ref{ch:Synthestic2Real} is published as {\color{blue} C Zheng, TJ Cham, J Cai. T2net: Synthetic-to-realistic translation for solving single-image depth estimation tasks. Proceedings of the European Conference on Computer Vision (ECCV). 2018.}

\vspace{2.0ex}
\noindent The contributions of the co-authors are as follows:
\begin{itemize}
    \item Chuanxia Zheng proposed the initial idea, designed the experiments, and prepared the manuscript.
    \item Tat-Jen Cham and Jianfei Cai discussed the idea, improved the experiments and revised the manuscript.
\end{itemize}
\vspace{3.0ex}

\noindent The work in Chapter \ref{ch:F(L)SeSim} is published as {\color{blue} C Zheng, TJ Cham, J Cai. The Spatially-Correlative Loss for Various Image Translation Tasks. Proceedings of the IEEE/CVF Conference on Computer Vision and Pattern Recognition (CVPR). 2021.
}

\vspace{2.0ex}
\noindent The contributions of the co-authors are as follows:
\begin{itemize}
    \item Chuanxia Zheng proposed the initial idea, designed the experiments, and prepared the manuscript.
    \item Tat-Jen Cham and Jianfei Cai discussed the idea, improved the experiments and revised the manuscript.
\end{itemize}
\vspace{3.0ex}

\noindent The work in Chapter \ref{ch:PICNet} is published as {\color{blue} C Zheng, TJ Cham, J Cai. Pluralistic image completion. Proceedings of the IEEE/CVF Conference on Computer Vision and Pattern Recognition (CVPR). 2019, 
} and published as {\color{blue} C Zheng, TJ Cham, J Cai. Pluralistic free-form image completion. International Journal of Computer Vision (IJCV). 2021. }

\vspace{2.0ex}
\noindent The contributions of the co-authors are as follows:
\begin{itemize}
    \item Chuanxia Zheng proposed the initial idea, designed the experiments, and prepared the manuscript.
    \item Tat-Jen Cham and Jianfei Cai discussed the idea, improved the experiments and revised the manuscript.
\end{itemize}
\vspace{3.0ex}

\noindent The work in Chapter \ref{ch:TFill} is reviewed as {\color{blue} C Zheng, TJ Cham, J Cai. TFill: Image Completion via a Transformer-Based Architecture (arXiv). 2021. }

\vspace{2.0ex}
\noindent The contributions of the co-authors are as follows:
\begin{itemize}
    \item Chuanxia Zheng proposed the initial idea, designed the experiments, and prepared the manuscript.
    \item Tat-Jen Cham and Jianfei Cai discussed the idea, improved the experiments and revised the manuscript.
\end{itemize}
\vspace{3.0ex}

\noindent The work in Chapter \ref{ch:VIV} is published as {\color{blue} C Zheng, DS Dao, G Song, TJ Cham, J Cai. Visiting the Invisible: Layer-by-Layer Completed Scene Decomposition. International Journal of Computer Vision (IJCV). 2021. }

\vspace{2.0ex}
\noindent The contributions of the co-authors are as follows:
\begin{itemize}
    \item Chuanxia Zheng proposed the initial idea, designed the experiments, and prepared the manuscript.
    \item Guoxian Song rendered the synthetic data.
    \item Duy-Son Dao worked for the amodal instance segmentation.
    \item Tat-Jen Cham and Jianfei Cai discussed the idea, improved the experiments and revised the manuscript.
\end{itemize}
\vspace{3.0ex}

}{06 June 2021}{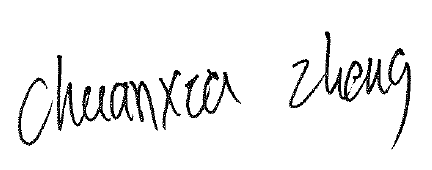}


\setstretch{1.3} 

\acknowledgements{\addtocontents{toc}{\vspace{0.8em}} 

Eleven years ago, for the first time, I was away from my hometown with a dream to at least see a bit of the world. I would like to thank many people who helped me a lot along my path. Without them, I may have no possibility of writing this thesis. 

I would like to express my greatest gratitude to my advisor, Tat-Jen Cham, for taking me under his wing. He is not only my academic advisor who guides me with encouraging, constructive and insightful comments on every research topic, but also the life coach that provides suggestions on how to think deeper, how to be a better man, and how to walk toward a humane and inspired life. In the past four years, I enjoyed a fantastic research journey under his supervision.

I would especially like to thank my co-advisor, Jianfei Cai, for his helpful advice and the immense knowledge he has shared with me. Without him, I would not have the opportunity to pursue the research position at NTU, and I can not come to Singapore. His guidance helped me in all the time of research and writing of this thesis. 

I would like to thank Nadia Magnenat Thalmann, the director of the institute for media innovation (IMI), for awarding me the Ph.D. Scholarship. In IMI, I always enjoyed the freedom to spend time on my interested research topics. 

I would like to thank Irene Goh for providing the powerful AI supercomputer system and helping the deep learning model running smoothly on the corresponding platform. Without these supercomputers in SCSE, I would not have finished many complex experiments. 

I would also thank all the members of the IMI lab and MICL who provided tremendous support for my study and research. Special thanks to Teng Deng for guiding me on the study when I first arrived at the lab. I would like to thank Guoxian Song for rendering the high-quality synthetic dataset on our projects. I would like to thank Duy-Son Dao for setting the baseline benchmark on amodal instance segmentation. I would like to thank Xingxing Xia and Frank Guan for providing helpful suggestions on the translation and completion task. I would like to thank many other folks including Xu Yang, Yuedong Cheng, Bo Hu, Zhonghua Wu, Zhijie Zhang, Junwu Weng, Yujun Cai, Jyothsna Vasudevan, and Ayan Kumar Bhunia for many interesting discussions. 

I would like to thank my old friends, \eg Zhongxia Xiong, Yikang Guo, Han Zhang, Shijie Zhang, Boyu Yang, Xiang Wen, Wei Zhao, among others, for their discussing, listening, and sharing. Special thanks to Zhongxia Xiong for providing suggestions and feedback on every project and publicly available code. 

Lastly, I am grateful to my parents, my parents-in-law, and my wife for their love and support during this wonderful journey. Special thanks to my wife, Mengping, who always accompanies me and gives me confidence and encouragement in these years. She and our lovely daughter, Keyu, are the most precious treasure in my life. 

\begin{flushright}
\emph{Chuanxia Zheng}

\emph{Nanyang Technological University}

\emph{June 2021}
\end{flushright}
}




\cleardoublepage 

\pagestyle{fancy} 
\fancyhead{}   
\fancyhead[LO]{\sl{\leftmark}}
\fancyhead[RE]{\sl{\rightmark}}
\fancyhead[LE,RO]{\thepage}


\addtotoc{Summary} 

\abstract{\addtocontents{toc}{\vspace{0.8em}}}


The goal of this thesis is to present my research contributions towards solving various visual synthesis and generation tasks, comprising image translation, image completion, and completed scene decomposition. This thesis consists of five pieces of work, each of which presents a new learning-based approach for synthesizing images with plausible content as well as visually realistic appearance. Each work demonstrates the superiority of the proposed approach on image synthesis, with some further contributing to other tasks, such as depth estimation.   

\textbf{Part \uppercase\expandafter{\romannumeral1}} describes methods for \textbf{changing visual appearance}. In particular, in Chapter \ref{ch:Synthestic2Real}, a \emph{synthetic-to-realistic} translation system is presented to address the real-world \emph{single-image depth estimation}, where only synthetic image-depth pairs and unpaired real images are used for training. This model provides a new perspective on a real-world estimation task by utilizing low-cost, yet high-reusable synthetic data. In Chapter \ref{ch:F(L)SeSim}, the focus is on general image-to-image (I2I) translation tasks, instead of narrowly synthetic-to-realistic image translation. A novel \emph{spatially-correlative loss} is proposed that is simple, efficient and yet effective for preserving scene structure consistency, while supporting large appearance changes. Spatial patterns of self-similarity are exploited as a means of defining scene structure, with this spatially-correlative loss geared towards only capturing spatial relationships within an image, rather than domain appearance. The extensive experiment results demonstrate significant improvements using this content loss on several I2I tasks, including single-modal, multi-modal, and even single-image translation. Furthermore, this new loss can easily be integrated into existing network architectures and thus allows wide applicability.

\textbf{Part \uppercase\expandafter{\romannumeral2}} presents approaches that \textbf{generate semantically reasonable content} for masked regions. Instead of purely modifying the local appearance as in Part \uppercase\expandafter{\romannumeral1}, two approaches are presented to create new content as well as realistic appearance for a given image. In Chapter \ref{ch:PICNet}, a new task is introduced, called \emph{pluralistic image completion} --- the task of generating \emph{multiple} and \emph{diverse} plausible results, which is as opposed to previous works that attempt to create only a single ``guess'' for this highly subjective problem. In this Chapter, a novel probabilistically principled framework is proposed, which achieved state-of-the-art results for this new task and has become the benchmark for later works. However, my subsequent observation is that architectures based on convolutional neural networks (CNN) model long-range dependencies via many stacked layers, where holes are progressively influenced by neighboring pixels, resulting in some artifacts. To mitigate this issue, in Chapter \ref{ch:TFill}, I propose treating image completion as a directionless sequence-to-sequence prediction task, and deploy a \emph{transformer} to directly capture long-range dependencies in the encoder in a first phase. Crucially, a \emph{restrictive CNN} with small and non-overlapping receptive fields (RF) is employed for token representation, which allows the transformer to explicitly model long-range context relations with equal importance in all layers, without implicitly confounding neighboring tokens when larger RFs are used. Extensive experiments demonstrate superior performance compared to previous CNN-based methods on several datasets. 

\textbf{Part \uppercase\expandafter{\romannumeral3}} combines recognitive learning and the latest generative modeling into a holistic scene decomposition and completion framework, where a network is trained to \emph{decompose} a scene into individual objects, \emph{infer} their underlying occlusion relationships, and moreover \emph{imagine} what the originally occluded objects may look like, \emph{while using only a single image as input}. In Chapter \ref{ch:VIV}, the aim is to derive a higher-level structural decomposition of a scene, automatically recognizing objects and generating intact shapes as well as photorealistic appearances for occluded regions, without requiring manual masking as in Part \uppercase\expandafter{\romannumeral2}. To achieve this goal, a new pipeline is presented that interleaves the two tasks of instance segmentation and scene completion through multiple iterations, solving for objects in a layer-by-layer manner. The proposed system shows significant improvement over the state-of-the-art methods and enables some interesting applications, such as scene editing and recomposition. 

In summary, the thesis introduces a series of works to synthesize photorealistic images by changing the appearance, imagining the semantic content, and inferring the invisible shape and appearance automatically. 



\pagestyle{fancy} 

\tableofcontents 

\listoffigures 

\listoftables 


%
%


%
%


\cleardoublepage  

\listofnomenclature{ll} 
{


RBM & Restricted Boltzmann Machine \\
AE & AutoEncoder \\
CNN & Convolutional Neural Network \\
MLP & Multi-Layer Perceptron \\
GAN & Generative Adversarial Network \\
VAE & Variational AutoEncoder \\
CVAE & Conditional Variational AutoEncoder \\
ResNet & Residual Network \\
RF & Receptive Field \\
NLP & Natural Language Processing \\
VQ & Vector Quantization \\
[0.618cm]
PSNR & Peak Signal-to-Noise Ratio \\
SSIM & Structural SIMilarity \\
IS   & Inception Score \\
FID  & Fr\'echet Inception Distance \\
LPIPS & Learned Perceptual Image Patch Similarity \\
AMT & Amazon Mechanical Turk \\
2AFC & 2-Alternative-Forced-Choice \\
D\&C & Density and Coverage \\
[0.618cm]
DoF     & degree of freedom \\
i.i.d.           & independent and identically distributed\\
\wrt    & with respect to \\
}


%
%
%


\mainmatter       
\pagenumbering{arabic}
\setstretch{1.3}  

\fancyhead{}   
\fancyhead[LO]{\sl{\leftmark}}
\fancyhead[RE]{\sl{\rightmark}}
\fancyhead[LE,RO]{\thepage}



\chapter{Introduction} 
\chaptermark{Introduction}
\label{ch:introduction}

This thesis focuses on building intelligent algorithms to synthesize visually realistic images for various computer vision tasks. This chapter provides general background on visual synthesis and realism evaluation. The subsequent chapters will provide more background on each of the specific \emph{visual synthesis} tasks, and present new methods to address them. 

\section{Visual Synthesis and Generation}\label{ch:intro_task}

Visual imagery is one of the most important aspects of the computer world, being part of our modern life via various media applications, such as TikTok, WeChat, Facebook, and YouTube. As a result, people freely create millions of photos and videos per day on the internet, making it possible for researchers to collect astounding amounts of visual content to tame the artificial intelligence model for various computer vision tasks \cite{goodfellow2016deep}. 

\begin{figure}[tb!]
    \centering
    \includegraphics[width=\linewidth]{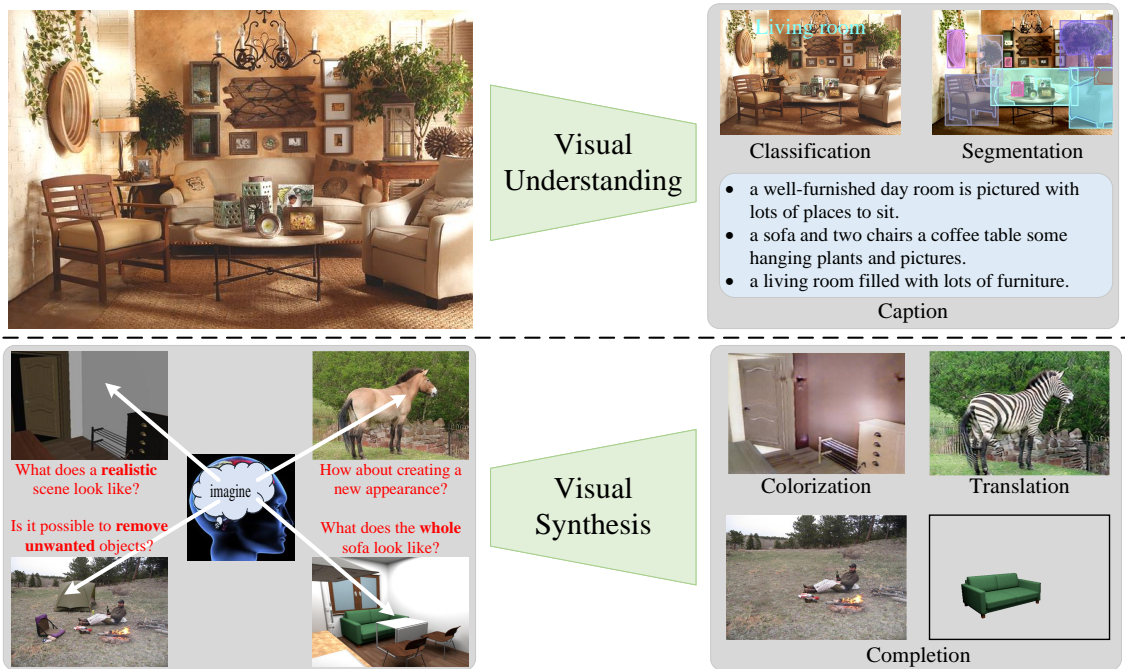}
    \caption[Visual synthesis as compared to visual understanding]{\textbf{The overall exhibition of the goal in this thesis.} In the top row, we first show the general \emph{visual understanding} tasks, which have achieved rapid advances, such as in image classification, instance segmentation, and image captioning, due to vast amounts of visual data along with deep learning networks \cite{goodfellow2016deep}. In this thesis, we attempt to explore the opposite direction, \emph{visual synthesis}, where we empower the model to imagine and generate new photorealistic images by estimating the data distribution. }
    \label{fig:intro_re_ge}
\end{figure}

In the computer vision community, researchers conventionally focus on recognition tasks \cite{lecun2015deep}. Due to the availability of these vast amounts of visual data and the advances of deep learning algorithms, the community has rapidly improved recognition results over a short period of time. For instance, in Figure \ref{fig:intro_re_ge} (top), we can now build powerful intelligent systems that accurately recognize the category of a scene \cite{vgg}, localize \cite{girshick2014rich} and segment \cite{pinheiro2015learning} object instances in it, and even describe the scene in natural language \cite{you2016image}. However, there is also the opposite research direction, \emph{visual synthesis}, which aims to create new visual content based on partial observation of real data. As shown in Figure \ref{fig:intro_re_ge} (bottom), we would like to teach machines to learn the capacity of imagination that humans are capable of, and be able to generate visual data with reasonable content and realistic appearance. For example, when a sofa is occluded by other furniture, can machines figure out what does the \textbf{whole} sofa look like? How would machines learn to imagine the missing content?

Why would \emph{visual synthesis and generation} be important for the computer vision community? One potential relevance is to \emph{self-supervised representation learning} \cite{bengio2013representation}. As we know, building datasets with extensive labels is a high-effort and high-cost undertaking, while the world is full of unlabeled, free data, particularly on the Internet. The traditional representation learning methods, such as Restricted Boltzmann Machines (RBM) \cite{smolensky1986information} and Auto-Encoders (AE) \cite{bourlard1988auto}, learn robust features without using labels by attempting to reconstruct the raw input. More recently, some methods \cite{kim2018learning,misra2020self} further attempted the more challenging label free tasks, such as colorization, completion,  solving jigsaws, and rotation prediction, resulting in more robust features for downstream tasks. Alternatively, the synthesized images can be used as augmented data for deep learning \cite{su2015render,zheng2018t2net}, especially for 3D-related tasks. For instance, as more and more high-quality 3D CAD models become available online \cite{song2017semantic, fu20203d}, it is possible to render an unlimited number of photorealistic images to support real-world tasks, \eg depth estimation, object detection and segmentation, and 3D reconstruction.

In addition to promoting machine understanding of the real world, \emph{visual synthesis and generation} can also create visual content that improves human-to-machine and computer-mediated human-to-human interaction. As mentioned above, people upload millions of images and videos per day on the internet, but often they are not entirely satisfied with the quality of such content. Supposing you have taken a photo of camping as shown in Figure \ref{fig:intro_re_ge}, but you would like to remove the unwanted objects, or create some new elements, or change the color and lighting. We desire to have an intelligent visual synthesis system that can be used to easily improve the picture, \eg removing unwanted objects in Figure \ref{fig:intro_re_ge}. In this way, we can help users easily synthesize more visually appealing photos to ideally express themselves better.

We investigate a number of data-driven visual synthesis and generation tasks for various applications in this thesis. In the following section, we will briefly define the tasks and give some background on the solutions. 

\subsection{Deep Generative Models}

Our methods are mainly built upon \emph{deep generative models}, which estimate complex high-dimensional data distributions using a set of variables in deep layers. Currently, the emerging powerful frameworks, including Generative Adversarial Networks (GANs) \cite{goodfellow2014generative} and Variational Autoencoders (VAEs) \cite{kingma2013auto}, have made impressive advances in many generation tasks. These models try to learn a function that maps the unknown distribution of training data $\textbf{x}$ from a predefined probability distribution of latent variable $\textbf{z}$. Formally,
\begin{equation}
    \textbf{x} = g(\textbf{z};\theta).
\end{equation}

In the VAE framework, an encoder $q(\textbf{z}|\textbf{x})$, acting as an approximate inference network, is used to obtain the latent variable $\textbf{z}$ from training instances. Once the model is learned, it can generate arbitrary data by resampling from the latent distribution. In a GAN framework, an auxiliary discriminator network is trained adversarially with the generator network, which should ideally lead to minimizing the distribution distance between the generated data and original data. 

\subsection{Image-to-Image Translation}

\emph{Image-to-Image (I2I) translation} involves designing algorithms that can learn to modify an input image $\textbf{x}$ to fit the \emph{style / appearance} of the target domain, while preserving the original \emph{content}, as shown in Figure \ref{fig:intro_re_ge}: \emph{horse} $\rightarrow$ \emph{zebra}. The process is to learn such a mapping:
\begin{equation}
    f: \textbf{x} \rightarrow \textbf{y}
\end{equation}
where the input image $\textbf{x}$ is translated to another image $\textbf{y}$ in the target domain. In this thesis, I2I refers to the task of only modifying the appearance, while the content / structure is preserved. 

One of the simplest forms is \emph{paired} I2I translation \cite{isola2017image}. In this case, the paired training examples $\{x_i,y_i\}_{i=1}^N$ are given, where the $y_i$ corresponds to each input $x_i$. However, obtaining such paired training data is difficult, expensive or even impossible in some situations. Therefore, following \cite{zhu2017unpaired}, we focus on unpaired I2I translation work that learns to translate between domains without paired input-output examples.  

\subsection{Image Completion}

\emph{Image completion} refers to the task of filling alternative reasonable content for missing or deleted parts in images, which can be used for restoring damaged paintings, removing unwanted objects, and generating new content for incomplete scenes. This task is a further development of traditional image ``inpainting'' \cite{bertalmio2000image}, which only works for narrow or small holes, due to the lack of deeper semantic understanding. Here, we investigate data-driven visual synthesis approaches to fill in semantic reasonable content with photorealistic appearance into arbitrary missing regions. In particular, given a masked image $\textbf{I}_m$ that is degraded by a number of missing pixels, the goal is to learn a model $\Phi$ to infer the content, conditioned on partially visible information:
\begin{equation}
    \textbf{I}_g = \Phi(\textbf{I}_m;\theta)
\end{equation}
where the input masked image $\textbf{I}_m$ is combined with newly synthesized content to become a completed image $\textbf{I}_g$.

The earlier learning-based approaches use the conventional convolutional operation to train a model in a deterministic way. In this thesis, I introduce a new direction, pluralistic image completion, that aims to generate multiple and diverse results for this highly subjective task. As it is important to explore the global visible information for missing content inference, a transformer-based image completion network is also later investigated in this thesis. 

\subsection{Completed Scene Decomposition}

The goal of \emph{completed scene decomposition} is to build an intelligent system that automatically \emph{decomposes} a scene into individual objects, \emph{infers} their underlying occlusion relationships and moreover \emph{imagines} what occluded objects may look like. This means that the learning algorithm must be able to \emph{understand} the scene to predict the geometry and categories of all objects in it (as shown in Figure \ref{fig:intro_re_ge} (top)), and also \emph{synthesize} invisible parts of objects and backgrounds (as shown in Figure \ref{fig:intro_re_ge} (bottom)). 

To do so, we aim at deriving a higher-level structure decomposition of a scene. As humans, we are highly aware of the shapes of individual objects and their ordering relationships, and we can generally imagine what occluded objects may look like. For instance, as shown in Figure \ref{fig:intro_re_ge} (bottom), humans can easily recognize the sofa and the table, and deduce the former is occluded by the latter, and even guess what the whole sofa looks like, based on global visible information and prior knowledge. 

\begin{figure}[tb!]
    \centering
    \includegraphics[width=\linewidth]{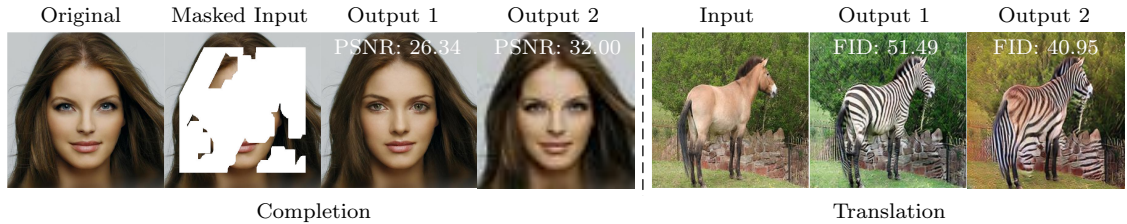}
    \begin{picture}(0,0)
		\put(-195,80){\scriptsize{Original}}
		\put(-146,80){\scriptsize{Masked Input}}
		\put(-80,80){\scriptsize{Output 1}}
		\put(-20,80){\scriptsize{Output 2}}
		\put(-85,68){\color{white}\scriptsize{PSNR: 26.34}}
		\put(-25,68){\color{white}\scriptsize{PSNR: 32.00}}
		\put(52,80){\scriptsize{Input}}
		\put(104,80){\scriptsize{Output 1}}
		\put(164,80){\scriptsize{Output 2}}
		\put(102,68){\color{white}\scriptsize{FID: 51.49}}
		\put(162,68){\color{white}\scriptsize{FID: 40.95}}
		\put(-114,6){\scriptsize{Completion}}
		\put(102,6){\scriptsize{Translation}}
	\end{picture}
	\vspace{-0.2cm}
    \caption[Evaluation metrics for visual image quality]{\textbf{Which output is the ``better'' result for each input in these examples?} In each case, the existing metrics disagree with human judgments. The traditional metrics, including $\ell_1$, PSNR, and SSIM, support the blurry ``Output 2'' in the first case because the latter is optimized only by $\ell_1$ reconstruction loss that encourages the same content to the original image. In the unpaired I2I task, the learning-based metrics, such as IS and FID, agree with the ``Output 2'' due to many results in this setting have repeated zebra's texture.}
    \label{fig:intro_eval}
\end{figure}

\section{Evaluation of Image Visual Realism}

\emph{Image quality evaluation} is a difficult research problem in computer vision. As laypeople, we may not be able to create realistic images just like artists, but we can easily judge whether a given image is ``realistic'', and we are able to correctly recognize which parts make a ``fake'' image appear unreal.

\paragraph{Classic Metrics} However, ``what makes a real image look realistic?'' has no clear answer in computer vision. In traditional works, researchers investigated a lot of factors, \eg color, texture, boundary, structure, and illumination, yet it is still hard to precisely define the impact of these factors mathematically. While various classic metrics, such as $\ell_1$ loss, Peak Signal-to-Noise Ratio (PSNR), and Structural SIMilarity (SSIM) \cite{wang2004image}, are proposed to assess image quality via unambiguous formulae, they are poorly related to human judgment due to independent pixel- and patch-level evaluation \cite{zhang2018unreasonable}. 

A well-known example is shown in Figure \ref{fig:intro_eval} (left). Compared to the high-quality completed ``Output 1'' \cite{zheng2021tfill}, the blurry completed ``Output 2'' has smaller $\ell_1$ reconstruction error (0.0144 \emph{vs} 0.0255), and larger PSNR (32.00 \emph{vs} 26.34) and SSIM score (0.9153 \emph{vs} 0.8248), with respect to the original image. This is because ``Output 2'' is trained using only $\ell_1$ reconstruction loss to the original unmasked image, for which blurry content can be smaller than content that is almost identical but slightly misaligned. Therefore, designing a ``perceptual metric'' that measures image quality similar to human judgment has been a longstanding goal. 

\paragraph{Learned Metrics} In more recent work, researchers have started focusing on learning-based feature-level distances, \eg Learned Perceptual Image Patch Similarity (LPIPS) metric \cite{zhang2018unreasonable}, Inception Score (IS) \citep{salimans2016improved} and Fr\'echet Inception Distance (FID) \citep{heusel2017gans}. These learned metrics mitigate the above-mentioned issue by evaluating the image quality in a deep neural network layer with large receptive field, instead of assuming pixel-wise independence in traditional metrics. For example, the LPIPS strongly agrees with human judgment that ``Output 1'' is more perceptual similar to the original image than the blurry ``Output 2''. However, these learned metrics are also not perfectly matched the human judgment as the currently pretrained networks tend to base their decisions much more on texture than shape \cite{geirhos2018imagenet}, while humans are more strongly focused on image structure \cite{landau1988importance} and related-context \cite{zhang2018unreasonable}. 

As an example depicted in Figure \ref{fig:intro_eval} (right), a horse is translated to the zebra domain, where there is no ground truth for evaluation. As humans, we can easily judge that ``Output 1'' is more realistic than ``Output 2''. However, the FID score, which compares the distance between distributions of translated and real images in a deep feature domain, evaluates ``Output 2'' as having a lower distribution distance, because all results in the second scenario have more obvious zebra stripes. The currently learned metrics are therefore not yet the perfect solutions for image quality evaluation.

\paragraph{Human Perceptual Metrics } Finally, we briefly introduce the human perceptual metrics, as proposed in \cite{zhang2016colorful} and widely adopted for image generation \cite{isola2017image,zhu2017unpaired,zhu2017toward,park2019semantic,Nazeri_2019_ICCV}. These are online metrics that the authors developed based on user studies, in which they provided some generated results and ground truth real images on Amazon Mechanical Turk (AMT)\footnote{https://www.mturk.com/}, and asked the participants to manually distinguish ``real'' and ``fake'' images. 

In this thesis, following existing state-of-the-art approaches, we report the corresponding evaluation metrics for different tasks. However, we would like to remind the reader that none of these are perfect for assessing generated image quality.

\section{Dissertation Overview}

The main research objective in this dissertation is to create new intelligent systems that can imagine and generate visually realistic natural photographs, which can be used in artistic creation, image editing, and further help real-world tasks (\eg depth estimation \cite{zheng2018t2net} and semantic segmentation \cite{zheng2021visiting}). As introduced in Section \ref{ch:intro_task}, in this dissertation, we explore three kinds of synthesis tasks: image translation \cite{zheng2018t2net,zheng2021spatiallycorrelative}, image completion \cite{zheng2019pluralistic,zheng2021tfill}, and completed scene decomposition \cite{zheng2021visiting}. 

\begin{itemize}
    \item \textbf{Part \uppercase\expandafter{\romannumeral1}. Changing Visual Appearance\quad} Chapters \ref{ch:Synthestic2Real} and \ref{ch:F(L)SeSim} describe methods for unpaired I2I translation that converts the visual appearance of the input image. In this task, deep learning is applied to learn a function $f:\bf{x} \rightarrow \bf{y}$, where $\bf{x}$ from a particular image domain $\mathcal{X}$, and $\bf{y}$ is the corresponding output that should belong to the target image domain $\mathcal{Y}$. Chapter \ref{ch:Synthestic2Real} focuses on \emph{synthetic-to-realistic} translation, in which I aim to bridge the gap between virtual and real scenes. This method is further integrated with a single-image depth estimation task to circumvent the challenges of obtaining accurate and sufficient 3D data of real scenes. In Chapter \ref{ch:F(L)SeSim}, a new content loss is introduced for arbitrary unpaired I2I translation tasks, in which I use self-similar correlations to better separate scene structure and appearance. 
    \item \textbf{Part \uppercase\expandafter{\romannumeral2}. Generating Semantic Reasonable Content\quad} Chapters \ref{ch:PICNet} and \ref{ch:TFill} present approaches that learn to fill alternative reasonable content into missing regions of degraded images. In Chapter \ref{ch:PICNet}, a new goal is introduced, \textbf{pluralistic image completion}, in which \emph{multiple} and \emph{diverse} plausible results are generated in a mathematically principled manner for this highly subjective process problem. Chapter \ref{ch:TFill}, further presents a newer framework that aims to generate single ``best'' result by directly modeling long-range dependencies in the masked image via a Transformer-based architecture. 
    \item \textbf{Part \uppercase\expandafter{\romannumeral3}. Modeling shape and appearance\quad} Chapter \ref{ch:VIV} combines the classical recognition task and the latest generation task into an end-to-end scene decomposition network, where a network is trained to \emph{decompose} a scene into individual objects, \emph{infer} their underlying occlusion relationships, and moreover \emph{imagine} what occluded objects may look like. In this task, a layer-by-layer algorithm is presented that can predict the geometry and categories of all objects in a scene, as well as generate their realistic appearance for originally occluded parts. 
    \item \textbf{Discussion\quad} In Chapter \ref{ch:Discussion}, we summarize the contributions of this thesis and discuss several future directions in image synthesis using deep learning. 
\end{itemize}

\part{Changing Visual Appearance: \\ Image-to-Image Translation}

\chapter{Synthetic-to-Realistic Translation} 
\chaptermark{Synthetic2Realistic}
\label{ch:Synthestic2Real} 

The main research goal presented in this chapter is to generate photorealistic images, which can be used to contribute the real-world single depth estimation task. The depth estimation is a classic research topic in computer vision, which has many different applications, such as autonomous driving, augmented reality, and scene reconstruction. As we live in a 3D world, humans are able to judge relative distances well even when only a single photograph is provided. However, it is still a challenge for a machine to accurately evaluate the depth from a \emph{single} RGB image. A main limitation is that 3D data is difficult to collect compared to the 2D images, due to requiring more specialized equipment. To address this issue, we aim to provide an alternative perspective by utilizing \emph{synthetic} image-depth pairs instead of real paired data. As more and more 3D CAD indoor scene models, such as in the SUNCG \cite{song2017semantic}\footnote{This work was published as \emph{T$^2$Net: Synthetic-to-Realistic Translation for Solving Single-Image Depth Estimation Tasks} in ECCV, 2018 \cite{zheng2018t2net}. At the time of publication, the SUNCG is still publicly available online.} and 3D-FRONT \cite{fu20203d} datasets, become publicly available on the internet, researchers can cheaply and effectively render a vast number of paired datasets. To bridge the gap between synthetic and real images, a \emph{wide-spectrum} translation network is proposed to convert synthetic-looking images with different levels of realism into realistic ones, such that after we train the depth estimation network on the translated images, the model can be directly applied to real images. 

The rest of this chapter is structured as follows: Sections \ref{ch2:intro} and \ref{ch2:back} describe the motivation and related works. Next, I explain the proposed framework in Section \ref{ch2:apporach}. Section \ref{ch2:data} introduces the synthetic datasets used. I then describe and discuss the experiments in Section \ref{ch2:result} and conclude in Section \ref{ch2:diss}.

\section{Introduction}\label{ch2:intro}

Single-image depth estimation is a challenging ill-posed problem for which good progress has been made in recent years, using supervised deep learning techniques \cite{eigen2014depth,eigen2015predicting,liu2016learning,laina2016deeper} that learn the mapping between image features and depth maps from large training datasets comprising image-depth pairs. An obvious limitation, however, is the need for vast amounts of paired training data for each scene type. Building such extensive datasets for specific scene types is a high-effort, high-cost undertaking \cite{saxena2009make3d,silberman2012indoor,Geiger2012we} due to the need for specialized depth-sensing equipment. The limitation is compounded by the difficulty that traditional supervised learning models face in generalizing to new datasets and environments \cite{liu2016learning}.

To mitigate the cost of acquiring large paired datasets, a few unsupervised learning methods \cite{garg2016unsupervised,godard2017unsupervised,kuznietsov2017semi} have been proposed, focused on estimating accurate disparity maps from easier-to-obtain binocular stereo images. Nonetheless, stereo imagery are still not as readily available as individual images, and systems trained on one dataset will find difficulty in generalizing well to other datasets (observed in \cite{godard2017unsupervised}), unless camera parameters and rigs are identical in the datasets.

\begin{figure}[tb!]
    \centering
    \includegraphics[width=\linewidth]{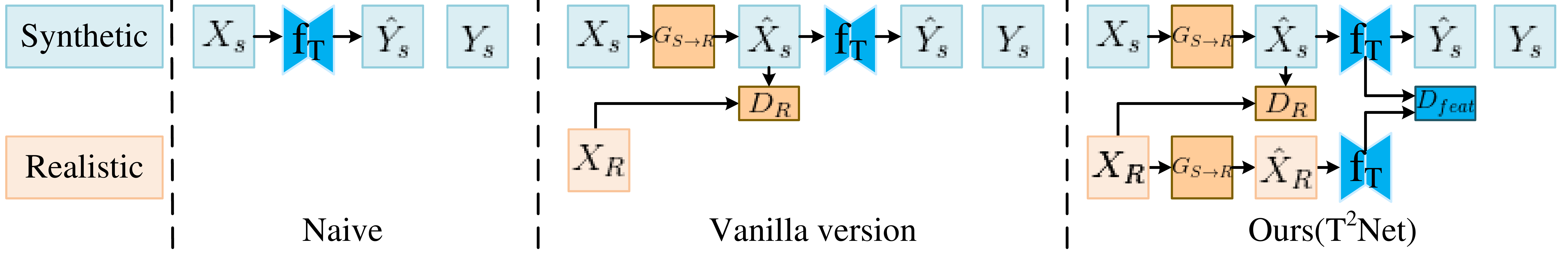}
    \caption[Various depth predication strategies]{\textbf{Depth prediction strategies given synthetic image-depth pairs $(x_s, y_s)$.} (Naive) structure directly trains an estimator using only synthetic image and depth maps. (Vanilla version) translates the synthetic image to the real domain and then trains the depth estimator. (Ours T$^2$Net) introduces a \emph{wide-spectrum} translation network that simultaneously maps the arbitrary images to target domains.}
    \label{fig:ch2_structure}
\end{figure}

A recent trend that has emerged from the challenge of real data acquisition is the approach of training on synthetic data for use on real data \cite{qiu2016unrealcv,shrivastava2017learning,hoffman2018cycada}, particularly for scenarios in which synthetic data can be easily generated. Inspired by these methods, we have researched a single-image depth estimation method that utilizes synthetic image-depth pairs instead of real paired data, but which also exploits the wide availability of unpaired real images. In short, our scenario is thus: we have a large set of real imagery, but these do not have any corresponding ground-truth depth maps. We also have access to a large set of synthetic 3D scenes\footnote{One 3D CAD model can be rendered to a vast number of paired data by setting different camera parameters.}, from which we can render multiple synthetic images from different viewpoints and their corresponding depth maps. The main goal then is to learn a depth map estimator when presented with a real image.
Consider two of the more obvious approaches: 1) Train an estimator using only synthetic image and depth maps, and hope that the estimator applies well to real imagery ({\bf Naive} in Figure \ref{fig:ch2_structure}). 2) Use a two-stage framework in which synthetic imagery is first translated into the real-image domain using a GAN, and then train the estimator as before ({\bf Vanilla version} in Figure \ref{fig:ch2_structure}).

The problem with 1) is that it is unlikely the estimator is oblivious to the differences between synthetic and real imagery. In 2), while a GAN may encourage synthetic images to map to the distribution of real images, it does not explicitly require the translated realistic image to have any physically-correct relationship to its corresponding depth map, meaning that the learned estimator will not apply well to actual real input. This may be somewhat mediated by introducing some regularization loss to try and keep the translated image ``similar'' in content to the original synthetic image (as in SimGAN \cite{shrivastava2017learning}), but we cannot identify any principled regularization loss functions, only heuristic ones.

In this chapter, an interesting perspective is introduced on the approach of 2). We propose to have the entire inference pipeline be agnostic as to whether the input image is real or synthetic, \ie it should work equally well regardless. To do so, we want the synthetic-to-realistic translation network to also behave as an identity transform when presented with real images, which is effected by including a reconstruction loss when training with real images.

The broad idea here is that, in a whole spectrum of synthetic images with differing levels of realism, \emph{the network should modify a realistic image less than a more obviously synthetic image}. This is not true of original GANs, which may transform a realistic image into a different realistic image. In short, for the synthetic-to-real translation portion, real training images are challenged with a reconstruction loss, while synthetic images are challenged with a GAN-based adversarial loss \cite{goodfellow2014generative}. This real-synthetic agnosticism is the principled formulation that allows us to dispense with an ad hoc regularization loss for synthetic imagery. When coupled with a task loss for the image-to-depth estimation portion, it leads to an end-to-end trainable pipeline that works well, and does not require the use of any real image-depth pairs nor stereo pairs ({\bf Ours(T$^2$Net)} in Figure~\ref{fig:ch2_structure}).

\section{Background}\label{ch2:back}

This task is related to two sets of work: \emph{single image depth estimation} and \emph{unpaired I2I translation}. Here, we briefly review these approaches. 

\paragraph{Single Image Depth Estimation} After classical learning techniques were earlier applied to single-image depth estimation \cite{Hoiem2005,Saxena2008,saxena2009make3d,karsch2012depth,ladicky2014pulling}, deep learning approaches took hold. In \cite{eigen2014depth}, a two-scale CNN architecture was proposed to learn the depth map from raw pixel values. This was followed by several CNN-based methods, which included combining deep CNN with continuous CRFs for estimating depth values \cite{liu2016learning}, simultaneously predicting semantic labels and depth maps \cite{wang2015towards}, and treating the depth estimation as a classification task \cite{cao2017estimating}. One common drawback of these methods is that they rely on large quantities of paired images and depths in various scenes for training. Unlike RGB images, real RGB-depth pairs are much scarcer. 

To overcome the above-mentioned problems, some unsupervised and semi-supervised learning methods have recently been proposed that do not require image-depth pairs during training.
In \cite{garg2016unsupervised}, the autoencoder network structure is translated to predict depths by minimizing the image reconstruction loss of image stereo pairs. More recently, this approach has been extended in \cite{godard2017unsupervised,kuznietsov2017semi}, where left-right consistency was used to ensure both good quality image reconstruction and depth estimation. While the data availability for these cases was perhaps not as challenging since special capture devices were not needed, nevertheless they depend on the availability or collection of stereo pairs with highly accurate rigs for consistent camera baselines and relative poses. This dependency makes it particularly difficult to cross datasets (\ie training on one dataset and testing on another), as evidenced by the results presented in \cite{godard2017unsupervised}. To alleviate this problem, an unsupervised adaption method \cite{tonioni2017unsupervised} was proposed to fine-tune a stereo network to a different dataset from which it was pre-trained on. This was achieved by running conventional stereo algorithms and confidence measures on the new dataset, but on much fewer images and at sparser locations.

\paragraph{Unpaired I2I Translation} Separately, several other works have explored image-to-image translation without using paired data. The earlier style-translation networks \cite{gatys2016image,johnson2016perceptual} would synthesize a new image by combining the "content" of one image with the "style" of another image. 
In \cite{liu2016coupled}, the weight-sharing strategy was introduced to learn a joint representation across domains. This framework was extended in \cite{liu2017unsupervised} by integrating variational autoencoders and generative adversarial networks. Other concurrent works \cite{zhu2017unpaired,kim2017learning,yi2017dualgan} utilized cycle consistency to encourage a more meaningful translation. However, these methods were focused on generating visually pleasing images, whereas for us image translation is an intermediate goal, with the primary objective being depth estimation, and thus the fidelity of 3D shape semantics in the translation has overriding importance.

In \cite{shrivastava2017learning}, a SimGAN was proposed to render realistic images from synthetic images for gaze estimation as well as human hand pose estimation. A self-regularization loss is used to force the generated target images to hold the similar content to the original source images. However, we consider this loss to be somewhat ad hoc and runs counter to the translation effort; it may work well in small domain shifts, but is too limiting for large style translation in our problem. As such, we use a more principled reconstruction loss as detailed in Section \ref{ch2:apporach}. More recently, a cycle-consistent adversarial domain adaption method was proposed \cite{hoffman2018cycada} to generate target domain training images for digit classification and semantic segmentation. However this method is too complex for end-to-end training, which we consider to be an important requirement to achieve good results.

\section{Overview}\label{ch2:over}
The main research goal is to train an image-to-depth network $f_T$, such that when presented with a single RGB image, it predicts the corresponding depth map accurately.

In terms of data availability for training, we assume that we have access to a collection of individual real-world images $x_r$, \emph{without} stereo pairing nor corresponding ground truth depth maps. Instead, we assume that we have access to a collection of synthetic 3D models, from which it is possible to render numerous synthetic images and corresponding depth maps, denoted in pairs of $(x_s, y_s)$.

Instead of directly training $f_T$ on the synthetic $(x_s, y_s)$ data, we expect that the synthetic images are insufficiently similar to the real images, to require a prior image translation network $G_{S\to R}$ for domain adaptation to make the synthetic images more realistic.  However, as discussed previously, existing image translation methods do not adequately preserve the geometric content for accurate depth prediction, or require heuristic regularization loss functions.

The \emph{key novel insight} is this: instead of training $G_{S\to R}$ to be a narrow-spectrum translation network that translates one specific domain to another, we will train it as a \emph{wide-spectrum} translation network, to which we can feed a range of input domains, \ie synthetic imagery as well as actual real images. The intention is to have $G_{S\to R}$ implicitly learn to apply the minimum change needed to make an image realistic, and consider this the most principled way to regularize a network for preserving shape semantics needed for depth prediction.

\begin{figure}[tb!]
	\centering
	\includegraphics[width=\textwidth]{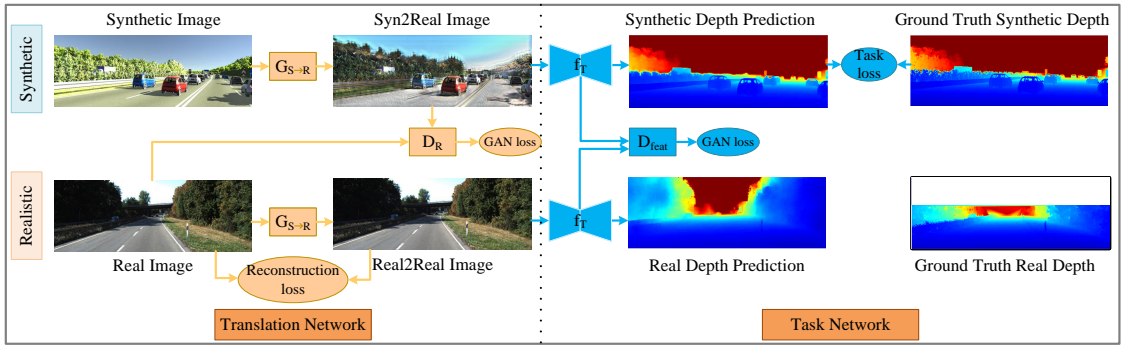}
	\caption[The overall pipeline of the proposed method]{\textbf{The overall pipeline of the proposed method.} The proposed $T^2$Net consists of the Translation part (left, orange) and Task prediction part (right, blue). The $G_{S\to R}$ is a generator to translate images from synthetic domain to real domain, and $D_{R}$ is the corresponding discriminator to judge whether the translated image is real or fake. $f_T$ is depth task estimation network and $D_{feat}$ is a discriminator on feature domain. }
	\label{fig:ch2_framework}
\end{figure}

Next, I introduce the proposed approach (Section \ref{ch2:apporach}) and the data collection (Section \ref{ch2:data}) in details. 

\section{Approach}\label{ch2:apporach}

To achieve the above-mentioned goal, a twin pipeline training framework is proposed (shown in Figure~\ref{fig:ch2_framework}), which is named as T$^2$Net to highlight the combination of an image \emph{\textbf{t}}ranslation network and a \emph{\textbf{t}}ask prediction network. The upper portion shows the training pipeline with synthetic $(x_s, y_s)$ pairs, while the lower portion shows the training pipeline with real images $x_r$. Note that both pipelines share identical weights for the $G_{S\to R}$ network, and likewise for the $f_T$ network. More specifically:
\begin{itemize}
	\item For real images, we want $G_{S\to R}$ to behave as an AE \cite{bourlard1988auto} and apply minimal change to the images, and thus use a \emph{reconstruction loss}.
	\item For synthetic data, we want $G_{S\to R}$ to translate synthetic images into the real-image domain, and use a \emph{GAN loss} via discriminator $D_R$ on the output. The translated images are next passed through $f_T$ for depth prediction, and then compared to the synthetic ground truth depths $y_s$ via a \emph{task loss}.
	\item In addition, we also propose that the inner feature representations of $f_T$ should share similar distributions for both real and translated images, which can be implemented through a feature-based GAN via $D_{\text{feat}}$. 
\end{itemize}
Note that one key benefit of this framework is that it can and should be trained end-to-end, with the weights of $G_{S\to R}$ and $f_T$ simultaneously optimized. 

\subsection{Synthesis Loss}

Intuitively, the gap between synthetic and realistic imagery comes from low-level differences such as color and texture (\eg of trees, roads), rather than high-level geometric and semantic differences. To bridge this gap between the two domains, an ideal translator network, for use within an image-to-depth framework, needs to output images that are impossible to be distinguished from real images and yet retain the original scene geometry present in the synthetic input images. The distribution of real-world images can be replicated using adversarial learning, where a generator $G_{S\to R}$ tries to transform a synthetic image $x_s$ to be indistinguishable from real images of $x_r$, while a discriminator $D_R$ aims to differentiate between the generated image $\hat{x}_s$ and real images $x_r$. Following the typical GAN approach \cite{goodfellow2014generative}, we model this minimax game using an \emph{adversarial loss} given by
\begin{equation} \label{eq:1}
\small
\mathcal{L}_{\text{GAN}}(G_{S\to R}, D_R) = \mathbb{E}_{x_r\sim X_R}[\log D_R(x_r)] + \mathbb{E}_{x_s\sim X_S}[\log(1-D_R(G_{S\to R}(x_s)))]
\end{equation}
where generator and discriminator parameters are updated alternately.

However, a vanilla GAN is insufficiently constrained to preserve scene geometry \cite{isola2017image}. To regularize this in a principled manner, we want generator $G_{S\to R}$ to behave as a \emph{wide-spectrum} translator, able to take in both real and synthetic imagery, and in both cases produce real imagery. When the input is a real image, we would want the image to remain as much unchanged perceptually, and a \emph{reconstruction loss}
\begin{equation}\label{eq:2}
\mathcal{L}_{r}(G_{S\to R}) = ||G_{S\to R}(x_r) - x_r||_1
\end{equation}
is applied when the input to $G_{S\to R}$ is a real image $x_r$. Note that while this may bear some resemblance to the use of reconstruction losses in CycleGAN \cite{zhu2017unpaired} and $\alpha$-GAN \cite{rosca2017variational}, ours is a unidirectional forward loss, and not a cyclical loss.

\subsection{Task Loss}
After a synthetic image $x_s$ is translated, we obtain a generated realistic image $\hat{x}_s$, which can still be paired to the corresponding synthetic depth map $y_s$. This paired translated data $(\hat{x}_s, y_s)$ can be used to train the task network $f_T$. Following convention, we directly measure per-pixel difference between the predicted depth map and the synthetic (ground truth) depth map as a task loss:
\begin{equation}\label{eq:3}
\mathcal{L}_{t}(f_T) = ||f_T(\hat{x}_s)-y_s||_1.
\end{equation}

We also regularize $f_T$ for real training images. Since real ground truth depth maps are not available during training, a locally smooth loss is introduced to guide a more reasonable depth estimation, in keeping with \cite{heise2013pm,garg2016unsupervised,godard2017unsupervised,kuznietsov2017semi}. As depth discontinuities often occur at object boundaries, we use a robust penalty with an edge-aware term to optimize the depths, similar to \cite{godard2017unsupervised}:
\begin{equation}\label{eq:4}
\mathcal{L}_{s}(f_T) = |\partial_xf_T(x_r)|e^{-|\partial_xx_r|} + |\partial_yf_T(x_r)|e^{-|\partial_yx_r|}
\end{equation}
where $x_r$ is the real-world image, and noting that $f_T$ share identical weights in both real and synthetic input pipelines.

In addition, we also want the internal feature representations of real and translated synthetic images in the encoder-decoder network of $f_T$ to have similar distributions \cite{ganin2015unsupervised}. In theory, the decoder portion of $f_T$ should generate similar prediction results from the two domains when their feature distributions are similar. Thus we further define a feature-level GAN loss as follows:
\begin{equation}\label{eq:5}
\small
\mathcal{L}_{\text{GAN}_f}(f_T, D_{\text{feat}}) =
\mathbb{E}_{f_{\hat{x}_s}\sim f_{\hat{X}_s}}[\log D_{\text{feat}}(f_{\hat{x}_s})] +
\mathbb{E}_{f_{x_r}\sim f_{X_r}}[\log(1-D_{\text{feat}}(f_{x_r}))]
\end{equation}  
where $f_{\hat{x}_s}$ and $f_{x_r}$ are features obtained by the encoder portion of $f_T$ for translated-synthetic images and real images respectively. As noted in \cite{goodfellow2014generative}, the optimal solution measures the Jensen-Shannon divergence between the two distributions.

\subsection{Full Objective}

Taken together, our full objective is:
\begin{align}\label{eq:6}
\mathcal{L}_{\text{T}^2\text{Net}}(G_{S\to R}, f_T, D_R,D_\text{feat}) =
& \mathcal{L}_\text{GAN}(G_{S\to R}, D_R) +  \alpha_{f}\mathcal{L}_{\text{GAN}_{f}}(f_T,D_\text{feat}) \nonumber\\
& + \alpha_{r}\mathcal{L}_{r}(G_{S\to R}) + \alpha_{t}\mathcal{L}_{t}(f_T) + \alpha_{s}\mathcal{L}_{s}(f_T) 
\end{align}
where $\mathcal{L}_\text{GAN}$ encourages translated synthetic images to appear realistic, $\mathcal{L}_{r}$ spurs translated real images to appear identical, $\mathcal{L}_{\text{GAN}_{f}}$ enforces closer internal feature distributions, $\mathcal{L}_t$ promotes accurate depth prediction for synthetic pairs, and $\mathcal{L}_s$ prefers an appropriate local depth variation for real predictions. In our end-to-end training, this objective is used in solving for optimal $f_T$ parameters:
\begin{equation}
f_T^* = \arg \min_{f_T} \min_{G_{S\to R}} \max_{D_R,D_\text{feat}}
\mathcal{L}_{\text{T}^2\text{Net}}(G_{S\to R}, f_T, D_R,D_\text{feat}).
\end{equation}

\subsection{Network Architecture}

The transform network, $G_{S\to R}$, is a residual network (ResNet) \cite{he2016deep} similar to SimGAN \cite{shrivastava2017learning}. Limited by memory constraints and the large size of scene images, one down-sampling layer is used in our model and the output is only passed through 6 blocks. For the image discriminator networks, we use PatchGANs \cite{shrivastava2017learning,zhu2017unpaired}, which have produced impressive results by discriminating locally whether image patches are real or fake.

The task prediction network is inspired by \cite{godard2017unsupervised}, which outputs four predicted depth maps of different scales. Instead of encoding input images into very small dimensions to extract global information, we instead use multiple dilation convolutions \cite{YuKoltun2016} with a large feature size to preserve fine-grained details. In addition, we employ different weights for the paths with skip connections \cite{ronneberger2015u}, which can simultaneously process larger-scale semantic information in the scene and yet also predict detailed depth maps. The use of these techniques allows our task prediction network $f_T$ to achieve state-of-the-art performance in our own real-supervised benchmark method (training $f_T$ on pairs of real images and depth), even when the encoder portion of $f_T$ is primarily based on VGG, as opposed to a more typical ResNet50-type network used in other methods \cite{godard2017unsupervised,kuznietsov2017semi}. 

\section{Data Collection}\label{ch2:data}

As mentioned above, in this work, we assume the synthetic images and the corresponding depth maps are easily obtained due to more and more 3D CAD models are publicly available. In the experiments, we collected both indoor and outdoor synthetic datasets.

\paragraph{Synthetic Indoor Dataset} To generate the paired synthetic training data, we rendered RGB images and depth maps from the SUNCG dataset \cite{song2017semantic}, which contains 45,622 3D houses with various room types and all 3D CAD models are publicly available at the time of publication. We chose the camera locations, poses, and parameters based on the distribution of real NYUDv2 dataset \cite{silberman2012indoor}. We retained valid depth maps using the criteria presented in \cite{song2017semantic}: a) valid depth area (depth values in the range of 1m to 10m) larger than 70\% of the image area, and b) more than two object categories in the scene. The RGB images are rendered using the default OpenGL-rendered method, resulting in low realism\footnote{In Chapter \ref{ch:VIV}, a high-quality image rendering pipeline is introduced using Maya \cite{Maya}, yet it still can not fully match a given real dataset.}. Therefore, our synthetic to realistic translation approach is applied to fit the synthetic images to real. Theoretically, we can render infinite numbers of paired images for training with the whole 3D CAD models of scenes. In this work, we generated 130,190 valid views from 4,562 different houses.

\paragraph{Synthetic Outdoor Dataset} We used Virtual KITTI (vKITTI) \cite{gaidon2016virtualworlds}, a photo-realistic synthetic dataset that contains 21,260 image-depth paired frames generated from different virtual urban worlds. The scenes and camera viewpoints are similar to the real KITTI dataset \cite{Menze2015CVPR}. However, the ground truth depths in vKITTI and KITTI are quite different. The maximum sensed depth in a real KITTI image is typically on the order of 80m, whereas vKITTI has precise depths to a maximum of 655.3m because it is rendered from Unity game engine without equipment limitation. To reduce the effect of ground truth differences, the vKITTI depth maps were clipped to 80m.

\section{Experiment}\label{ch2:result}

We evaluated our model on the outdoor KITTI dataset \cite{Geiger2012we} and the indoor NYU Depth v2 dataset \cite{silberman2012indoor}. During the training process, we only used unpaired real images from these datasets in conjunction with synthetic image-depth pairs, obtained via SUNCG \cite{song2017semantic} and vKITTI \cite{gaidon2016virtualworlds} datasets, in our proposed framework.

\subsection{Implementation Details}\label{sec:ch2_implementation}

\paragraph{Training Details} In order to control the effect of GAN loss, we substituted the vanilla negative log likelihood objective with a least-squares loss \cite{mao2016multi}, which has proven to be more stable during adversarial learning \cite{zhu2017unpaired}. Hence, for GAN loss $\mathcal{L}_\text{GAN}(G_{S\to R}, D_R)$ in (\ref{eq:1}), we trained $G_{S\to R}$ by minimizing
\begin{equation*}
\mathbb{E}_{x_s\sim X_s}[(D_R(G_{S\to R}(x_s))-1)^2]
\end{equation*}
and trained $D_R$ by minimizing
\begin{equation*}
\mathbb{E}_{x_r\sim X_r}[(D_R(x_r)-1)^2] + \mathbb{E}_{x_s\sim X_s}[D_R^2(G_{S\to R}(x_s))].
\end{equation*}
A similar procedure was also applied for the GAN loss in (\ref{eq:5}).

\paragraph{Our $f_T$-only Benchmark Models} Besides our full T$^2$Net model, we also tested our partial model, which comprised solely the $f_T$ task prediction network. We evaluated this in two scenarios: (1) an \textbf{``all-real''} scenario, in which we used real image and depth map pairs for training, for which we would expect to \emph{upper bound} our full model performance, and (2) an \textbf{``all-synthetic'' (naive version)} scenario, in which we used only synthetic image-depth pairs and eschewed even unpaired real images, for which we would expect to \emph{lower bound} our full model performance.

\paragraph{Evaluation Metrics} We evaluated the performance of our approach using the depth evaluation metrics reported in \cite{eigen2014depth}:

\begin{equation}
\footnotesize
\renewcommand{\arraystretch}{1.5}
\setlength{\arraycolsep}{0pt}
\begin{array}{ll}
{\bf RMSE(log):} \sqrt{\frac{1}{|T|}\sum_{i=1}^{T}||\log \hat{y}_{r,i} - \log y_{r,i} ||^2} & {\bf RMSE:} \sqrt{\frac{1}{|T|}\sum_{i=1}^{T}||\hat{y}_{r,i} - y_{r,i}||^2}\\
{\bf Sq.\ relative:} \frac{1}{|T|}\sum_{i=1}^{T}{||\hat{y}_{r,i} - y_{r,i}||^2}/{y_{r,i}} & {\bf Abs\ relative:} \frac{1}{|T|}\sum_{i=1}^{T}{|\hat{y}_{r,i} - y_{r,i}|}/{y_{r,i}} \\
{\bf Accuracy:\ \% \ of\ y_{r,i}\ s.t.}\ max(\frac{\hat{y}_{r,i}}{y_{r,i}}, \frac{y_{r,i}}{\hat{y}_{r.i}})=\delta < thr
\end{array}
\end{equation}

\subsection{NYUDv2 Dataset}

\begin{figure}[tb!]
	\centering
	\includegraphics[width=\textwidth]{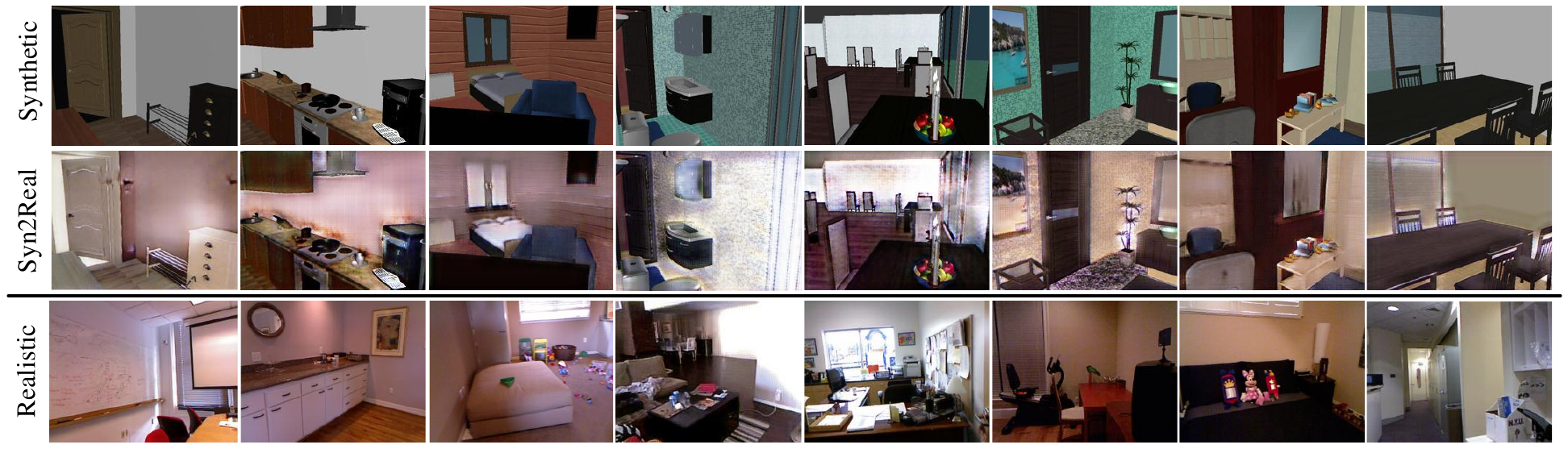}
	\caption[Example outputs for indoor scene]{\textbf{Example outputs of our translation network for indoor scene.} Top: synthetic images rendered from SUNCG. Middle: corresponding images after $G_{S\to R}$ translation. Bottom: real images from NYUDv2 \cite{silberman2012indoor} (no correspondence to above rows). 
	}
	\label{fig:ch2_syn2real_nyu}
\end{figure}

\paragraph{Translated Results} Figure~\ref{fig:ch2_syn2real_nyu} shows sample output from translation through $G_{S\to R}$. We observe that the visual differences between synthetic and real images are obvious: colors, textures, illumination and shadows in real scenes are more complex than in synthetic ones. Compared to synthetic images, the translated images are visually more similar to real images in terms of low-level appearance.

\begin{table}[tb!]
    \centering
    \footnotesize
    \renewcommand{\arraystretch}{1.1}
    \setlength\tabcolsep{3pt}
    \begin{tabular}{@{}lcccccccc@{}}
         \hlineB{3.5} 
         \multirow{2}{*}{Method}& \multicolumn{4}{c}{$\downarrow$} && \multicolumn{3}{c}{$\uparrow$} \\
         \cline{2-5}\cline{7-9}
         & Abs Rel & Sq Rel & RMSE & RMSE log && $\delta<1.25$ & $\delta<1.25^2$ & $\delta<1.25^3$ \\
         \hlineB{2.5}
         Ladicky et al. \cite{ladicky2014pulling} & - & - & - & - && 0.542 & 0.829 & 0.940 \\
		 Eigen et al. \cite{eigen2014depth} Fine & 0.215 & 0.212 & 0.907 & 0.285 && 0.611 & 0.887 & 0.971 \\
		 Liu et al. \cite{liu2016learning} & 0.213 & - & 0.759 & - && 0.650 & 0.906 & 0.976 \\
		 Eigen et al. \cite{eigen2015predicting} (VGG) & 0.158 & 0.121$^*$ & 0.641 & 0.214 && 0.769 & 0.950$^*$ & 0.988$^*$ \\
		 \hline
		 \rowcolor[rgb]{0.9,0.9,0.9}
			Baseline, train set mean & 0.439  & 0.641 & 1.148 & 0.415 && 0.412 & 0.692 & 0.856 \\  
		 \hline
		 Our $f_T$, all-real &  0.157$^*$ & 0.125 & 0.556$^*$ & 0.199$^*$ && 0.779$^*$ & 0.943 & 0.983 \\
		\rowcolor[rgb]{0.9,0.9,0.9}
		Our $f_T$, all-synthetic & 0.304 & 0.394 & 1.024 & 0.369 && 0.458 & 0.771 & 0.916 \\
		\rowcolor[rgb]{0.9,0.9,0.9}
		Our T$^2$Net, $D_\text{feat}$ only & 0.320 & 0.405 & 0.991 & 0.343 && 0.480 & 0.792 & 0.933 \\
		\rowcolor[rgb]{0.9,0.9,0.9}
		Our T$^2$Net, $D_\text{image}$ only & 0.274 & 0.336 & 1.001 & 0.325 && 0.496 & 0.814 & 0.938 \\
		\rowcolor[rgb]{0.9,0.9,0.9}
		Our full T$^2$Net & {\bf 0.257} & {\bf 0.281} & {\bf 0.915} & {\bf 0.305}  && {\bf 0.540} & {\bf 0.832} & {\bf 0.948} \\
		\hlineB{2.5}
    \end{tabular}
    \caption[Depth estimation results on indoor scene]{\textbf{Depth estimation results on NYUDv2 dataset \cite{silberman2012indoor}.} \emph{Gray rows indicate methods in which training is conducted \textbf{without} real image-depth pairs. Best supervised results are marked with *, while best unsupervised results are in bold}. $\downarrow$ = lower is better. $\uparrow$ = higher is better.}
    \label{tab:ch2_indoor}
\end{table}

\begin{figure}[tb!]
	\centering
	\includegraphics[width=\textwidth]{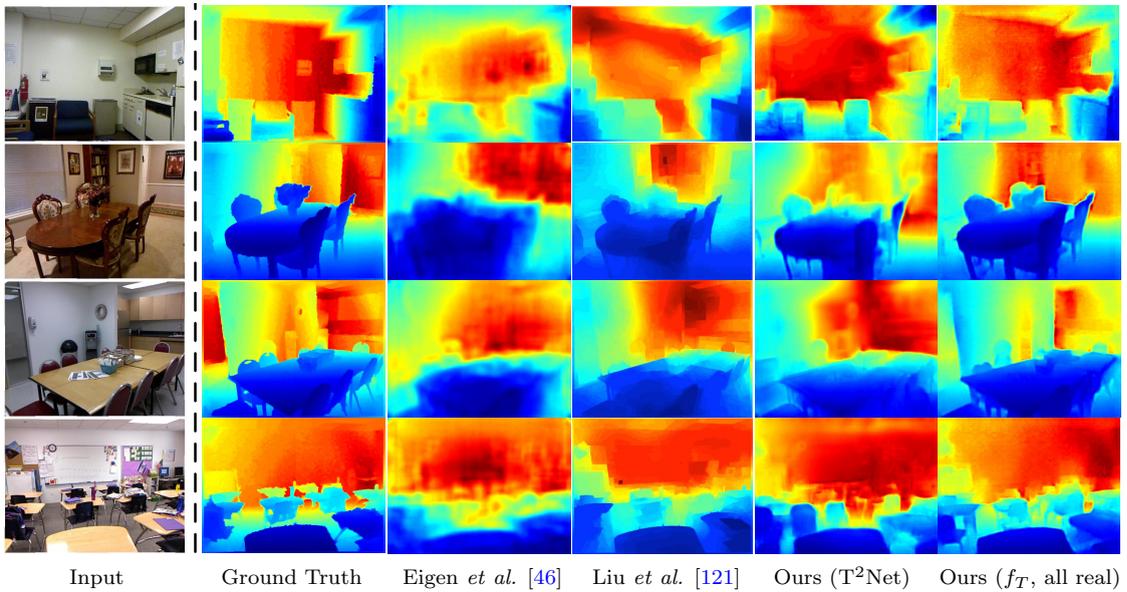}
	\begin{picture}(0,0)
		\put(-183,4){\scriptsize Input}
		\put(-126,4){\scriptsize Ground Truth}
		\put(-58,4){\scriptsize Eigen \etal \cite{eigen2014depth}}
		\put(13,4){\scriptsize Liu \etal \cite{liu2016learning}}
		\put(81,4){\scriptsize Ours (T$^2$Net)}
		\put(143,4){\scriptsize Ours ($f_T$, all real)}
	\end{picture}
	\vspace{-0.2cm}
	\caption[Qualitative results on NYUDv2]{\textbf{Qualitative results on NYUDv2.} All results are shown as relative depth maps (red = far, blue = close).}
	\label{fig:ch2_depth_indoor}
\end{figure}

\paragraph{Depth Estimation Results} In Table \ref{tab:ch2_indoor}, we report the performance of our models (varying different applications of the two GANs) as compared to latest state-of-the-art methods on the public NYUDv2 dataset. In the indoor dataset, these previous works were all based on supervised learning with real image-depth pairs. The gray rows highlight methods in which real image-depth pairs were \emph{not} used in training. The {\bf train-set-mean} baseline used the mean synthetic depth map in the training dataset as prediction, with the results providing an indication of the correlation between depth maps in the synthetic and real datasets.
We also present results from our $f_T$-only benchmark models in the ``all-real'' and ``all-synthetic'' setups, which we expect to provide the upper bound and lower bound of our model respectively.

Our proposed models produced a clear gap to the train-set-mean baseline and the synthetic-only benchmark. While our models were unable to outperform the latest fully-supervised methods trained on real paired data, the full T$^2$Net model was even able to outperform the earlier supervised learning method of \cite{ladicky2014pulling} on two of the three metrics, despite not using real paired data.

We also show qualitative results in Figure~\ref{fig:ch2_depth_indoor}. Although the absolute values of our predicted depths were not as accurate as the latest supervised learning methods, we observe that our T$^2$Net model generates reasonably good relative depths with distinct furniture shapes, even without using real paired training data. 

\begin{figure}[tb!]
	\centering
	\includegraphics[width=\textwidth]{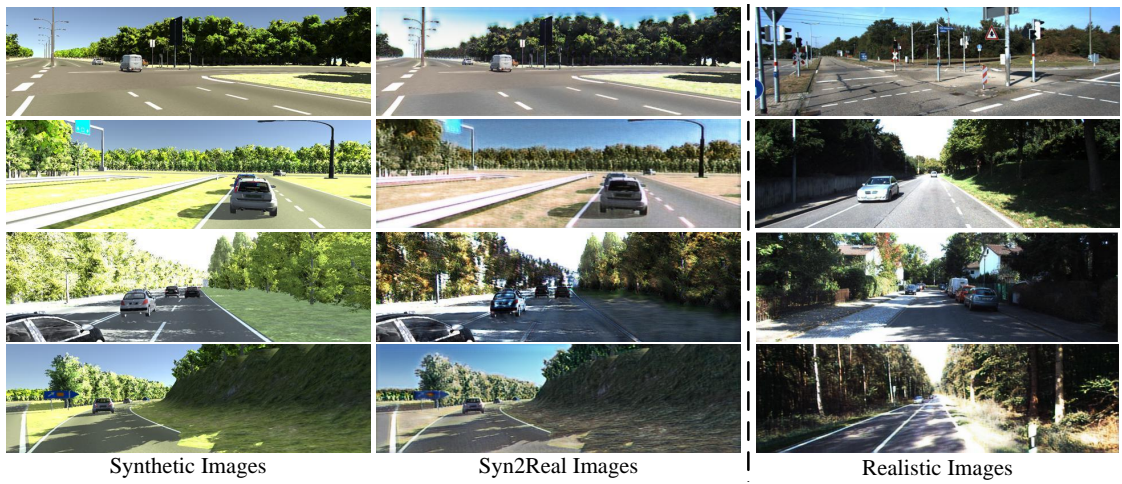}
	\caption[Example translated images for the outdoor scene]{\textbf{Example translated images for the outdoor vKITTI dataset \cite{gaidon2016virtualworlds}.} (Left) synthetic images from vKITTI and translated images. (Right) images in real KITTI.}
	\label{fig:ch2_syn2real_kitti}
\end{figure}

\subsection{KITTI Dataset}

\paragraph{Translated Results} Figure \ref{fig:ch2_syn2real_kitti} shows examples of synthetic, translated, and real images from the outdoor datasets. As shown, the translated images have substantially greater resemblance to the real images than the synthetic images. Our translation network can visually replicate the distributions of colors, textures, shadows and other low-level features present in the real images, and meanwhile preserve the scene geometry of the original synthetic images.

\paragraph{Depth Estimation Results} In order to compare with previous work, we used the test split of 697 images proposed in \cite{eigen2014depth}. Following \cite{godard2017unsupervised}, we chose 22,600 RGB images from the remaining 32 scenes for training the translation network. As before, we did not use real depths nor stereo pairs in our T$^2$Net models. The ground truth depth maps in KITTI were obtained by aligning laser scans with color images, which produced less than $5\%$ depth values and introduced sensor errors. For fair comparison with state-of-the-art single view depth estimation methods, we evaluated our results based on the cropping given in \cite{garg2016unsupervised} and clamping the predicted depth values within the range of 1--50m.

\begin{table}[tb!]
    \centering
    \tiny
    \renewcommand{\arraystretch}{1.3}
    \setlength\tabcolsep{4pt}
    \begin{tabular}{@{}lcccccccccc@{}}
         \hlineB{3.5} 
         \multirow{2}{*}{Method} & \multirow{2}{*}{Dataset} & \multirow{2}{*}{Cap} & \multicolumn{4}{c}{$\downarrow$} && \multicolumn{3}{c}{$\uparrow$} \\
         \cline{4-7}\cline{9-11}
         & & & Abs Rel & Sq Rel & RMSE & RMSE log && $\delta<1.25$ & $\delta<1.25^2$ & $\delta<1.25^3$ \\
         \hlineB{2.5}
         	Eigen et al.\cite{eigen2014depth} Fine & K(I+D) & 0-80m & 0.190 & 1.515 & 7.156 & 0.270 && 0.692 & 0.899 & 0.967 \\
			\hline 
			Garg et al.\cite{garg2016unsupervised} L12 Aug.8x & K(L+R) & 1-50m & 0.169 & 1.080 & 5.104 & 0.273 && 0.740 & 0.904 & 0.962 \\  
			Godard et al. \cite{godard2017unsupervised} & CS+K(L+R) & 1-50m & 0.117 & 0.762 & 3.972 & 0.206 && 0.860 & 0.948 & 0.976 \\
			Kuznietsov et al. \cite{kuznietsov2017semi} & K(D+L+R) & 1-50m & 0.108$^*$ & 0.595$^*$ & 3.518$^*$ & 0.179 && 0.875$^*$ & 0.964$^*$ & 0.988$^*$ \\
			\hline
			\rowcolor[rgb]{0.9,0.9,0.9}
			Baseline, train set mean &  vK(I+D) & 1-50m & 0.521 & 11.024 & 10.598 & 0.473 && 0.638 & 0.755 & 0.835\\
			\hline
			Our $f_T$, all-real  & K(I+D) & 1-50m & 0.114	& 0.627& 3.549 & 0.178$^*$ && 0.867 & 0.960 & 0.986 \\
			\rowcolor[rgb]{0.9,0.9,0.9}
			Our $f_T$, all-synthetic  & vK(I+D) & 1-50m & 0.278 & 3.216 & 6.268 & 0.322 && 0.681 & 0.854 & 0.929\\
			\rowcolor[rgb]{0.9,0.9,0.9}
			Our T$^2$Net, $D_\text{feat}$ only & vK(I+D) + K(I) & 1-50m &  0.233 & 2.902 & 6.285 & 0.300 && 0.743 & 0.880 & 0.938\\
			\rowcolor[rgb]{0.9,0.9,0.9}
			Our T$^2$Net, $D_\text{image}$ only&  vK(I+D) + K(I) & 1-50m & {\bf 0.168} & {\bf 1.199} & {\bf 4.674} & {\bf 0.243} && {\bf 0.772} & {\bf 0.912} & {\bf 0.966}\\
			\rowcolor[rgb]{0.9,0.9,0.9}
			Our full T$^2$Net &vK(I+D) + K(I) & 1-50m & 0.169 & 1.230 & 4.717& 0.245 && 0.769 & {\bf 0.912} & 0.965 \\
			\hlineB{2.5}
    \end{tabular}
    \caption[Depth prediction results on KITTI 2015]{\textbf{Results on KITTI 2015 \cite{Menze2015CVPR}} using the split of Eigen \etal \cite{eigen2014depth}. For dataset, K is the real KITTI dataset \cite{Menze2015CVPR}, CS is Cityscapes \cite{Cordts2016Cityscapes} and vK is the synthetic KITTI dataset \cite{gaidon2016virtualworlds}. L, R are the left and right stereo images, and I, D are the images and depths. \emph{The gray rows highlight methods that did not use real image-depth pairs nor stereo pairs for training. Best real-supervised or stereo-based results are marked with *, while best unsupervised results are in bold.} $\downarrow$ = lower is better. $\uparrow$ = higher is better. }
    \label{tab:ch2_outdoor}
\end{table}

\begin{figure}[tb!]
	\centering
	\includegraphics[width=\textwidth]{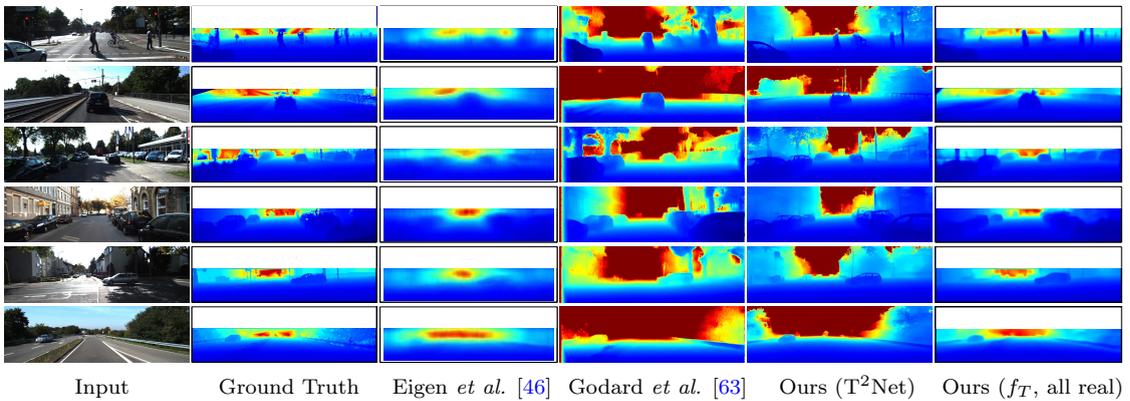}
	\begin{picture}(0,0)
		\put(-181,4){\scriptsize Input}
		\put(-127,4){\scriptsize Ground Truth}
		\put(-62,4){\scriptsize Eigen \etal \cite{eigen2014depth}}
		\put(4,4){\scriptsize Godard \etal \cite{godard2017unsupervised}}
		\put(83,4){\scriptsize Ours (T$^2$Net)}
		\put(144,4){\scriptsize Ours ($f_T$, all real)}
	\end{picture}
	\vspace{-0.2cm}
	\caption[Qualitative results on KITTI]{\textbf{Qualitative results on KITTI} with Eigen split \cite{eigen2014depth}. The ground truth depths in the original dataset were very sparse and have been interpolated for visualization. We converted the disparity maps provided in \cite{godard2017unsupervised} to depth maps.}
	\label{fig:ch2_depth_outdoor}
\end{figure}

Table \ref{tab:ch2_outdoor} shows quantitative results of testing with real images of the KITTI dataset. We can observe that the performance of T$^2$Net has a substantial 9.1\% absolute improvement compared to our all-synthetic trained model. Unlike the indoor results, the best performance comes from without $D_{feat}$. This is likely due to the translated images much closer to real KITTI, which does not need to match the feature distribution using $D_{feat}$ adversarial learning. We also observe that our model. despite training without real paired data, is able to outperform the method of \cite{eigen2014depth} trained on real paired image-depth data, as well as the method of  \cite{garg2016unsupervised} trained on real left-right stereo data.

We also qualitatively compared the performance of the proposed model with the state-of-the-art in Figure~\ref{fig:ch2_depth_outdoor}. We only chose two representatives that either used real paired color-depth images \cite{eigen2014depth}, or real left-right stereo images \cite{godard2017unsupervised}. Compared to \cite{eigen2014depth}, our model can generate full dense depth maps of input image size. Our method is also able to detect more detail at object boundaries than \cite{godard2017unsupervised}, with a likely reason being that the synthetic training depth maps preserved object details better. Another interesting observation is the predicted depth maps were treating glass windows as permeable based on synthetic data, while they were mostly sensed as opaque in the laser-based ground truth.

\begin{figure}[tb!]
	\centering
	\includegraphics[width=\textwidth]{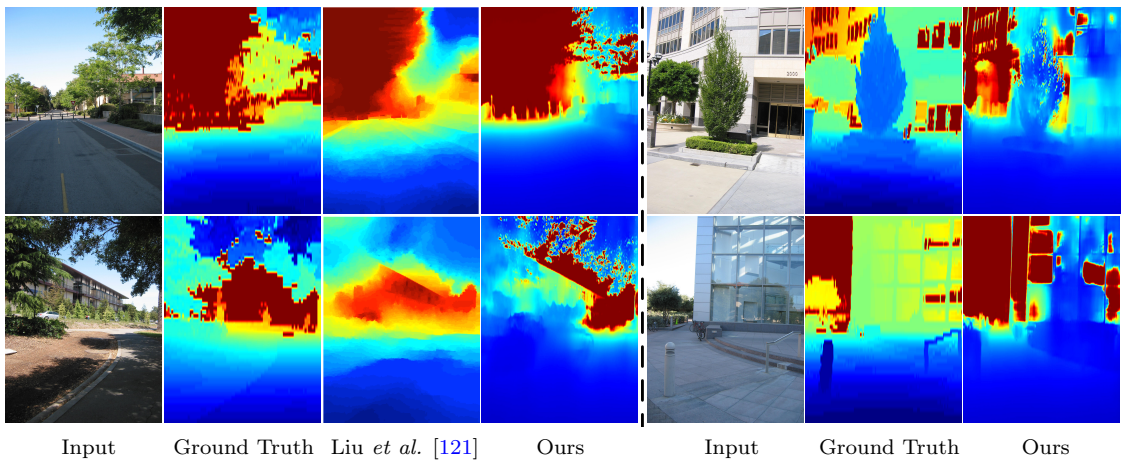}
	\begin{picture}(0,0)
		\put(-186,4){\scriptsize Input}
		\put(-144,4){\scriptsize Ground Truth}
		\put(-85,4){\scriptsize Liu \etal \cite{liu2016learning}}
		\put(-8,4){\scriptsize Ours}
		\put(55,4){\scriptsize Input}
		\put(98,4){\scriptsize Ground Truth}
		\put(174,4){\scriptsize Ours}
	\end{picture}
	\vspace{-0.2cm}
	\caption[Qualitative results on Make3D]{\textbf{Qualitative results on Make3D \cite{saxena2009make3d}}. For most cases the model generated reasonable depths except scenes with new object types not present in the synthetic data. }
	\label{fig:ch2_depth_outdoor3D}
\end{figure}

\subsection{Performance on Make3D}

To compare the generalization ability of our T$^2$Net to a different test dataset, we used our full T$^2$Net model, trained only on vKITTI paired data and (unpaired) real KITTI images, for testing on the Make3D dataset \cite{saxena2009make3d}. We evaluated our model quantitatively on Make3D using the standard C1 metric. The RMSE(m) accuracy is 8.935, Log-10 is 0.574, Abs Rel is  0.508 and Sqr Rel is 6.589. The qualitative results presented in Figure \ref{fig:ch2_depth_outdoor3D} show that our model can generate reasonable depth maps in most situations. The right part of Figure \ref{fig:ch2_depth_outdoor3D} displays some failure cases, likely due to large building windows not being widely observed in the vKITTI datasets.

\begin{figure}[tb!]
	\centering
	\includegraphics[width=\textwidth]{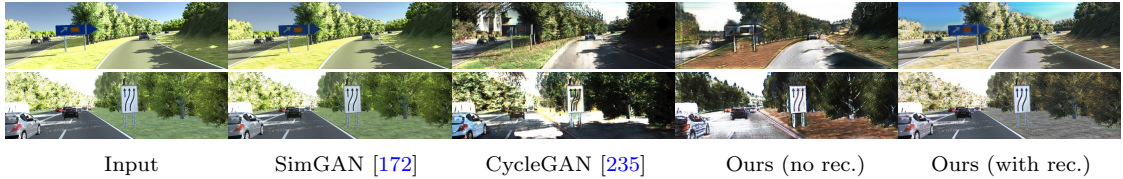}
	\begin{picture}(0,0)
		\put(-170,4){\scriptsize Input}
		\put(-106,4){\scriptsize SimGAN \cite{shrivastava2017learning}}
		\put(-27,4){\scriptsize CycleGAN \cite{zhu2017toward}}
		\put(63,4){\scriptsize Ours (no rec.)}
		\put(140,4){\scriptsize Ours (with rec.)}
	\end{picture}
	\vspace{-0.2cm}
	\caption[Ablation study for different translation networks]{\textbf{Ablation study for different translation networks}. The qualitative results of different unpaired image-to-image translation methods trained using vKITTI and real KITTI dataset. }
	\label{fig:ch2_syn2real_ablation}
\end{figure}

\begin{table}[]
    \centering
    \scriptsize
    \renewcommand{\arraystretch}{1.2}
    \setlength\tabcolsep{4pt}
    \begin{tabular}{@{}lcccccccc@{}}
       \hlineB{3.5} 
       \multirow{2}{*}{Method}& \multicolumn{4}{c}{$\downarrow$} && \multicolumn{3}{c}{$\uparrow$} \\
       \cline{2-5}\cline{7-9}
        & Abs Rel & Sq Rel & RMSE & RMSE log && $\delta<1.25$ & $\delta<1.25^2$ & $\delta<1.25^3$ \\
        \hlineB{2.5}
            baseline, synthetic only & 0.278 & 3.216 & 6.268 & 0.322 && 0.681 & 0.854 & 0.929 \\
			vanilla task network, synthetic only & 0.295 & 3.793 & 8.403 & 0.363 && 0.600 & 0.817 & 0.912 \\
			vanilla task network, full approach & 0.259 & 2.891 & 6.380 & 0.324 && 0.694 & 0.853 & 0.927\\
			\hline
			separated training & 0.234 & 2.706 & 6.068 & 0.293 && 0.747 & 0.882 & 0.942 \\
			separated training with CycleGAN & 0.212 & 1.973 &  5.340 & 0.269 && 0.750 & 0.895 & 0.952\\
			self-domain reconstruction &  0.199 & 1.517 & 5.349 & 0.298 && 0.695 & 0.866 & 0.9420 \\
			\hline
			No reconstruction loss(epoch 3)& 0.201 & 1.941 & 5.619 & 0.286 && 0.741 & 0.882 & 0.945 \\
			No feature loss & {\bf 0.168} & {\bf 1.199} & {\bf 4.674} & {\bf 0.243} && {\bf 0.772} & {\bf 0.912}& {\bf 0.966}\\
			No image GAN loss & 0.233 & 2.902 & 6.285 & 0.300 && 0.743 & 0.880 & 0.938 \\
			\hline
			our full approach & 0.169 & 1.230 & 4.717& 0.245 && 0.769 & 0.912 & 0.965 \\
	   \hlineB{2.5}
    \end{tabular}
    \caption[Quantitative results of different variants of our T$^2$Net]{\textbf{Quantitative results of different variants of our T$^2$Net} on KITTI using the split of \cite{eigen2014depth}. All methods are trained without the real-world ground truth depth map. $\downarrow$ = lower is better. $\uparrow$ = higher is better. }
    \label{tab:ch2_ablation}
\end{table}

\begin{table}[]
    \centering
    \scriptsize
    \renewcommand{\arraystretch}{1.2}
    \setlength\tabcolsep{4pt}
    \begin{tabular}{@{}lcccccccc@{}}
       \hlineB{3.5} 
       \multirow{2}{*}{Method}& \multicolumn{4}{c}{$\downarrow$} && \multicolumn{3}{c}{$\uparrow$} \\
       \cline{2-5}\cline{7-9}
        & Abs Rel & Sq Rel & RMSE & RMSE log && $\delta<1.25$ & $\delta<1.25^2$ & $\delta<1.25^3$ \\
        \hlineB{2.5}
            baseline, synthetic only & 0.304 & 0.394 & 1.024 & 0.369 && 0.458 & 0.771 & 0.916 \\
			\hline
			separated training & 0.288 & 0.364 & 1.095 & 0.352 && 0.463 & 0.768 & 0.902 \\
			separated training with CycleGAN   & 0.280 & 0.362 & 0.971 & 0.355 && 0.478 & 0.777 & 0.921 \\
			self-domain reconstruction  & 0.287 & 0.352 & 0.968 & 0.351 && 0.491 & 0.782 & 0.934 \\
			\hline
			No reconstruction loss(epoch 3) & 0.278 & 0.341 & 0.942 & 0.345 && 0.514 & 0.808 & 0.929  \\
			No feature loss & 0.274 & 0.336 & 1.001 & 0.325 && 0.496 & 0.814 & 0.938 \\
			No image GAN loss & 0.320 & 0.405 & 0.991 & 0.343 && 0.480 & 0.792 & 0.933 \\
			\hline
			our full approach & {\bf 0.257} & {\bf 0.281} & \bf{0.915} & {\bf0.305} && {\bf 0540} & {\bf 0.832} & {\bf 0.948} \\
	   \hlineB{2.5}
    \end{tabular}
    \caption[Quantitative results of different variants of our T$^2$Net]{\textbf{Quantitative results of different variants of our T$^2$Net} on NYUv2 dataset \cite{silberman2012indoor}. All methods are trained without the real-world ground truth depth map. $\downarrow$ = lower is better. $\uparrow$ = higher is better. }
    \label{tab:ch2_ablation_indoor}
\end{table}

\subsection{Ablation Study}

We evaluated the contribution of different design choices in the proposed T$^2$Net. Table \ref{tab:ch2_ablation} shows the quantitative results and Figure~\ref{fig:ch2_syn2real_ablation} shows some example outputs of different methods for unpaired image translation.

\paragraph{End-to-End \emph{vs} Separated} We began by evaluating the effect of end-to-end learning. We found that end-to-end training outperformed separated training of the translation network and task prediction network. One reasonable explanation is that task loss is a form of supervised loss for synthetic-to-realistic translation. This incentivizes the translation network to preserve geometric content present in a synthetic image.

We also experimented with the unpaired image translation network CycleGAN \cite{zhu2017unpaired}. This model has two encoder-decoder translation networks and two discriminators, but we were limited by machine memory and trained the CycleGAN and task network separately. From Figure \ref{fig:ch2_syn2real_ablation}, we found that while this model generated very visually realistic images, it also created some realistic-looking details that significantly distorted scene geometry. The quantitative performance is close to our separated training results.

\paragraph{No Image Reconstruction} We studied what happens when training without real-image reconstruction loss. In Figure \ref{fig:ch2_syn2real_ablation}, we may surmise that the task loss in the depth domain is able to encourage reasonable depiction of scene geometry in the translation network. However, the lack of a real image reconstruction loss appears to make it harder to generate high-resolution images. In addition, we noticed that while the removal of reconstruction loss still led to relatively good results as seen in Table \ref{tab:ch2_ablation} and \ref{tab:ch2_ablation_indoor}, this was only true in early training with best results in epoch 3, with accuracy dropping after more training epochs.

\paragraph{ Target Reconstruction \emph{vs} Self-Regularization} Since the self-regularization component of SimGAN is closest to our target-domain reconstruction concept, we also trained our full model with L1 reconstruction loss for synthetic imagery, which forces the generated target images to be similar to original input images. From Figure \ref{fig:ch2_syn2real_ablation}, we observe that this is unable to work well for large domain shifts, for the GAN loss and self-domain reconstruction loss play opposite roles in the translation.  

\section{Limitations and Discussion}\label{ch2:diss}

A novel, end-to-end trainable T$^2$Net deep neural network is presented for single-image depth estimation, that requires only synthetic image-depth pairs and unpaired real images for training. The overall system comprises an image translation network and a depth prediction network. It is able to generate realistic images via a learning framework that combines adversarial loss for synthetic input and target-domain reconstruction loss for real input in the translation network, and a further combination of a task loss and feature GAN loss in the depth prediction network. The T$^2$Net can be trained end-to-end, and does not require real image-depth pairs nor stereo pairs for training. It is able to produce good results on the NYUDv2 and KITTI datasets despite the lack of access to real paired training data, and even outperformed early deep learning methods that were trained on real paired data. Many recent works \cite{zhao2019geometry,chen2021s2r,ranftl2020towards} have also begun to explore the single-image depth estimation on different datasets. In particular, Zhao \etal \cite{zhao2019geometry} and Chen \etal \cite{chen2021s2r} follow our experiment setting to address the gap between synthetic and real domain, and consistently consider our method as a state-of-the-art benchmark for single image depth estimation using only synthetic ground truth depth. In the future, we intend to explore mechanisms that provide greater generalization capability across different datasets.

While the proposed \emph{wide-spectrum} translation network works well on this synthetic-to-realistic translation task, it requires joint training with the task network, which ensures \emph{depth / structure} consistency during the end-to-end training. However, it is not always the case that a complementary task is available to support an I2I problem. For example, it remains challenging to explicitly model the \emph{content} and \emph{style} for I2I translation. In Chapter \ref{ch:F(L)SeSim}, we will introduce a spatially-correlative loss, which can explicitly extract the structure representation to allow preservation of scene structure consistency during the translation when appearance may dramatically change. 
\chapter{Spatially-Correlative Loss for Various Image Translation Tasks} 
\chaptermark{F(L)SeSim}
\label{ch:F(L)SeSim} 

The previous \emph{wide-spectrum} translation network works well only when translation and task networks are jointly optimized, in which the task network can provide a geometry loss to support synthesis with depth consistency. However, it is not scalable to many I2I translation scenarios, where only unpaired images in the two domains are available. As the goal in I2I translation is to modify the input image to fit the \emph{style / appearance} of the target domain, while preserving the original \emph{content / structure}, learning to assess the \emph{content} and \emph{style} correctly is thus of central importance. In this chapter, a novel spatially-correlative loss is proposed that is simple, efficient, and yet effective for preserving scene structure consistency. Previous methods attempt this by using pixel-level cycle-consistency or feature-level matching losses, but the domain-specific nature of these losses hinder translation across large domain gaps. To address this, we exploit the spatial patterns of self-similarity as a means of defining scene structure. The spatially-correlative loss is geared towards only capturing spatial relationships within an image, rather than domain appearance. A new self-supervised learning method is also introduced to explicitly learn spatially-correlative maps for each specific translation task. We show distinct improvement over baseline models in all three modes of unpaired I2I translation: single-modal, multi-modal, and even single-image translation.

We first introduce the motivation in Section \ref{ch3:intro} and review previous works in Section \ref{ch3:back}. Section \ref{ch3:appro} explains how to calculate the spatially-correlative loss for I2I translation tasks and Section \ref{ch3:experiments} demonstrates the superiority of the proposed loss. We discuss the loss in Section \ref{ch3:conc}.

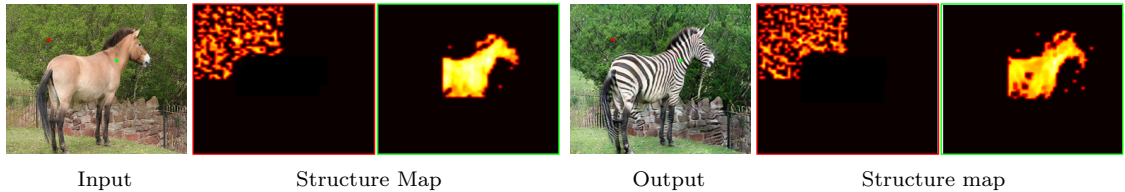
\begin{figure}[tb!]
	\centering
	\includegraphics[width=\linewidth]{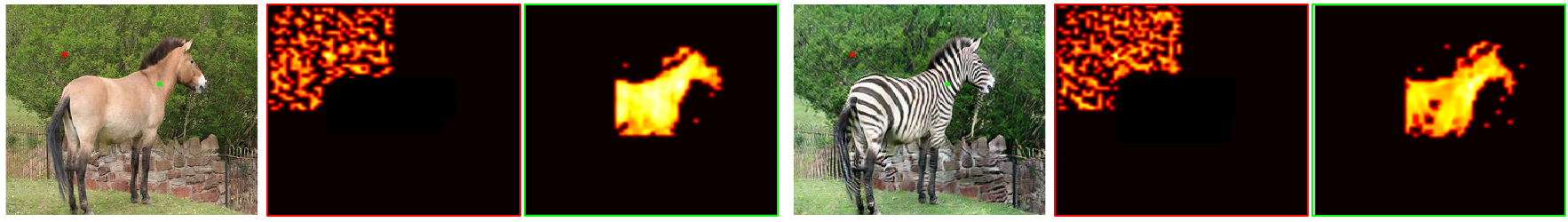}
	\begin{picture}(0,0)
	\put(-180,4){\scriptsize Input}
	\put(-98,4){\scriptsize Structure Map}
	\put(28,4){\scriptsize Output}
	\put(114,4){\scriptsize Structure map}
	\end{picture}
	\vspace{-0.2cm}
	\caption[Example for learned spatially-correlative representation]{\textbf{Our learned spatially-correlative representation} encodes local scene structure based on self-similarities. Despite vast appearance differences between the \emph{horse} and \emph{zebra}, when the scene structures are identical (\ie same poses), the spatial patterns of self-similarities are as well.} 
	\label{fig:ch3_example}
\end{figure}

\section{Introduction}\label{ch3:intro}

I2I translation refers to the task of modifying an input image to fit the \emph{style / appearance} of the target domain, while preserving the original \emph{content / structure } (as shown in Figure~\ref{fig:ch3_example}: \emph{horse} $\rightarrow$ \emph{zebra}); learning to assess the \emph{content} and \emph{style} correctly is thus of central importance. While GANs~\cite{goodfellow2014generative} have the ability to generate images that adhere to the overall dataset distribution, it is still difficult to preserve scene structure during translation when image-conditional GANs are optimized with purely adversarial loss.

To mitigate the issue of scene structure discrepancies, a few loss functions for comparing the content between input and output images have been proposed, including (a) \emph{pixel-level} image reconstruction loss~\cite{isola2017image, shrivastava2017learning,chen2017photographic} and cycle-consistency loss~\cite{kim2017learning,zhu2017unpaired,yi2017dualgan}; (b) \emph{feature-level} perceptual loss~\cite{dosovitskiy2016generating,johnson2016perceptual} and PatchNCE loss~\cite{park2020cut}. However, these losses still have several limitations. First, pixel-level losses do \emph{not} explicitly decouple structure and appearance. Second, feature-level losses help but continue to conflate domain-specific structure and appearance attributes. Finally, most feature-level losses are calculated using a fixed ImageNet~\cite{deng2009imagenet} pre-trained network (\eg VGG16~\cite{vgg}), which will not correctly adapt to arbitrary domains.

In this chapter, we aim to design a \emph{domain-invariant} representation to precisely express scene structure, rather than using original pixels or features that couple both appearance and structure. To achieve this, we propose to revisit the idea of \emph{self-similarity}. Classically, low-level self-similarity has been used for matching~\cite{shechtman2007matching} and image segmentation~\cite{shi2000normcuts}, while feature-level self-similarity in deep learning manifests as self-attention maps~\cite{xu2015show}. We propose to go further, to advance an assumption that \emph{all} regions within same categories exhibit some form of self-similarity. For instance, while the horse and zebra in Figure~\ref{fig:ch3_example} appear very different, there is obvious visual self-similarity in their own regions. We believe a network can learn deeper representations of self-similarities (beyond just visual ones) that can encode intact object shapes, even when there are variations in appearances within an object. Then through estimating such co-occurrence signals in self-similarity, we can explicitly represent the structure as multiple \emph{spatially-correlative} maps, visualized as heat maps in Figure \ref{fig:ch3_example}. Based on this within-shape self-similarity, we propose then that \emph{a structure-preserving image translation will retain the patterns of self-similarity in both the source and translated images, even if appearances themselves change dramatically}. 

Our basic spatially-correlative map, called \emph{FSeSim}, is obtained by computing the \textbf{F}ixed \textbf{Se}lf-\textbf{Sim}ilarity of features extracted from a pre-trained network. While this basic version achieved comparative or even better results than state-of-the-art methods~\cite{zhu2017unpaired,fu2019geometry,park2020cut} on some tasks, the generality is limited because features extracted from an ImageNet pre-trained network are biased towards photorealistic imagery. Hence, this will not optimally work with images in non-realistic styles.

To obtain a more general spatially-correlative map, the \textbf{L}earned \textbf{Se}lf-\textbf{Sim}ilarity, called \emph{LSeSim}, is presented by using a form of contrastive loss, in which we explicitly encourage homologous structures to be closer, regardless of their appearances, and reciprocally dissociate dissimilar structures even they have similar appearances. To do this, the model learns a domain-invariant spatially-correlative map, where having the same scene structure leads to similar maps, even if the images are from different domains.

There are several advantages of using the proposed F/LSeSim loss: (a) In contrast to the existing losses that directly compare the loss on pixels \cite{zhu2017unpaired} or features \cite{johnson2016perceptual}, F/LSeSim captures the domain-invariant structure representation, regardless of the absolute pixel values; (b) Through contrastive learning, the LSeSim learns a spatially-correlative map for a specific image translation task, rather than features extracted from a fixed pre-trained network, as in \eg perceptual loss~\cite{johnson2016perceptual}, contextual loss \cite{mechrez2018contextual}; (c) The translation model is more efficient and faster than the widely used cycle-consistency architectures, because our F/LSeSim explicitly encodes the structure, bypassing the expensive multi-cycle looping; (d) As we show in Figure~\ref{fig:ch3_error_map}, our F/LSeSim correctly measures the structural distance even when the two images are in completely different domains; (e) Finally, our F/LSeSim can easily be integrated into various frameworks. In our experiments, we directly used the generator and discriminator architectures of CycleGAN~\cite{zhu2017unpaired}, MUNIT~\cite{huang2018multimodal} and StyleGAN~\cite{karras2019style,karras2020analyzing} for extensive I2I translation tasks. The experimental results show that our model outperformed the existing both one-sided translation methods~\cite{benaim2017one,amodio2019travelgan,fu2019geometry,park2020cut} and two-sided translation methods~\cite{zhu2017unpaired,yi2017dualgan,huang2018multimodal}.

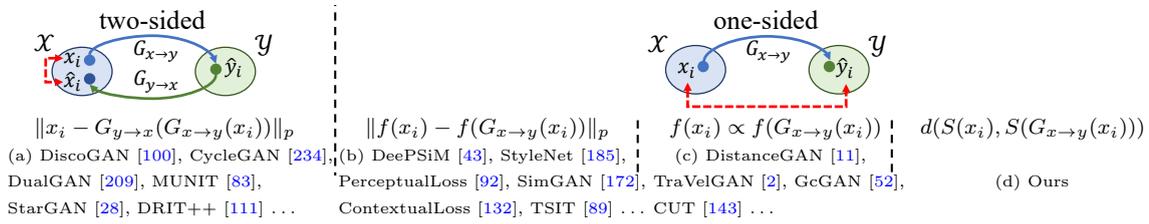
\begin{figure}[tb!]
	\centering
	\includegraphics[width=\linewidth]{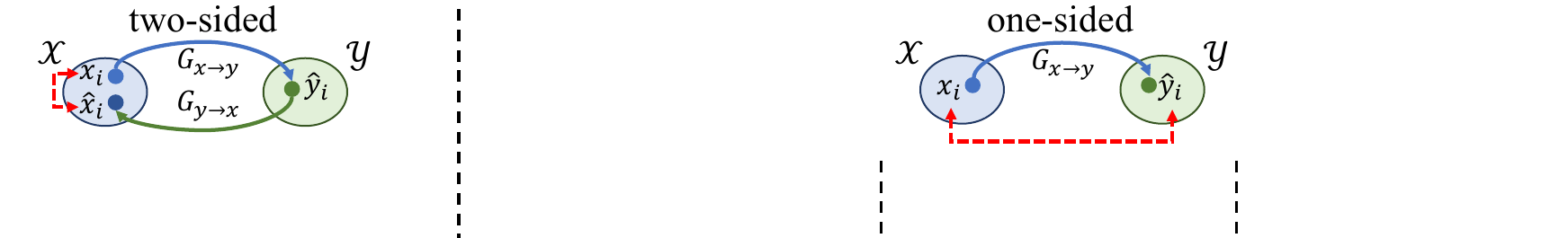}
	\begin{picture}(0,0)
		\put(-200,30){\scriptsize $\Vert x_i-G_{y\rightarrow x}(G_{x\rightarrow y}(x_i))\Vert_p$}
		\put(-210,20){\tiny (a) DiscoGAN~\cite{kim2017learning}, CycleGAN~\cite{zhu2017unpaired}, }
		\put(-210,10){\tiny DualGAN~\cite{yi2017dualgan}, MUNIT~\cite{huang2018multimodal}, }
		\put(-210,0){\tiny StarGAN~\cite{choi2018stargan}, DRIT++~\cite{lee2020drit++} $\dots$}
		\put(-76,30){\scriptsize $\Vert f(x_i)-f(G_{x\rightarrow y}(x_i))\Vert_p$}
		\put(-86,20){\tiny (b) DeePSiM~\cite{dosovitskiy2016generating}, StyleNet~\cite{ulyanov2017improved}, }
		\put(-86,10){\tiny PerceptualLoss~\cite{johnson2016perceptual}, SimGAN~\cite{shrivastava2017learning}, }
		\put(-86,0){\tiny ContextualLoss~\cite{mechrez2018contextual}, TSIT~\cite{jiang2020tsit} $\dots$}
		\put(38,30){\scriptsize $f(x_i) \propto f(G_{x\rightarrow y}(x_i))$}
		\put(40,20){\tiny (c) DistanceGAN~\cite{benaim2017one},}
		\put(32,10){\tiny TraVelGAN~\cite{amodio2019travelgan}, GcGAN~\cite{fu2019geometry}, }
		\put(32,0){\tiny CUT~\cite{park2020cut} $\dots$}
		\put(132,30){\scriptsize $d(S(x_i),S(G_{x\rightarrow y}(x_i)))$}
		\put(160,10){\tiny (d) Ours}
	\end{picture}
	\caption[Comparison of I2I translation methods with various content losses]{\textbf{Comparison of unpaired I2I translation methods with various content losses}. (a) The cycle-consistency loss~\cite{kim2017learning,zhu2017unpaired,yi2017dualgan} in a two-sided framework. (b) Pixel-level image reconstruction loss~\cite{shrivastava2017learning} and feature-level matching loss~\cite{johnson2016perceptual}. (c) Various indirect relationships~\cite{benaim2017one,fu2019geometry} between the input and output. (d) Our spatially-correlative loss based on a learned spatially-correlative map. } 
	\label{fig:ch3_losses}	
\end{figure}

\section{Background}\label{ch3:back}

Existing unpaired I2I translation either use  cycle-consistency loss in a two-sided framework~\cite{kim2017learning,zhu2017unpaired,yi2017dualgan}, or other forms of pixel-level and feature-level losses in a one-sided framework~\cite{benaim2017one,amodio2019travelgan,fu2019geometry} for preserving content (Figure~\ref{fig:ch3_losses}).

\paragraph{Two-Sided Unsupervised Image Translation} \emph{Cycle-consistency} has become a de facto loss in most works, whether the cycles occur in the image domain~\cite{zhu2017unpaired,kim2017learning,yi2017dualgan,choi2018stargan,hoffman2018cycada,lee2018diverse}, or in latent space~\cite{zhu2017toward,huang2018multimodal,lee2020drit++}. However, without explicit constraints, the content in a translated image can be easily distorted~\cite{lee2018diverse}. Furthermore, the cycle-based methods require auxiliary generators and discriminators for the reverse mapping.

\paragraph{One-Sided Unsupervised Image Translation} To avoid cycle-consistency artifacts, 
DistanceGAN~\cite{benaim2017one} and GcGAN~\cite{fu2019geometry} pre-define an implicit distance in a one-sided framework. In contrast, the feature-level losses~\cite{johnson2016perceptual,mechrez2018contextual} evaluate the content distance in a deep feature space, which have been applied in both style transfer~\cite{gatys2016image,johnson2016perceptual,mechrez2018contextual,zhang2018unreasonable} and image translation~\cite{chen2017photographic,wang2018high,park2019semantic,jiang2020tsit}. However, the underlying assumption that high-level semantic information is solely determined in feature space does not always hold. Furthermore, these features are extracted from a fixed pre-trained network (\eg VGG16~\cite{vgg}). While the latest CUT~\cite{park2020cut} learns a PatchNCE loss for a specific task, the distance used is directly computed from extracted features, and will still be affected by domain-specific peculiarities. 

\paragraph{Contrastive Representation Learning} Driven by the potential of discriminative thought, a series of self-supervised methods~\cite{hjelm2018learning,wu2018unsupervised,oord2018representation,bachman2019learning,henaff2019data,chen2020simple,he2020momentum} have emerged in recent years. These self-supervised methods learn robust features by associating ``positive'' pairs and dissociating ``negative'' pairs. CUT~\cite{park2020cut} first introduced contrastive learning for unsupervised I2I translation. While we utilize a patch-wise contrastive loss within an image in a similar manner to CUT, we propose a systematic way to learn a structure map that excludes appearance attributes. As described below, our LSeSim method learns a domain-independent structure representation.

\section{Approach}\label{ch3:appro}

As shown in Figure~\ref{fig:ch3_losses}, given a collection of images $\mathcal{X}\subset\mathbb{R}^{H\times W\times C}$ from a particular domain (\eg horse), our main goal is to learn a model $\Phi$ that receives the image $x\in \mathcal{X}$ as input and transfers it into the target domain $\mathcal{Y}\subset\mathbb{R}^{H\times W\times C}$ (\eg zebra), in a manner that retains the original scene structure but converts the appearance appropriately. 

Here, we focus on designing a loss function that measures the structural similarity between the input image $x$ and the translated image $\hat{y}=\Phi(x)$.
However, unlike most existing approaches that directly attempt to evaluate the structural similarity between input and translated images at some deep feature level, we will instead compute the \emph{self-similarity} of deep features \emph{within each image}, and then \emph{compare the self-similarity patterns} between the images.

In subsequent sections, we investigate two losses, \emph{fixed self-similarity (FSeSim)} and \emph{learned self-similarity (LSeSim)}. In the first instance, we directly compare the self-similarity patterns of features extracted from a fixed pre-trained network (\eg VGG16~\cite{vgg}). In the second instance, we additionally introduce a structure representation model that learns to correctly compare the self-similarity patterns, in which we use the contrastive infoNCE loss~\cite{oord2018representation} to learn such a network without label supervision.

\begin{figure}[tb!]
	\centering
	\includegraphics[width=0.8\linewidth]{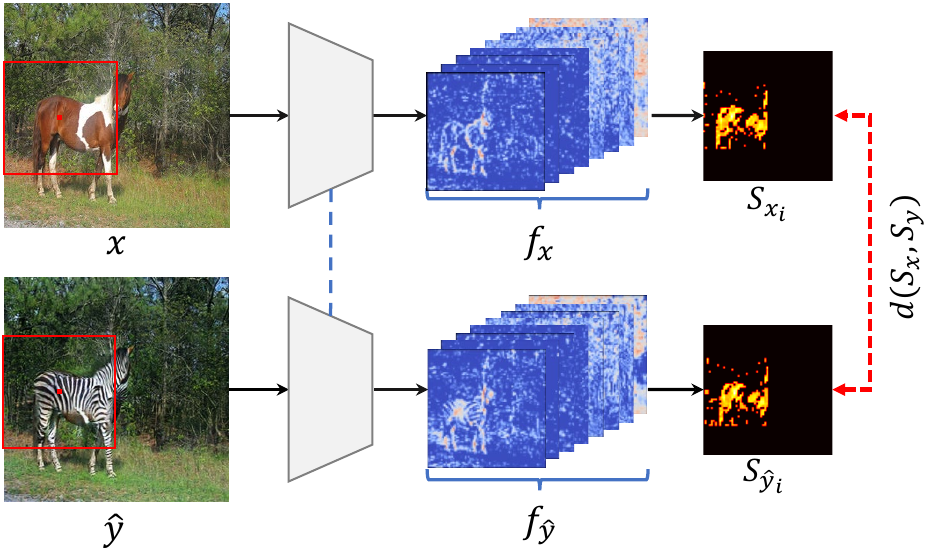}
	\begin{picture}(0,0)
	\put(-232,94){\rotatebox{90}{\footnotesize shared}}
	\put(-312,-9){\footnotesize Image}
	\put(-237,2){\footnotesize Feature}
	\put(-239,-9){\footnotesize extractor}
	\put(-163,-9){\footnotesize Features}
	\put(-90,-9){\footnotesize Similarity map}
	\end{picture}
	\vspace{0.4cm}
	\caption[Fixed Self-similarity]{\textbf{An example of computing spatially-correlative loss from self-similarity maps.} The image $x$ and corresponding translated image $\hat{y}$ are first fed into the feature extractor. We then compute the local self-similarity for each query point. Here, we show one example for the red query point. }
	\label{fig:ch3_framework}
\end{figure}
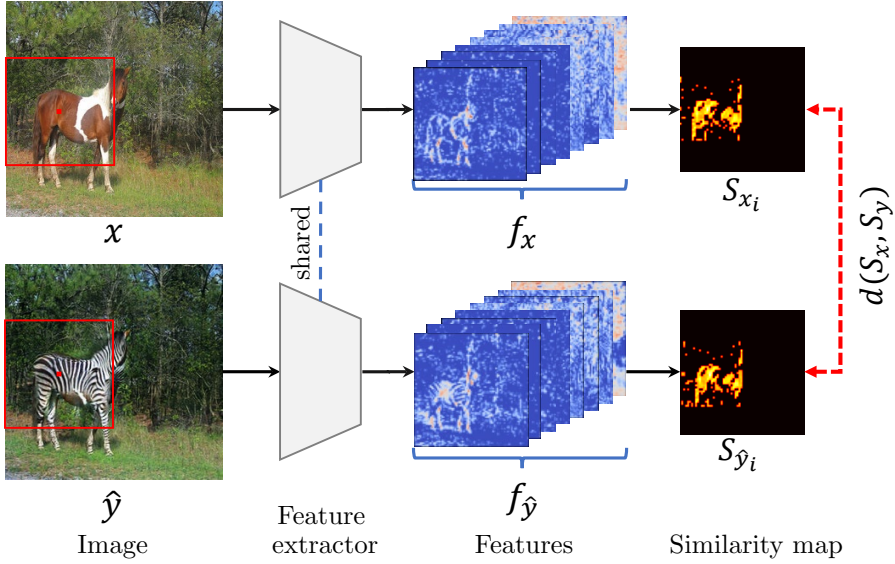 

\subsection{Fixed Self-Similarity (FSeSim)}\label{sec:ch3_FSeSim}

We first describe our fixed spatially-correlative loss. Given an image $x$ in one domain and its corresponding translated image $\hat{y}$ in another, we extract the features $f_x$ and $f_{\hat{y}}$ using a simple network (\eg VGG16~\cite{vgg}). Instead of directly computing the feature distance $\Vert f_x-f_{\hat{y}} \Vert_p $, we compute the self-similarity in the form of a map. We call this a \emph{spatially-correlative map}, formally:
\begin{equation}
	S_{x_i} = (f_{x_i})^T(f_{x_*})
\end{equation}
where $f_{x_i}^T\in\mathbb{R}^{1\times C}$ is the feature of a query point $x_i$, $f_{x_*}\in\mathbb{R}^{C\times N_p}$ contains corresponding features in a patch of $N_p$ points, and $S_{x_i}\in\mathbb{R}^{1\times N_p}$ captures the feature spatial correlation between the query point and other points in the patch. We show one query example in Figure~\ref{fig:ch3_framework}, where the spatially-correlative map for the query patch is visualized as a heat map. Note that unlike the original features that would still encode domain-specific attributes such as color, lighting and texture, the self-similarity map only captures the spatially-correlative relationships. 

Next, we represent the structure of the whole image as a collection of multiple spatially-correlative maps $S_x=[S_{x_1};S_{x_2};\dots;S_{x_s}]\in\mathbb{R}^{N_s\times N_p}$, where $N_s$ is the numbers of sampled patches. This is a semi-sparse representation, but is more computationally efficient. We then compare the multiple structure similarity maps between the input $x$ and the translated image $\hat{y}$, as follows:
\begin{equation}\label{eq:FSeSimL}
	\mathcal{L}_{s} = d(S_x, S_{\hat{y}})
\end{equation} 
where $S_{\hat{y}}$ are corresponding spatially-correlative maps in the target domain. Here, we consider two forms for $d(\cdot)$, the $L_1$ distance $\Vert S_x-S_{\hat{y}} \Vert_1$ and the cosine distance $\Vert 1-\cos(S_x, S_{\hat{y}})\Vert$. The former term strongly encourages the spatial similarity to be consistent at all points in a patch, while the latter term supports pattern correlation without concern for differences in magnitude. 

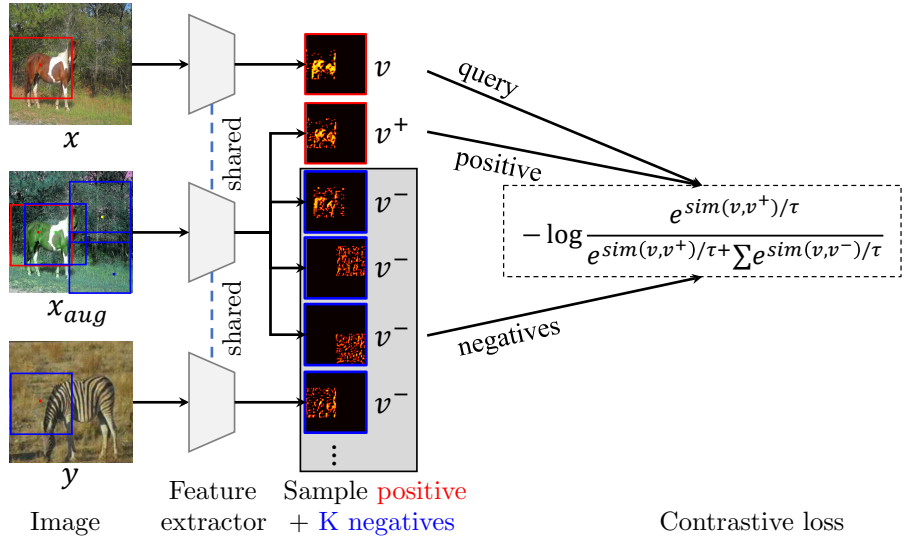
\begin{figure}[tb!]
	\centering
	\includegraphics[width=0.8\linewidth]{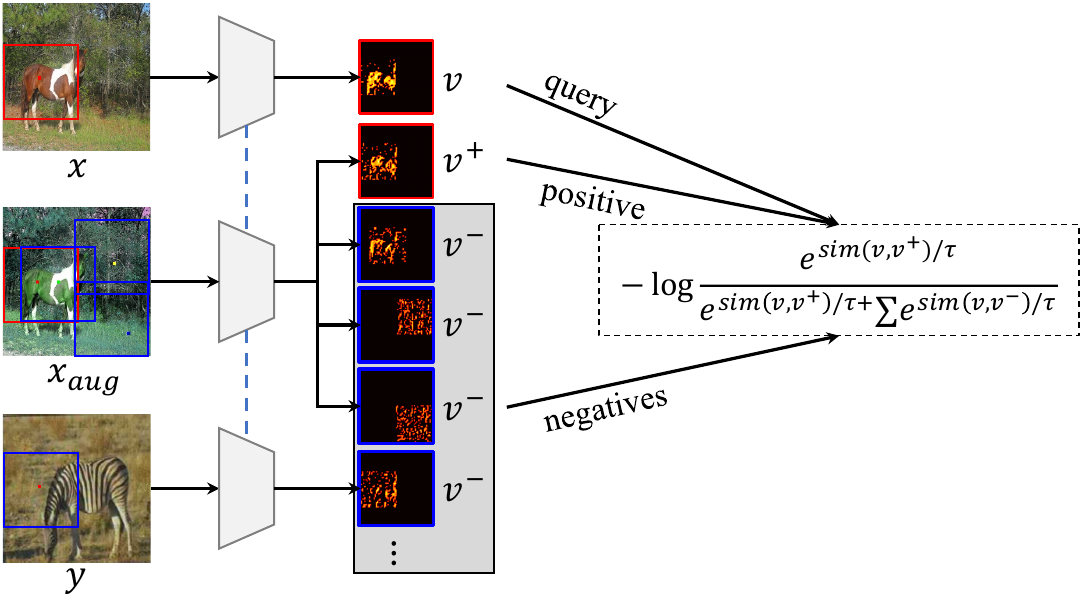}
	\begin{picture}(0,0)
	\put(-260,117){\rotatebox{90}{\footnotesize shared}}
	\put(-260,55){\rotatebox{90}{\footnotesize shared}}
	\put(-332,-11){\footnotesize Image}
	\put(-280,1){\footnotesize Feature}
	\put(-283,-11){\footnotesize extractor}
	\put(-237,1){\footnotesize Sample \textcolor{red}{positive}}
	\put(-235,-11){\footnotesize + \textcolor{blue}{K negatives}}
	\put(-96,-11){\footnotesize Contrastive loss}
	\end{picture}
	\vspace{0.2cm}
	\caption[Learned Self-Similarity]{\textbf{Patchwise contrastive learning for the learned self-similarity.} Three images are fed into the feature extractor, in which two images, $x$ and $x_{aug}$, are homologous with the same structure but varied appearances, and $y$ is another randomly sampled image. For each query patch in $x$, the ``positive'' sample is the corresponding patch in $x_{aug}$, and all other patches are considered as ``negative'' samples.}
	\label{fig:ch3_learned_attn}
\end{figure} 

\subsection{Learned Self-Similarity (LSeSim)}\label{sec:LSeSim}

Although our FSeSim provides strong supervision for structure consistency, it does \emph{not} explicitly learn a structure representation for a specific translation task. As opposed to existing feature-level losses~\cite{johnson2016perceptual,mechrez2018contextual} that only utilize the features from a fixed pre-trained network, we propose to additionally learn a structure representation network for each task that expresses the learned self-similarity, or LSeSim.

In order to learn such a model \emph{without supervision}, we consider the self-supervised contrastive learning that associates similar features, while simultaneously dissociates different features. Following PatchNCE~\cite{park2020cut}, we build our contrastive loss at patch level, except here the pairs for comparison are our spatially-correlative maps, rather than the original features in existing works~\cite{hjelm2018learning,chen2020simple,he2020momentum,park2020cut}. To help generate pairs of similar patch features for self-supervised learning, we create augmented images by applying structure-preserving transformations. 

Formally, let $\boldsymbol{v}=S_{x_i}\in\mathbb{R}^{1\times N_p}$ denotes the spatially-correlative map of the ``query'' patch. Let $\boldsymbol{v}^+=S_{\hat{x}_i}\in\mathbb{R}^{1\times N_p}$ and $\boldsymbol{v}^-\in\mathbb{R}^{K \times N_p}$ be ``positive'' and ``negative'' patch samples, respectively. The query patch is positively paired with a patch in the same position $i$ within an augmented image $x_{aug}$, and negatively paired to patches sampled from other positions in $x_{aug}$, or patches from other images $y$. The number of negative patches used is $K=255$.

Our LSeSim design is illustrated in Figure~\ref{fig:ch3_learned_attn}. The contrastive loss is given by:
\begin{equation}
	\mathcal{L}_c = -\log\frac{e^{sim(\boldsymbol{v},\boldsymbol{v}^+)/\tau}}
	{e^{sim(\boldsymbol{v},\boldsymbol{v}^+)/\tau}+\sum_{k=1}^{K}e^{sim(\boldsymbol{v},\boldsymbol{v}_k^-)/\tau}}
\end{equation}
where $sim(\boldsymbol{v},\boldsymbol{v}^+)=\boldsymbol{v}^T\boldsymbol{v}^+/\Vert\boldsymbol{v}\Vert\Vert\boldsymbol{v}^+\Vert$ is the cosine similarity between two spatially-correlative maps, and $\tau$ is a temperature parameter. To minimize this loss, our network encourages the corresponding patches with the same structure to be close even they have very different visual appearances, which fits in with the goal of image translation. Note that, this contrastive loss is only used for optimizing the structure representation network. The spatially-correlative loss for the generator is always the loss in equation \ (\ref{eq:FSeSimL}).

\subsection{Full Objective}
Overall, we train the networks by jointly minimizing the following losses:
\begin{equation}\label{eq:all}
	\begin{split}
	\mathcal{L}_D &= -\mathbb{E}_{y\sim p_{d}}[\log D(y)]-\mathbb{E}_{\hat{y}\sim p_{g}}[\log(1-D(\hat{y}))] \\
	\mathcal{L}_S &= \mathcal{L}_c \\
	\mathcal{L}_G &= \mathbb{E}_{\hat{y}\sim p_{g}}[\log(1-D(\hat{y}))] + \lambda d(S_x, S_{\hat{y}}) \\
	\end{split}
\end{equation}
where $\mathcal{L}_D$ is the adversarial loss for the discriminator $D(\cdot)$, $\hat{y}$ is the translated image, and $\mathcal{L}_S$ is the contrastive loss for the structure representation network $f(\cdot)$. $\mathcal{L}_G$ is the loss for the generation (translation) network $G(\cdot)$, which consists of the style loss term and the structure loss term. $\lambda$ is a hyper-parameter to trade off between style and content.

\begin{figure}[tb!]
	\centering
	\includegraphics[width=\linewidth]{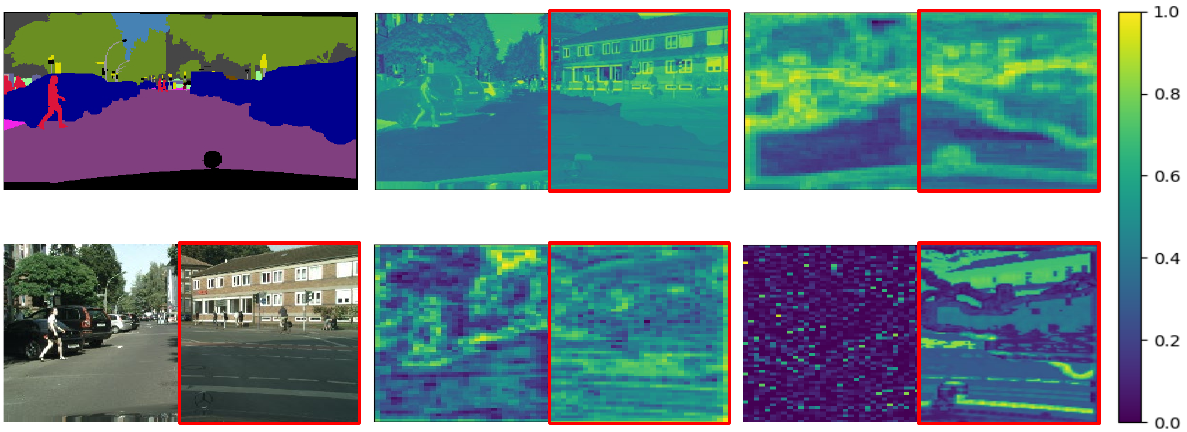}
	\begin{picture}(0,0)
		\put(-150,90){\footnotesize $x$}
		\put(-52,90){\footnotesize $l_1$ pixel loss~\cite{shrivastava2017learning}}
		\put(78,90){\footnotesize Perceptual loss~\cite{johnson2016perceptual}}
		\put(-186,6){\footnotesize $y_{align}$}
		\put(-130,6){\footnotesize $y_{unalign}$}
		\put(-60,6){\footnotesize PatchNCE loss~\cite{park2020cut}}
		\put(92,6){\footnotesize LSeSim loss}
	\end{picture}
	\vspace{-0.2cm}
	\caption[Error maps]{\textbf{Error map visualization.} Our LSeSim has small errors on the left where ground truth paired data is provided, while having large errors on the right for obviously unpaired data.}
	\label{fig:ch3_error_map}
\end{figure}
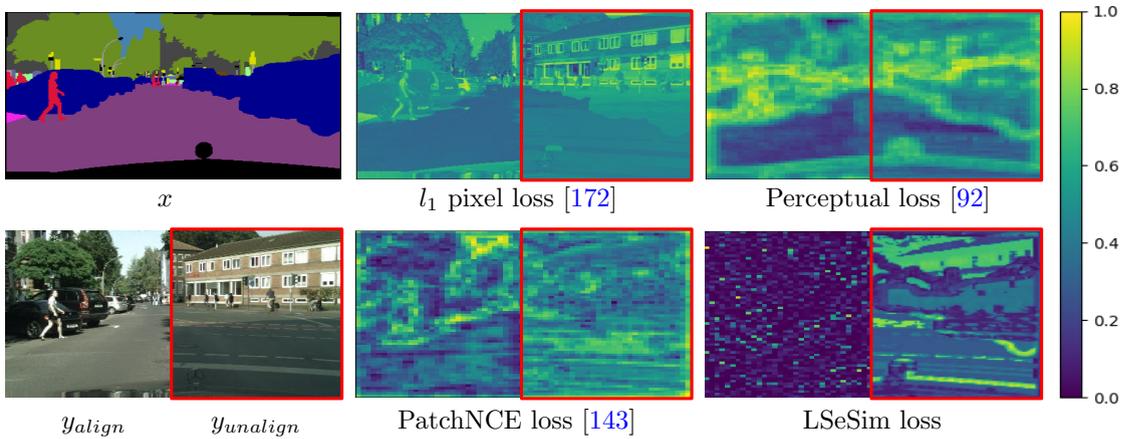

\subsection{Analysis}\label{sec:analysis}

Readers may wonder why the proposed F/LSeSim losses would perform better than existing feature-level losses~\cite{johnson2016perceptual,mechrez2018contextual,park2020cut}. An intuitive interpretation is that \emph{self-similarity deals only with spatial relationships of co-occurring signals, rather than their original absolute values}.

To provide further clarity, we consider a scenario where given a semantic map $x$ (Figure~\ref{fig:ch3_error_map}), the task is to translate it to a photorealistic image $y$. We consider an ideal result (the paired ground truth $y_{align}$ in the dataset) and a wrong result (another image $y_{unalign}$), respectively. Under such a setting, a good structure loss should penalize the wrong result, while supporting the ideal result. To visualize the error maps, for each corresponding pair of query patches in $x$ and $y$ we computed the error at that patch location for different losses. As can be seen, pixel-level loss~\cite{shrivastava2017learning} is naturally unsuitable when there are large domain gaps, and while Perceptual loss~\cite{johnson2016perceptual} will report significant errors for both aligned and unaligned results. PatchNCE~\cite{park2020cut} mitigates the problem by calculating the cosine distance of features, but it can be seen the loss map still retains high errors in many regions within the aligned result, due to extracted features consisting of appearance attributes, such as color and texture.

In contrast, appearance attributes are ignored in LSeSim by representing scene structure as a spatially-correlative map. Figure~\ref{fig:ch3_error_map} shows that our LSeSim leads to low errors for the aligned image (left), even when they are in quite different domains, but large errors for the non-aligned image (right). Even for $y_{unalign}$, LSeSim differentiates between related structures (\eg roads) and unrelated structures (trees vs windows), with lower errors for the former. Hence LSeSim can better help preserve scene structure even across large domain gaps. 

\begin{figure}[tb!]
	\centering
	\setlength{\abovecaptionskip}{0.cm}
	\setlength{\belowcaptionskip}{-0.cm}
	\includegraphics[width=\linewidth]{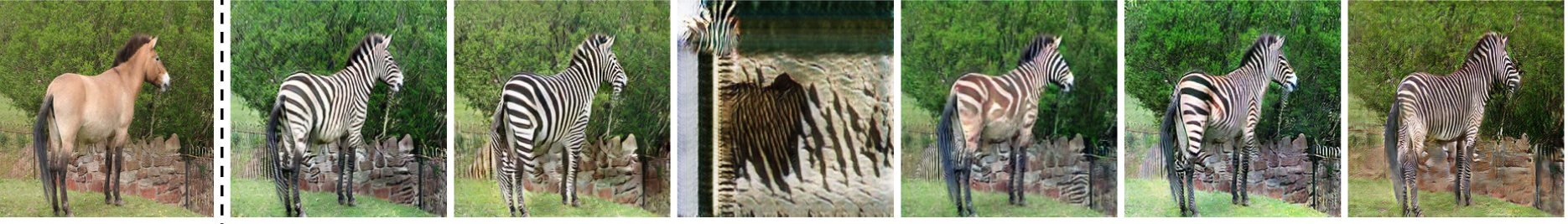}
	\begin{picture}(0,0)
		\put(-186,7){\tiny Input}
		\put(-128,7){\tiny LSeSim}
		\put(-71,7){\tiny FSeSim }
		\put(-31,7){\tiny GAN loss only~\cite{zhu2017unpaired}}
		\put(36,7){\tiny Cycle loss~\cite{zhu2017unpaired}}
		\put(90,7){\tiny Perceptual loss~\cite{johnson2016perceptual}}
		\put(152,7){\tiny PatchNCE loss~\cite{park2020cut}}
	\end{picture}
	\caption[Comparing results under different content losses]{\textbf{Comparing results under different content losses}. All results are reported following the same setting of CycleGAN~\cite{zhu2017unpaired}, except using different content losses. Our model generates much better visual results with only loss modification. }
	\label{fig:ch3_loss_comp}	
\end{figure}
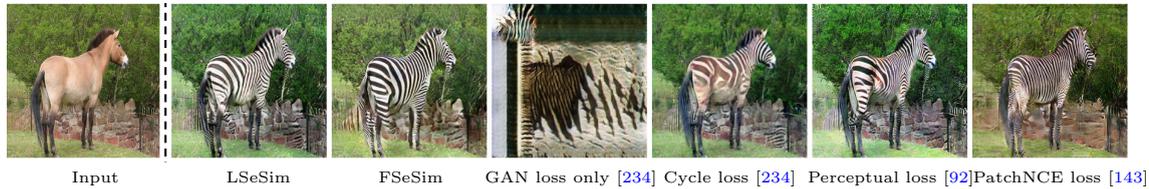

In Figure~\ref{fig:ch3_loss_comp}, we report a qualitative comparison of various losses that be applied to a same translation network architecture. All methods following the setting in CycleGAN~\cite{zhu2017toward}, except that the content loss is changed. Cycle-consistency is achieved using the auxiliary generator and discriminator, and all other methods are one-sided translation. We find that our method produces results with much better visual quality.

\paragraph{Discussion} Similar to conventional feature-level losses \cite{johnson2016perceptual,mechrez2018contextual}, our F/LSeSim is computed in a deep feature space. However, we represent the structure as multiple spatially-correlative maps. So rather than directly at feature level which is not free from domain-specific attributes, our comparison is done at a more abstract level that is intended to transcend domain specificity.

While attention maps have been used in previous image translation works~\cite{chen2018attention,alami2018unsupervised}, it is fundamentally different from our F/LSeSim in concept --- their attention maps effectively function as saliency maps to guide the translation, but content preservation is primarily still dependent on cycle-consistency loss. In our case, the multiple spatially-correlative maps are used to encode and determine invariance in scene structure. Our F/LSeSim also differs from the content loss used in~\cite{kolkin2019style}, in which the self-similarity was calculated at random positions without a clear purpose. Our F/LSeSim is on the other hand organized at a local patch level to explicitly represent the scene structure. As shown in Section \ref{res:ch3_multi}, our local structure representation is better than just using random spatial relationships. Furthermore, our LSeSim is a metric learned from the infoNCE loss, which generalizes well robustly on various tasks. While PatchNCE loss~\cite{park2020cut} can also learn feature similarity using contrastive loss, it directly compares features in two domains. 

\section{Experiment}\label{ch3:experiments}
To demonstrate the generality of our method, we instantiated F/LSeSim in multiple frameworks on various I2I translation tasks, including \emph{single-modal}, \emph{multi-modal}, and even \emph{single-image} translation. For each task, we used a suitable baseline architecture, but replaced their content losses with our F/LSeSim loss. In addition, we are only interested in scenarios where scene structure is preserved during the translation~\cite{isola2017image,zhu2017unpaired,zhu2017toward}, rather investigating translations incorporating shape modification \cite{choi2018stargan,choi2020stargan,nizan2020breaking,Kim20DST,baek2020rethinking}.

\subsection{\emph{Single-Modal} Unpaired Image Translation}\label{res:ch3_singlemodal}

We first evaluated our loss on the classical single-modal unpaired I2I translation. 

\paragraph{Implementation details} In this task, we chose CycleGAN \cite{zhu2017unpaired} as the reference architecture, but only used half of their pipeline and replaced the cycle-consistency loss with our F/LSeSim loss. Specifically, we used the ResNet-based generator with PatchGAN discriminator~\cite{isola2017image}. Details can be found on their \href{https://github.com/junyanz/pytorch-CycleGAN-and-pix2pix}{website}.

Our FSeSim is based on the ImageNet-pretrained VGG16 \cite{vgg}, where we used features from layers \texttt{relu}3\_1 and \texttt{relu}4\_1. While the LSeSim employs the same structure as FSeSim, the weights are not fixed and additionally two convolution layers, implemented as $1\times1$ kernels, are included to select better features. As for the selection of patches to build the contrastive loss, we found that random sampling the patch locations performed much better than uniform sampling on a grid, leading to better convergence when training the structure representation network. We set $\lambda=10$ in FSeSim and $\tau=0.07$ in LSeSim.

\paragraph{Metrics} Our evaluation protocols are adopted from previous work~\cite{heusel2017gans,park2019semantic,park2020cut}. We first used the popular Fr\'echet Inception Distance (FID)~\cite{heusel2017gans}\footnote{As claimed in StyleGANv2~\cite{karras2020analyzing}, ImageNet-pretrained classifiers tend to evaluate the distribution on texture than shape, while humans focus on shape. The best FID score does \emph{not} ensure the best image quality for translated images. As such, for a fair comparison, we reported the best FID score from all trained epochs for all methods, rather than the score in the last epoch. } to assess the visual quality of generated images by comparing the distance between distributions of generated and real images in a deep feature domain. For \emph{semantic image synthesis}, we further applied semantic segmentation to the generated images to estimate how well the predicted masks match the ground truth segmentation masks as in~\cite{chen2017photographic,wang2018high,park2019semantic,park2020cut}. Following~\cite{park2020cut,jiang2020tsit}, we used the pre-trained DRN~\cite{yu2017dilated}.

\begin{figure}[tb!]
	\centering
	\includegraphics[width=\linewidth]{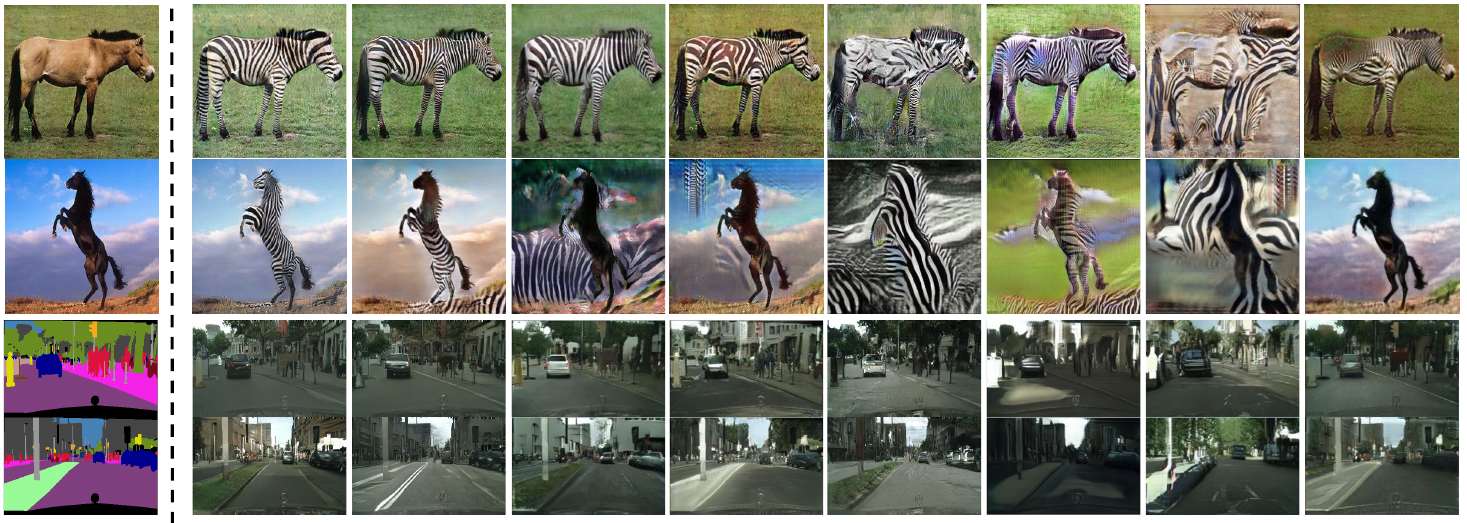}
	\begin{picture}(0,0)
		\put(-196,7){\tiny Input}
		\put(-144,7){\tiny LSeSim}
		\put(-99,7){\tiny FSeSim}
		\put(-56,7){\tiny CUT \cite{park2020cut}}
		\put(-21,7){\tiny CycleGAN \cite{zhu2017unpaired}}
		\put(32,7){\tiny MUNIT \cite{huang2018multimodal}}
		\put(75,7){\tiny DRIT++ \cite{lee2020drit++}}
		\put(126,7){\tiny DisGAN \cite{benaim2017one}}
		\put(168,7){\tiny GcGAN \cite{fu2019geometry}}
	\end{picture}
	\vspace{-0.2cm}
	\caption[Qualitative comparison on single-modal image translation]{\textbf{Qualitative comparison on single-modal image translation.} Here, we show results for \emph{horse}$\rightarrow$\emph{zebra} and \emph{label}$\rightarrow$ \emph{image}.}
	\label{fig:ch3_results_sin_mod}	
\end{figure}
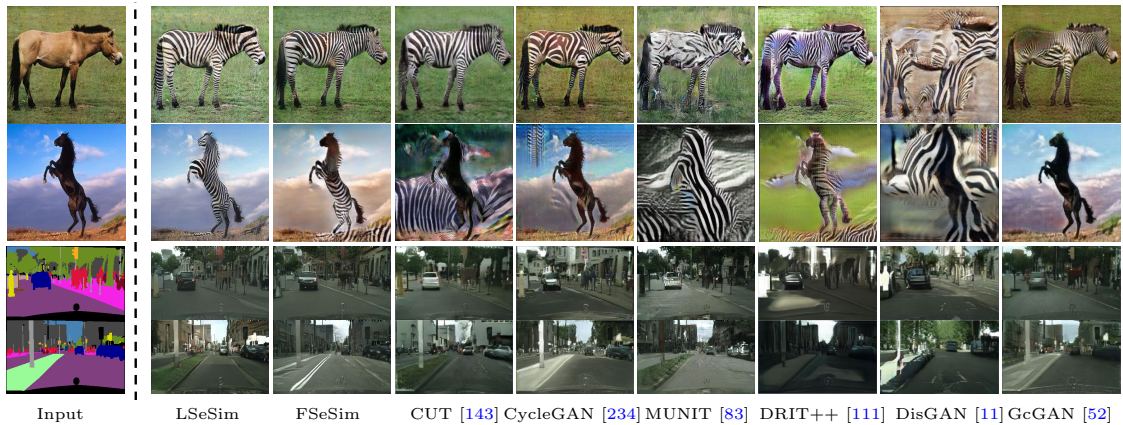

\begin{table}[tb!]
	\centering
	\renewcommand{\arraystretch}{1.1}
	\setlength\tabcolsep{8pt}
	\begin{tabular}{@{}lccccc@{}}
		\hlineB{3.5}
		\multirow{2}{*}{\textbf{Method}} & \multicolumn{2}{c}{\textbf{Cityscapes}}&& \multicolumn{2}{c}{\textbf{Horse$\rightarrow$Zebra}}\\
		\cline{2-3}\cline{5-6}
		& pixAcc$\uparrow$& FID$\downarrow$ && FID$\downarrow$ & Mem$\downarrow$\\
		\hlineB{2}
		CycleGAN~\cite{zhu2017unpaired} & 57.2 & 76.3 && 77.2 & 4.81 \\
		MUNIT~\cite{huang2018multimodal}  & 58.4 & 91.4 && 98.0 & 9.43\\
		DRIT++~\cite{lee2020drit++} & 60.3 & 96.2 && 88.5 & 11.2\\
		\cline{1-6}
		Distance~\cite{benaim2017one} & 47.2 & 75.9 && 67.2& 2.72\\
		GcGAN~\cite{fu2019geometry} & 65.5 & 57.4 && 86.7 & 4.68 \\
		CUT~\cite{park2020cut} & 68.8 & 56.4 && 45.5 & 3.33 \\
		\cline{1-6}
		FSeSim & 69.4 & 53.6 && 40.4 & \textbf{2.65} \\
		LSeSim & \textbf{73.2} & \textbf{49.7} && \textbf{38.0} & 2.92 \\
		\hlineB{2}
	\end{tabular}
	\caption[Quantitative comparison on single-modal image translation]{\textbf{Quantitative comparison on single-modal image translation}. FID~\cite{heusel2017gans} measures the distance between distributions of generated images and real images. ``Mem'' denotes the memory cost during training.}
	\label{table:ch3_single_mod}
\end{table}

\paragraph{Results}  In Table~\ref{table:ch3_single_mod}, we reported either published results or our reproductions with publicly-available code, choosing the better. Our simple, inexpensive losses substantially outperformed state-of-the-art methods, including two-sided frameworks with multiple cycle-consistency losses~\cite{zhu2017toward,huang2018multimodal,lee2020drit++}, and one-sided frameworks using self-distance~\cite{benaim2017one}, geometry consistency~\cite{fu2019geometry} and contrastive loss~\cite{park2020cut}. 

When compared to CycleGAN~\cite{zhu2017unpaired} and CUT~\cite{park2020cut}, although we used the same settings for the generator and discriminator, our method led to significant improvement. Unlike CUT~\cite{park2020cut} that depends on an identity pass for good performance, our results were achieved by training with only one pass using F/LSeSim and GAN losses. This suggests that once we explicitly decouple scene structure and appearance, it is easier for the model to modify the visual appearance correctly. As our model belongs to one-sided image translation that does \emph{not} require additional generators and discriminators, our model is also memory-efficient.  

Qualitative results are shown in Figures~\ref{fig:ch3_loss_comp} and \ref{fig:ch3_results_sin_mod}. In Figure~\ref{fig:ch3_loss_comp}, despite keeping the same settings and only comparing different content losses, our method translated the zebra appearance more cleanly. We also compared results using the same examples as \cite{park2020cut} in Figure~\ref{fig:ch3_results_sin_mod}, where our method achieved better visual results, even for some failure cases of \cite{park2020cut}.

\subsection{\emph{Multi-Modal} Unpaired Image Translation}\label{res:ch3_multi}

Our F/LSeSim is also naturally suited for multi-modal image translation, since the use of our spatial-correlative maps imposes only structural consistency and not appearance constraints.
We performed a thorough comparison of F/LSeSim to state-of-the-art methods, along with comprehensive ablation experiments. 

\paragraph{Implementation details} Our multi-modal setting is based on MUNIT~\cite{huang2018multimodal,lee2020drit++}, except our model uses only one generator and one discriminator of MUNIT~\cite{huang2018multimodal} without requiring the auxiliary generators and discriminators for multiple cycle training. Specifically, we used the ResNet-based generator with Instance Normalization (IN)~\cite{ulyanov2017improved} in the encoder and Adaptive Instance Normalization (AdaIN)~\cite{huang2017arbitrary,karras2019style} in the decoder, plus multi-scale discriminators~\cite{wang2018high}. The details of the architecture can be found on their \href{https://github.com/NVlabs/MUNIT}{website}. The F/LSeSim used is identical to that used in Section~\ref{res:ch3_singlemodal}. 

\paragraph{Metrics} Besides using FID to measure quality, we also used the average LPIPS distance~\cite{zhang2018unreasonable} to evaluate the diversity of generated results. The LPIPS distance is calculated by comparing the features of two images. Following~\cite{huang2018multimodal,zhu2017toward}, we computed the distances between 1900 pairs, sampling 100 images 19 times. We also report the latest metrics of Density and Coverage (D\&C)~\cite{ferjad2020icml}, which separately evaluate the diversity and fidelity of generated results. Likewise, we used the 1900 sampled pairs to compute D\&C scores. Higher scores here indicate larger diversity and better coverage to the ground-truth domain, respectively. 

\begin{figure}[tb!]
	\centering
	\includegraphics[width=\linewidth]{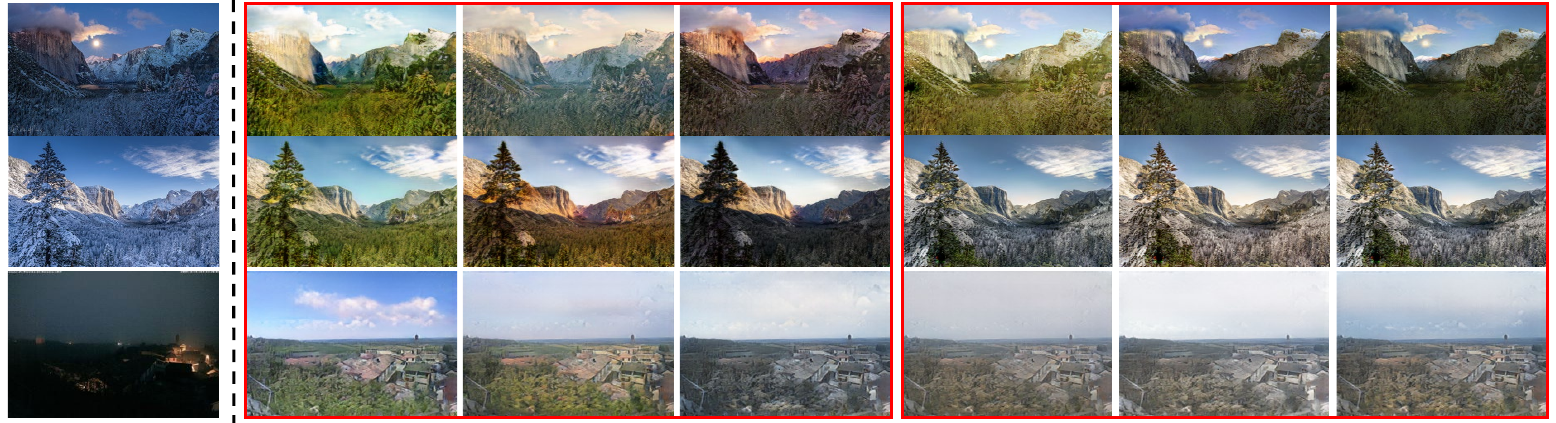}
	\begin{picture}(0,0)
		\put(-188,5){\footnotesize Input}
		\put(-70,5){\footnotesize LSeSim}
		\put(100,5){\footnotesize MUNIT \cite{huang2018multimodal}}
	\end{picture}
	\vspace{-0.2cm}
	\caption[Qualitative comparison on multi-modal image translation]{\textbf{Qualitative comparison on multi-modal image translation}. Here, we show the examples of \emph{winter}$\rightarrow$\emph{summer} and \emph{night}$\rightarrow$\emph{day}. Our model provides not only better visual results, but also produces larger diversity. }
	\label{fig:ch3_results_mul_mod}	
\end{figure}
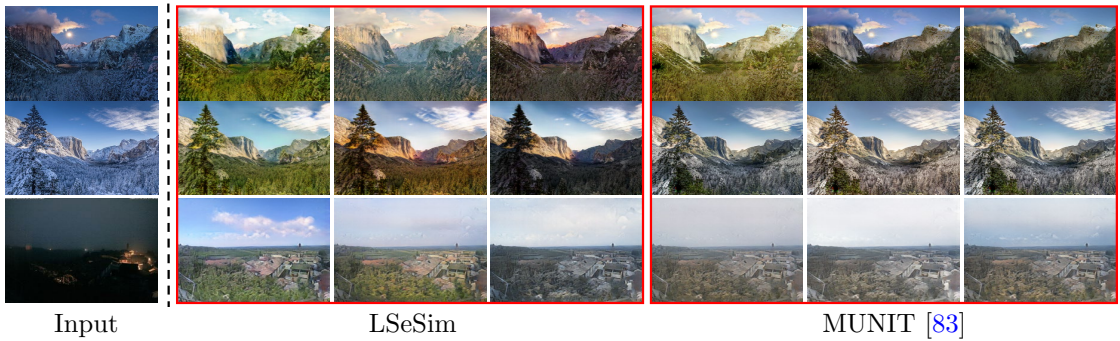

 \begin{table}[tb!]
	\centering
	\scriptsize
	\renewcommand{\arraystretch}{1.3}
	\setlength\tabcolsep{6pt}
	\begin{tabular}{@{}lccccccc@{}}
		\hlineB{3.5}
		\multirow{2}{*}{\textbf{Method}} & \multicolumn{3}{c}{\textbf{Winter$\rightarrow$Summer}}&& \multicolumn{3}{c}{\textbf{Night$\rightarrow$Day}}\\
		\cline{2-4}\cline{6-8}
		& LPIPS $\uparrow$& FID $\downarrow$ & D \& C $\uparrow$ && LPIPS $\uparrow$ & FID $\downarrow$ & D \& C $\uparrow$\\
		\hlineB{2}
		Real images & 0.770 & 44.1 & 0.997 / 0.986 && 0.684& 146.1 & 0.977 / 0.962 \\
		\cdashline{1-8}
		BicycleGAN~\cite{zhu2017toward} & \textbf{0.285} & 99.2 $\pm$ 3.2 & \textbf{0.536} / 0.667 && \textbf{0.349} & 290.9 $\pm$ 6.5 & \textbf{0.375} / 0.515 \\
		MUNIT~\cite{huang2018multimodal} & 0.160 & 97.4 $\pm$ 2.2 & 0.439 / 0.707&& 0.152 & 267.1 $\pm$ 2.7 & 0.271 / 0.548 \\
		DRIT++~\cite{lee2020drit++} & 0.186 & 93.1 $\pm$ 2.0 & 0.494 / 0.753 && 0.167 & 258.5 $\pm$ 2.3 & 0.298 / 0.631 \\
		\cdashline{1-8}
		\textbf{FSeSim} & 0.216 & 90.5 $\pm$ 1.9 & 0.501 / 0.779 && 0.203 & 234.3 $\pm$ 2.8 & 0.332 / 0.638 \\
		\textbf{LSeSim} & 0.232 & \textbf{89.4 $\pm$ 1.9} & 0.516 / \textbf{0.793} && 0.215 & \textbf{224.9 $\pm$ 2.0}  & 0.347 / \textbf{0.652} \\
		\hlineB{2}
	\end{tabular}
	\caption[Quantitative evaluation on multi-modal image translation task]{\textbf{Quantitative evaluation on multi-modal image translation task}. LPIPS distance~\cite{zhang2018unreasonable} measures the diversity of generated images by comparing the features of two images, while (D\&C)~\cite{ferjad2020icml} evaluates the diversity and fidelity by matching whole features in the generated and real datasets. }
	\label{table:ch3_multi_modal}
\end{table}

\paragraph{Results} We compared our F/LSeSim to state-of-the-art methods in multi-modal image translation in Table~\ref{table:ch3_multi_modal}. Our method outperformed the two baselines, MUNIT~\cite{huang2018multimodal} and DRIT++~\cite{lee2020drit++}, although we deployed the same network architecture. In Table~\ref{table:ch3_multi_modal}, our method achieved larger diversity with higher LPIPS score and better image quality with lower FID score. Besides, BicycleGAN~\cite{zhu2017toward} achieved larger diversity (higher LPIPS score) on all tasks by adding noise to all decoders through the U\_Net~\cite{ronneberger2015u}, but the tradeoffs are worse visual results (the highest FID score), due to the larger noise being directly added to the last generative layer. In contrast, we only added noise to the middle layers of generation, through AdaIN.

In Figure~\ref{fig:ch3_results_mul_mod}, we show qualitative comparisons of our method to MUNIT~\cite{huang2018multimodal} on \emph{winter $\rightarrow$ summer}, and \emph{night $\rightarrow$ day} tasks. As can be seen, our model not only generated higher quality translated results, but also produced more diverse solutions for these multi-modal tasks. We believe this is because the formulation of our F/LSeSim will only maintain structural fidelity, and does not impose penalties on appropriate appearance modifications in the target domain.

\begin{table}[tb!]
	\centering
	\scriptsize
	\renewcommand{\arraystretch}{1.3}
	\setlength\tabcolsep{4pt}
	\begin{tabular}{@{}llcccccc@{}}
		\hlineB{3.5}
		& \multirow{2}{*}{\textbf{Configuration}} & \multicolumn{2}{c}{\textbf{Horse $\rightarrow$ Zebra}} && \multicolumn{3}{c}{\textbf{Night $\rightarrow$ Day}} \\
		\cmidrule{3-4}\cmidrule{6-8}
		& & FID $\downarrow$ & Mem(GB) $\downarrow$ && FID $\downarrow$ & LPIPS $\uparrow$ & D \& C $\uparrow$ \\
		\hlineB{2}
		A & STROTSS~\cite{kolkin2019style} (random SeSim) & 70.1 & 2.68 && 262.7 $\pm$ 3.6  & 0.162 & 0.289 / 0.554 \\
		B & Baseline (global SeSim on single layer) & 53.7 & 2.97 && \textbf{173.2 $\pm$ 2.2} & 0.168 &  0.303 / \textbf{0.664} \\
		\cdashline{1-8}
		C & (B): Global $\rightarrow$ Patch ($32\times32$) & 45.8 & \textbf{2.61} && 231.3 $\pm$ 2.5 & 0.181 &  0.317 / 0.634 \\
		D & (C): Single $\rightarrow$ Multi (\texttt{relu3}\_1, \texttt{relu}4\_1) & 43.3 & 2.65 && 229.4 $\pm$ 2.1 & 0.177 &  0.311 / 0.646 \\
		E & (D): l1loss $\rightarrow$ 1 - $consine$& 40.4 & 2.65 && 234.3 $\pm$ 2.8 & 0.203 & 0.332 / 0.638  \\
		\cdashline{1-8}
		F & Ours LSeSim & \textbf{38.0} & 2.92 && 224.9 $\pm$ 2.0 &  \textbf{0.215} & \textbf{0.347} / 0.652 \\
		\hlineB{2}
	\end{tabular}
	\caption[Ablation study on both single- and multi-modal image translation]{\textbf{Ablation study on both single- and multi-modal image translation}. Refer to ablation experiments in main text for details.}
	\label{table:ch3_ablation}
\end{table}

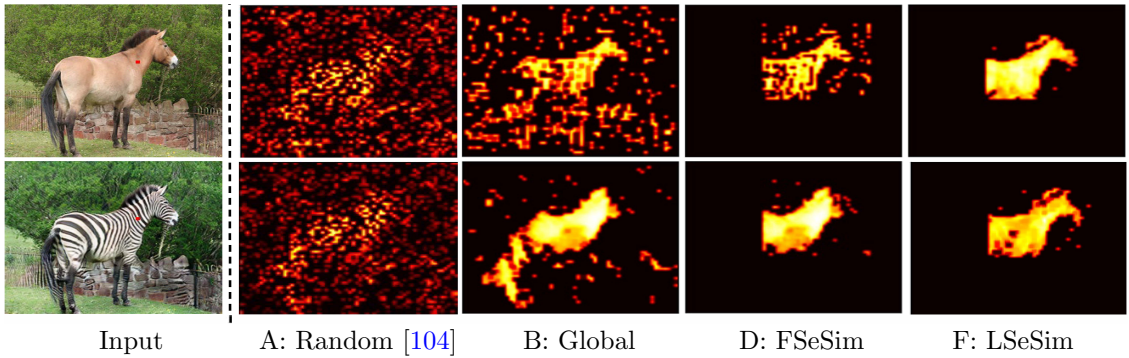
\begin{figure}[tb!]
	\centering
	\includegraphics[width=\linewidth,height=0.18\textheight]{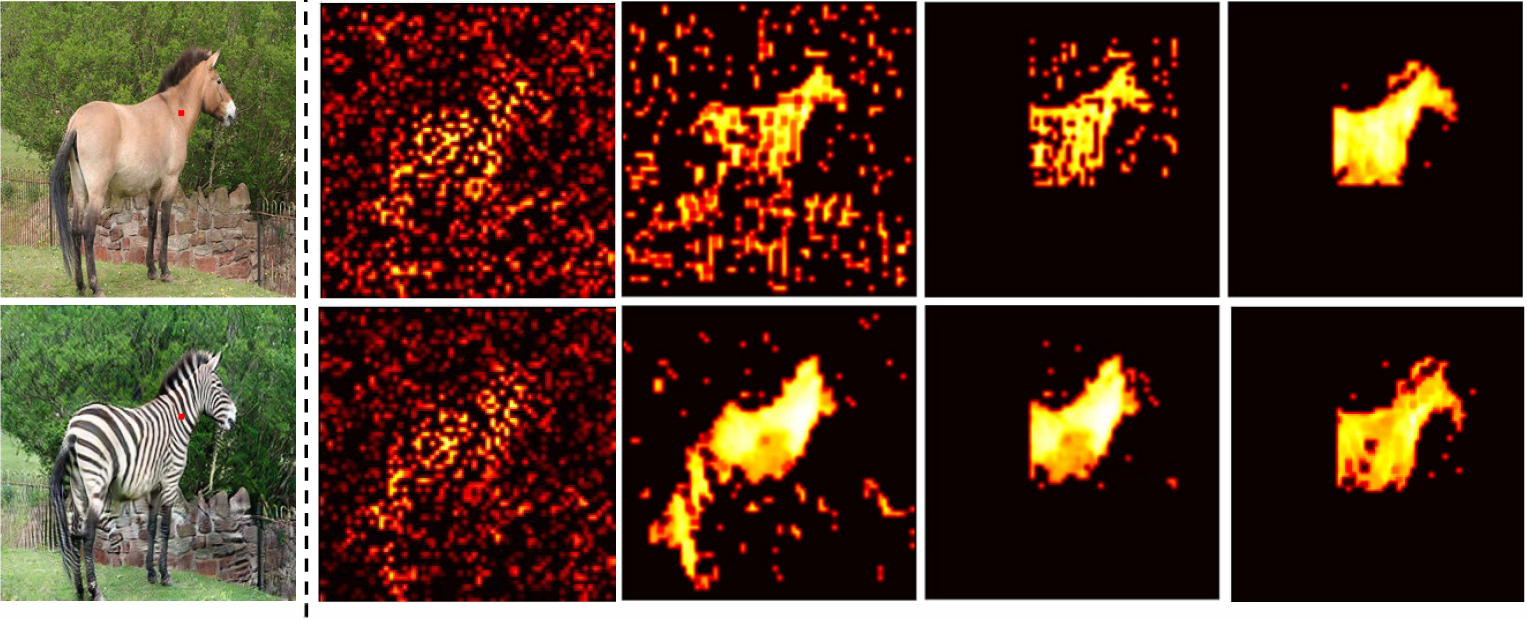}
	\begin{picture}(0,0)
		\put(-175,5){\footnotesize Input}
		\put(-115,5){\footnotesize A: Random~\cite{kolkin2019style}}
		\put(-16,5){\footnotesize B: Global}
		\put(65,5){\footnotesize D: FSeSim}
		\put(145,5){\footnotesize F: LSeSim}
	\end{picture}
	\caption[Ablation study on self-similarity maps]{\textbf{Ablation study on self-similarity maps}. A, B, D and F correspond to the settings in Table~\ref{table:ch3_ablation}, respectively.}
	\label{fig:ch3_results_ablation}	
\end{figure}

\paragraph{Ablation Experiments} To understand the influence of different components for the proposed spatially-correlative loss, we ran a number of ablations. The quantitative results are reported in Table~\ref{table:ch3_ablation} for both single- and multi-modal image translation. In this table, \textbf{row A} shows the performance of the baseline method~\cite{kolkin2019style} which utilizes self-similarity as content loss. However, it calculates the similarity using random sampled features and does not have an explicit connection to spatial structure. \textbf{Row B} is a global attention map. While this version performed well and ran faster by avoiding sampling, it has two main limitations. First, the original global attention module is memory intensive and \emph{cannot} be applied to multiple scales nor to large feature spaces. Second, as evident from Figure~\ref{fig:ch3_results_ablation}, the spatially distant correlation is essentially noise (as is also the case for the Random baseline of \cite{kolkin2019style}), which is detrimental to the results. Compared to the global version, \textbf{row C} largely improved the performance as clearer shapes are captured in the local patches. In \textbf{row D}, we applied local attention to multiple layers with a fixed path size. This results in the spatially-correlative maps having different receptive fields, which further improves the performance. \textbf{Row E} replaces the $l_1$ distance with cosine distance. While the improvement in image quality is not obvious, the diversity scores increased substantially. This is due to the cosine similarity supporting only the correlation between the two spatially-correlative maps without encouraging the maps to be fully same. \textbf{Row E} shows the performance of the full model (same as in Tables~\ref{table:ch3_single_mod} and~\ref{table:ch3_multi_modal}), where LSeSim of \textbf{row F} improved on many metrics, and had better visual results.

 \begin{figure}[tb!]
	\centering
	\includegraphics[width=\linewidth]{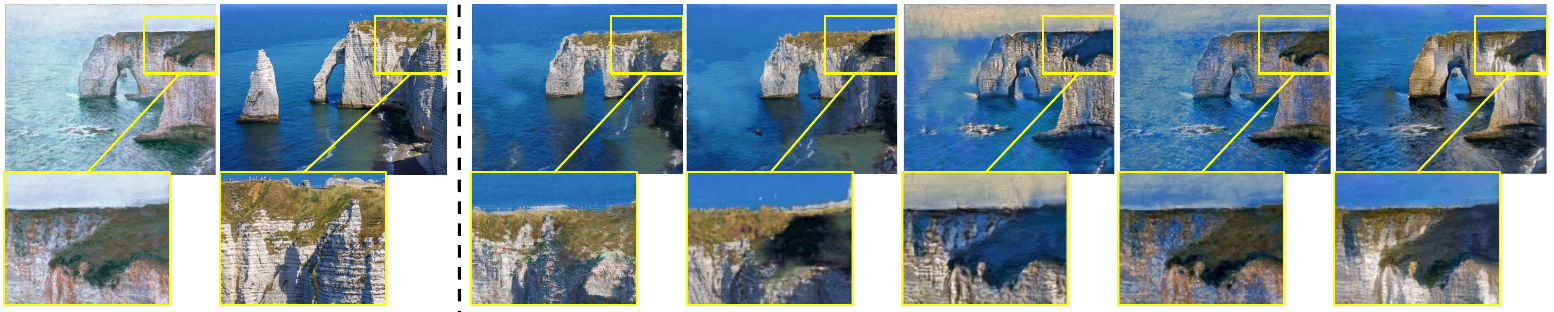}
	\begin{picture}(0,0)
		\put(-186,5){\scriptsize Input}
		\put(-130,5){\scriptsize Target}
		\put(-65,5){\scriptsize FSeSim}
		\put(-10,5){\scriptsize CUT~\cite{park2020cut}}
		\put(35,5){\scriptsize Gatys \etal~\cite{gatys2016image}}
		\put(102,5){\scriptsize WCT$^2$~\cite{yoo2019photorealistic}}
		\put(155,5){\scriptsize STRORSS~\cite{kolkin2019style}}
	\end{picture}
	\caption[High-resolution painting to photorealistic image]{\textbf{High-resolution painting to photorealisitc image} on single-image translation.}
	\label{fig:ch3_results_sin_img_real}	
\end{figure}
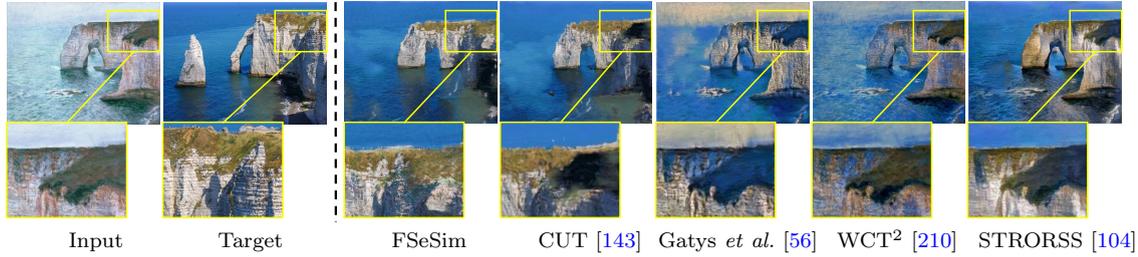

\subsection{\emph{Single-Image} Unpaired Image Translation}

To further test the generalization ability, we applied the FSeSim to a high-resolution single-image translation task. Here, only one source and one target image are provided for training, but they are unpaired. This task is conceptually similar to the style transfer~\cite{gatys2016image,johnson2016perceptual,luan2017deep}, except that here we trained a SinGAN-like~\cite{shaham2019singan,InGAN} model that captures the distribution of a single image through the adversarial learning, rather than using a fixed style loss \cite{johnson2016perceptual}.

\paragraph{Implementation details} The single-image translation setting is based on the CUT~\cite{park2020cut} method, except that the PatchNCE loss is replaced by our FSeSim loss. In detail, the StyleGAN2-based generator and discriminator~\cite{karras2020analyzing} with the gradient penalty~\cite{mescheder2018training,karras2019style} are used. To further increase simplicity, we removed the identity loss in CUT~\cite{park2020cut}, and only used a GAN loss in conjunction with the proposed FSeSim loss to assess the appearance and structure separately. As these $64\times64$ cropped patches have to be taken from a high-resolution image for training here, it becomes less useful to further subsample ``positive'' and ``negative'' patches. Therefore, we only use our FSeSim to train the model, without using LSeSim with contrastive loss.

\paragraph{Results} In Figure~\ref{fig:ch3_results_sin_img_real}, we show qualitative results from the CUT~\cite{park2020cut} paper on the \emph{painting}$\rightarrow$\emph{photo} task. As evident in the highlighted regions, our model generated not only higher quality results, but they were also closer to the target image style than existing methods, including classical style transfer models, such as WCT$^2$~\cite{yoo2019photorealistic} and STRORSS~\cite{kolkin2019style}, as well as the latest single-image translation CUT~\cite{park2020cut} model.  

\section{Limitations and Discussion}\label{ch3:conc}

In this chapter, we introduced F/LSeSim, a new structure consistency loss that focuses only on spatially-correlative relationships, without regard to visual appearances. The proposed F/LSeSim is naturally suitable for tasks that require structure consistency, and can be easily applied to existing architectures. We demonstrated its generality to various unpaired I2I translation tasks, where a simple replacement of the existing content losses with F/LSeSim led to solid performance improvements.

As demonstrated in experiments \ref{ch3:experiments}, the proposed spatially-correlative loss can easily be integrated into existing network architectures and thus allows wide applicability. However, the proposed structure loss only models the \emph{content / structure} representation in this work, leaving the \emph{style / appearance} to be judged by an auxiliary discriminator, which is not always stable in some situations. A future step is to better model the \emph{style}, and to effectively incorporate \emph{content} and \emph{style} losses in translation. 

So far, in the last two chapters, I have investigated the specific problem of unpaired I2I translation. In Chapter \ref{ch:Synthestic2Real}, a system for synthetic-to-realistic translation was proposed that solves for single image depth estimation. In this chapter, a general spatially-correlative loss was introduced for various I2I translation tasks, where the \emph{structure} representation was explicitly modeled. These works mainly focus on modifying the appearance, which is a basic operation in visual synthesis. Next, in Part \uppercase\expandafter{\romannumeral2}, we go further to explore the content modification in visual synthesis, which requires a high-level semantic perception of a scene, instead of purely changing low-level appearance. 
\part{Generating Semantic Content: \\ Image Completion} %

\chapter{Pluralistic Image Completion} 
\chaptermark{PICNet}
\label{ch:PICNet} 

This chapter covers a classic generative task: image inpainting / completion \cite{bertalmio2000image}. At the time of publication\footnote{This work was published as \emph{Pluralistic Image Completion} in CVPR, 2019 \cite{zheng2019pluralistic}.}, the previous approaches produced only one result for a given masked image, although there may be many reasonable possibilities. In this chapter, a new perspective is presented, \textbf{pluralistic image completion} -- the task of generating \emph{multiple} and \emph{diverse} plausible solutions. Although there had been some earlier works on multiple solutions in image generation and translation, it is significantly harder for image completion as all the multiple solutions need to seamlessly fit with the unmasked regions of the input image. A novel and probabilistically principled framework with two parallel paths is proposed. One is a reconstructive path that utilizes the only one ground truth to get a prior distribution of missing patches and rebuild the original image from this distribution. The other is a generative path for which the conditional prior is coupled to the distribution obtained in the reconstructive path. Experiments show that our method not only yields better results in various datasets than existing state-of-the-art methods, but also provides multiple and diverse outputs. This work was followed by many researchers \cite{deng2020image,zhao2020uctgan,peng2021generating,wan2021high} to explore the multiple and diverse results for this highly subjective task. 

The rest of this chapter is organized as follows: We introduce and discuss the motivation and previous works in Sections \ref{ch4:intro} and \ref{ch4:back}. Next, we describe the proposed probabilistically principled framework and the improved attention module in Section \ref{ch4:approch}. Section \ref{ch4:user} presents the interface for users to freely edit images. We then describe and discuss the experiments in Section \ref{ch4:experiment}, and conclude in Section \ref{ch4:conc}.

\section{Introduction}\label{ch4:intro}

Image completion involves filling alternative content into the missing parts of images. It can be used for restoring damaged paintings, removing unwanted objects, and generating new content for incomplete scenes. Many approaches have been proposed for this non-trivial task, including diffusion-based methods \cite{bertalmio2000image,ballester2001filling,levin2003learning,bertalmio2003simultaneous}, patch-based methods \cite{criminisi2003object,criminisi2004region,jia2004inference,barnes2009patchmatch}) and learning-based methods \cite{pathak2016context,iizuka2017globally,yu2018generative,Liu_2018_ECCV,Nazeri_2019_ICCV,yi2020contextual}. While these approaches rapidly improve the completion results, they produce only one ``optimal'' result for a given masked image and do \emph{not} have the capacity to generate a variety of semantically meaningful results. It remains a challenging problem to provide \emph{multiple} and \emph{diverse} plausible results for this highly subjective problem. 

\begin{figure*}[tb!]
	\centering
	\includegraphics[width=\linewidth]{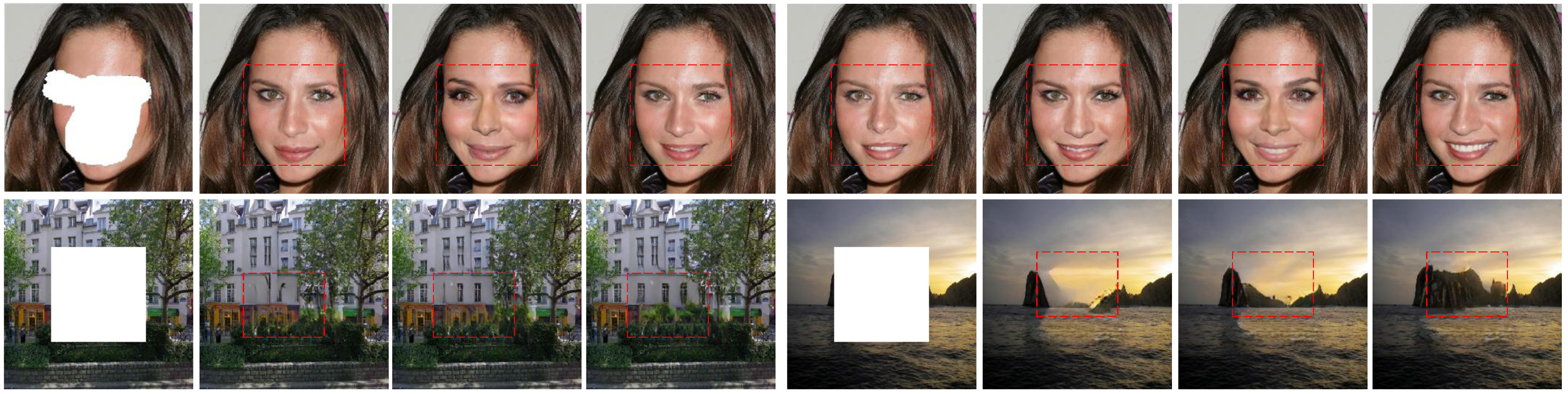}
	\caption[Example completion results]{\textbf{Example completion results of our method} on images of a face, a building, and natural scenery with various masks (masks shown in white only for visual purpose). For each group, the masked input image is shown left, followed by sampled multiple results from our model without any post-processing.}
	\label{fig:ch4_example}
\end{figure*}

Supposing you were shown the images with various missing regions in Figure~\ref{fig:ch4_example}, what would you \emph{imagine} to be occupying these holes? Bertalmio \etal \cite{bertalmio2000image} related how expert conservators would restore damaged art by: 1) imagining the semantic content to be filled based on the overall scene; 2) ensuring structural continuity between the masked and unmasked regions; and 3) filling in visually realistic content for missing regions. Nonetheless, each expert will independently end up creating \emph{substantially different details}, such as various shapes and colors of eyes, even if they may universally agree on high-level semantics, such as general placement of eyes and mouth on a damaged portrait.

Based on this observation, the main research goal in this chapter is thus to generate \emph{multiple} and \emph{diverse} plausible results when presented with a masked image. We refer to this task as \textbf{pluralistic image completion} (depicted in Figure~\ref{fig:ch4_example}). This is as opposed to existing works that attempt to generate only a single ``guess'' for this ill-posed problem.

To obtain a diverse set of results for a given input, some methods utilize conditional variational auto-encoders (CVAE) \cite{sohn2015learning, walker2016, bao2017cvae, Eslami2018}, a conditional extension of variational auto-encoders (VAE) \cite{kingma2013auto}, which explicitly code for a distribution that can be sampled. However, specifically for an image completion scenario, the standard single-path formulation usually leads to grossly underestimating variances. This is because when \emph{the condition label is itself a masked image}, the number of ground truth instances in the training data that match the label is \emph{typically only one -- the original complement of the masked image}. Hence, the estimated original conditional distributions tend to have very limited variation since they were trained to reconstruct the single original image.

An important insight we will use is that \emph{partial images (patches)}, as a superset of full images, may also be considered as generated from \emph{a latent space with smooth prior distributions} \cite{shaham2019singan}. This provides a mechanism for alleviating the problem of having scarce samples per conditional masked image. To do so, we introduce a {\bf P}luralistic {\bf I}mage {\bf C}ompletion {\bf Net}work, called {\bf PICNet}, with two parallel but linked training pipelines. The first pipeline is a VAE-based reconstructive path that not only utilizes the full instance ground truth, but also imposes smooth priors for the latent space of missing partial image. The second pipeline is a generative path that learns to predict the latent prior distribution for the missing regions only based on the visible pixels, from which can be sampled to generate diverse results. The training process for the latter path does \emph{not} attempt to steer the output towards reconstructing the instance-specific results at all, instead allowing the reasonableness of results being driven by an auxiliary discriminator network \cite{goodfellow2014generative}. This leads to substantially great variability in generation. To further utilize the information from the visible partial images as much as possible \cite{barnes2009patchmatch,yu2018generative}, we also introduce an enhanced \emph{short+long term patch attention} layer, a generic attention mechanism that allows information flowing from visible regions to missing holes. 

We comprehensively evaluate and compare our approach with existing state-of-the-art methods on a large variety of scenes (Section \ref{sec:ch4_comp_results}), where various masks, including regular and free-form irregular masks, are used to erode the images. We additionally present many interesting applications of our model on free-form image editing (Section \ref{sec:ch4_add_result}), \eg object removal, face editing, and scene content-aware move. The extensive experimental results demonstrate that our proposed PICNet not only generates higher-quality completion results, but also produces multiple diverse solutions for this subjective processing task.

\section{Background}\label{ch4:back}

Existing work on image completion either uses information from within the image  \cite{bertalmio2000image, bertalmio2003simultaneous}, or information from a large image dataset \cite{hays2007scene, pathak2016context}. Most approaches generate only one result per masked image, which is precisely the downside we want to address in this chapter.

\subsection{Intra-Image Completion}

Traditional intra-image completion works (also known as ``inpainting'' \cite{bertalmio2000image}) mainly propagate, copy and realign the background regions to missing regions, focusing only on the steps 2 and 3 mentioned above, by assuming that the holes should be filled with similar appearance to that of the visible regions. One category of intra-image completion methods are diffusion-based image synthesis \cite{bertalmio2000image,ballester2001filling,levin2003learning,bertalmio2003simultaneous}. These methods fill the surrounded backgrounds to the missing regions by propagating the local colors. They only work well on the small and narrow holes. Another category of intra-image completion methods are patch-based approaches \cite{criminisi2003object,criminisi2004region,jia2004inference,barnes2009patchmatch}. They fill the holes by copying information from similar visible regions, which produce high-quality texture-consistent result. However, these intra-image methods cannot capture global semantics to hallucinate new content for large holes (as in step 1), which is significant for real image completion. 

\subsection{Inter-Image Completion}

To hallucinate semantically new content, inter-image completion borrows information from a large dataset. Hays and Efros~\cite{hays2007scene} first present an image completion method using millions of images. Recently, learning-based approaches are proposed. Initial works \cite{kohler2014mask, ren2015shepard} focus on small and thin holes. Then, Pathak \etal \cite{pathak2016context} proposed the Context Encoders (CE) to handle 64$\times$64-sized holes. Iizuka \etal \cite{iizuka2017globally} built upon \cite{pathak2016context} by combining global and local discriminators (GL) as adversarial loss. Wang \etal \cite{wang2018image} designed Multi-column CNNs and a cosine similarity based loss for high-quality image inpainting. More recent, Liu \etal \cite{Liu_2018_ECCV} introduced ``partial convolution'' for free-form irregular mask image completion. 

Some work has also explored additional information for semantically image completion. In \cite{yeh2017semantic}, the ``closest'' features in the latent space for the masked image are searched to generate an image. Li \etal \cite{li2017generative} introduced additional face parsing loss to ensure the semantic consistency of completed images. Song \etal \cite{song2018spg} proposed SPG-Net that simultaneously does semantic map and RGB appearance completion. Moreover, sketches and color are used in the latest Faceshape \cite{portenier2018faceshop}, DeepFillv2~\cite{yu2019free}, EdgeConnect~\cite{Nazeri_2019_ICCV} and SC-FEGAN~\cite{jo2019sc}. A common drawback of these methods is that they utilize the visible information only through local convolutional operations, which creates distorted structures and blurry textures inconsistent with the visible regions, especially for large holes. 

\subsection{Combing Intra- and Inter-Image Completion}

To mitigate the blurry problems, Yang \etal \cite{yang2017high} proposed multi-scale neural patch synthesis, which generates high-frequency details by copying patches from mid-layer features. More recently, several works \cite{yu2018generative,Yan_2018_ECCV,song2018contextual,yi2020contextual} exploit spatial attention \cite{jaderberg2015spatial,zhou2016view} to get high-frequency details. Yu \etal~\cite{yu2018generative} proposed a contextual attention layer to produce high-frequency details by copying similar features from visible regions to missing regions. Yan \etal \cite{Yan_2018_ECCV} and Song \etal \cite{song2018contextual} proposed PatchMatch-like ideas on feature domain. Yi \etal \cite{yi2020contextual} proposed contextual residual aggregation for very high resolution (8K) image inpainting. However, these methods identify similar features by comparing features of holes and visible regions, which is somewhat contradictory as feature transfer is unnecessary when two features are very similar, but when needed the features are too different to be matched easily. Furthermore, distant information is not used for new content that differs from visible regions. Our model solves it by extending self-attention to harness abundant context.

\subsection{Image Generation}

Image generation has progressed significantly using methods such as VAE \cite{kingma2013auto} and GANs \cite{goodfellow2014generative}. These have been applied to conditional image generation tasks, such as image translation \cite{isola2017image,zhu2017unpaired}, synthetic to realistic \cite{shrivastava2017learning,zheng2018t2net}, future prediction \cite{mathieu2015deep}, and 3D models \cite{park2017transformation}. Perhaps most relevant in spirit to us are conditional VAEs (CVAE) \cite{sohn2015learning,walker2016} and CVAE-GAN \cite{bao2017cvae}, but these are not specially targeted for image completion. CVAE-based methods are most useful when the conditional labels are few and discrete, and there are sufficient training instances per label. Some recent work utilizing these in image translation can produce diverse output~\cite{zhu2017toward,lee2018diverse}, but in such situations the condition-to-sample mappings are more local (\eg pixel-to-pixel), and only change the visual appearance without generating new content. This is untrue for image completion, where the conditional label is the masked image itself, with only one training instance of the original holes. In \cite{chen2018high}, different outputs were obtained for face completion by specifying facial attributes (\eg smile), but this method is very domain specific, requiring targeted attributes. In contrast, our proposed probabilistically principled framework produces multiple and diverse plausible in various datasets, which does not need any label information for training.

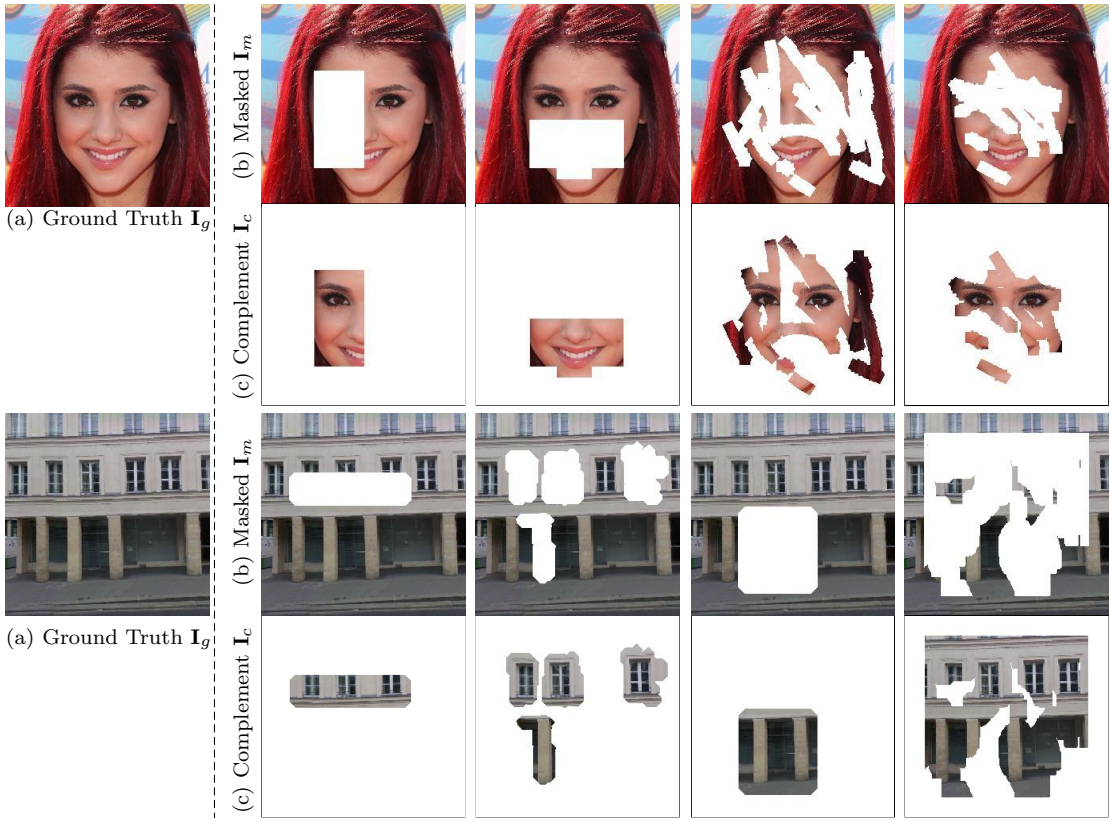
\begin{figure}[tb!]
	\centering
	\includegraphics[width=\linewidth]{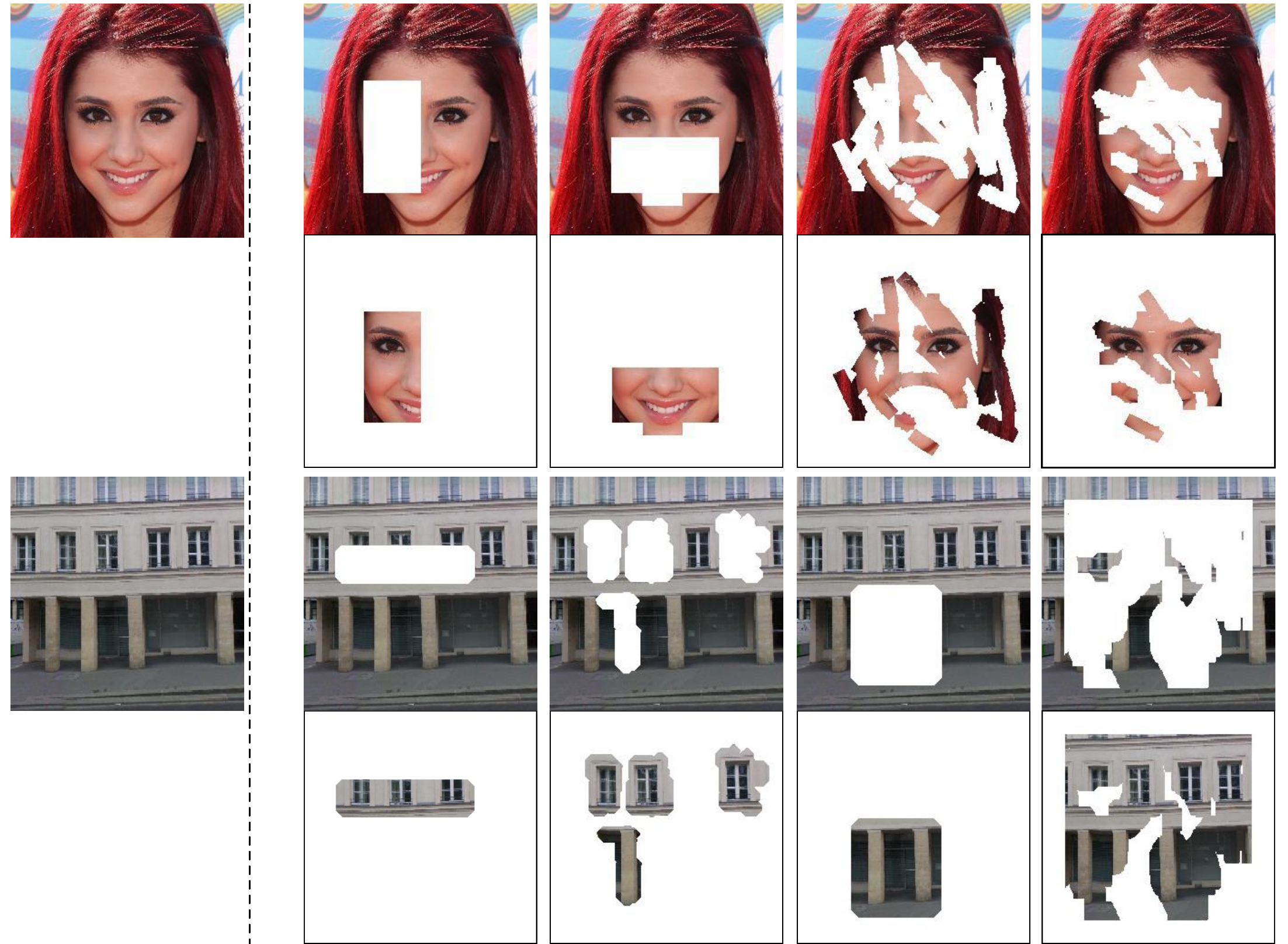}
	\begin{picture}(0,0)
	    \put(-205, 238){\scriptsize (a) Ground Truth $\textbf{I}_g$}
	    \put(-120,255){\rotatebox{90}{\scriptsize (b) Masked $\textbf{I}_m$}}
	    \put(-120,172){\rotatebox{90}{\scriptsize (c) Complement $\textbf{I}_c$}}
	    \put(-205, 80){\scriptsize (a) Ground Truth $\textbf{I}_g$}
	    \put(-120,102){\rotatebox{90}{\scriptsize (b) Masked $\textbf{I}_m$}}
	    \put(-120,16){\rotatebox{90}{\scriptsize (c) Complement $\textbf{I}_c$}}
	\end{picture}
	\vspace{-0.3cm}
	\caption[Examples of different degraded images]{\textbf{Examples of different degraded images.} (a) Ground truth image $\mathbf{I}_g$. (b) Masked image $\mathbf{I}_m$. (c) The corresponding complement image $\mathbf{I}_c$ to each top masked image $\mathbf{I}_g$. It is often not reasonable to strongly enforce the completed masked regions to be identical to the ground truth, especially in cases when large variations in the completed content can still be perfectly consistent to the visible regions, \eg when the entire mouth or both eyes are masked. } 
	\label{fig:ch4_degraded_image}	
\end{figure}

\section{Approach}\label{ch4:approch}

Suppose we have an image, originally ground truth $\mathbf{I}_g$ (Figure~\ref{fig:ch4_degraded_image} (a)), but degraded by a number of missing pixels to become $\mathbf{I}_m$ (Figure~\ref{fig:ch4_degraded_image} (b)), \emph{masked partial image} comprising the visible pixels. We also define $\mathbf{I}_c$ (Figure~\ref{fig:ch4_degraded_image} (c)) as its \emph{complement partial image} comprising the missing pixels. 

\begin{figure}[tb!]
	\centering
	\includegraphics[width=0.8\linewidth]{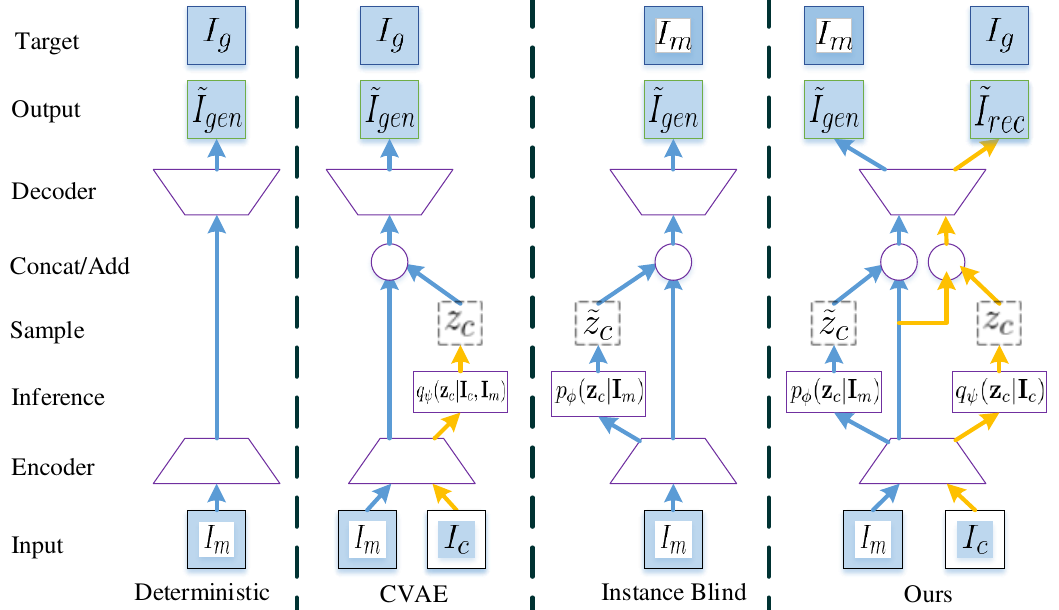}
	\caption[Completion strategies given masked image]{\textbf{Completion strategies given masked image.} (Deterministic) structure directly predicts the ground truth instance. (CVAE \cite{walker2016}) adds in random sampling to diversify the output, but is still trained on the single ground truth. (Instance Blind) only matches the masked instance, but training is unstable. (Ours) uses a generative path during testing, but is guided by a parallel reconstructive path during training. Note that, yellow path is only used for training.}
	\label{fig:ch4_coarse_framework}	
\end{figure}

Prior image completion methods \cite{yu2018generative,pathak2016context,iizuka2017globally,Nazeri_2019_ICCV} attempt to reconstruct the original unmasked image $\mathbf{I}_g$ in a deterministic fashion from $\mathbf{I}_m$ (see Figure~\ref{fig:ch4_coarse_framework} ``Deterministic''). However, this rigid approach has several limitations. First, while it is fine to rebuild the original image $\mathbf{I}_g$ when visible regions tightly constrain the completed content, \eg when only the left half of a face is masked in Figure~\ref{fig:ch4_degraded_image}, it is unnecessarily limiting when visible regions allow a much greater range of perceptually consistent completion, \eg with many different mouth expressions or building door appearances equally acceptable in Figure~\ref{fig:ch4_degraded_image}. Second, deterministic methods can only generate a single solution and are not able to recover a richer distribution of reasonable possibilities. Instead, our goal is to \emph{sample} from $p(\mathbf{I}_c|\mathbf{I}_m)$ and we reconstruct the original image only when the corresponding complement partial images $\mathbf{I}_c$ are provided during the training. 

\subsection{Pluralistic Image Completion Network}\label{sec:ch4_picnet}

\subsubsection{Probabilistic Framework}\label{ch4_probabilistic}

In order to have a distribution to sample from, an approach is to employ the CVAE \cite{sohn2015learning} which estimates a parametric distribution over a latent space, from which sampling is possible. This involves a variational lower bound of the conditional log-likelihood:
\begin{equation}\label{eq:ch4_CVAE}
	\log p(\mathbf{I}_c|\mathbf{I}_m)\ge -\text{KL}(q_\psi(\mathbf{z}_c|\mathbf{I}_c,\mathbf{I}_m)||p_\phi(\mathbf{z}_c|\mathbf{I}_m)) + \mathbb{E}_{q_\psi(\mathbf{z}_c|\mathbf{I}_c,\mathbf{I}_m)}[\log p_\theta(\mathbf{I}_c|\mathbf{z}_c,\mathbf{I}_m)]
\end{equation}
where $\mathbf{z}_c$ is the latent vector of missing patches,  $q_\psi(\mathbf{z}_c|\mathbf{I}_c,\mathbf{I}_m)$ is the recognition network, $p_\phi(\mathbf{z}_c|\mathbf{I}_m)$ is the conditional prior, and $p_\theta(\mathbf{I}_c|\mathbf{z}_c,\mathbf{I}_m)$ is the likelihood, with $\psi$, $\phi$ and $\theta$ being the deep network parameters of their corresponding functions. This lower bound is maximized \wrt all parameters. The detail proofs are provided in Appendix \ref{AppendixA}.

For our purposes, the chief difficulty of using CVAE \cite{sohn2015learning} directly is that the high DoF of recognition network $q_\psi(\mathbf{z}_c|\mathbf{I}_c,\mathbf{I}_m)$ and conditional prior network $p_\phi(\mathbf{z}_c|\mathbf{I}_m)$ are not easily separable in equation (\ref{eq:ch4_CVAE}). Besides, since the conditional prior network $p_\phi(\mathbf{z}_c|\mathbf{I}_m)$ is sufficiently unconstrained in equation (\ref{eq:ch4_CVAE}), it will lean a narrow delta-like prior distribution of $p_\phi(\mathbf{z}_c|\mathbf{I}_m)\rightarrow \delta(\mathbf{z}_c-\mathbf{z}^*_c)$, where $\mathbf{z}^*_c$ is the maximum latent likelihood point of $p_\theta(\mathbf{I}_c|\mathbf{z}_c,\mathbf{I}_m)$. In this way, the variance $\sigma^2$ of the learned latent distribution is easily driven towards zero. Then it is approximately equivalent to maximizing $\mathbb{E}_{p_\phi(\mathbf{z}_c|\mathbf{I}_m)}[\log p_\theta(\mathbf{I}_c|\mathbf{z}_c,\mathbf{I}_m)]$, the ``GSNN'' variant in \cite{sohn2015learning}, in which they directly set the recognition network the same as the prior network, \ie, $q_\psi(\mathbf{z}_c|\mathbf{I}_c,\mathbf{I}_m)$ $=$ $p_\phi(\mathbf{z}_c|\mathbf{I}_m)$. While this low variance prior may be useful in estimating a single solution, sampling from it will lead to \emph{negligible diversity} in image completion results. When the CVAE variant of \cite{walker2016}, which assumes conditional prior $p_\phi(\mathbf{z}_c|\mathbf{I}_m)=\mathcal{N}(\mathbf{0},\mathbf{I})$, is used instead, the network learns to ignore the latent sampling and directly estimates $\mathbf{I}_c$ from $\mathbf{I}_m$ for a fixed ground truth, also resulting in similar solutions. A possible way to diversify the output is simply to not incentivize the output to reconstruct the instance-specific $\mathbf{I}_g$ during training, only needing it to fit in with the training set distribution as deemed by a learned adversarial discriminator (see Figure~\ref{fig:ch4_coarse_framework} ``Instance Blind''). However, this approach is unstable, especially for large and complex scenes \cite{song2018contextual}. A detail analysis is presented in Section~\ref{sec:ch4_ana_net}.

\paragraph{Latent Priors of Holes} In our approach, we require that missing partial images (patches), as a superset of full images, \emph{to also arise from a latent space distribution} \cite{shaham2019singan}, with a smooth prior of $p(\mathbf{z}_c)$. The variational lower bound is:
\begin{equation}\label{eq:ch4_VAE}
	\log p(\mathbf{I}_c)\ge -\text{KL}(q_\psi(\mathbf{z}_c|\mathbf{I}_c)||p(\mathbf{z}_c)) + \mathbb{E}_{q_\psi(\mathbf{z}_c|\mathbf{I}_c)}[\log p_\theta(\mathbf{I}_c|\mathbf{z}_c)]
\end{equation}
where in \cite{kingma2013auto} the prior is set as $p(\mathbf{z}_c)=\mathcal{N}(\mathbf{0},\mathbf{I})$. However, we can be more discerning when it comes to partial images (patches) since they have different numbers of pixels. In particular, \emph{a complement image $\mathbf{I}_c$ with more pixels (large holes for the masked image $\mathbf{I}_m$, as shown in the last column in Figure \ref{fig:ch4_degraded_image}) should have greater prior variance than a complement image $\mathbf{I}_c$ with fewer pixels (small holes)} and in fact a masked partial image $\mathbf{I}_m$ with no pixels missing should be completely deterministic! Hence we generalize the prior $p(\mathbf{z}_c)=\mathcal{N}_m(\mathbf{0},\sigma^2(n)\mathbf{I})$ to adapt to the number of  missing pixels $n$, where $\sigma^2(n)=\frac{n}{H\times W}\in$(0,1]. 

\paragraph{Prior-Conditional Coupling} Next, we combine the latent priors into the conditional lower bound of (\ref{eq:ch4_CVAE}). Since $\mathbf{z}_c$ represents the distributions of target missing partial image $\mathbf{I}_c$, $\mathbf{z}_c$ can be naturally inferred using the target missing image $\mathbf{I}_c$, that $q_\psi(\mathbf{z}_c|\mathbf{I}_c,\mathbf{I}_m)$ $\approx$ $q_\psi(\mathbf{z}_c|\mathbf{I}_c)$ when $\mathbf{I}_c$ is available in the training. Updating (\ref{eq:ch4_CVAE}):
\begin{equation}\label{eq:ch4_CVAE_with_prior}
	\log p(\mathbf{I}_c|\mathbf{I}_m)\ge -\text{KL}(q_\psi(\mathbf{z}_c|\mathbf{I}_c)||p_\phi(\mathbf{z}_c|\mathbf{I}_m)) + \mathbb{E}_{q_\psi(\mathbf{z}_c|\mathbf{I}_c)}[\log p_\theta(\mathbf{I}_c|\mathbf{z}_c,\mathbf{I}_m)].
\end{equation}
However, unlike in (\ref{eq:ch4_CVAE}), notice that $q_\psi(\mathbf{z}_c|\mathbf{I}_c)$ \emph{is no longer freely learned during training, yet is tied to its presence in (\ref{eq:ch4_VAE})}. Intuitively, the learning of $q_\psi(\mathbf{z}_c|\mathbf{I}_c)$ is regularized by the prior $p(\mathbf{z}_c)$ in (\ref{eq:ch4_VAE}), while the learning of the conditional prior $p_\phi(\mathbf{z}_c|\mathbf{I}_m)$ is in turn regularized by $q_\psi(\mathbf{z}_c|\mathbf{I}_c)$ in (\ref{eq:ch4_CVAE_with_prior}).

\paragraph{Reconstruction vs Creative Generation} One issue with (\ref{eq:ch4_CVAE_with_prior}) is that the sampling is taken from $q_\psi(\mathbf{z}_c|\mathbf{I}_c)$ during training, but is not available during testing, whereupon sampling must come from $p_\phi(\mathbf{z}_c|\mathbf{I}_m)$ which may not be adequately learned for this role. In order to mitigate this problem, we modify (\ref{eq:ch4_CVAE_with_prior}) to have a blend of formulations \emph{with and without importance sampling}.

As is typically the case for image completion, there is only one training instance of $\mathbf{I}_c$ for each unique $\mathbf{I}_m$. This means that for function $q_\psi(\mathbf{z}_c|\mathbf{I}_c,\mathbf{I}_m)$, $\mathbf{I}_c$ can be learned into the network as a hard-coded dependency of the input $\mathbf{I}_m$, so $q_{\psi}(\mathbf{z}_{c}|\mathbf{I}_{c},\mathbf{I}_{m})\cong\hat{q}_{\psi}(\mathbf{z}_{c}|\mathbf{I}_{m})$. Assuming that the network for $p_{\phi}(\mathbf{z}_{c}|\mathbf{I}_{m})$ has similar or higher modeling power and there are no other explicit constraints imposed on it, then in training $p_{\phi}(\mathbf{z}_{c}|\mathbf{I}_{m})\rightarrow\hat{q}_{\psi}(\mathbf{z}_{c}|\mathbf{I}_{m})$, and the KL divergence in (\ref{eq:ch4_CVAE}) goes to zero. Then we get the following function:
\begin{align}\label{eq:ch4_GSNN}
	\log p(\mathbf{I}_c|\mathbf{I}_m)\ge\mathbb{E}_{p_{\phi}(\mathbf{z}_{c}|\mathbf{I}_{m})}[\log p_\theta(\mathbf{I}_c|\mathbf{z}_c,\mathbf{I}_m)]
\end{align}
the ``GSNN'' version in \cite{sohn2015learning}. However, unlike \cite{sohn2015learning}, the variance $\sigma^2$ of the learned distribution $p_{\phi}(\mathbf{z}_{c}|\mathbf{I}_{m})$ in our method will not be zero as mentioned above. This $\mathbf{z}_c$ for missing regions is sampling from the visible regions $\mathbf{I}_m$, we call this \emph{without importance sampling}, contrary to the \emph{importance sampling} $q_\psi(\mathbf{z}_c|\mathbf{I}_c)$. Finally, we combine (\ref{eq:ch4_CVAE_with_prior}) and (\ref{eq:ch4_GSNN}) to obtain the reconstruction and creative generation function:
\begin{equation}\label{eq:ch4_mixed_models}
	\log p(\mathbf{I}_c|\mathbf{I}_m) \geq 
	\lambda \left\{ \mathbb{E}_{q_\psi}[\log p^r_\theta(\mathbf{I}_c|\mathbf{z}_c,\mathbf{I}_m)]
	-  \text{KL}(q_\psi || p_\phi) \right\} + (1-\lambda)\,  \mathbb{E}_{p_\phi}[
	\log p^g_\theta(\mathbf{I}_c|\mathbf{z}_c,\mathbf{I}_m)]
\end{equation}
where $\lambda\in$ [0,1] is implicitly set by training loss coefficients in Section~\ref{sec:ch4_training} (see details in Appendix \ref{AppendixA}). When sampling from the importance function $q_\psi(\mathbf{z}_c|\mathbf{I}_c)$, the missing instance information is available and we formulate the likelihood $p^r_\theta(\mathbf{I}_c|\mathbf{z}_c,\mathbf{I}_m)$ to be focused on \emph{reconstructing} $\mathbf{I}_c$. Conversely, when sampling from the learned distribution $p_\phi(\mathbf{z}_c|\mathbf{I}_m)$ which does not contain $\mathbf{I}_c$, we will facilitate \emph{creative generation} by having the likelihood model $p^g_\theta(\mathbf{I}_c|\mathbf{z}_c,\mathbf{I}_m) \cong \ell^g_\theta(\mathbf{z}_c,\mathbf{I}_m)$ be \emph{independent of the original instance} of $\mathbf{I}_c$. Instead it only \emph{encourages generated samples to fit in with the overall training distribution.} 

\paragraph{Joint Unconditional and Conditional Variational Lower Bounds} Our overall training objective may then be expressed as jointly maximizing the lower bounds in (\ref{eq:ch4_VAE}) and (\ref{eq:ch4_mixed_models}). This can be done by unifying the likelihood in (\ref{eq:ch4_VAE}) to that in (\ref{eq:ch4_mixed_models}) as $p_\theta(\mathbf{I}_c|\mathbf{z}_c)\cong p^r_\theta(\mathbf{I}_c|\mathbf{z}_c,\mathbf{I}_m)$, in which the $\mathbf{z}_c$ is sampling from the \emph{important sampling} $q_\psi(\mathbf{z}_c|\mathbf{I}_c)$ that can be used for rebuild the original missing regions $\mathbf{I}_c$. We can then define a combine function as our maximization goal:
\begin{align}\label{eq:ch4_total_bound}
	\mathcal{B}=& \beta \, \mathcal{B}_{1}+\mathcal{B}_{2}\nonumber\\
	=& -\left[\beta\mathrm{KL}(q_{\psi}||p_{z_{c}})+\lambda\mathrm{KL}(q_{\psi}||p_{\phi})\right] + (\beta+\lambda)\mathbb{E}_{q_{\psi}}\log p_{\theta}^{r}+(1-\lambda)\mathbb{E}_{p_{\phi}}\log p_{\theta}^{g}
\end{align}
where $\mathcal{B}_{1}$ is the lower bound related to the unconditional log likelihood of missing partial image $\mathbf{I}_c$, and $\mathcal{B}_{2}$ relates to the log likelihood of missing regions $\mathbf{I}_c$ conditioned on $\mathbf{I}_m$. Note that this function holds a key different with hybrid objective function in \cite{sohn2015learning} that \emph{the conditional prior network $p_\phi(\mathbf{z}_c|\mathbf{I}_m)$ and the recognition network $q_\psi(\mathbf{z}_c|\mathbf{I}_c)$ are no longer freely learned, but are constrained by a mask related prior} $p(\mathbf{z}_c)=\mathcal{N}_m(\mathbf{0},\sigma^2(n)\mathbf{I})$. Furthermore, our \emph{without importance sampling}, also the testing sampling, does not learn to predict a fixed instance during the training, which encourages larger diversity.

\begin{figure}[tb!]
	\centering
	\includegraphics[width=\textwidth]{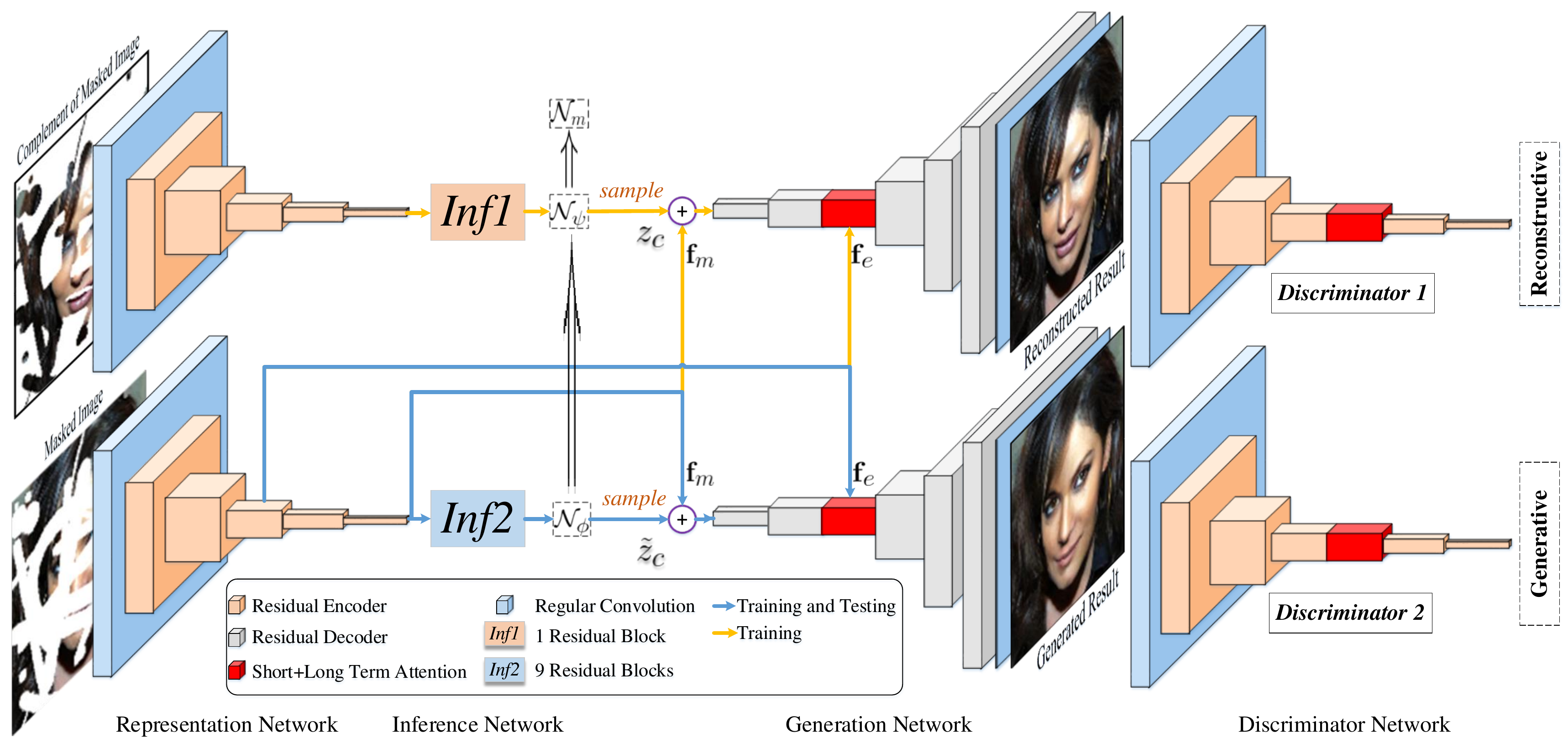}
	\caption[Overview of the proposed architecture ]{\textbf{Overview of our architecture with two parallel pipelines.} The top \textbf{reconstructive} pipeline (yellow line) combines information from $\mathbf{I}_m$ and $\mathbf{I}_c$, which is used only for training. The bottom \textbf{generative} pipeline (blue line) infers the conditional distribution of hidden regions, that can be sampled during testing. The two representation networks and generation networks in top and bottom share identical weights.}
	\label{fig:ch4_framework}	
\end{figure}

\subsubsection{Network Structure and Training Loss}\label{sec:ch4_training}

The formula in (\ref{eq:ch4_total_bound}) is implemented as our dual pipeline, illustrated in Figure~\ref{fig:ch4_framework}. This consists of representation, inference, generation, and auxiliary discriminator networks in two paths. The upper pipeline is the reconstruction path used in training that corresponds to the lower bound $\mathcal{B}_{1}$, in which $\textbf{z}_c$ contains information of missing image $\textbf{I}_c$. Hence when combined with the conditional feature $\textbf{f}_m$, we can easily train this path to rebuild the original image $\textbf{I}_g$. In contrast, the lower path, used in both training and testing, is responsible for the lower bound $\mathcal{B}_{2}$, where the missing information is inferred only from the masked image $\textbf{I}_m$, resulting in a less restrictive prediction. 

We transfer the lower bound terms in (\ref{eq:ch4_total_bound}) as the corresponding loss function. During training, jointly maximizing the lower bounds is then minimizing a total loss $\mathcal{L}$, which consists of three groups of component losses:
\begin{equation}
\begin{split}
	\mathcal{L} = & \alpha_\text{KL}(\mathcal{L}_\text{KL}^r + \mathcal{L}_\text{KL}^g) + \alpha_\text{app}(\mathcal{L}_\text{app}^r + \mathcal{L}_\text{app}^g) + \alpha_\text{ad}(\mathcal{L}_\text{ad}^r + \mathcal{L}_\text{ad}^g)
	\label{eq:ch4_total_loss}
\end{split}
\end{equation}
where the $\mathcal{L}_\text{KL}$ group regularizes consistency between pairs of distributions in terms of KL divergences, the $\mathcal{L}_\text{app}$ group encourages appearance matching fidelity, and the $\mathcal{L}_\text{ad}$ group forces sampled images to fit in with the training set distribution. Each of the groups has a separate term for the reconstructive and generative paths. 

\paragraph{Distributive Regularization} The typical interpretation of the KL divergence term in a VAE is that it regularizes the learned importance sampling function $q_\psi(\mathbf{z}_c|\mathbf{I}_c)$ to a latent prior $p(\mathbf{z}_c)$. Defining both as Gaussians, we get:
\begin{equation}
	\mathcal{L}_\text{KL}^{r,(i)} = \text{KL}(q_\psi(\mathbf{z}|I_c^{(i)})||\mathcal{N}_m(\mathbf{0},\sigma^{2,(i)}(n)\mathbf{I})).
\end{equation} 

For the generative path, the appropriate interpretation is \emph{reversed}: the learned conditional prior  $p_\phi(\mathbf{z}_c|\mathbf{I}_m)$, also a Gaussian, is regularized to $q_\psi(\mathbf{z}_c|\mathbf{I}_c)$.
\begin{equation}
	\mathcal{L}_\text{KL}^{g,(i)} = \text{KL}(q_\psi(\mathbf{z}|I_c^{(i)}))||p_\phi(\mathbf{z}|I_m^{(i)}))).
\end{equation}
Note that the conditional prior uses $\mathbf{I}_m$, while the importance function has access to the missing regions $\mathbf{I}_c$.

\paragraph{Appearance Matching Loss} The likelihood term $p^r_\theta(\mathbf{I}_c|\mathbf{z}_c)$ is interpreted as probabilistically encouraging appearance matching to the missing regions $\mathbf{I}_c$. However, our framework also auto-encodes the masked image $\mathbf{I}_m$ (via $\mathbf{f}_m$) deterministically, and the loss function needs to cater for this reconstruction. As such, the per-instance loss here is:
\begin{equation}
	\mathcal{L}_\text{app}^{r, (i)} = ||I_\text{rec}^{(i)}-I_g^{(i)}||_1
	\label{eq:ch4_app_loss_rec}
\end{equation} 
where $I_\text{rec}^{(i)}$=$G(z_c,f_m)$ and $I_g^{(i)}$ are the reconstructed and original full images, respectively. The purpose of this loss is to bias the representation towards the actual visible information.
In contrast, for the generative path, the latent distribution $\mathcal{N}_\phi$ of the missing regions $\mathbf{I}_c$ is inferred based only on the visible $\mathbf{I}_m$. This would be significantly less accurate than the inference in the upper path. Thus, we ignore instance-specific appearance matching for $\mathbf{I}_c$, and only focus on reconstructing $\mathbf{I}_m$:
\begin{equation}
	\mathcal{L}_\text{app}^{g, (i)} = ||M*(I_\text{gen}^{(i)}-I_g^{(i)})||_1
	\label{eq:ch4_app_loss_gen}
\end{equation}
where $I_\text{gen}^{(i)}$=$G(\tilde{z}_c,f_m)$ is the generated image, and $M$ is the binary mask selecting visible pixels. 

\paragraph{Adversarial Loss} The formulation of $p^r_\theta(\mathbf{I}_c|\mathbf{z}_c,\mathbf{I}_m)$ and the instance-blind $p^g_\theta(\mathbf{I}_c|\mathbf{\tilde z}_c,\mathbf{I}_m)$ also incorporates the use of adversarially learned discriminators $D_1$ and $D_2$ to judge whether the generated images fit into the training set distribution. Inspired by \cite{bao2017cvae}, we use a mean feature match loss in the reconstructive path for the generator,
\begin{equation}
	\mathcal{L}_\text{ad}^{r, (i)} = ||f_{D_{1}}(I_\text{rec}^{(i)}) - f_{D_{1}}(I_g^{(i)})||_2
	\label{eq:ch4_ad_loss_rec}
\end{equation}
where $f_{D_{1}}(\cdot)$ is the feature output of the final layer of $D_1$. This encourages the original and reconstructed features in the discriminator to be close together. Conversely, the adversarial loss in the generative path for the generator is:
\begin{equation}
	\mathcal{L}_\text{ad}^{g, (i)} = [D_2(I_\text{gen}^{(i)})-1]^2.
	\label{eq:ch4_ad_loss_gen}
\end{equation}
This is based on the generator loss in LSGAN \cite{mao2017least}, which performs better than the original GAN loss \cite{goodfellow2014generative} in our scenario. The discriminator loss for both $D_1$ and $D_2$ is also based on LSGAN.

\begin{table}[tb!]
	\centering
	\footnotesize
	\renewcommand{\arraystretch}{1.2}
	\setlength\tabcolsep{6pt}
	\begin{tabular}{@{}lccccccc@{}}
		\hlineB{3.5}
		& \multicolumn{2}{c}{Diversity (LPIPS)} && \multicolumn{4}{c}{Image Quality ($\mathbf{I}_{out}$)}\\
		\cline{2-3}\cline{5-8}
		& $\mathbf{I}_{out} \uparrow$ & $\mathbf{I}_{out(m)} \uparrow$ && $\ell_1$ loss $\downarrow$ & SSIM $\uparrow$ & PSNR $\uparrow$  & FID $\downarrow$\\
		\hlineB{2.5}
		CA~\cite{yu2018generative} & - & - && 0.031 & 0.820 & 23.57 & 9.53 \\
		EC~\cite{Nazeri_2019_ICCV} & - & - && 0.030 & 0.819 & 23.47 & 8.01 \\
		MEDFE~\cite{Liu2019MEDFE} & - & - && 0.028 & 0.830 & 24.38 & 7.85 \\
		\cdashline{1-8}
		CVAE~\cite{sohn2015learning} & 0.004 & 0.014&&  0.023 & 0.847 & 24.02 & 9.96\\
		Instance Blind & 0.015& 0.049 &&  0.025 & 0.852 & 23.77 & 9.48 \\
		BicycleGAN~\cite{zhu2017toward}& 0.020 & 0.060&& 0.026 & 0.845 & 23.71 & 11.56 \\
		PICNet & {\bf 0.024} & {\bf 0.071}&& {\bf 0.021 } & {\bf 0.867} & {\bf 24.69} & {\bf 6.43}\\
		\hlineB{2.5}
	\end{tabular}
	\caption[Quantitative comparisons of different network structures]{\textbf{Quantitative comparisons of different network structures} on CelebA-HQ testing set \cite{liu2015deep,karras2017progressive} with center masks. $\downarrow$ = lower is better, $\uparrow$ = higher is better. $\mathbf{I}_{out}$ is the completed output image and $\mathbf{I}_{out(m)}=(1-M)\times\mathbf{I}_{out}$ is extracted for the missing regions.}
	\label{tab:ch4_diversity_comparisons}
\end{table}

\begin{figure}[tb!]
	\centering
	\includegraphics[width=0.78\linewidth]{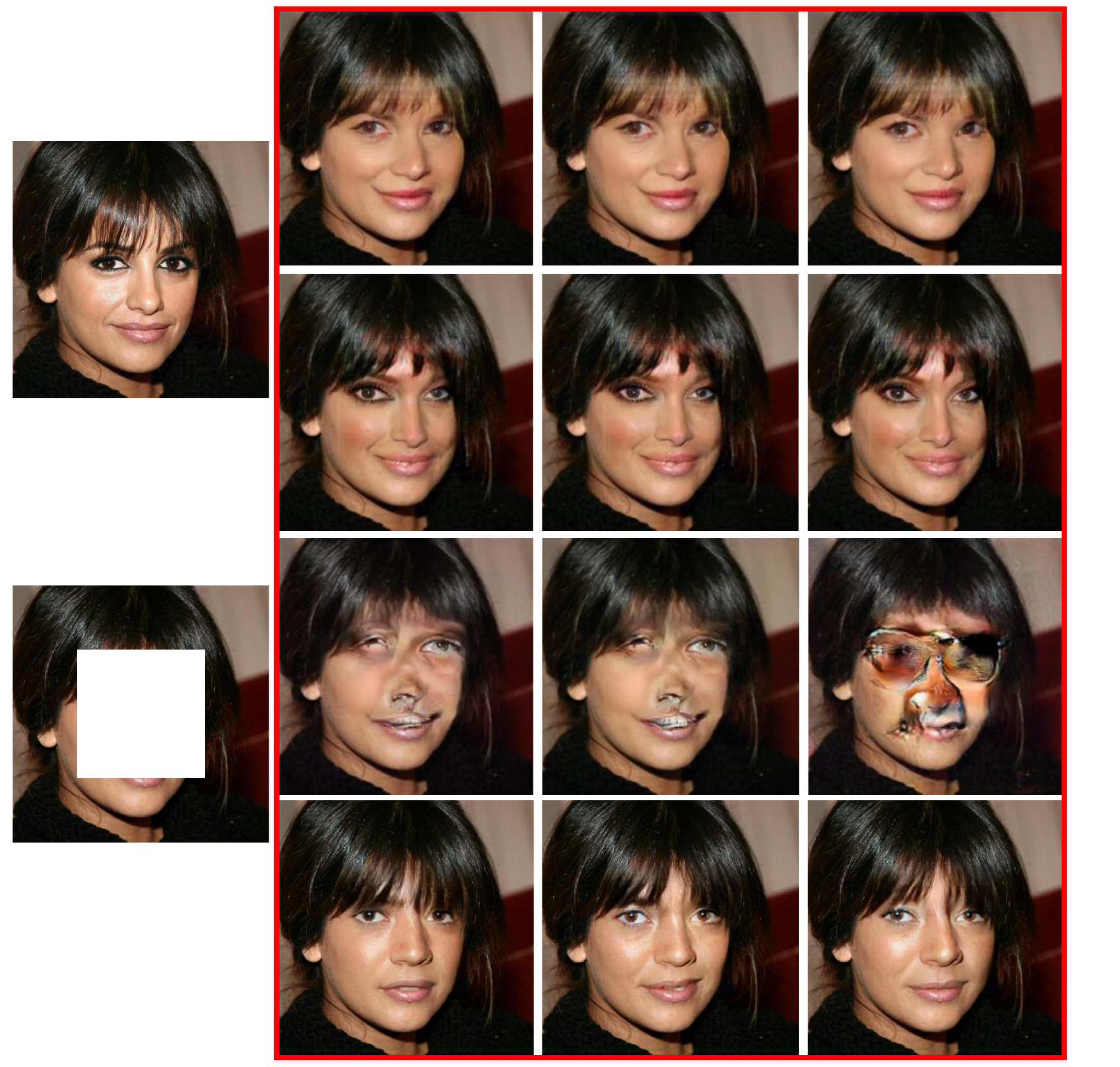}
	\begin{picture}(0,0)
	    \put(-15, 248){\rotatebox{90}{\footnotesize CVAE \cite{walker2016}}}
	    \put(-15, 164){\rotatebox{90}{\footnotesize Instance-Blind}}
	    \put(-15, 78){\rotatebox{90}{\footnotesize BicycleGAN \cite{zhu2017toward}}}
	    \put(-15, 24){\rotatebox{90}{\footnotesize PICNet}}
	    \put(-310,185){\footnotesize Original}
	    \put(-306,55){\footnotesize Input}
	\end{picture}
	\vspace{-0.2cm}
	\caption[Qualitative comparison results of different training strategies]{\textbf{Qualitative comparison results of different training strategies.} \textbf{First Column:} Original and Masked image. \textbf{Others:} the completed results of different methods. Our method provides diverse results, \ie different hairstyles and mouth expressions, with realistic appearance.} 
	\label{fig:ch4_ablation_paths}	
\end{figure}
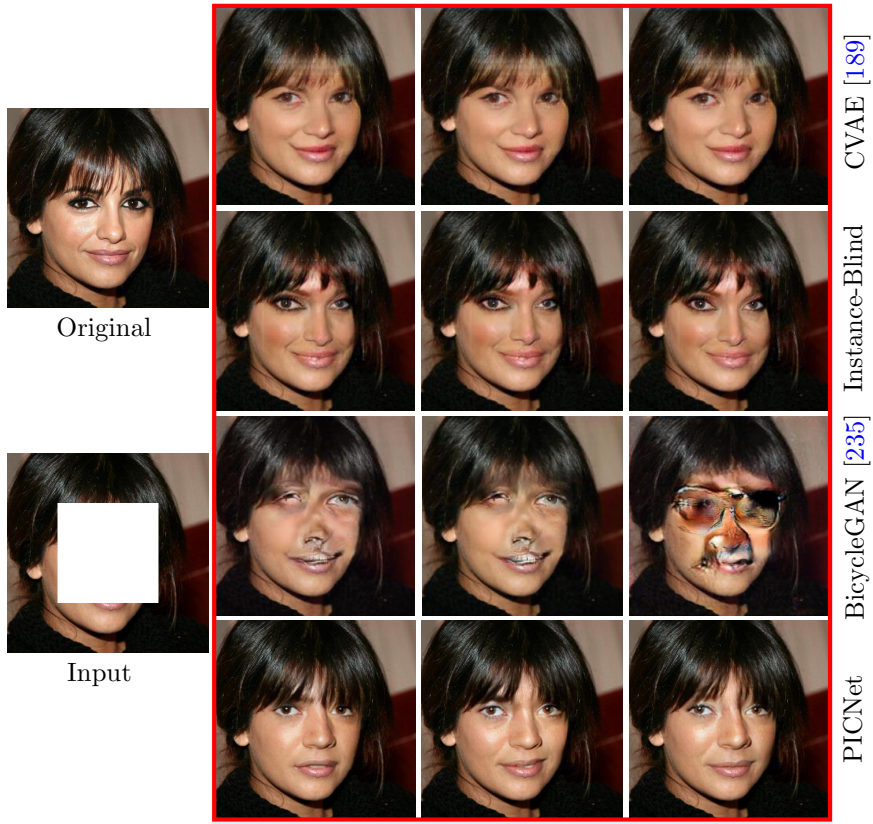

\subsubsection{Analysis}\label{sec:ch4_ana_net}

\paragraph{Effect of Network Structure} We first investigated the influence of using our two-path training structure in comparison to other variants such as the CVAE of~\cite{walker2016} and the ``Instance Blind'' structures in Figure~\ref{fig:ch4_coarse_framework}. We also trained the state-of-the-art multi-model BicycleGAN~\cite{zhu2017toward} on Celeba-HQ dataset \cite{liu2015deep,karras2017progressive} by setting $\mathbf{A} = \mathbf{I}_m$, $\mathbf{B} = \mathbf{I}_c$ with center mask. 

We first computed the diversity score using the Learned Perceptual Image Patch Similarity (LPIPS) metric reported in \cite{zhu2017toward}. LPIPS metric~\cite{zhang2018unreasonable} calculates the average distance of samples in a deep feature domain. For each random pairs, a pre-trained deep network (\eg VGG \cite{vgg}) is used to extract the features of images. Then, the distance of two vectors is calculated using $\ell_1$ distance. The larger distance indicates the results are much more diverse, as the generated pairs far from each other. For each method, we sampled 50K pairs of randomly generated images from 1K center masked images. $\mathbf{I}_{out}$ and $\mathbf{I}_{out(m)}$ are the full output and the masked-regions' output, respectively. Furthermore, we used the popular Fr\'echet Inception Distance (FID) \cite{heusel2017gans} to assess the visual quality of completed images by comparing the distance between distributions of completed and real images in a deep feature domain. As for the traditional pixel-level and patch-level image quality metrics, including the mean $\ell_1$ loss, structural similarity (SSIM), and peak signal-to-noise ratio (PSNR), we select the closest generated image to the ground truth image for calculation, as these metrics are based on one-to-one pairing.

Table~\ref{tab:ch4_diversity_comparisons} shows diversity and image quality analysis for different network structures. We note that our method not only improved the image quality significantly (relative 18\% improvement for FID), but also generated multiple and diverse completion results. Here, BicycleGAN obtained relatively higher diversity scores than our baseline framework by using cycle loss instead of reconstruction loss. However, the completed images are of low quality (as shown in Figure~\ref{fig:ch4_ablation_paths}), which suggests that despite increased diversity, its network structure is not directly suitable for image completion. 

Figure~\ref{fig:ch4_ablation_paths} shows some sampled examples of each structure. We observe that CVAE~\cite{walker2016} obtains reasonable results, yet with little variation. The framework has likely learned to ignore the sampling and predicted a deterministic outcome as it always tries to rebuild the original ground truth during the training no matter what masks are used to degrade the input. As for ``Instance Blind'', If we enforced the generated image back to the original ``ground truth'' $I_g$, the experience will be similar to the CVAE~\cite{walker2016}. The visual results of BicycleGAN are much worse than other methods. In their model, the latent code $\mathbf{z}$ to the encoder is replicated from $1\times 1\times Z$ to $ H\times W\times Z$, where the different spatial position holds the same random value that does not represent any semantic meaning. On the contrary, our latent code $\mathbf{z}$ is inferred from the visible pixels during the testing, which includes the predicted semantic information from the visible pixels.

\subsection{Short+Long Term Patch Attention}\label{sec:ch4_attention}

\begin{figure}[tb!]
	\centering
	\includegraphics[width=0.8\linewidth]{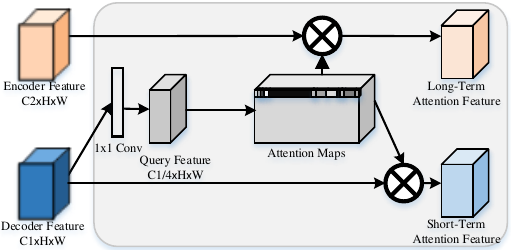}
	\caption[Short + Long Term Attention Layer]{\textbf{Our short+long term patch attention layer}. The attention map is directly computed on the decoder features to estimate the content similarity  in the same domain. After obtaining the self-attention scores, we use these to compute self-attention on decoder features, as well as contextual flow on encoder features.}
	\label{fig:ch4_attention}
\end{figure}

A weakness of purely convolutional operations is that they have limited spatial ranges, and cannot efficiently exploit distant correlation. Extending beyond the Self-Attention in SAGAN \cite{zhang2019self}, we propose a novel short + long term patch attention layer that not only to use the self-attention within a decoder layer to harness \emph{distant spatial context}, but also to further capture \emph{feature-feature context} between encoder and decoder layers. Our \emph{key novel insight} is: doing so would allow the network a choice of attending to the finer-grained visible features in the encoder or the more semantically generative features in the decoder, depending on circumstances. Our proposed structure is shown in Figure~\ref{fig:ch4_attention}.

\subsubsection{Self-Patch-Attention Map} 

Feature attention has been widely used in image completion task \cite{yu2018generative,Yan_2018_ECCV,song2018contextual,yi2020contextual}. They calculate the attention map by comparing low-frequency decoder features of holes and high-frequency encoder feature of visible regions. Then, the high-frequency features are copied from visible regions to the missing holes based on the similarity score. However, this is a little contradictory as \emph{feature transfer is unnecessary when two features are very similar, but when needed the features are too difficult to be matched easily}. 

To address this, we calculate the content similarity in itself feature domain, the decoder feature. Our attention map calculates the response at a position in a sequence by paying attention to other position in the \emph{same sequence}. Given the features $\mathbf{f}_d$ from the previous decoder layer, we first calculate the point attention score of:
\begin{equation}
	\textbf{A}_{j,i} = \frac{\exp(s_{i,j})}{\sum_{i=1}^{N}\exp(s_{i,j})},\mbox{where } s_{i,j}=\theta(f_{di})^\top\theta(f_{dj}),
\end{equation}
where $\textbf{A}_{j,i}$ represents the similarity of $i^{th}$ location to the $j^{th}$ location. $N=H\times W$ is the number of pixels, while $\theta$ is a 1x1 convolution filter for refining the feature.

Inspired by PatchMatch~\cite{barnes2009patchmatch}, we further ensure the consistency of attention maps by fusing the similarity score in a square patch:
\begin{equation}
	{\bf\hat A}_{j,i} = \sum_{{j}'\in U_j,{i}'\in U_i}\textbf{A}_{{j}',{i}'}
\end{equation}
where $U_j$ and $U_i$ are the neighborhood patch sets at $j^{th}$ and $i^{th}$ locations separately. We fixed the square size as $3\times3$ throughout this chapter. 

\subsubsection{Short-Term Attention from Decoder Full Regions}
After we obtain the attention map, the non-local information is fused in the decoder features. This leads to the short-term intra-layer attention feature (\textbf{Short-Term Attention} in Figure~\ref{fig:ch4_attention}) and the output $\mathbf{y}_d$:
\begin{equation}
	c_{dj} = \sum_{i=1}^{N}{\bf\hat A}_{j,i}f_{di}
	\;, \hspace{0.5cm}
	\mathbf{y}_d= \gamma_d\mathbf{c}_d+\mathbf{f}_d
\end{equation}
where, we use a scale parameter $\gamma_d$ to balance the weights between attention feature $\mathbf{c}_d$ and decoder feature $\mathbf{f}_d$. The initial value of $\gamma_d$ is set to zero. 

\subsubsection{Long-Term Attention from Encoder Visible Regions}

In addition, specifically for image completion task, we not only need the high-quality results for missing holes, but also need to ensure the appearance consistency of the generated patches of missing parts and the original patches of visible parts. Then, we introduce a long-term inter-layer attention feature (\textbf{Long-Term Attention} in Figure~\ref{fig:ch4_attention}), in which the response attends to visible encoded features $\mathbf{f}_e$. Therefore, the output $\mathbf{y}_e$ is given by:
\begin{equation}
	c_{ej}=\sum_{i=1}^{N}{\bf\hat A}_{j,i}f_{ei}
	\;, \hspace{0.5cm}
	\mathbf{y}_e= \gamma_e(1-M)\mathbf{c}_e+M\mathbf{f}_e.
\end{equation}
As before, a scale parameter $\gamma_e$ is used to combine the encoder feature $\mathbf{f}_e$ and the attention feature $\mathbf{c}_e$. However, unlike the decoder feature $\mathbf{f}_d$ which has information for generating a full image, the encoder feature $\mathbf{f}_e$ only represents visible parts $\mathbf{I}_m$. Hence, a binary mask $M$ (1 denotes visible regions, and 0 represents the holes) is used. In this way, the high-quality visible features are flowed to the holes based on the content similarity. Finally, both the short- and long-term attention features are aggregated and fed into further decoder layers. 

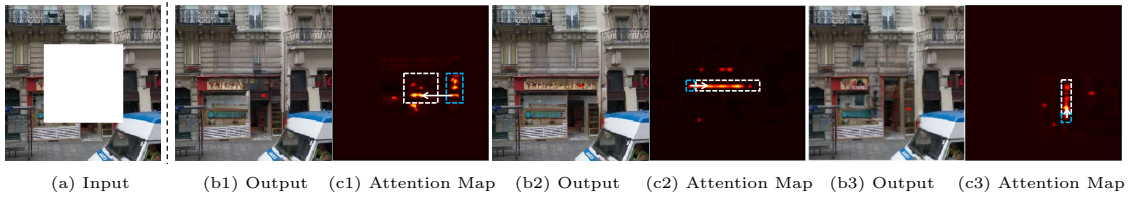
\begin{figure}[tb!]
	\centering
	\includegraphics[width=\linewidth]{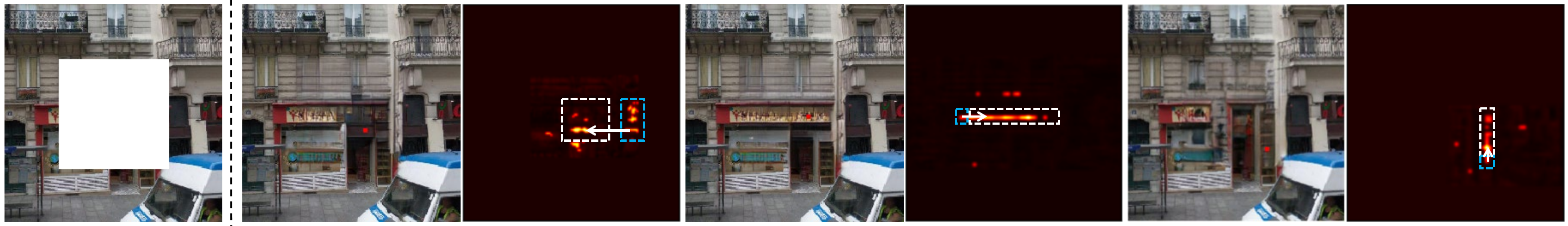}
	\begin{picture}(0,0)
		\put(-190,5){\tiny (a) Input}
		\put(-133,5){\tiny (b1) Output}
		\put(-86,5){\tiny (c1) Attention Map}
		\put(-16,5){\tiny (b2) Output}
		\put(33,5){\tiny (c2) Attention Map}
		\put(102,5){\tiny (b3) Output}
		\put(150,5){\tiny (c3) Attention Map}
	\end{picture}
	\vspace{-0.3cm}
	\caption[Texture flow for diversely generated contents]{\textbf{Texture flow (white arrow) for diversely generated contents} with the same mask. (a) Masked input image. (b*) Multiple and diverse results as well as one query point (red dot). (c*) The corresponding attention maps (unsampled to original image size for visualization) for the query points in the output. The high-quality textures are copied from different visible regions (blue rectangles) to the generated regions (white rectangles), depending on what content has been generated. }
	\label{fig:ch4_ablation_attention1}	
\end{figure}

\begin{figure}[tb!]
	\centering
	\includegraphics[width=\linewidth]{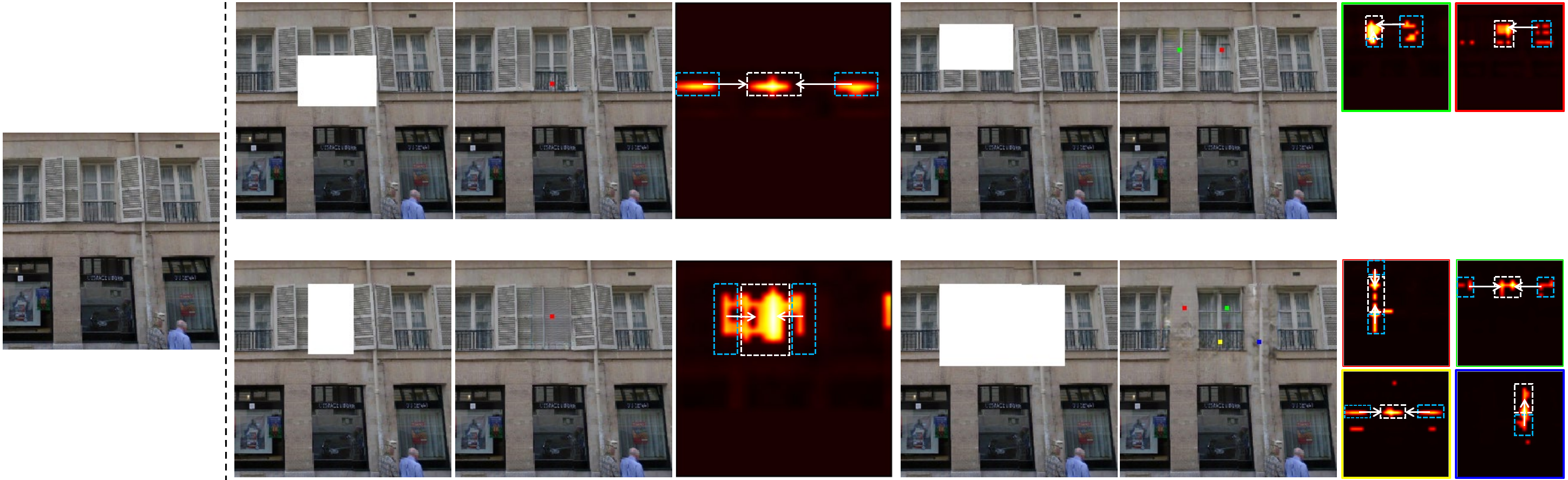}
	\begin{picture}(0,0)
		\put(-202,42){\scriptsize (a) Original}
		
		\put(-136,76){\scriptsize (b1) Input}
		\put(-83,76){\scriptsize (c1) Output}
		\put(-34,76){\scriptsize (d1) Attention Map}
		\put(44,76){\scriptsize (b2) Input}
		\put(100,76){\scriptsize (c2) Output}
		\put(148,76){\scriptsize (d2) Attention Map}
		
		\put(-136,3){\scriptsize (b3) Input}
		\put(-83,3){\scriptsize (c3) Output}
		\put(-34,3){\scriptsize (d3) Attention Map}
		\put(44,3){\scriptsize (b4) Input}
		\put(100,3){\scriptsize (c4) Output}
		\put(148,3){\scriptsize (d4) Attention Map}
	\end{picture}
	\caption[Texture flow for different masked regions]{\textbf{Texture flow (white arrow) for different masked regions.} (a) Original image. (b*) Masked input images with different degraded regions. (c*) The completed results as well as query points (denoted by color dots). (d*) The corresponding attention maps for the query points in the output. The results attend to different visible regions (blue rectangles) based on the different visible content.}
	\label{fig:ch4_ablation_attention2}	
\end{figure}
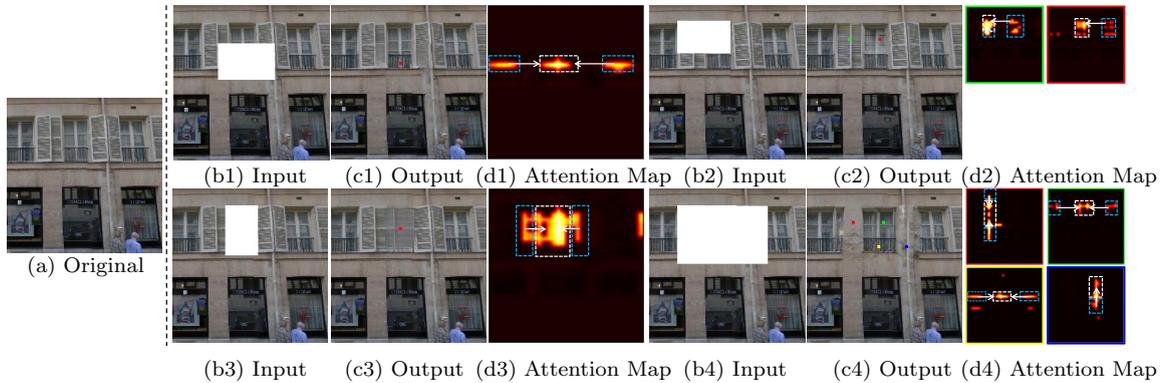

\subsubsection{Analysis}

Readers may wonder why the proposed short-long term attention layer would achieve better performance than existing contextual attention layers \cite{yu2018generative,yi2020contextual}. Here, we show that the proposed module is able to exploit non-local information from \emph{both} visible and generated regions for the holes, instead of purely copying high-frequency information from visible regions. 

In Figures~\ref{fig:ch4_ablation_attention1} and~\ref{fig:ch4_ablation_attention2}, completed results, along with corresponding attention maps for query points, are presented. Here, only points with the highest attention scores are highlighted. We use white arrows to explicitly show the texture flow, or how the attention layer copies information from high-quality visible features (blue rectangles) to the originally masked regions (white rectangles). In Figure~\ref{fig:ch4_ablation_attention1}, we find that the proposed attention layer attends to different visible regions for differently generated content, as sampled from our model. In this way, the model ensures appearance consistency between the diversely generated appearance and the visible pixels. Figure~\ref{fig:ch4_ablation_attention2} shows other examples of texture flow from visible regions to masked regions. When we mask different regions of the window, the proposed attention layer learns to copy high-quality pixels from corresponding visible regions (blue rectangles) to the missing holes (white rectangles). 

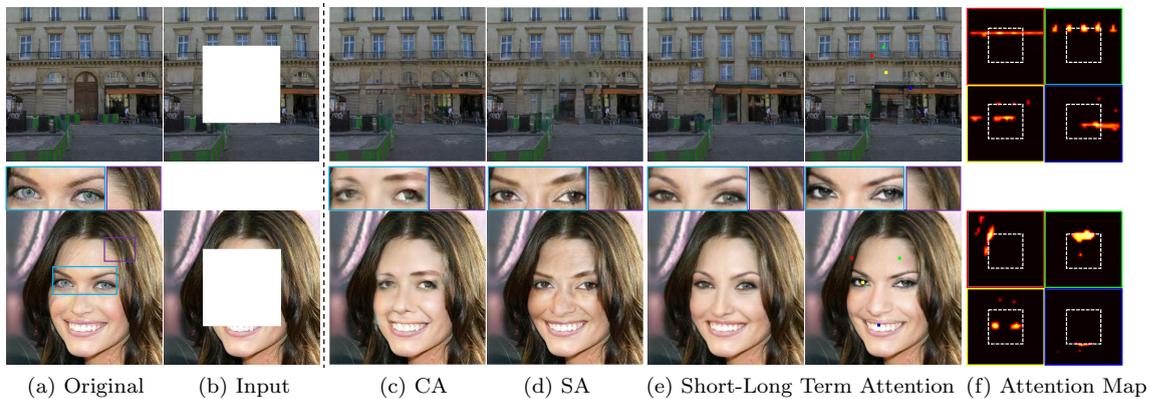
\begin{figure}[tb!]
	\centering
	\includegraphics[width=\linewidth]{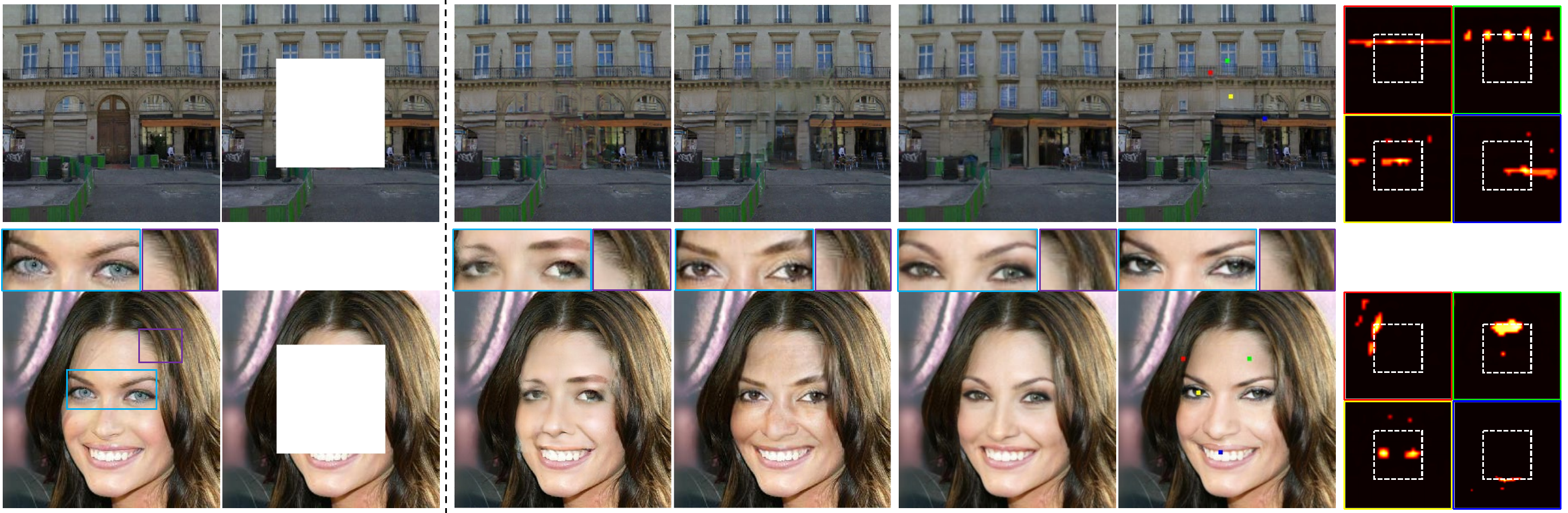}
	\begin{picture}(0,0)
		\put(-200,5){\scriptsize (a) Original}
		\put(-136,5){\scriptsize (b) Input}
		\put(-68,5){\scriptsize (c) CA}
		\put(-14,5){\scriptsize (d) SA}
		\put(32,5){\scriptsize (e) Short-Long Term Attention}
		\put(152,5){\scriptsize (f) Attention Map}
	\end{picture}
	\vspace{-0.2cm}
	\caption[Comparison of various attention modules]{\textbf{Comparison of various attention modules.} (a) Original Image. (b) Masked input image. (c) Results of contextual attention \cite{yu2018generative}. (d) Results of self-attention \cite{zhang2019self}. (e) Multiple results of our  method with short-long term patch attention. (f) The corresponding attention maps for the query points, \eg hair (red), skin (green), eye (yellow) and teeth (blue) on the face.}
	\label{fig:ch4_ablation_attention3}	
\end{figure}

We also compare the proposed attention layer to previous methods, including contextual attention (\textbf{CA}) \cite{yu2018generative} and self-attention (\textbf{SA}) \cite{zhang2019self} for image completion. As shown in Figure~\ref{fig:ch4_ablation_attention3}, our proposed attention layer borrows features from different positions, rather than directly copying similar features from one visible position like CA. In the building scene, CA's result is of similar high quality to our method, due to the presence of repeated structures. However, in the case of faces, if the mask regions are large, both CA and SA are unable to generate high-quality results. It is worth mentioning that CA can copy high-quality pixels for skin (purple rectangle) from the visible skin, yet obtaining unrealistic eyes (blue rectangle). This is because when two eyes are masked, they cannot copy non-local similar patches from other visible parts. Conversely, SA only copies features in the decoder network, ignoring high-quality visible features. While it generates plausible appearances for skin and eyes, the generated skin is inconsistent to the visible skin. Our attention module is able to utilize both decoder features (which do not have masked parts) and encoder features appropriately. In completing the left eye, information is distantly shared from the decoded right eye. When it comes to completing a point in a masked hair region, it will focus on encoded features from visible hairs.

\section{User Interface}\label{ch4:user}

We designed a real-time interactive system\footnote{The local version is available on \href{https://github.com/lyndonzheng/Pluralistic-Inpainting}{https://github.com/lyndonzheng/Pluralistic-Inpainting}, while the online real-time system is available on \href{http://www.chuanxiaz.com/project/pluralistic/}{http://www.chuanxiaz.com/project/pluralistic/}.} that allows the user to easily explore and edit the image by creating free-form or regular masks.

\begin{figure}
    \centering
    \includegraphics[width=\linewidth,height=0.3\textheight]{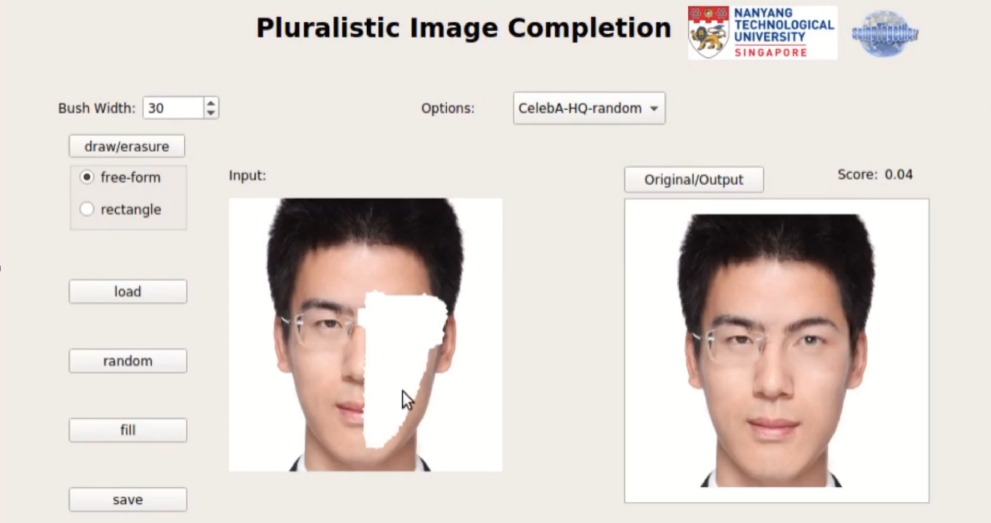}
    \caption[Local interface for image editing]{\textbf{Local interface for free-form image editing.} We produce a local interface on the \href{https://github.com/lyndonzheng/Pluralistic-Inpainting}{GitHub}.}
    \label{fig:ch4_local_interface}
\end{figure}

\begin{figure}
    \centering
    \includegraphics[width=\linewidth]{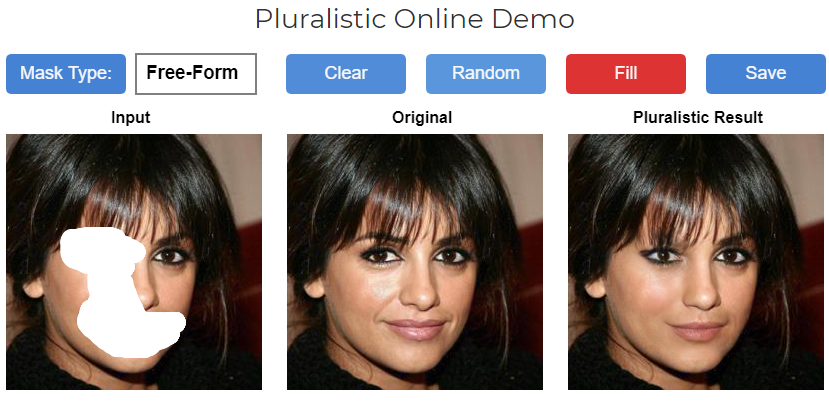}
    \caption[Online interface for image editing]{\textbf{Online interface for free-form image editing.} We produce an online interface on the \href{http://www.chuanxiaz.com/project/pluralistic/}{Website} that can be used to edit the image for diverse outputs.}
    \label{fig:ch4_online_interface}
\end{figure}

As shown in Figures \ref{fig:ch4_local_interface} and \ref{fig:ch4_online_interface}, the interface is composed of a button (``Random'') to load in an input image, a button (``Mask Type'') to select the mask type (\emph{free-from} or \emph{regular}), a button (``Fill'') to fill into reasonable content as well as visually realistic appearance. After the user makes an edit, the interface delivers the corresponding output. If the user hits the ``Fill'' button many times, it will randomly output a different result each time. Some of our results are presented in Section \ref{ch4:experiment} by using this user interface.

\section{Results and Applications}\label{ch4:experiment}

\subsection{Experimental Details}\label{sec:ch4_implementation}

\paragraph{Datasets} We evaluated the proposed \textbf{PICNet} with arbitrary mask types on various datasets, including Paris \cite{doersch2012makes}, CelebA-HQ \cite{liu2015deep,karras2017progressive}, ImageNet \cite{russakovsky2015imagenet} and Places2 \cite{zhou2018places}. Here, we only train one model to evaluate both the general free-form irregular masks and the center regular mask.

\paragraph{Metrics} Quantitative evaluation is tricky for the pluralistic image completion task, as our goal is to get diverse but reasonable solutions for a given masked image. The original image is only one solution of many, and comparisons should not be made only based on this image. Therefore, we first used the Fr\'echet Inception Distance (FID) \cite{heusel2017gans} and Inception Score (IS) \cite{salimans2016improved} to assess the quality of the completed image, as they are measured on learned features over the whole test set. Following \cite{Liu_2018_ECCV,Nazeri_2019_ICCV}, we then reported the traditional pixel- and patch-level image quality metrics, including $\ell_1$ loss, structure similarity index (SSIM) and peak signal-to-noise ratio (PSNR). We additionally compared the visual realism of all results using human judgment, as previously proposed \cite{zhang2016colorful} and widely adopted for image generation \cite{isola2017image,zhu2017unpaired,zhu2017toward,park2019semantic,Nazeri_2019_ICCV}. 

\paragraph{Training}  PICNet is implemented in PyTorch v1.4. The missing regions take value 0 in the input. We highlight the missing regions as white in the figures only for visual purposes. Each mini-batch has 16 images per NVIDIA V100 GPU and each input has 1 reconstructive and 1 generative output. For the binary masks, we used randomly regular and irregular holes. However, allowing unrestricted mask sizes is more difficult than keeping to center masks as in our prior work~\cite{zheng2019pluralistic}. In order to train the networks to convergence, two training steps were used: first, the completion network was trained using only the losses for the top reconstructive path, which has full information from both visible and missing regions.  To do this, we estimated the missing regions' distributions that relate to different mask sizes. After we obtained the distribution of missing regions through the reconstructive path, the bottom generative path was trained to infer the distribution of missing holes based on the visible parts, from which we can generate multiple results. 

\paragraph{Inference} At test time, only the bottom generation path will be applied to generate \emph{multiple} and \emph{diverse} results based on the visible information. We sampled 50 images for each masked input image ${\bf I}_\text{m}$. Note that the distribution we sampled from is also learned from the visible regions, rather than a fixed distribution used in previous works~\cite{sohn2015learning,walker2016}. The visual results were automatically selected based on the higher discriminator scores.

\subsection{Comparison with Existing Work}\label{sec:ch4_comp_results}

We mainly compare our method with 6 methods:

\begin{itemize}
	\item \textbf{PM:} PatchMatch \cite{barnes2009patchmatch}, the state-of-the-art non-learning based approach.
	\item \textbf{CE\footnote{\href{https://github.com/pathak22/context-encoder}{https://github.com/pathak22/context-encoder}}:} Context Encoder \cite{pathak2016context}, the first learning-based method for large holes.
	\item \textbf{GL\footnote{\href{https://github.com/satoshiiizuka/siggraph2017\_inpainting}{https://github.com/satoshiiizuka/siggraph2017\_inpainting}}:} Globally and Locally \cite{iizuka2017globally}, the first learning-based method for arbitrary regions. 
	\item \textbf{CA\footnote{\href{https://github.com/JiahuiYu/generative\_inpainting}{https://github.com/JiahuiYu/generative\_inpainting}}:} Contextual Attention \cite{yu2018generative}, the first method combining learning- and patch-based methods.
	\item \textbf{PConv\footnote{\href{https://github.com/NVIDIA/partialconv}{https://github.com/NVIDIA/partialconv}}:} Partial Convolution \cite{Liu_2018_ECCV}, the first learning-based method for free-form irregular holes.
	\item \textbf{EC\footnote{\href{https://github.com/knazeri/edge-connect}{https://github.com/knazeri/edge-connect}}:} EdgeConnect \cite{Nazeri_2019_ICCV}, the latest works using auxiliary edge information.
\end{itemize}

Compared to these approaches, our {\bf PICNet} is the first work considering multiple solutions on various datasets for this ill-posed problem. For fair comparison among learning-based methods, we mainly reported the results with \emph{each model trained on the corresponding dataset}. We consider the released models on the respective authors' websites to be their best performing models.

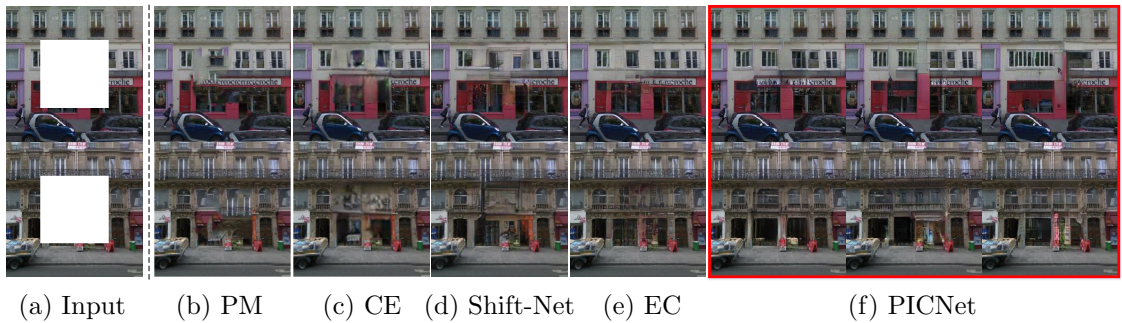
\begin{figure}[tb!] 
	\centering
	\includegraphics[width=\textwidth]{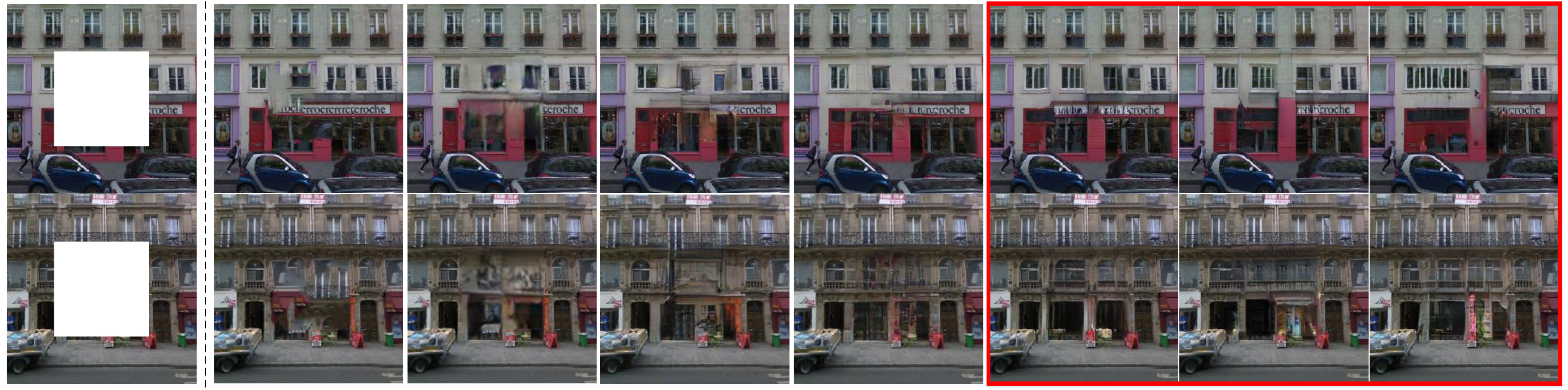}
	\begin{picture}(0,0)
		\put(-204,2){\footnotesize (a) Input}
		\put(-145,2){\footnotesize (b) PM}
		\put(-90,2){\footnotesize (c) CE}
		\put(-52,2){\footnotesize (d) Shift-Net}
		\put(15,2){\footnotesize (e) EC}
		\put(107,2){\footnotesize (f) PICNet}
	\end{picture}
	\caption[Qualitative results on Paris]{\textbf{Qualitative results on Paris val set} \cite{doersch2012makes} for center region completion. Here, we compare with \textbf{PM} \cite{barnes2009patchmatch}, \textbf{CE} \cite{pathak2016context}, \textbf{Shift-Net} \cite{Yan_2018_ECCV}  and \textbf{EC} \cite{Nazeri_2019_ICCV}. Note that, our \textbf{PICNet} generates different numbers of windows and varying door size.}
	\label{fig:ch4_result_building}	
\end{figure}

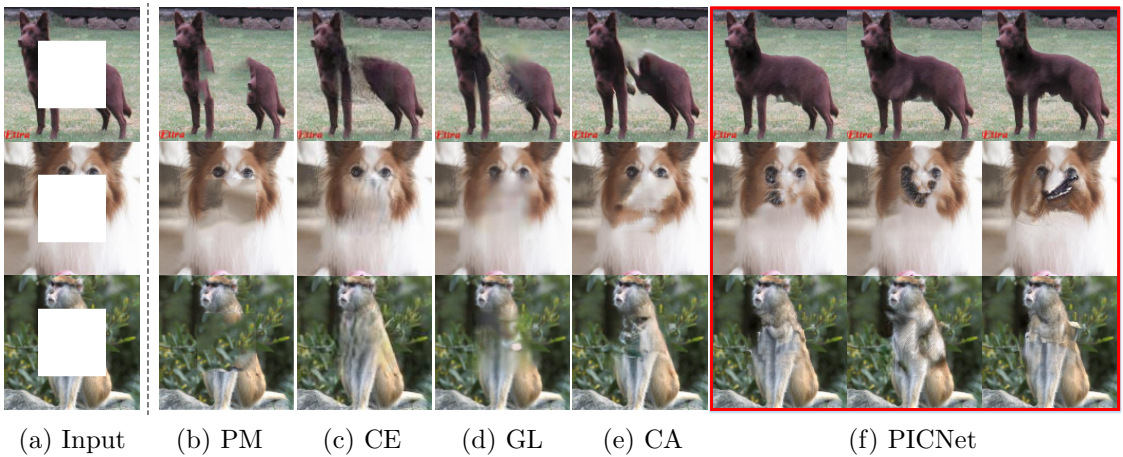
\begin{figure}[tb!]
	\centering
	\includegraphics[width=\textwidth]{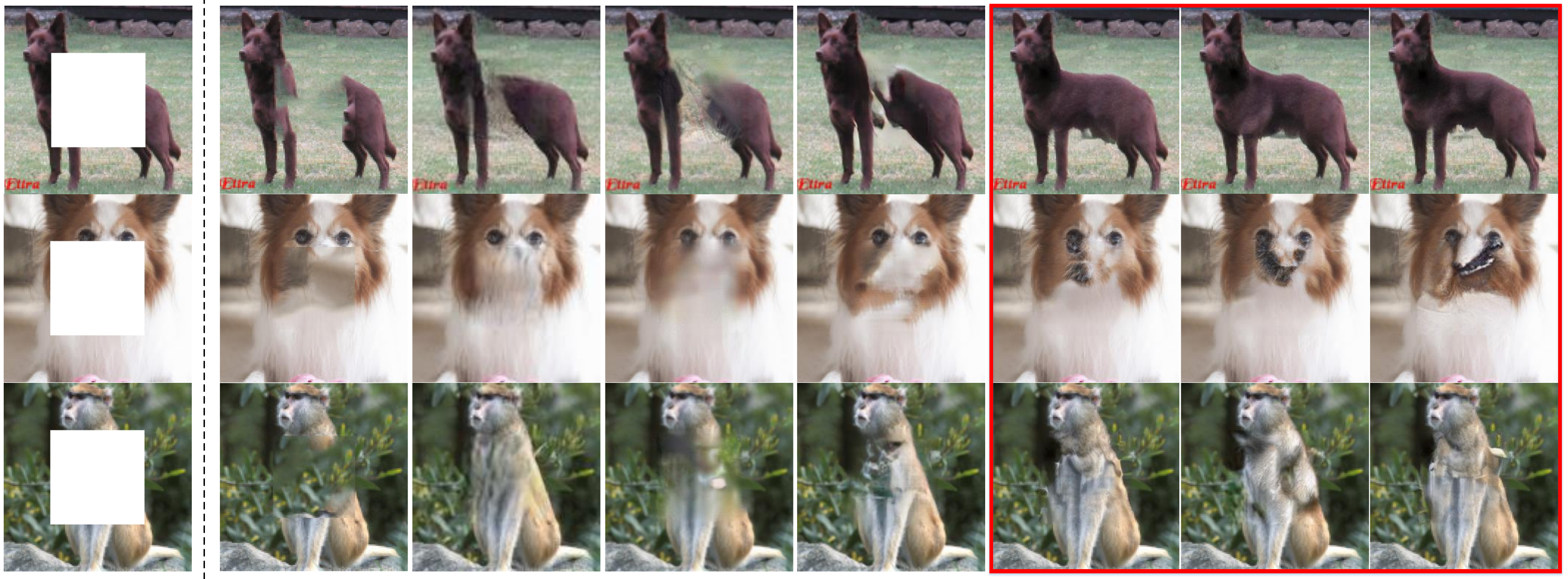}
	\begin{picture}(0,0)
		\put(-204,2){\footnotesize (a) Input}
		\put(-145,2){\footnotesize (b) PM}
		\put(-90,2){\footnotesize (c) CE}
		\put(-38,2){\footnotesize (d) GL}
		\put(15,2){\footnotesize (e) CA}
		\put(107,2){\footnotesize (f) PICNet}
	\end{picture}
	\caption[Qualitative results on the ImageNet validation set]{\textbf{Qualitative results and comparisons with the \textbf{PM} \cite{barnes2009patchmatch}, \textbf{CE} \cite{pathak2016context}, \textbf{GL} \cite{iizuka2017globally} and \textbf{CA} \cite{yu2018generative} on the ImageNet \cite{russakovsky2015imagenet}}. Our \textbf{PICNet} tries to generate some semantic result for the animals, when the significant semantic information is missing.}
	\label{fig:ch4_result_scene}	
\end{figure}

\subsubsection{Center Region Completion}

\paragraph{Qualitative Results} In Figure~\ref{fig:ch4_result_building}, we first show the visual results on the Paris dataset \cite{doersch2012makes}. \textbf{PM} works by coping similar patches from visible regions and obtains good results on this dataset with repetitive structures. \textbf{CE} generates reasonable structures with blurry textures. \textbf{Shift-Net} produces better results by copying feature from visible regions to holes, which is similar to \textbf{CA} (\textbf{CA} did not release model for Paris). \textbf{EC} provides a single reasonable solution. Compared to these, our \textbf{PICNet} model not only generates more natural images with high-quality, but also provides multiple results, \eg different numbers of windows and varying door sizes. 

Next, we report the performance on the more challenging ImageNet dataset~\cite{russakovsky2015imagenet}. For a fair comparison, we also used a subset of 100K training images of ImageNet to train our model as previous works \cite{iizuka2017globally}. Visual results on a variety of objects from the validation set are shown in Figure~\ref{fig:ch4_result_scene}. These visual test images are those chosen in \cite{iizuka2017globally}. We note that, while learning-based methods \textbf{CE}, \textbf{GL} and \textbf{CA} provide correctly semantic results, our model is able to infer the content quite effectively. We observe that our model tries to generate full body for the first dog, and the mouth for the second dog. Meanwhile, our \textbf{PICNet} provides \emph{multiple} and \emph{diverse} results, from which we can choose different realistic results. 

\subsubsection{Free-form Region Completion}

We further evaluate our model on various datasets with irregular holes as proposed by Liu \etal \cite{Liu_2018_ECCV}. In this testing dataset, they generated 6 categories of free-form masks with different hole-to-image area ratios: [0.01, 0.1], (0.1, 0.2], (0.2, 0.3], (0.3, 0.4], (0.4, 0.5], (0.5, 0.6]. Each has 2,000 irregular masks. Results are compared against the current state-of-the-art approaches both qualitatively and quantitatively. Results of \textbf{GL} and \textbf{CA} were obtained from their released models, which were trained only on regular random masks. Results of \textbf{EC} were also generated from their released model, which was trained on the same images and masks as ours. As \textbf{PConv} only provided the partial convolutional operation, we reproduced the model with the same masks. 

\begin{table}[tb!]
	\centering
	\footnotesize
	\renewcommand{\arraystretch}{1.2}
	\setlength\tabcolsep{6pt}
	\begin{tabular}{@{}c|c|ccccc@{}}
		\hlineB{3.5}
		& \textbf{Size} & {\bf GL} \cite{iizuka2017globally} & {\bf CA} \cite{yu2018generative} & {\bf PConv} \cite{Liu_2018_ECCV} & {\bf EC} \cite{Nazeri_2019_ICCV} & {\bf PICNet}\\
		\hlineB{2.5}
		\multirow{6}{1em}{\rotatebox[origin=c]{90}{$\text{FID}^{\dagger}$}} 
		& [0.01, 0.1]& 10.40  & 12.63 & 11.59 & {\bf 8.78} & 9.33 \\
		& (0.1, 0.2]& 26.42 & 24.63 & 26.46 & 16.75 & {\bf 15.93} \\
		& (0.2, 0.3]& 50.37 & 39.87 & 47.32 & 28.37 & {\bf 22.74} \\
		& (0.3, 0.4]& 79.01 & 57.44 & 77.16 & 43.74 & {\bf 36.23} \\
		& (0.4, 0.5]& 108.37 & 76.10 & 91.29 & 63.15 & {\bf 53.14} \\
		& (0.5, 0.6]& 125.41 & 93.55 & 113.62 & 93.43 & {\bf 78.53} \\
		\hline
		\multirow{6}{1em}{\rotatebox[origin=c]{90}{IScore$^{\star}$}} 
		& [0.01, 0.1]& 34.66 & 37.33 & {\bf 38.62} & 38.57 & 38.18 \\
		& (0.1, 0.2]& 31.94 & 34.95 & 31.97 & {\bf 35.59} & 35.36 \\
		& (0.2, 0.3]& 24.26 & 28.79 & 25.53 & 31.06 & {\bf 32.95 }\\
		& (0.3, 0.4]& 17.00 & 22.52 & 18.43 & 26.27 & {\bf 28.73}\\
		& (0.4, 0.5]& 12.13 & 18.35 & 12.43 & 18.94 & {\bf 21.20}\\
		& (0.5, 0.6]& 8.12 & 13.37 & 10.2 & 12.84 & {\bf 16.99}\\
		\hlineB{2.5}
	\end{tabular}
	\caption[Quantitative comparisons on ImageNet]{\textbf{Quantitative comparisons on ImageNet~\cite{russakovsky2015imagenet}} with free-form masks provided in~\cite{Liu_2018_ECCV}. $^{\dagger}$ = lower is better. $^{\star}$ = higher is better. Here, we used the top 10 samples (ranked by the discriminator score) in our models for the latest learning-based feature-level image quality evaluation.}
	\label{tab:ch4_free-form-is-fid}
\end{table}

\paragraph{Quantitative Results} In Table \ref{tab:ch4_free-form-is-fid}, we first report the FID and IS results on the ImageNet test set \cite{russakovsky2015imagenet}. In this setting, we used our top 10 samples of the 50 generated images for the evaluation (automatically voted using the discriminator score). As can be seen, while our multiple results are slight worse than \textbf{EC} on small mask sizes, we improve FID and IS significantly on large mask ratios, \eg ``78.53'' \emph{vs} ``93.43'' (\emph{16\% relative improvement}) FID for mask ratio (0.5, 0.6]. This suggests that when the mask ratios are small, it is sufficient to predict a \emph{single} best result based on the neighboring visible pixels, yet it is not reasonable when the mask ratios are large. The latter requires our approach of generating multiple and diverse results that match the testing set distribution.

\begin{table}[tb!]
	\centering
	\footnotesize
	\renewcommand{\arraystretch}{1.2}
	\setlength\tabcolsep{6pt}
	\begin{tabular}{@{}c|c|ccccc@{}}
		\hlineB{3.5}
		& \textbf{Size} & {\bf GL} \cite{iizuka2017globally} & {\bf CA} \cite{yu2018generative}& {\bf PConv} \cite{Liu_2018_ECCV} & {\bf EC} \cite{Nazeri_2019_ICCV} & {\bf PICNet}\\
		\hlineB{2.5}
		\multirow{6}{1em}{\rotatebox[origin=c]{90}{$\ell_1(\%)^{\dagger}$}} 
		& [0.01, 0.1]& 0.023  & 0.024 & 0.021 & 0.020 & {\bf 0.010}\\
		& (0.1, 0.2]& 0.035 & 0.034 & 0.030 & 0.025 & {\bf 0.016}\\
		& (0.2, 0.3]& 0.050 & 0.047 & 0.042 & 0.033 & {\bf 0.025}\\
		& (0.3, 0.4]& 0.066 & 0.061 & 0.057 & 0.042 & {\bf 0.035}\\
		& (0.4, 0.5]& 0.081 & 0.075 & 0.073 & 0.051 & {\bf 0.046}\\
		& (0.5, 0.6]& 0.095 & 0.093 & 0.099 & 0.068& {\bf 0.064}\\
		\hline
		\multirow{6}{1em}{\rotatebox[origin=c]{90}{SSIM$^{\star}$}} 
		& [0.01, 0.1]& 0.915 & 0.908 & 0.917 & 0.923 & {\bf 0.963}\\
		& (0.1, 0.2]& 0.853 & 0.845 & 0.859 & 0.878 & {\bf 0.914}\\
		& (0.2, 0.3]& 0.767 & 0.765 & 0.782 & 0.820 & {\bf 0.852}\\
		& (0.3, 0.4]& 0.682 & 0.691 & 0.704 & 0.760 & {\bf 0.785}\\
		& (0.4, 0.5]& 0.600 & 0.613 & 0.622 & 0.693 & {\bf 0.712}\\
		& (0.5, 0.6]& 0.529 & 0.532 & 0.513 & 0.599 & {\bf 0.618}\\
		\hline
		\multirow{6}{1em}{\rotatebox[origin=c]{90}{PSNR$^{\star}$}}
		& [0.01, 0.1]& 28.42 & 26.85 & 28.79 & 29.47 & {\bf 32.26}\\
		& (0.1, 0.2]& 24.41 & 23.18 & 24.67 & 26.25 & {\bf 27.33}\\
		& (0.2, 0.3]& 21.33 & 20.44 & 21.63 & 23.82 & {\bf 24.44}\\
		& (0.3, 0.4]& 19.11 & 18.63 & 19.39 & 21.95 & {\bf 22.32}\\
		& (0.4, 0.5]& 17.56 & 17.30 & 17.75 & 20.44 & {\bf 20.71}\\
		& (0.5, 0.6]& 16.48 & 16.08 & 15.68 & 18.53 & {\bf 18.72}\\
		\hlineB{2.5}
	\end{tabular}
	\caption[Quantitative comparisons over Places2]{\textbf{Quantitative comparisons over Places2}~\cite{zhou2018places} on free-form masks provided in~\citep{Liu_2018_ECCV}. $^{\dagger}$ = lower is better. $^{\star}$ = higher is better. Here, the closest to the original ground truth samples in our method are selected for the traditional pixel- and patch-level image quality evaluation. }
	\label{tab:ch4_free-form-pp}
\end{table}

Traditional pixel- and patch-level comparison results are reported on the Places2 test set \cite{zhou2018places} in Table \ref{tab:ch4_free-form-pp}. As these metrics require one-to-one matched images for the evaluation, we selected one sample from our multiple results, with the best balance of quantitative measures for comparison. Without bells and whistles, all instantiations of our model outperform the existing state-of-the-art models, indicating that our random samples include the close example to the original image. While the prior works \cite{iizuka2017globally,yu2018generative,Liu_2018_ECCV,Nazeri_2019_ICCV} strongly enforce the generated images to be the same as the original images via a reconstruction loss, the testing images are not in the training set. 

\begin{figure}[tb!]
	\centering
	\includegraphics[width=\textwidth]{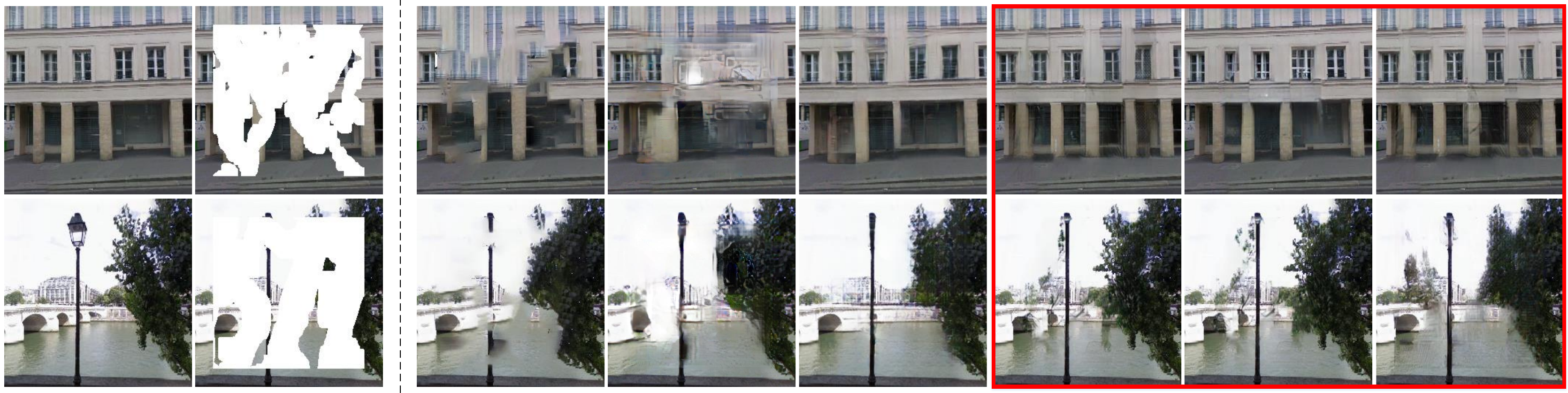}
	\begin{picture}(0,0)
		\put(-210,2){\footnotesize (a) Original}
		\put(-153,2){\footnotesize (b) Input}
		\put(-90,2){\footnotesize (c) PM}
		\put(-46,2){\footnotesize (d) PConv}
		\put(14,2){\footnotesize (e) EC}
		\put(107,2){\footnotesize (f) PICNet}
	\end{picture}
	\caption[Comparison of qualitative results on Paris val set]{\textbf{Comparison of qualitative results on Paris val set \cite{doersch2012makes} with free-form masks from PConv \cite{Liu_2018_ECCV}}. (a) Original image. (b) Masked input. (c) Results of \textbf{PM} \cite{barnes2009patchmatch}. (d) Results of \textbf{PConv} \cite{Liu_2018_ECCV}. (e) Results of \textbf{EC} \cite{Nazeri_2019_ICCV}. (f). Our multiple and diverse results. }
	\label{fig:ch4_result_random_pari}	
\end{figure}
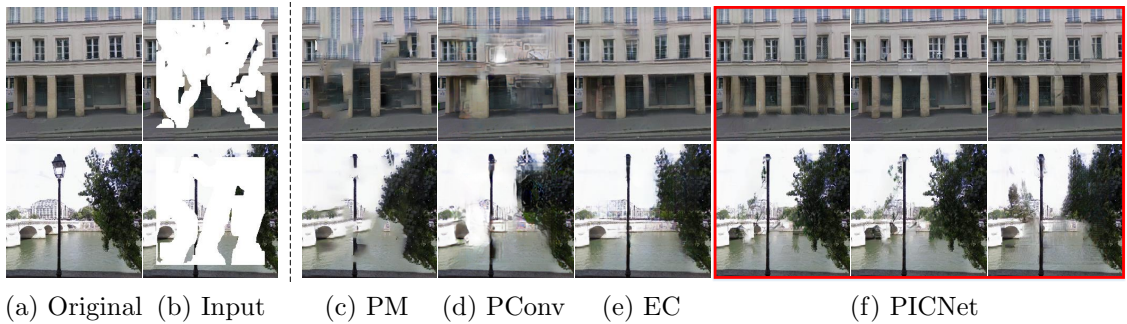

\begin{figure}[tb!]
	\centering
	\includegraphics[width=\textwidth]{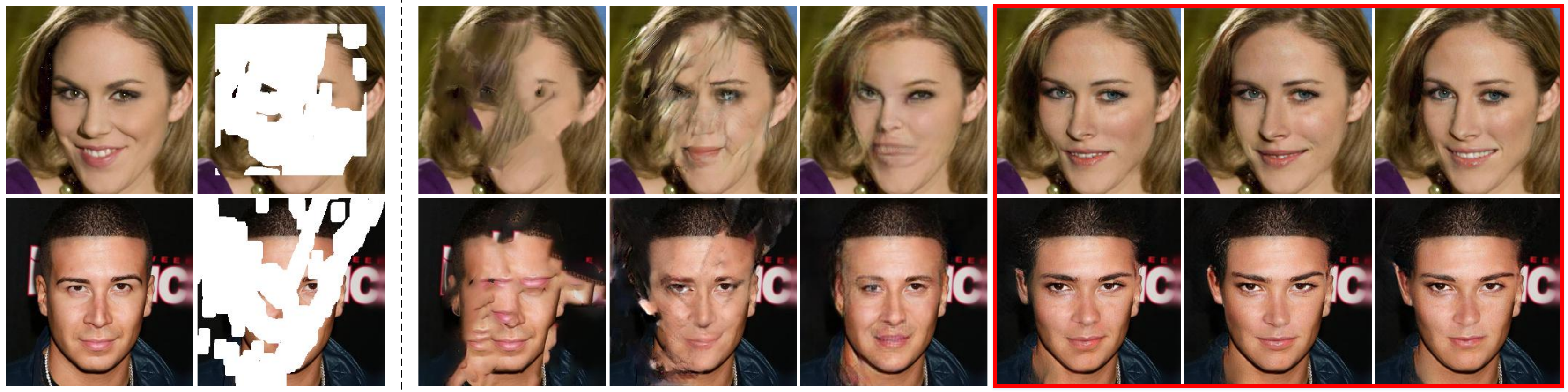}
	\begin{picture}(0,0)
		\put(-210,2){\footnotesize (a) Original}
		\put(-153,2){\footnotesize (b) Input}
		\put(-90,2){\footnotesize (c) PM}
		\put(-36,2){\footnotesize (d) CA}
		\put(14,2){\footnotesize (e) EC}
		\put(108,2){\footnotesize (f) PICNet}
	\end{picture}
	\caption[Qualitative results on CelebA-HQ testing set]{\textbf{Qualitative results on CelebA-HQ testing set \cite{liu2015deep,karras2017progressive} with free-form masks from PConv \cite{Liu_2018_ECCV}.} (a) Original image. (b) Masked input. (c) Results of \textbf{PM} \cite{barnes2009patchmatch}. (d) Results of \textbf{CA} \cite{yu2018generative}. (e) Results of \textbf{EC} \cite{Nazeri_2019_ICCV}. (f). Our multiple and diverse results. }
	\label{fig:ch4_result_random_celeba}	
\end{figure}
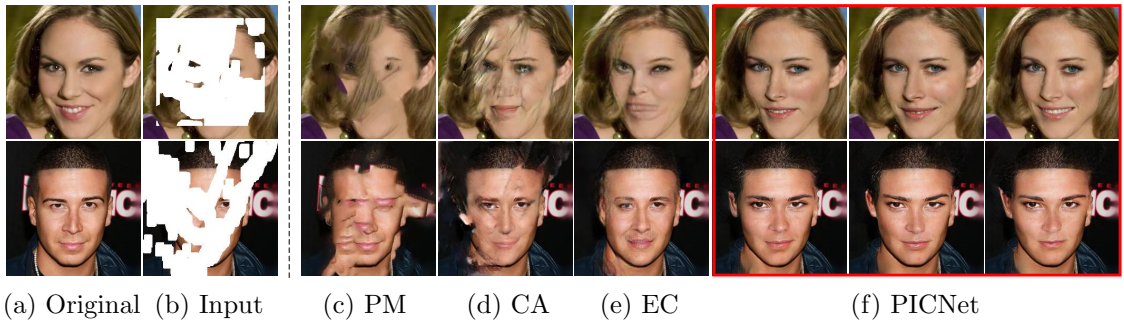

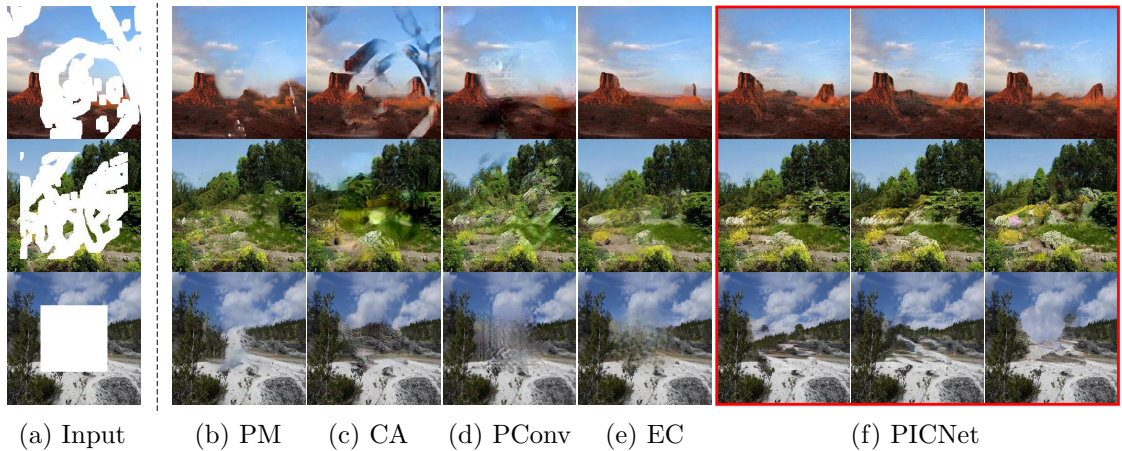
\begin{figure*}[tb!]
	\centering
	\includegraphics[width=\textwidth]{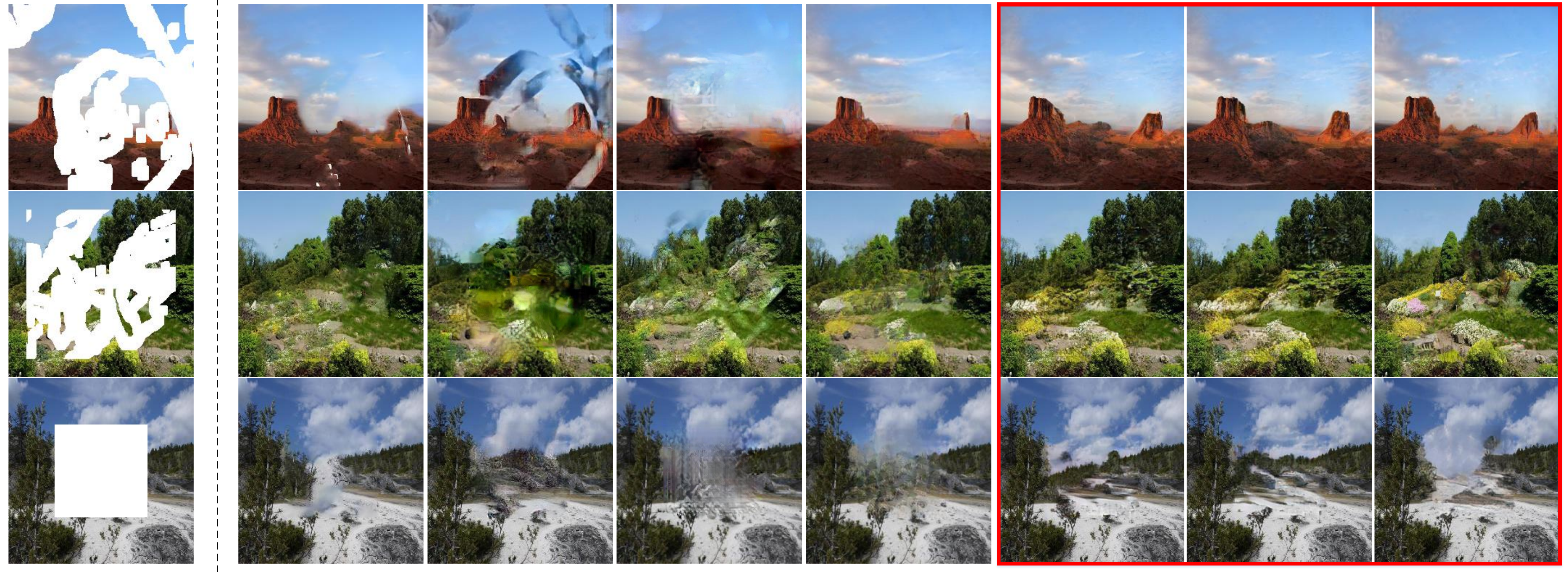}
	\begin{picture}(0,0)
		\put(-204,2){\footnotesize (a) Input}
		\put(-138,2){\footnotesize (b) PM}
		\put(-88,2){\footnotesize (c) CA}
		\put(-43,2){\footnotesize (d) PConv}
		\put(16,2){\footnotesize (e) EC}
		\put(108,2){\footnotesize (f) PICNet}
	\end{picture}
	\caption[Qualitative results on Place2 testing set]{\textbf{Qualitative results on Place2 testing set \cite{zhou2018places} with various masks.} (a) Masked input. (b) Results of \textbf{PM} \cite{barnes2009patchmatch}. (c) Results of \textbf{CA} \cite{yu2018generative}. (d) Results of \textbf{PConv} \cite{Liu_2018_ECCV}. (e) Results of \textbf{EC} \cite{Nazeri_2019_ICCV}. (f). Our multiple and diverse results. }
	\label{fig:ch4_result_random_place}	
\end{figure*}

\paragraph{Qualitative Results.} Qualitative comparison results are visualized in Figures~\ref{fig:ch4_result_random_pari}, \ref{fig:ch4_result_random_celeba} and \ref{fig:ch4_result_random_place}. Our PICNet is able to achieve good results for multiple solutions even under challenging conditions. 

In Figure~\ref{fig:ch4_result_random_pari}, we show some results on Paris dataset. We can see that \textbf{PM} and \textbf{PConv} fail to synthesize semantic structure for large holes. The \textbf{EC} works well on the obvious structure by utilizing the auxiliary edge. Our method was explicitly trained to copy information from visible parts, leading to better visual results on repetitive structures, \eg the window in the first row. Furthermore, our model provides multiple and diverse results for one given masked image. More results are available online\footnote{\href{https://github.com/lyndonzheng/Pluralistic-Inpainting}{https://github.com/lyndonzheng/Pluralistic-Inpainting}}.

Figure~\ref{fig:ch4_result_random_celeba} shows some results on the Celeba-HQ dataset. We can see that the non-learning-based method \textbf{PM} is unable to generate reasonable semantic content in the images. While the \textbf{CA} is able to generate novel content on the face, it is not as suitable for large holes. \textbf{EC} results in reasonable semantic structure but blurry and inconsistent images. Our approach was explicitly trained for variable results, rather than strongly enforcing the completed image to be close to the original. Hence, our \textbf{PICNet} can provide multiple plausible results with different expressions. The online demo is also provided on our project page\footnote{\href{http://www.chuanxiaz.com/project/pluralistic/}{http://www.chuanxiaz.com/project/pluralistic/}}.

In Figure~\ref{fig:ch4_result_random_place}, we further show results on the more challenging Places2 dataset. The non-learning-based \textbf{PM} fills in reasonable pixels for natural scenes by copying similar patches from visible parts to missing holes. The \textbf{CA} only works well on regular masks as their released model was only trained on random regular masks. \textbf{EC} results are not as realistic. Instead, we can select plausible images from PICNet's multiple sampled results. Furthermore, it is hard to identify the filled-in areas in our completed images, as our short-long term patch attention copies non-local information from visible regions based on correctly predicted content.

\subsubsection{Visual Turing Tests}

We additionally compared the perceived visual fidelity of our model against existing approaches using human perceptual metrics, as proposed in \cite{zhang2016colorful}. We conducted two types of user surveys: \emph{2 alternative forced choice} (2AFCs) and \emph{visual fidelity and perceived quality} (VFPQ). In particular, for 2AFCs, we randomly presented a generated image from an undisclosed method to the participants, and asked them to decide whether the presented image was real or fake. For quality control, we also inserted a number of real images to avoid negative testing. For VFPQ, we gave the participants a masked input and the corresponding results from all methods (blinded), and asked the participants to choose the image that was the most visually realistic. The participants were allowed to vote for multiple images simultaneously, if they felt the images were equally realistic. For each participant, we randomly presented 100 questions, consisting of 60 2AFCs examples and 40 VFPQ questions. We collected 47 valid surveys with 4,700 answers.

\begin{table}[tb!]
	\centering
	\footnotesize
	\renewcommand{\arraystretch}{1.2}
	\setlength\tabcolsep{6pt}
	\begin{tabular}{@{}l|ccccc@{}}
		\hlineB{2.5}
		& {\bf GL} \cite{iizuka2017globally} & {\bf CA} \cite{yu2018generative} & {\bf EC} \cite{Nazeri_2019_ICCV} & {\bf PICNet} & {\bf Real} \\
		\hlineB{2.5}
		2AFC(\%) & 15.1$\pm$1.8 & 17.8$\pm$2.0 & 44.2$\pm$3.7&  57.0$\pm$4.5 & 90.44$\pm$1.5 \\
		\hlineB{2}
	\end{tabular}
	\caption[2-alternative-forced-choice (2AFCs) score on CelebA-HQ testing set]{\textbf{2-alternative-forced-choice (2AFCs) score on CelebA-HQ \cite{liu2015deep,karras2017progressive} testing set.} All testing images were degraded by center masks. Here, the participants were required to judge whether a randomly displayed image was real \emph{or} fake. The reported values are the percentages of images generated by each method that were judged ``real''. }
	\label{tab:ch4_AFC}
\end{table}

\begin{table}[tb!]
	\centering
	\footnotesize
	\renewcommand{\arraystretch}{1.2}
	\setlength\tabcolsep{6pt}
	\begin{tabular}{@{}lccccc@{}}
		\hlineB{3.5}
		& \multicolumn{4}{c}{VFPQ(\%)} \\
		\cline{2-5}
		& [0.01, 0.1]& (0.1, 0.2]& (0.2, 0.3]& (0.3, 0.4]\\
		\hlineB{2.5}
		GL \cite{iizuka2017globally} & 23.3 $\pm$ 4.3 & 9.8 $\pm$ 1.6 & 6.4 $\pm$ 1.0 & 4.1 $\pm$ 0.4 \\
		CA \cite{yu2018generative} & 11.4 $\pm$ 2.1 & 9.7 $\pm$ 1.3 & 7.6 $\pm$ 0.9 & 8.6 $\pm$ 0.9 \\
		PConv \cite{Liu_2018_ECCV} & 27.8 $\pm$ 4.0 & 13.5 $\pm$ 1.6 & 11.0 $\pm$ 1.4 & 5.3 $\pm$ 0.7 \\
		EC \cite{Nazeri_2019_ICCV} & 42.3 $\pm$ 6.0 & 38.8 $\pm$ 4.9 & 33.6 $\pm$ 3.2 & 26.7 $\pm$ 3.3 \\
		PICNet& 57.5 $\pm$ 3.6 & 63.0 $\pm$ 4.0 & 69.9 $\pm$ 3.8 & 71.4 $\pm$ 3.4 \\
		\hlineB{2.5}
	\end{tabular}
	\caption[Visual fidelity and perceived quality (VFPQ) score on Places2 test set]{\textbf{Visual fidelity and perceived quality (VFPQ) score on Places2 \cite{zhou2018places} test set.} All testing images were degraded by free-form masks provided in PConv \cite{Liu_2018_ECCV}. Participants selected the most realistic image from among blinded methods for the same masked input, with multiple selections allowed. Headers are ranges of mask sizes (as fraction of image). For each method, we report the percentage of trials for which it was selected, and the 95\% margin of error.}
	\label{tab:ch4_JND}
\end{table}

We first show the 2FACs evaluation results in Table \ref{tab:ch4_AFC}. Most participants correctly identified the real image during the evaluation, showing that they made conscientious discerning judgement. Our model achieved better realism scores than existing state-of-the-art methods. Table \ref{tab:ch4_JND} shows the VFPQ evaluation results. We found that the participants strongly favored our completed results for all mask ratios, and especially so on the challenging large mask ratios. This suggests that once the visible regions do not impose strong constraints, our multiple and diverse results were naturally varied but mostly realistic and reasonable.

\subsection{Additional Results}\label{sec:ch4_add_result}

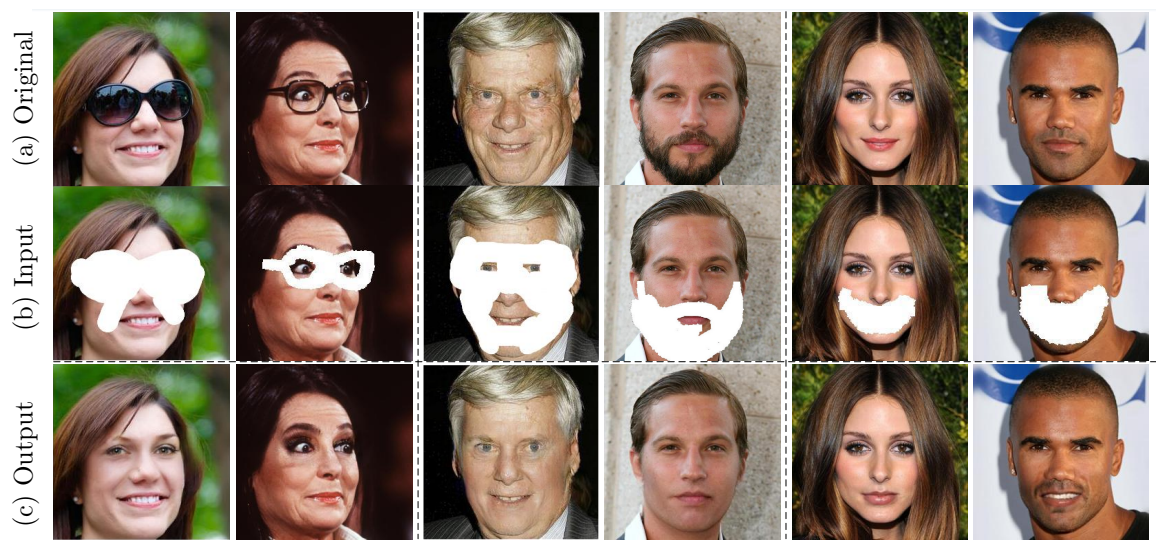
\begin{figure}[tb!]
    \centering
    \includegraphics[width=\linewidth]{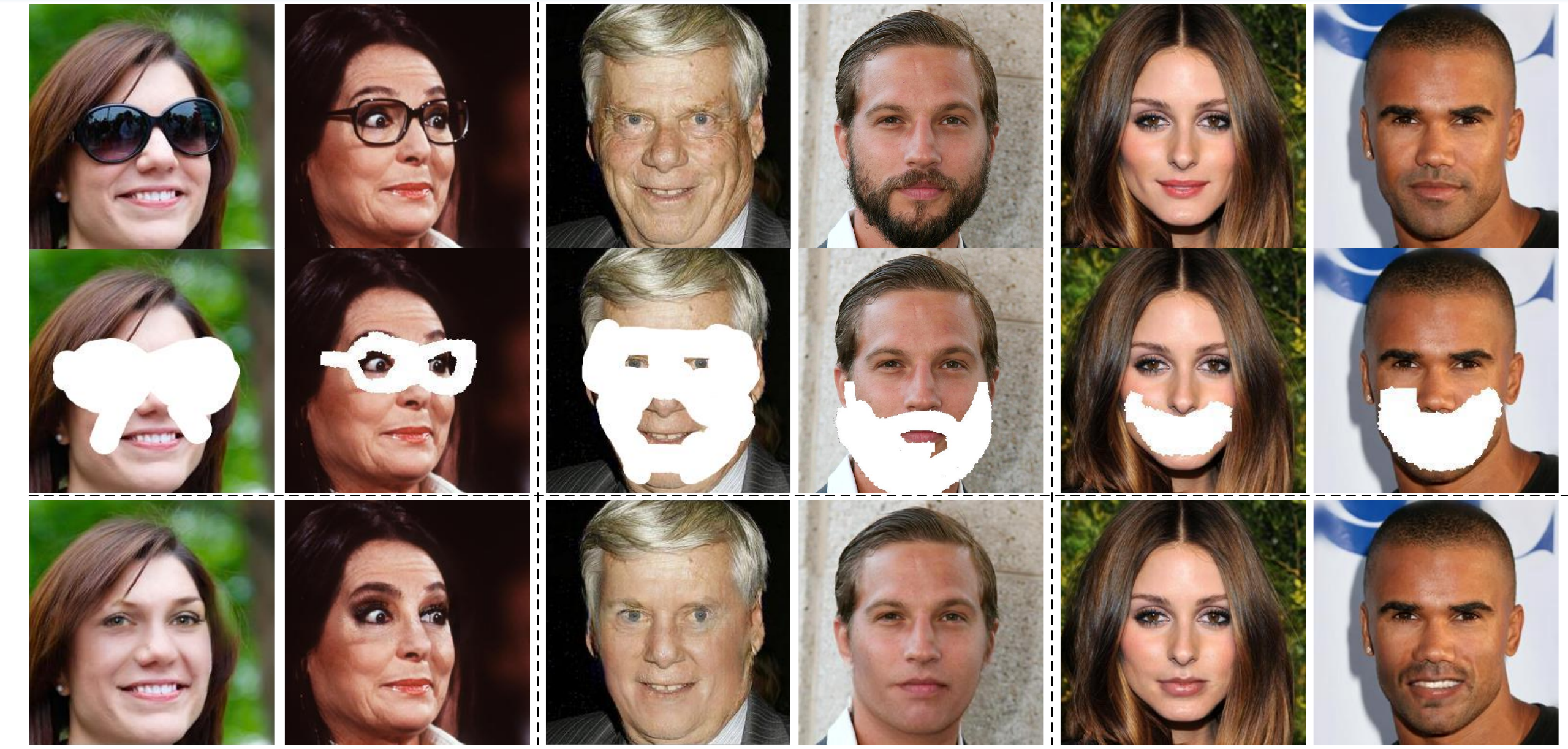}
    \begin{picture}(0, 0)
        \put(-216,156){\rotatebox{90}{\footnotesize (a) Original}}
        \put(-216,94){\rotatebox{90}{\footnotesize (b) Input}}
        \put(-216,20){\rotatebox{90}{\footnotesize (c) Output}}
    \end{picture}
    \vspace{-0.3cm}
    \caption[Additional results on the CelebA-HQ test set for free-form image editing]{\textbf{Additional results  of our \textbf{PICNet} on the CelebA-HQ test set \cite{liu2015deep,karras2017progressive} for free-form image editing.} (a) Original image. (b) Masked input image. (c) Output of our \textbf{PICNet}. In the first two columns, we erased eyeglasses. Wrinkles and facial hair were removed in the next two columns. Finally, we freely changed mouth expressions. Note that due to the provision of multiple and diverse results, the users can easily select their favorite result. We refer readers to our online demo for testing.}
    \label{fig:ch4_free_edit_face}
\end{figure}

\begin{figure}[tb!]
    \centering
    \includegraphics[width=\linewidth]{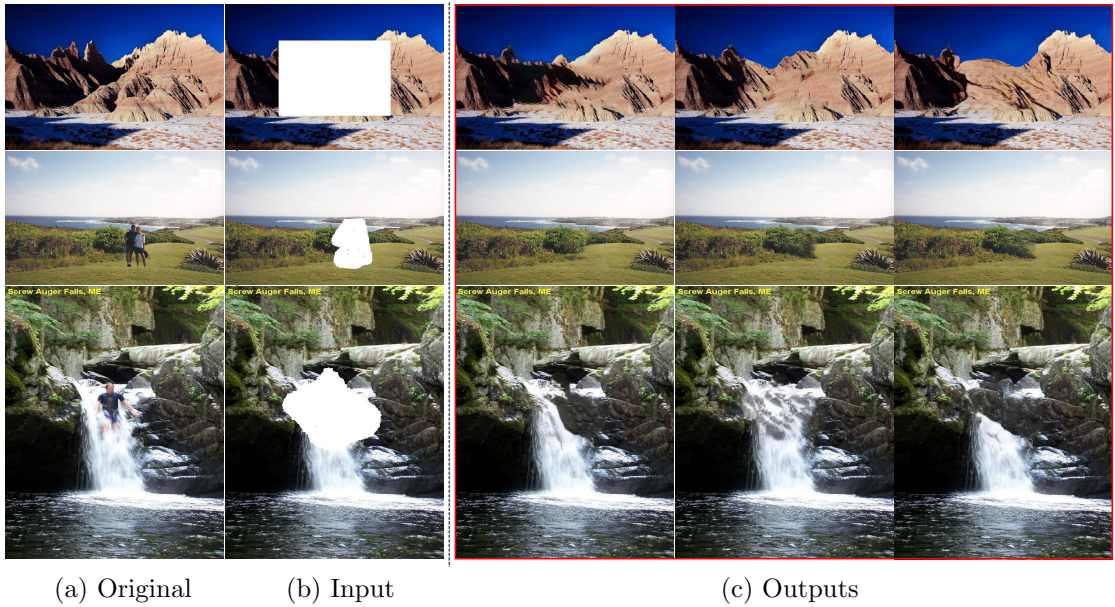}
    \begin{picture}(0, 0)
        \put(-190,3){\footnotesize (a) Original}
        \put(-103,3){\footnotesize (b) Input}
        \put(60,3){\footnotesize (c) Outputs}
    \end{picture}
    \caption[Additional results on the Places2 test set for free-form image editing]{\textbf{Additional results  of \textbf{PICNet} on the Places2 test set \cite{zhou2018places} for free-form image editing.} (a) Original image. (b) Input masked image. (c) Multiple and diverse outputs of our \textbf{PICNet}. Here, we show examples of reshaping the mountain ridge and subject removal, but, unlike conventional inpainting, we can provide multiple and diverse choices and on high-resolution images. }
    \label{fig:ch4_free_edit_place}
\end{figure}

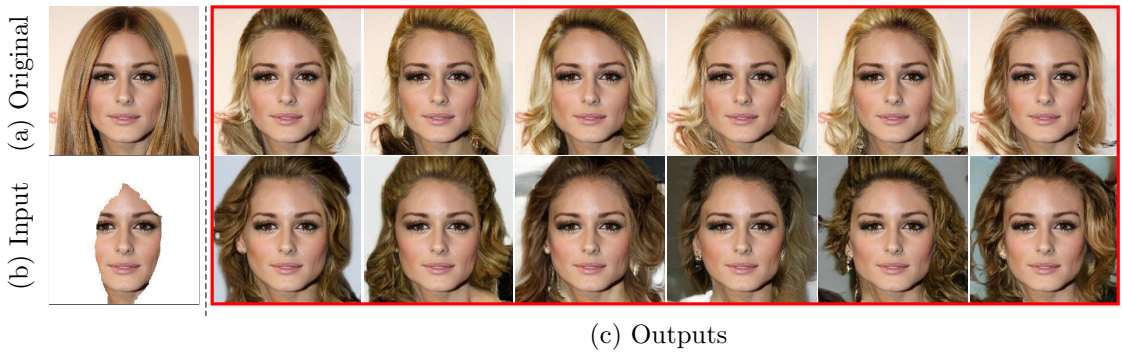
\begin{figure}[tb!]
	\centering
	\includegraphics[width=\linewidth]{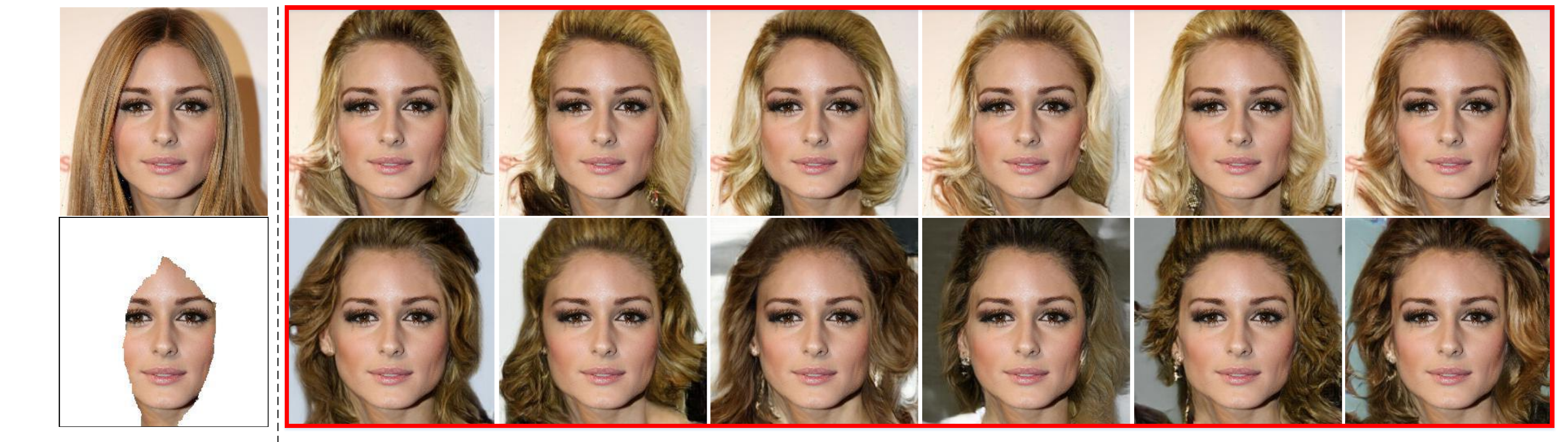}
	\begin{picture}(0,0)
		\put(-208,76){\rotatebox{90}{\footnotesize (a) Original}}
		\put(-208,24){\rotatebox{90}{\footnotesize (b) Input}}
		\put(10,4){\footnotesize (c) Outputs}
	\end{picture}
	\vspace{-0.2cm}
	\caption[Outpainting examples of our models.]{\textbf{Outpainting examples of our models.} (a) Original image. (b) Masked input. (c) Multiple and diverse results of our \textbf{PICNet}. Note that, it provides different hairstyles for the users. }
	\label{fig:ch4_result_visible_center}	
\end{figure}

We show additional results of our proposed PICNet in Figures \ref{fig:ch4_free_edit_face}, \ref{fig:ch4_free_edit_place} and \ref{fig:ch4_result_visible_center}. Our approach is suitable for a wide range of applications, \eg face editing, scene recomposition, object removal and outpainting. 

\paragraph{Face Editing} We first show free-form image editing on face images in Figure~\ref{fig:ch4_free_edit_face}. Our model works well for conventional object removal, \eg removing eyeglasses in the first two columns. Next, we smoothed faces by removing wrinkles and facial hair. Finally, we changed mouth expressions by selecting an example among our multiple and diverse completed results. 

\paragraph{High-Resolution Natural Image Editing} The basic PICNet did not handle high resolution (HR) image completion, because the generation from a random vector $\textbf{z}$ only works for a fixed feature size \cite{karras2020analyzing}. However, following the \emph{two-stage} image completion approaches \cite{yang2017high,yu2018generative,song2018contextual,yu2019free,Nazeri_2019_ICCV,yi2020contextual,zeng2020high}, we trained another encoder-decoder framework to refine the fixed resolution output of our PICNet. Since this work does \emph{not} focus on HR images, we used a simple  design for the refinement network by directly reapplying the PICNet framework in the second refinement stage, but without the sampling process. Note that the multiple and diverse solutions were seeded by the first content generation stage. 

As can be seen in Figure~\ref{fig:ch4_free_edit_place}, our approach produces diverse results as well as visually realistic appearance for HR natural image editing, \eg reshaping the mountain ridge and generating various mountain streams. This demonstrates that our model works well for HR images.

\paragraph{Outpainting} In our dual pipeline framework, the masked image $\mathbf{I}_m$ and its corresponding complement image $\mathbf{I}_c$ can be easily swapped. Therefore, we randomly reversed the input mask during training on Celeba-HQ. Figure~\ref{fig:ch4_result_visible_center} shows examples where information is missing from the image border regions. This ``outpainting'' is a challenging task as these regions have much larger uncertainty~\cite{iizuka2017globally}. Note that the subject's hair can be significantly varied during completion, suggesting that our model is applicable to style editing. Our structure has been extended to other related tasks, such as spherical image generation~\cite{hara2020spherical}.

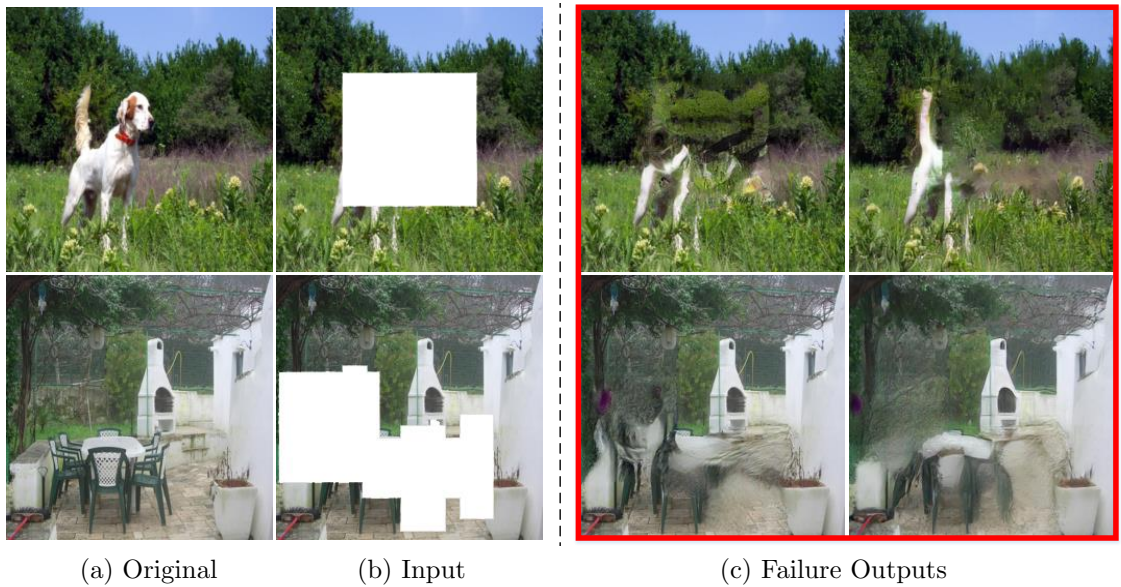
\begin{figure}[tb!]
	\centering
	\includegraphics[width=\linewidth]{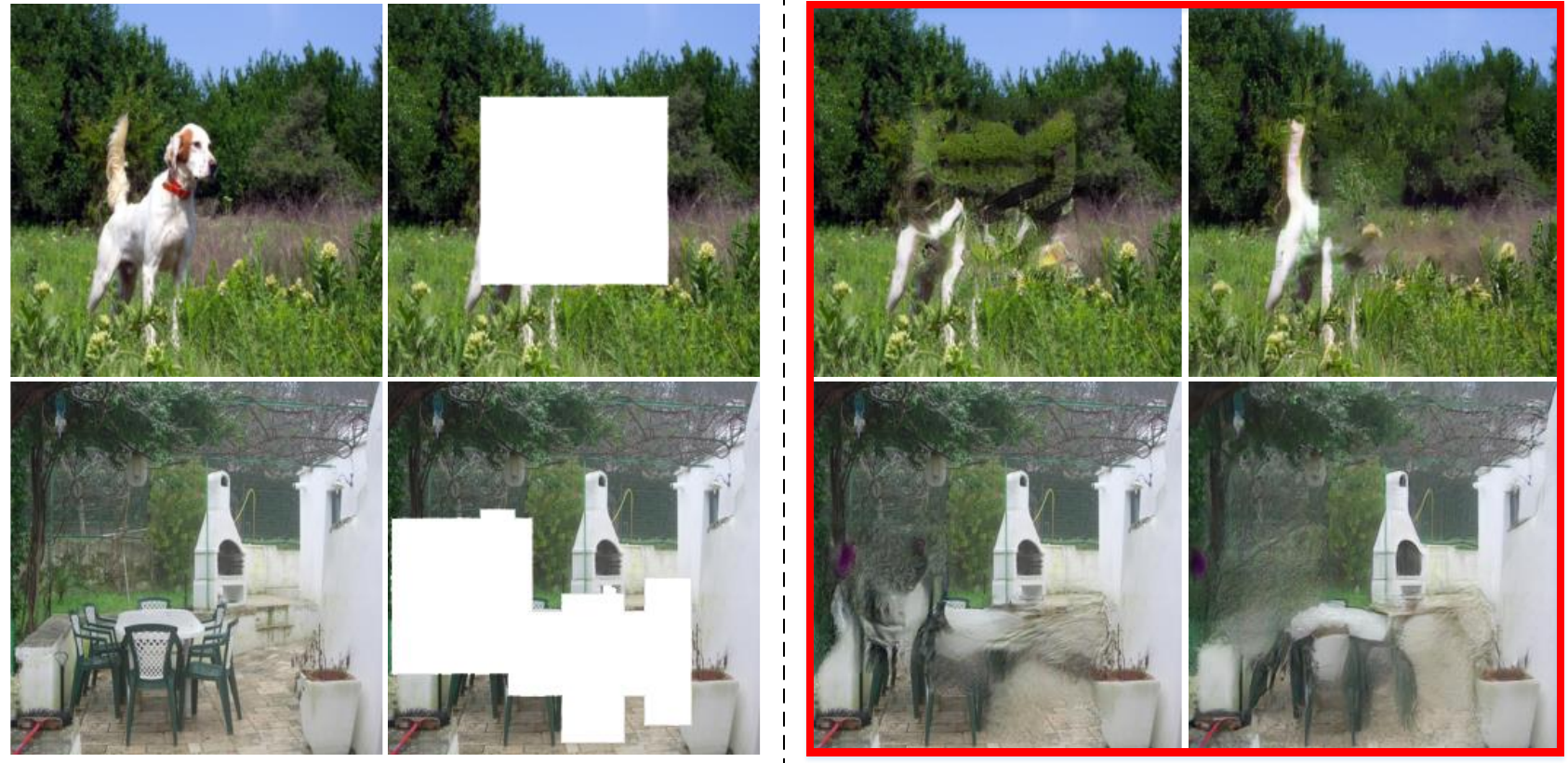}
	\begin{picture}(0,0)
		\put(-180,2){\footnotesize(a) Original}
		\put(-76,2){\footnotesize (b) Input}
		\put(60,2){\footnotesize (c) Failure Outputs}
	\end{picture}
	\caption[Failure cases of our PICNet]{\textbf{Failure cases of our \textbf{PICNet}.} (a) Original image. (b) Masked input. (c) Failure results of our \textbf{PICNet}, where the semantic information is heavily masked, \eg only four legs are visible of the dogs. }
	\label{fig:ch4_result_fail}	
\end{figure}

\section{Limitations and Discussion}\label{ch4:conc}

In this chapter, a novel solution is presented for the image completion task. Unlike existing methods \cite{pathak2016context,iizuka2017globally,yu2018generative,yu2019free,Nazeri_2019_ICCV,yi2020contextual}, our probabilistically principled framework can generate multiple and diverse solutions with plausible content for a given masked image. The resulting \textbf{PICNet} shows that prior-conditional lower bound coupling is significant for conditional image generation, leading to a more reasonable two-branch training than the current deterministic structure. We also introduce an enhanced short+long term patch attention layer, which improves realism by automatically attending to both high quality visible features and semantically correct generated features. 

Experiments on a variety of datasets demonstrated that the multiple solutions were diverse and of high quality. On the latest learning based feature-level metrics and traditional pixel- and patch-level metrics, we demonstrated that PICNet outperformed the single-solution approaches \cite{iizuka2017globally,yu2018generative,Liu_2018_ECCV,Nazeri_2019_ICCV}, especially for large mask ratios with large uncertainty.  We further showed in studies that users strongly favored our completed results when compared to the results in existing approaches. We additionally demonstrated that our PICNet is suitable for many interesting free-form image editing, \eg object removal, expression changing, and scene recomposition. These multiple and diverse results can also be easily extended to HR image editing.

Although the proposed model achieved better results than prior methods on various datasets by selecting images from the number of diverse sampling results, the model does not cope well with heavily structured objects with important information missing, as shown in Figure~\ref{fig:ch4_result_fail}. As semantic image completion is as yet an immature task that builds upon conventional image inpainting, a full understanding of semantic image content remains a challenge. In Figure~\ref{fig:ch4_result_fail}(top), we can see that although the four legs of the dog are visible, the model cannot generate a complete dog even after multiple sampling. This is a significant issue in current image completion task. When the important semantic regions, such as head or body of human, are missing heavily, the current deep learning models cannot correctly imagine these missed contents. In the bottom image, if the content is not correctly generated, our attention model fails to provide high-quality visual results. In Chapter \ref{ch:TFill}, I aim to address these issues.

\chapter{Image Completion via Transformer} 
\chaptermark{TFill}
\label{ch:TFill} 

The previous chapter introduces multiple and diverse results for the high subjective image completion task, but it has some failed cases, especially when the mask is very large. Bridging distant context interactions is important for high-quality image completion with large masks. However, previous methods that attempt this via deep or large receptive field (RF) convolutions cannot escape from the dominance of nearby interactions, which may tend to inferior results. In this chapter, I propose treating image completion as a directionless sequence-to-sequence prediction task, and deploy a transformer to directly capture long-range dependence in the encoder in a first phase. Crucially, a \emph{restrictive CNN} with small and non-overlapping RF is employed for token representation, which allows the transformer to explicitly model the long-range context relations with equal importance in all layers, without implicitly confounding neighboring tokens when larger RFs are used. In a second phase, to improve appearance consistency between visible and generated regions, a novel attention-aware layer (AAL) is introduced to better exploit distantly related features and also avoid the insular effect of standard attention.

This chapter is organized as follows: I first introduce the motivation in Section \ref{ch5:intro} and then discuss the related works in Section \ref{ch5:back}. Next, I describe the proposed transformer-based completion framework in Section \ref{ch5:appro}. Section \ref{ch5:experiment} demonstrates how models with a transformer can be used to improve the completion performance. Finally, I will discuss the limitations and further directions in Section \ref{ch5:conl}.

\begin{figure}[tb!]
    \centering
    \includegraphics[width=\linewidth,height=0.3\textheight]{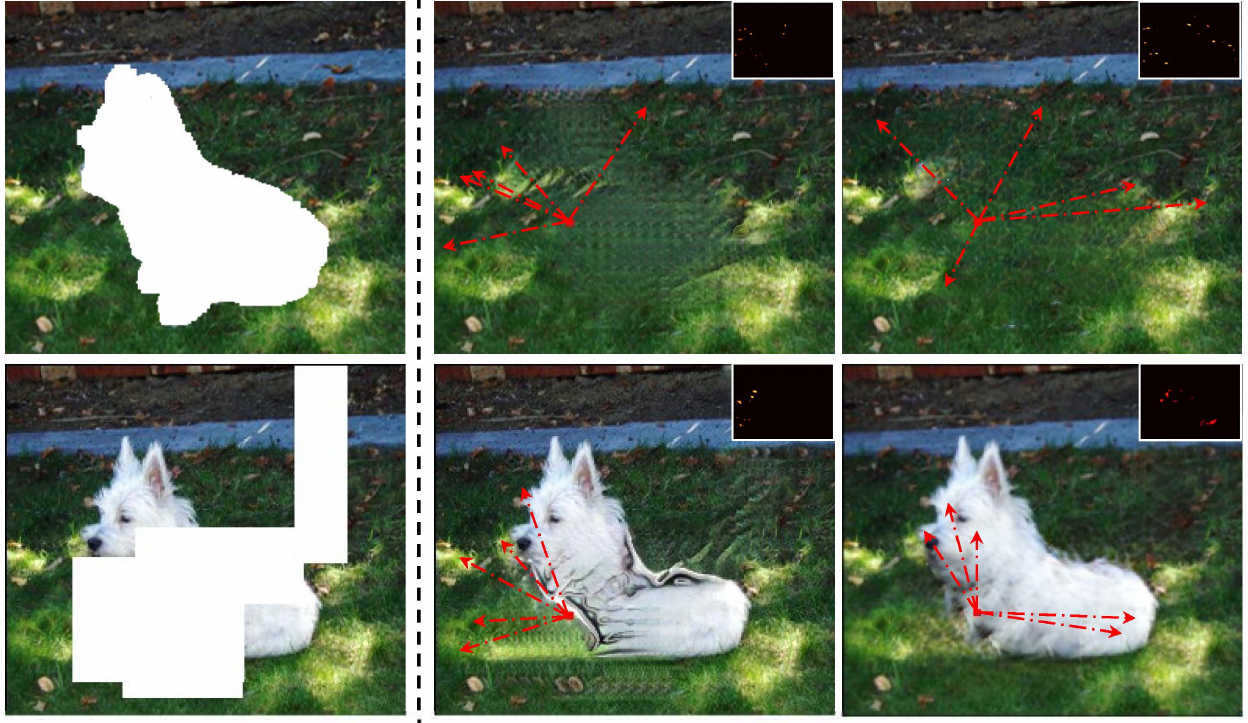}
    \begin{picture}(0,0)
    \put(-12,150){\footnotesize {\color{red}$\textbf{x}_i$}}
    \put(-12,40){\footnotesize {\color{red}$\textbf{x}_i$}}
    \put(106,150){\footnotesize {\color{red}$\textbf{x}_i$}}
    \put(106,40){\footnotesize {\color{red}$\textbf{x}_i$}}
    \put(-178,5){\footnotesize (a) Masked 1nput}
    \put(-30,5){\footnotesize (b) HiFill~\cite{yi2020contextual}}
    \put(106,5){\footnotesize (c) Ours-TFill}
    \end{picture}
    \caption[An example of information flow in image completion]{\textbf{An example of information flow in image completion with free-form masks.} The position $\textbf{x}_i$'s response (flow) is calculated by inferring the \emph{Jacobian} matrix between it to all pixels in the given masked input. Here, only the highest flows are shown. Our TFill correctly captures long-range visible context flow, even with a large mask splitting two semantically important zones.}
    \label{fig:ch5_example}
\end{figure}
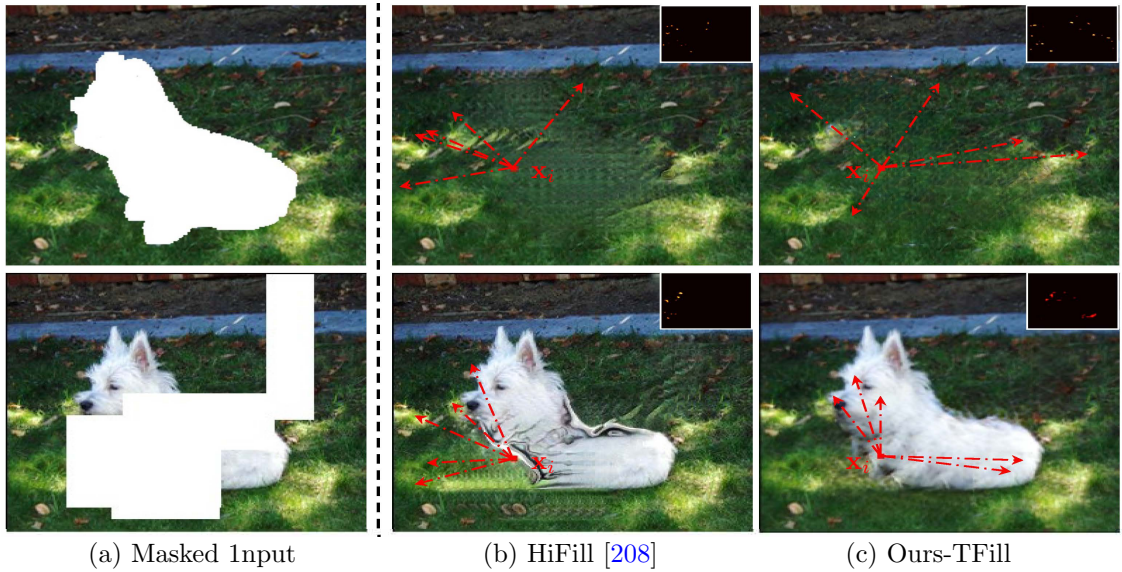

\section{Motivation}\label{ch5:intro}

In Section \ref{ch4:intro}, we introduce how expert conservators would restore damaged art, where the first and most important step is to imagine the semantic content to be filled based on the overall visible scene. However, \emph{bridging and exploiting visible information globally, after it had been degraded by arbitrary masks} remains a main challenge in this task. As depicted in Figure~\ref{fig:ch4_example} top row, when the entire dog is masked, the natural expectation is to complete the masked area based on the visible background context. In contrast, in the bottom row, when the free-form regular mask covers the bulk of the dog but leaves the head and tail visible, it is necessary but highly challenging to globally capture \emph{long-range} dependencies between the two separated foreground regions, so that the masked area can be completed in not just a photorealistic, but also semantically correct, manner.

To achieve this goal, many \emph{two-stage} approaches \cite{yu2018generative,Nazeri_2019_ICCV,yi2020contextual,zeng2020image} have been proposed, consisting of a \emph{content inference network} and an \emph{appearance refinement network}. They typically infer a coarse image or edge/semantic map based on globally visible information in a first phase, and then fill in visually realistic appearance in a second phase. However, this global perception is achieved by repeated \emph{local} convolutional operations, which have several limitations. First, due to translation equivariance in convolutions, the information flow tends to be predominantly local, with global information only shared gradually through heat-like propagation across multiple layers. Second, during inference, the elements between adjacent layers are connected via learned but fixed weights, rather than input-dependent adaptive weightings. These issues mean long-distance messages are only delivered inefficiently in a very deep layer, resulting in a strong inclination for the network to fill holes based on nearby rather than distant visible pixels (Figure~\ref{fig:ch5_example} (b)).

In this chapter, we propose an alternative perspective by treating image completion as a \emph{directionless sequence-to-sequence} prediction task. In particular, instead of modeling the global context using deeply stacked convolutional layers, we design a new content inference model, called TFill, that uses a \textbf{T}ransformer-based architecture to \textbf{Fill} reasonable content into the missing holes. An important insight here is that a transformer directly exploits long-range dependencies at every encoder layer through the attention mechanism, which \emph{creates an equal flowing opportunity for all visible pixels, regardless of their relative spatial positions}. This reduces the proximity-dominant influence that can lead to semantically incoherent results.

Our design is motivated by the transformer literature in natural language processing (NLP) \cite{vaswani2017attention,devlin2018bert,radford2018improving,radford2019language}. However, it remains a challenge to directly apply these transformer models to visual generation tasks. Particularly, unlike the NLP that naturally treats each word as a vector for token embedding, it is unclear what a good token representation should be for the visual task. If we use every pixel as a token, the memory cost will make this infeasible except for very small images \cite{chen2020generative}. To mitigate this issue, our model embeds the masked image into an intermediate latent space for token representation, an approach also broadly taken by recent vision transformer models \cite{carion2020end,zhu2020deformable,esser2020taming,SETR}. However, unlike these models that use traditional CNN-based encoders to embed the tokens, we propose a \emph{restrictive CNN} for token representation, which has a profound influence on how the visible information is connected in the network. To do so, we ensure the individual tokens represent visible information independently, each with a \emph{small} and \emph{non-overlapping} receptive field (RF). This forces \emph{the long-range context relationships between tokens to be explicitly and co-equally perceived in every transformer encoder layer}, without neighboring tokens being entangled by implicit correlation through overlapping RF. As a result, each token will \emph{not} be gradually affected by neighboring regions, retaining equal probability of being captured in every layer. 

While the proposed transformer-based architecture can achieve better results than state-of-the-art methods \cite{yu2018generative,zheng2019pluralistic,yi2020contextual,esser2020taming}, by itself it only works for a fixed sequence length because of the position embedding (Figure~\ref{fig:ch5_framework}(a)). To allow our approach to flexibly scale to images of different sizes, a fully convolutional encoder-decoder network (Figure~\ref{fig:ch5_framework}(b))) is subsequently applied to refine the visual appearance built upon the coarse content previously inferred.
We also design a novel \textbf{A}ttention-\textbf{A}ware \textbf{L}ayer (AAL) between the encoder and decoder that adaptively balances the attention paid to visible and generated content, leading to semantically superior feature transfer. 

\section{Background}\label{ch5:back}

\subsection{Image Completion}

As this related research has been discussed above, we refer readers to Section \ref{ch4:back}.

\subsection{The Transformer Family}

The transformer architecture is first proposed by Vaswani \etal \cite{vaswani2017attention}, and has later become the de facto standard backbone in NLP tasks \cite{devlin2018bert,radford2018improving,radford2019language}. It merges the information between the inputs solely through attention \cite{bahdanau2014neural}, which directly models the long-range dependence by calculating the similarity of two points, regardless of their spatially relative position. 

The early application of attention in computer vision is only in one deep layer to learn the long-range dependence, such as the attention in Non-local Neural Networks \cite{wang2018non} and self-attention GAN \cite{zhang2019self}. This is because the 2D image has a quadratic cost in the number of pixels than the 1D sequence sentence with words. It will require complex engineering and high GPU memory to implement the vision transformer efficiently on hardware accelerators. 

To mitigate this issue, recent works have explored different visual token representation methods to directly apply a standard transformer for visual tasks, such as image classification \cite{chen2020generative,dosovitskiy2020image}, object detection \cite{carion2020end,zhu2020deformable}, semantic segmentation \cite{wu2020visual,SETR}, image generation and translation \cite{esser2020taming, chen2020pre,hudson2021gansformer,jiang2021transgan}. As illuminated in Section \ref{ch5:appro}, compared to these general token representations, our \emph{restrictive CNN} is particularly well suited due to its compact representation that limits implicit correlation. 

\begin{figure}[tb!]
    \centering
    \includegraphics[width=\linewidth]{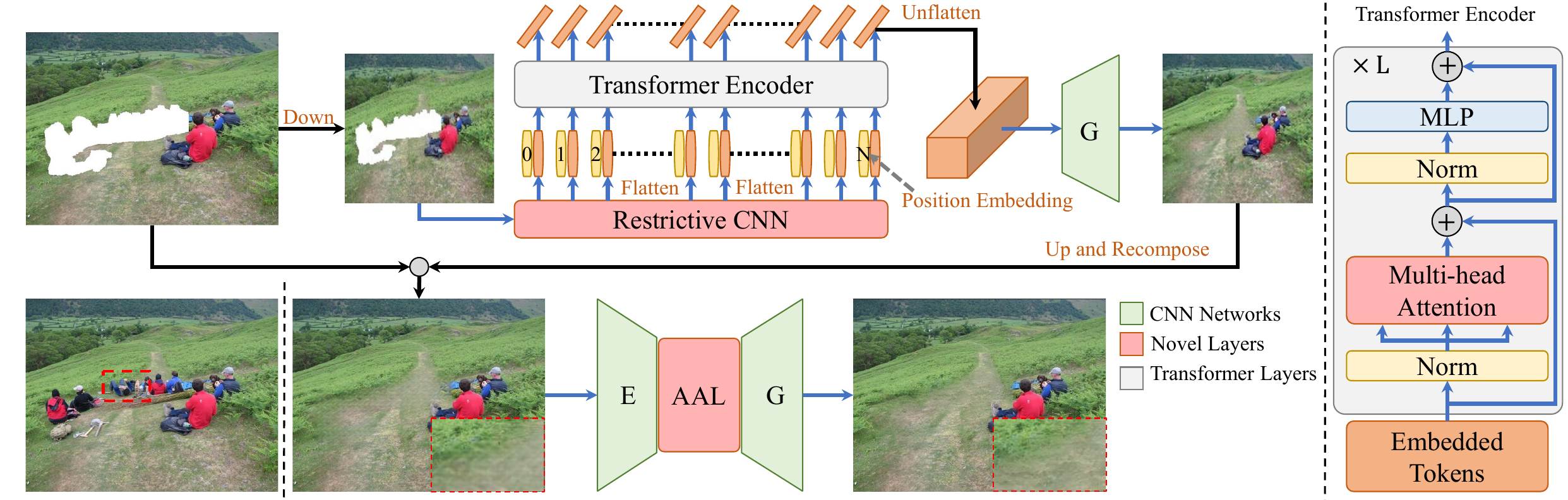}
    \begin{picture}(0,0)
    \put(-190,143){\tiny Masked Input}
    \put(-136, 136){\tiny Fixed Input ($256^2$)}
    \put(82, 136){\tiny Coarse Output ($256^2$)}
    \put(-216,90){\rotatebox{90}{\footnotesize (a) Coarse}}
    \put(-216,18){\rotatebox{90}{\footnotesize (b) Refine}}
    \put(-180, 7){\tiny Original}
    \put(-127, 7){\tiny Recomposed Input}
    \put(30, 7){\tiny Refined Output}
    \end{picture}
    \caption[The overall pipeline of the proposed method]{\textbf{The overall pipeline of the proposed method.} (a) Masked input is resized to a fixed low resolution ($256^2$) and it is then fed into the transformer to generate semantically correct content. (b) The inferred content is merged with the original high-resolution image and passed to a refinement network with an \textbf{A}ttention-\textbf{A}ware \textbf{L}ayer (\textbf{AAL}) to transfer high-quality information from both visible and masked regions. Note the recomposed input has repeating artifacts, which are resolved in our refined network. Zoom in to see the details.}
    \label{fig:ch5_framework}
\end{figure}
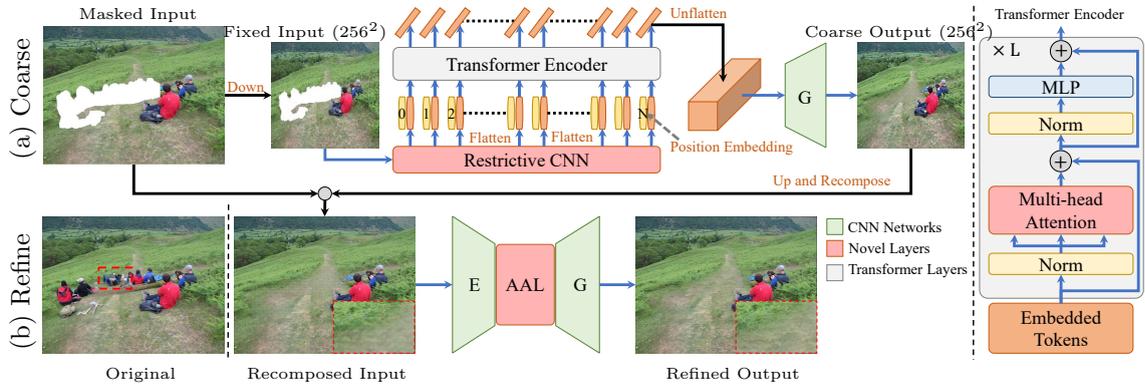

\section{Approach}\label{ch5:appro}

Given a masked image $\textbf{I}_m$, degraded from a real image $\textbf{I}$ by a free-form mask, our goal is to learn a model $\Phi$ to infer the content for missing regions, as well as filling in with visually realistic appearance. To achieve this, our image completion framework, illustrated in Figure~\ref{fig:ch5_framework}, consists of a content inference network and an appearance refinement network. The former is responsible for capturing the global context through a transformer encoder at a fixed scale. The embedded tokens have small receptive fields (RF) and limited capacity, preventing their states from being implicitly dominated by visible pixels nearby than far. While similar transformer-based architectures have recently been explored for visual tasks \cite{chen2020generative,dosovitskiy2020image,carion2020end,zhu2020deformable,esser2020taming,wu2020visual,chen2020pre,SETR}, we believe our work is the first to explore this for free-form image completion, where we discover \emph{how the token representation has a profound effect on the flow of visible information in the network, in spite of the supposedly global reach of transformers}. The latter network is designed to refine visual appearance by utilizing high-resolution visible features, and also frees the limitation to fixed image sizes.

\subsection{Transformer-based Architecture}

\paragraph{Background} We begin by briefly reviewing the transformer \cite{vaswani2017attention}. As depicted on Figure~\ref{fig:ch5_framework}~(right), a transformer encoder layer consists of multihead \texttt{self-attention} (MSA) and \texttt{Multi layer Perception} (MLP) blocks (see Appendix \ref{app:sec:ch5_msa}). The MSA is responsible for capturing  long-range dependencies, while the MLP is applied to further transform merged features. The \texttt{LayerNorm} (LN) is used for non-linear projection. These are expressed by:
\begin{alignat}{3}
\textbf{z}_0 & = [\mathbf{x}^1;\mathbf{x}^2;\dots;\mathbf{x}^N] + \mathbf{E}_{pos} \\
\textbf{z}^\prime_\ell & = \text{MSA}(\text{LN}(\textbf{z}_{\ell-1})) + \textbf{z}_{\ell-1} \\
\textbf{z}_\ell & = \text{MLP}(\text{LN}(\textbf{z}^\prime_\ell)) + \textbf{z}^\prime_\ell
\end{alignat}
where $\textbf{z}\in\mathbb{R}^{N\times C}$ is the 1D sequence of $N$ tokens $\textbf{x}$ with $C$ channels, and $\mathbf{E}_{pos}\in\mathbb{R}^{N\times C}$ is the position embedding. 

\paragraph{Transformer-Encoder} In order to feed a 2D masked image $\textbf{I}_m$ into the transformer, we first downsample the high-resolution image to a fixed size, \eg $256^2$. However, it is \emph{not} feasible to run the transformer model if we directly \emph{flatten} image pixels into a 1D sequence with 196,608 tokens. To achieve independent token representation and reduce its length, a projection is implemented using our proposed \emph{restrictive CNN}, a decision we will analyze in Section \ref{sec:ch5_analysis-trans}. After that, we obtain a 2D feature map with size $\frac{256}{16}$$\times$$\frac{256}{16}$$\times$$C$, and then flatten it to a 1D sequence of $256$$\times$$C$, where $256$ is the sequence length and $C$=512 is the feature dimension. As shown in Figure~\ref{fig:ch5_framework} (a), once we embed the image to a 1D sequence, a transformer encoder distills long-range relationships between all tokens in every layer. 

To encourage the model to \emph{bias} to the important visible values, we replace the self-attention layer with the \emph{masked} self-attention layer, in which a weight is applied to scale the attention scores. The initial weight $w_{key}\in(0.02,1.0]$ is obtained by calculating the fraction of visible pixels in a small RF, \eg $192/16^2$ means $3/4$ of the region in the $16^2$ RF contains visible pixels. It will then be gradually amplified by updating $w_{key}$$\gets$$\sqrt{w_{key}}$ after every encoder layer, to \emph{reflect} visible information flow. This initial ratio for each token is efficiently implemented in our restrictive CNN encoder using a modified partial convolution layer \cite{Liu_2018_ECCV}. The implementation details can be found in Appendix \ref{app:sec:ch5_experiment}. 

\paragraph{CNN-based Decoder} While a one-layer non-linear projection may be used to directly map the output features back to a completed image, the visual appearance is slightly worse than using a stacked decoder. Therefore, following existing works \cite{yu2018generative,zheng2019pluralistic,yi2020contextual}, a gradual upsampling decoder is implemented to generate photorealistic images. 

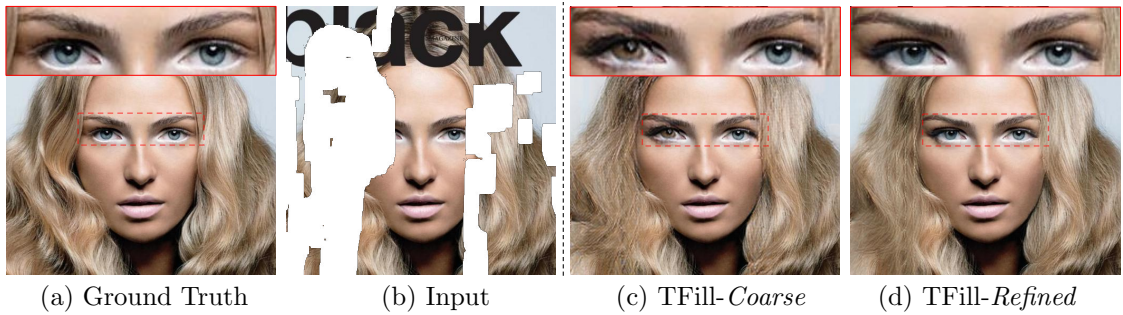
\begin{figure}[tb!]
    \centering
    \includegraphics[width=\linewidth]{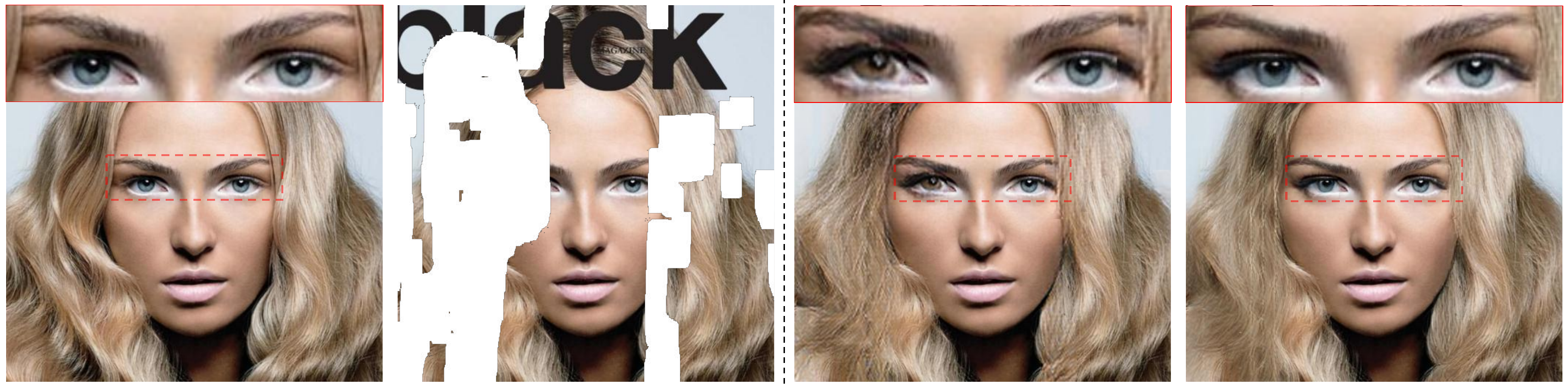}
    \begin{picture}(0,0)
    \put(-196, 4){\footnotesize (a) Ground Truth}
    \put(-68, 4){\footnotesize (b) Input}
    \put(20, 4){\footnotesize (c) TFill-\emph{Coarse}}
    \put(118, 4){\footnotesize (d) TFill-\emph{Refined}}
    \end{picture}
    \caption[Coarse and Refined results]{\textbf{Coarse and Refined results}. (a) Ground truth. (b) Masked input degraded by free-form masks. (c) Coarse output. (d) Refined output. We can see that the refinement network not only increased image quality to a high resolution ($256^2$ \emph{vs} $512^2$), but also encourages the left eyeball to be consistent with the visible right eyeball using our attention-aware layer.}
    \label{fig:ch5_attn-example}
\end{figure}

\subsection{Attention-Aware Layer (AAL)}\label{sec:ch5_attn-aware}

Although our TFill-\emph{Coarse} model correctly infers reasonable content by equally utilizing the global visible information in every layer, two limitations remain. First, it is \emph{not} suitable for high-resolution input due to the fixed length position embedding. One solution is to follow the directional sequence-to-sequence methods \cite{chen2020generative,esser2020taming} that only use the top-left context to predict the next token, in an auto-regressive manner. However, this will not adequately capture the global visible information needed for image completion. Second, the realistic completed results may not be fully consistent with the original visible appearances, \eg the generated left eye having a different shape and color to the visible right eye in Figure~\ref{fig:ch5_attn-example} (c). This is because the embedded tokens are extracted from a $16^2$ resolution feature map, where important high-frequency details may be lost. 

To mitigate these issues, a CNN-based encoder-decoder refinement network, trained on high-resolution images, is proposed (Figure~\ref{fig:ch5_framework} (b)). In particular, to further utilize the visible high-frequency details, an \textbf{A}ttention-\textbf{A}ware \textbf{L}ayer (AAL) is designed to capture long-range dependencies.

\begin{figure}[tb!]
    \centering
    \includegraphics[width=\linewidth]{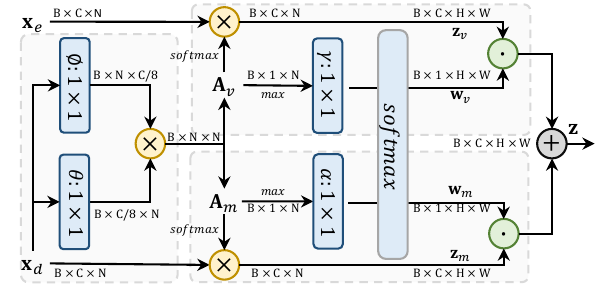}
    \caption[Attention-aware layer]{\textbf{Attention-aware layer.} The feature maps are shown as tensors. ``$\bigotimes$'' denotes matrix multiplication, ``$\bigodot$'' denotes element-wise multiplication and ``$\bigoplus$'' is element-wise sum. The blue boxes denote $1\times1$ convolution filters that are learned.}
    \label{fig:ch5_attn-layer}
\end{figure}

As depicted in Figure~\ref{fig:ch5_attn-layer}, given a decoded feature ${\bf x}_d$, we first calculate the attention score of:
\begin{equation}
    {\bf A} = \phi({\bf x}_d)^\intercal \theta({\bf x}_d)
\end{equation}
where ${\bf A}_{ij}$ represents the similarity of the $i^{\text{th}}$ patch to the $j^{\text{th}}$ patch, and $\mathbf{\phi}$, $\mathbf{\theta}$ are $1$$\times$$1$ convolution filters.

Interestingly, we discover that using ${\bf A}$ directly in a standard self-attention layer is suboptimal, because the ${\bf x}_d$ features for visible regions are generally distinct from those generated for masked regions. Consequently, \emph{the attention tends to be insular}, with masked regions preferentially attending to masked regions, and vice versa. To avoid this problem, we explicitly handled the attention to visible regions separately from masked regions. So before \texttt{softmax} normalization, ${\bf A}$ is split into two parts: ${\bf A}_v$ --- similarity to visible regions, and ${\bf A}_m$ --- similarity to generated masked regions. Next, we get long-range dependencies via:
\begin{equation}
    \begin{matrix}
    {\bf z}_v = \texttt{\small softmax}({\bf A}_v){\bf x}_e
    & ,
    & {\bf z}_m = \texttt{\small softmax}({\bf A}_m){\bf x}_d
    \end{matrix}
\end{equation}
where ${\bf z}_v$ contains features of contextual flow \cite{yu2018generative} for copying high-frequency details from the encoded high-resolution features ${\bf x}_e$ to masked regions, while ${\bf z}_m$ has features from the self-attention that is used in SAGAN \cite{zhang2019self} for high-quality image generation. 

Instead of learning fixed weights \cite{zheng2019pluralistic} to combine ${\bf z}_v$ and ${\bf z}_m$, we learn the \emph{weights mapping} based on the largest attention score in each position. Specifically, we first obtain the largest attention score of ${\bf A}_v$ and ${\bf A}_m$, respectively. Then, we use the $1$$\times$$1$ filter $\gamma$ and $\alpha$ to \emph{modulate} the ratio of the weights. \texttt{Softmax} normalization is applied to ensure ${\bf w}_v$$+$${\bf w}_m$$=$$1$ in every spatial position:
\begin{equation}
    [{\bf w}_v, {\bf w}_w] = \texttt{\small softmax}([\gamma(\texttt{\small max}({\bf A}_v)),\alpha(\texttt{\small max}({\bf A}_m)]))
\end{equation}
where \texttt{\small max} is executed on the attention score channel. Finally,  an attention-balanced output ${\bf z}$ is obtained by:
\begin{equation}
    {\bf z} = {\bf w}_v \cdot {\bf z}_v +  {\bf w}_m \cdot {\bf z}_m
\end{equation}
where ${\bf w}_v, {\bf w}_m \in\mathbb{R}^{B\times 1\times H\times W}$ hold different values for various positions, dependent on the largest attention scores in the visible and masked regions, respectively. 

\subsubsection{Discussion on prior art}

While contextual attention \cite{yu2018generative} has recently been widely applied in image completion  \cite{yu2018generative,song2018contextual,Yan_2018_ECCV,yi2020contextual}, it is fundamentally different from the attention in our transformer-based architecture --- the contextual attention is used to refine visual appearance by copying high-frequency information from visible regions to masked holes, rather than capturing and modeling long-range context for content inference. In addition, our AAL focuses on automatically selecting features from both visible and generated features, instead of copying only from visible regions \cite{yu2018generative,song2018contextual,Yan_2018_ECCV,yi2020contextual} or selecting through fixed weights \cite{zheng2019pluralistic}. 

\section{User Interface}\label{ch5:user}

\begin{figure}
    \centering
    \includegraphics[width=\linewidth]{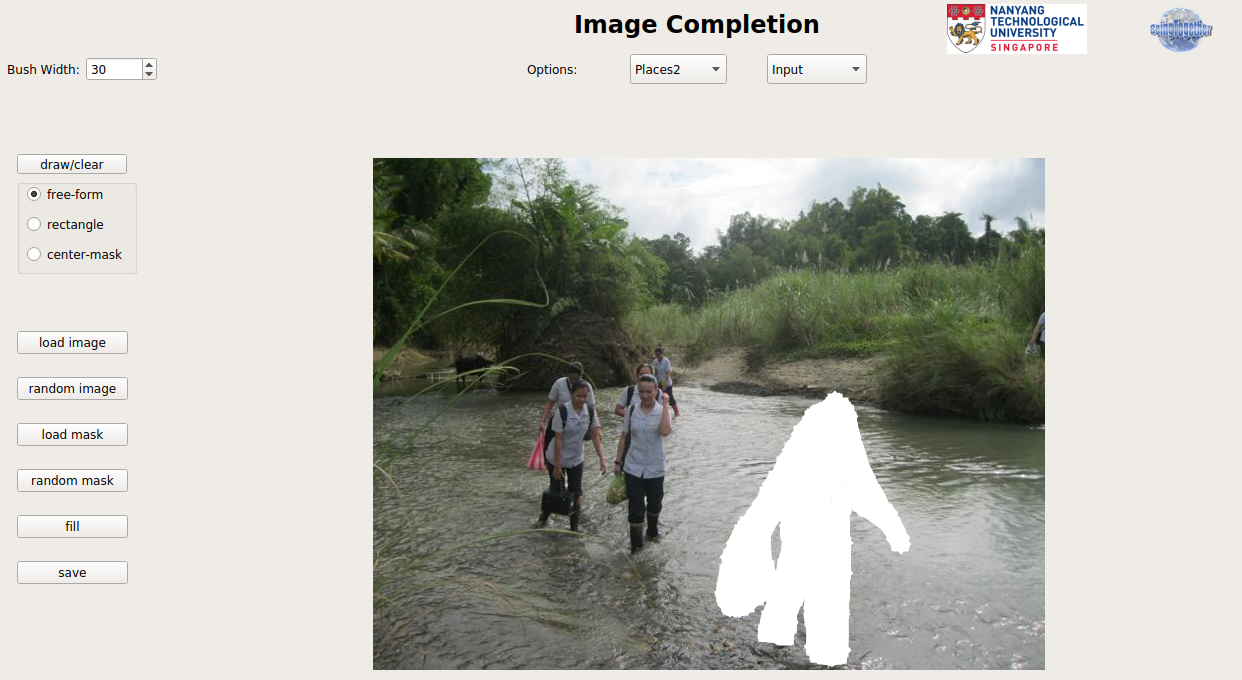}
    \caption[Local interface for high-resolution image editing]{\textbf{Local interface for free-form high-resolution image editing.} }
    \label{fig:ch5_local_interface}
\end{figure}

I designed a real-time interactive system that allows the user to easily explore and edit the high-resolution image by creating input masks. As shown in Figure \ref{fig:ch5_local_interface}, this user interface is built upon the interface in Section \ref{ch4:user}. Here, the resolution of the input image can be in multiples of $2^5=32$, \eg $960\times640$, instead of the fixed resolution ($256\times256$) in Chapter \ref{ch:PICNet}. In addition, two buttons, ``load mask'' and ``random mask'', are added to load the free-form masks provided by Liu \etal \cite{Liu_2018_ECCV}. In this way, the user can directly load the free-form masks to assess the robustness of various methods. 

\section{Results and Applications}\label{ch5:experiment}

\paragraph{Datasets} We evaluated our TFill with arbitrary mask types on various datasets, including CelebA-HQ \cite{liu2015deep,karras2017progressive}, FFHQ \cite{karras2019style}, Places2 \cite{zhou2018places}, and ImageNet \cite{russakovsky2015imagenet}.

\paragraph{Metrics} As proposed in previous works \cite{yu2018generative,zheng2019pluralistic}, it is not reasonable to require the completed image to be exactly the same as the original image. Hence, we only report the LPIPS \cite{zhang2018unreasonable} and the FID \cite{heusel2017gans} scores in the main text, leaving the traditional pixel- and patch-level evaluation results, \eg the mean $\ell_1$ loss, in Appendix \ref{app:sec:ch5_ana}.

\paragraph{Implementation details} Our model is trained in two stages: \textbf{1)} the content inference network is first trained for $256^2$ resolution; and \textbf{2)} the visual appearance network is then trained for $512^2$ resolution. Both networks are optimized using the loss $L = L_{pixel} + L_{per} + L_{GAN}$, where $L_{pixel}$ is the $\ell_1$ reconstruction loss, $L_{per}$ is the perceptual loss \cite{johnson2016perceptual}, and $L_{GAN}$ is the discriminator loss \cite{goodfellow2014generative}. 

\subsection{Comparison with Existing Work}

Here we compared with these image completion methods: \textbf{PM} \cite{barnes2009patchmatch}, a classical approach; \textbf{GL} \cite{iizuka2017globally}, the first learning-based method for arbitrary regions; \textbf{CA} \cite{yu2018generative}, the first method combining learning and patch-based methods; ours \textbf{PICNet} \cite{zheng2019pluralistic} in Chapter \ref{ch:PICNet}, the first work considering multiple solutions; \textbf{HiFill} \cite{yi2020contextual}, the latest very high-resolution (8K) method. Our TFill introduces a transformer-based architecture for this challenging image completion problem.

\begin{table}[tb!]
    \centering
    \footnotesize
    \renewcommand{\arraystretch}{1.2}
    \setlength\tabcolsep{6pt}
    \begin{tabular}{@{}l|c|ccccc@{}}
        \hlineB{3.5}
         &  \textbf{Size} &  {\bf GL} \cite{iizuka2017globally}  & {\bf CA} \cite{yu2018generative} & {\bf PICNet} \cite{zheng2019pluralistic}  & {\bf HiFill}  \cite{yi2020contextual} & \textbf{TFill}\\
         \hlineB{2}
        \multirow{6}{1em}{\rotatebox[origin=c]{90}{LPIPS}} & [0.01, 0.1] & 0.057 & 0.083 & 0.037 & 0.056 &  \bf{0.027} \\
        & (0.1, 0.2] & 0.112 & 0.134 & 0.074 & 0.105 & \bf{0.055} \\
        & (0.2, 0.3] & 0.185 & 0.195 & 0.118 & 0.163 &  \bf{0.092} \\
        & (0.3, 0.4] & 0.254 & 0.249 & 0.167 & 0.226 &  \bf{0.133} \\
        & (0.4, 0.5] & 0.319 & 0.306 & 0.225 & 0.305 &  \bf{0.180} \\
        & (0.5, 0.6] & 0.370& 0.364 & 0.330 & 0.412 &  \bf {0.259} \\
        \hline
        \multirow{6}{1em}{\rotatebox[origin=c]{90}{FID}} & [0.01, 0.1] & 16.86 & 10.21 & 7.04 & 9.10 &  \bf{5.22}\\
        & (0.1, 0.2] & 26.11 & 18.93 & 13.58 & 16.72 & \bf{9.67} \\
        & (0.2, 0.3] & 39.22 & 30.31 & 21.62 & 26.89 & \bf{15.28} \\
        & (0.3, 0.4] & 53.24 & 40.29 & 29.59 & 38.40 & \bf{19.99} \\
        & (0.4, 0.5] & 68.46 & 53.39 & 41.60 & 56.24 & \bf{25.88 }  \\
        & (0.5, 0.6] & 74.95 & 59.85 & 61.17 & 83.36 &  \bf{34.58} \\
        \hlineB{2}
    \end{tabular}
    \caption[Quantitative comparisons on Places2]{Quantitative comparisons on Places2~\cite{zhou2018places} with free-form masks \cite{Liu_2018_ECCV}. Without bells and whistles, TFill outperformed all traditional CNN-based models. The results are reported on $256^2$ resolution, as earlier works were trained only on this scale. }
    \label{tab:ch5_SOTA_comp}
\end{table}

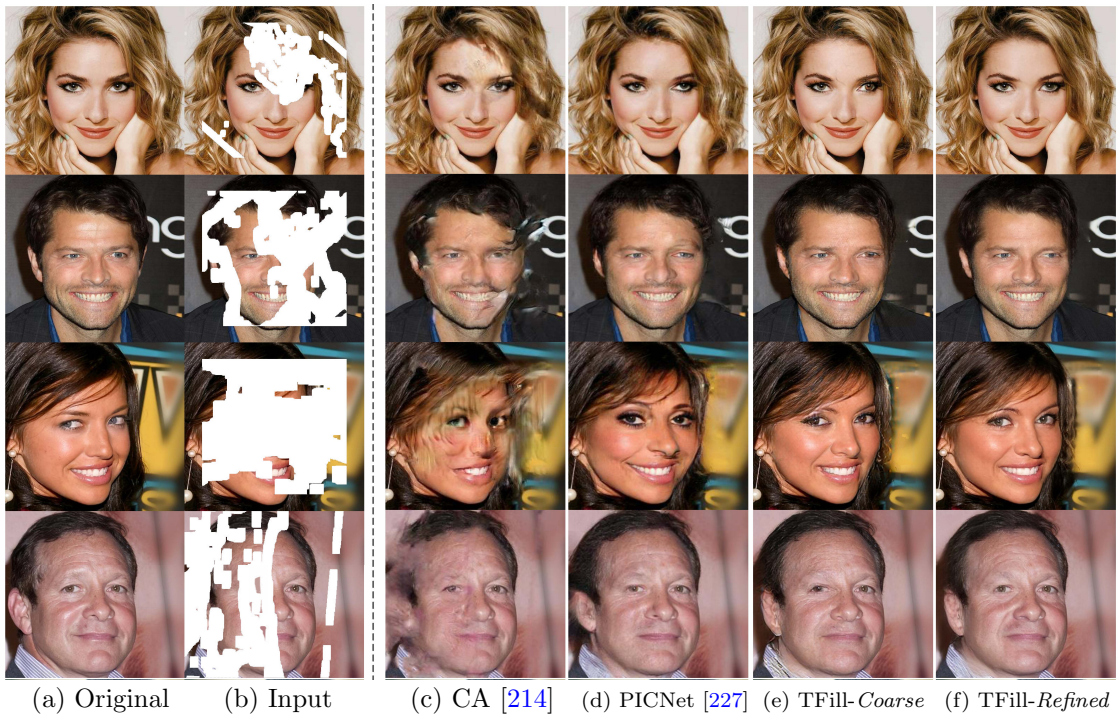
\begin{figure}[tb!]
    \centering
    \includegraphics[width=\linewidth,height=0.38\textheight]{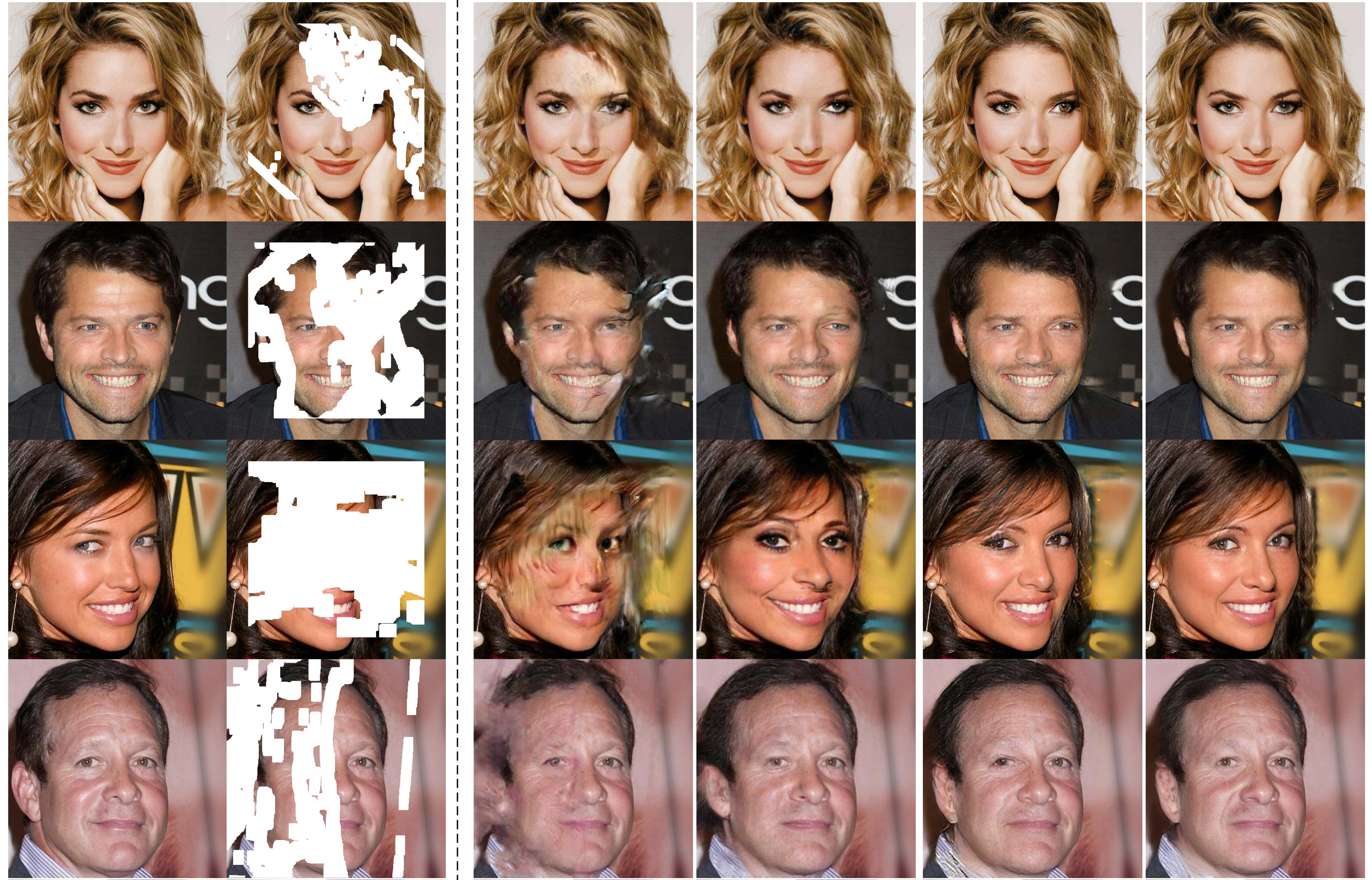}
    \begin{picture}(0,0)
    \put(-198,4){\footnotesize (a) Original}
    \put(-126,4){\footnotesize (b) Input}
    \put(-56,4){\footnotesize (c) CA~\cite{yu2018generative}}
    \put(8,4){\scriptsize (d) PICNet~\cite{zheng2019pluralistic}}
    \put(76,4){\scriptsize (e) TFill-\emph{Coarse}}
    \put(144,4){\scriptsize (f) TFill-\emph{Refined}}
    \end{picture}
    \caption[Completion results on CelebA-HQ testing set]{\textbf{Completion results on CelebA-HQ \cite{liu2015deep,karras2017progressive} testing set among CA~\cite{yu2018generative}, PICNet~\cite{zheng2019pluralistic} and Ours.} Our results are reported for $512^2$ resolution. While ours previous PICNet~\cite{zheng2019pluralistic} works well for frontal facing faces, it may generate more uncanny faces with mismatched features at larger angles, \eg the examples in third and last row.}
    \label{fig:ch5_sup-free-form-face}
\end{figure}

\begin{figure}[tb!]
    \centering
    \includegraphics[width=\linewidth,height=0.31\textheight]{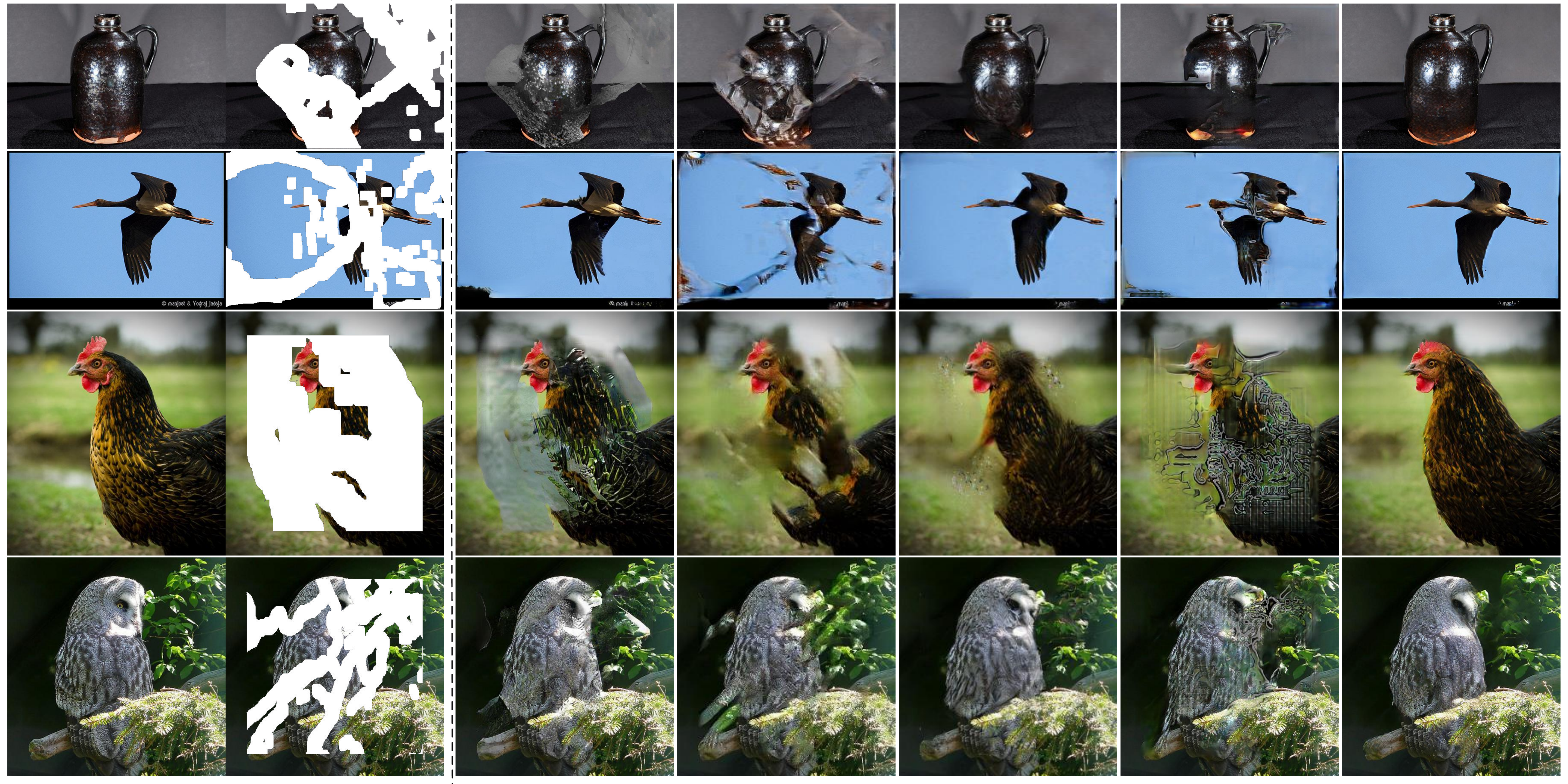}
    \begin{picture}(0,0)
    \put(-205,4){\footnotesize (a) Original}
    \put(-142,4){\footnotesize (b) Input}
    \put(-84,4){\footnotesize (c) GL~\cite{iizuka2017globally}}
    \put(-26,4){\footnotesize (d) CA~\cite{yu2018generative}}
    \put(30,4){\scriptsize (e) PICNet~\cite{zheng2019pluralistic}}
    \put(92,4){\scriptsize (f) HiFill~\cite{yi2020contextual}}
    \put(148,4){\scriptsize (g) TFill-\emph{Refined}}
    \end{picture}
    \caption[Completion results on ImageNet testing set]{\textbf{Completion results on ImageNet \cite{russakovsky2015imagenet} testing set among GL~\cite{iizuka2017globally}, CA~\cite{yu2018generative}, PICNet~\cite{zheng2019pluralistic}, HiFill~\cite{yi2020contextual} and Ours.} Our TFill model generated better visual results even under very challenging situations, \eg the heavily masked chicken in the second last row.}
    \label{fig:ch5_sup-free-form-imagenet}
\end{figure}
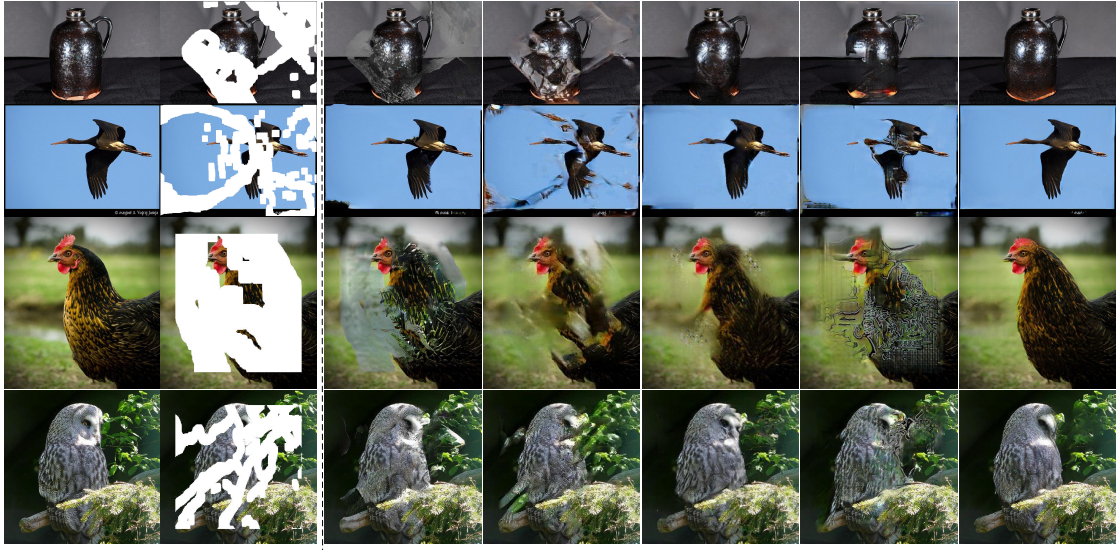

Table \ref{tab:ch5_SOTA_comp} shows quantitative evaluation results on Place2 \cite{zhou2018places}, in which the images were degraded by free-form masks provided in the PConv \cite{Liu_2018_ECCV} testing set. The size column denotes the range of masking proportion applied to the images. We observe that our transformer-based model improved both LPIPS and FID quite significantly over the CNN-based state-of-the-art models in all mask scales. Specifically, it achieves relative $27\%$ and $21\%$ improvement for LPIPS at scales of [0.01, 0.1] and (0.5, 0.6], respectively. Furthermore, our completed images form closer distributions to the real testing set, with FID scores averaging  $32\%$ relative improvement on all mask scales.

\begin{figure}[tb!]
    \centering
    \includegraphics[width=\linewidth,height=0.44\textheight]{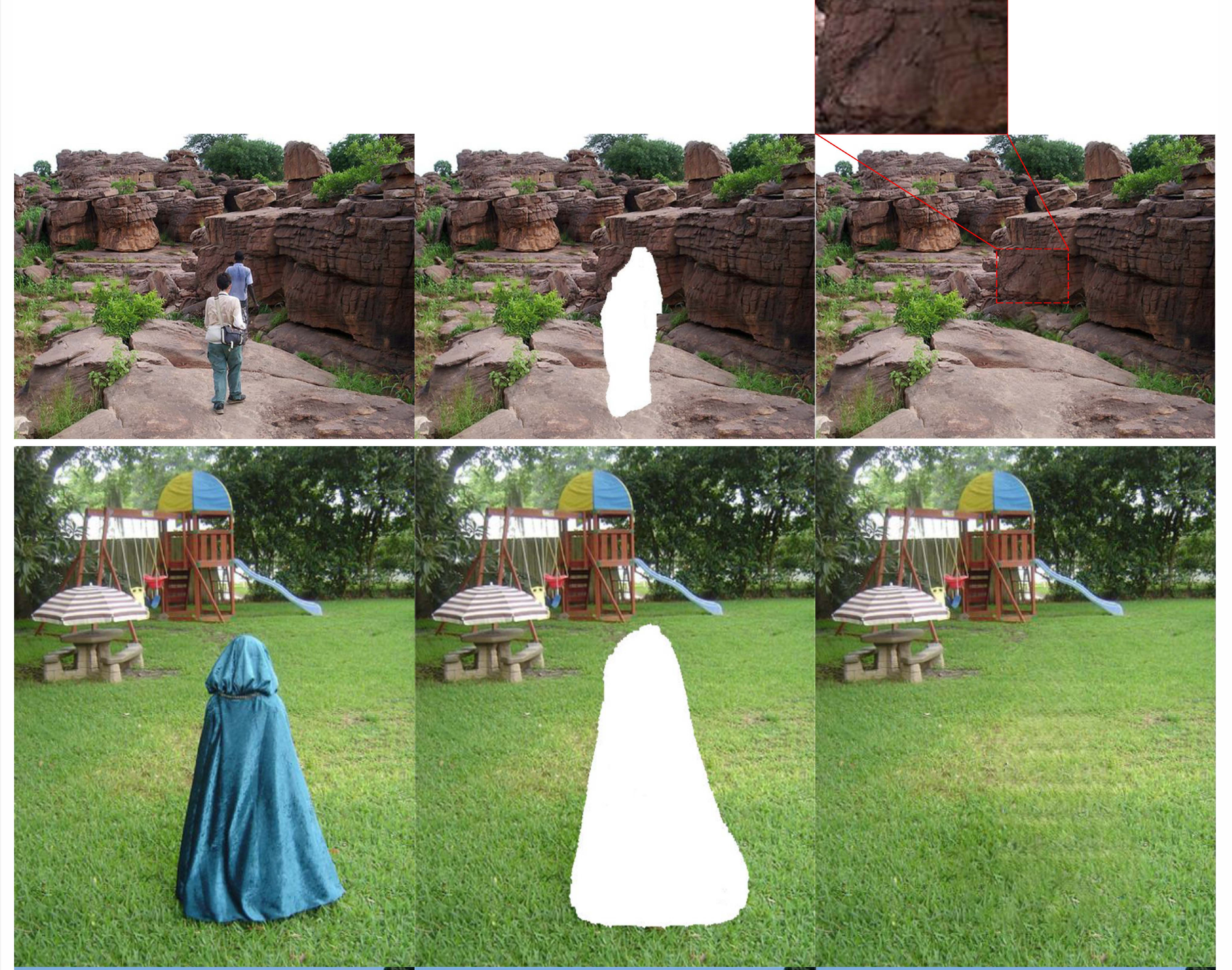}
    \begin{picture}(0,0)
    \put(-160,4){\footnotesize (a) Original}
    \put(-20,4){\footnotesize (b) Input}
    \put(96,4){\footnotesize (c) Edited Output}
    \end{picture}
    \caption[Free-form editing results on ImageNet]{\textbf{Free-form editing results on ImageNet \cite{russakovsky2015imagenet}.} }
    \label{fig:ch5_sup-free-edit-imagenet0}
\end{figure}
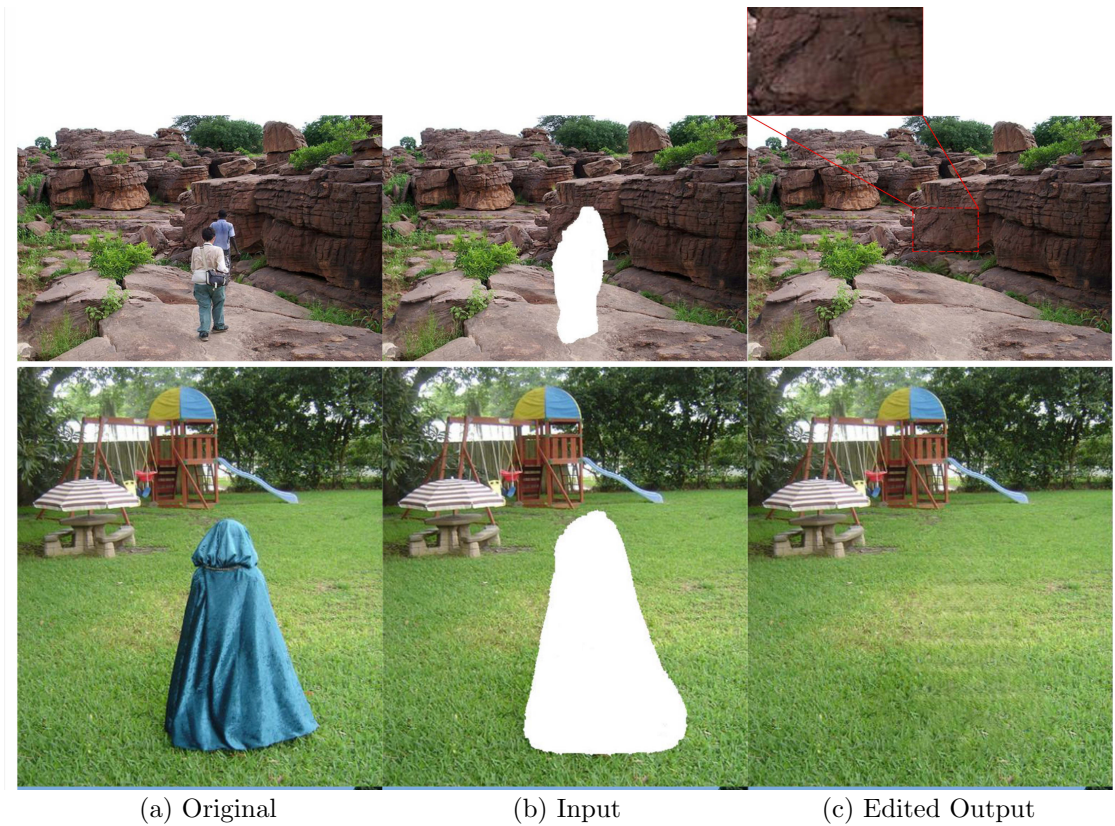

The qualitative comparisons are visualized in Figures \ref{fig:ch5_sup-free-form-face} and \ref{fig:ch5_sup-free-form-imagenet}. TFill achieved superior visual results even under challenging conditions. In Figure \ref{fig:ch5_sup-free-form-face}, we compare with CA and our previous PICNet trained on CelebA-HQ dataset. Our TFill generates photorealistic high-resolution ($512^2$) results, even when significant semantic information is missing due to large free-form masks. Figure \ref{fig:ch5_sup-free-form-imagenet} shows visual results on natural images that were degraded by random masks. GL and CA, while good at object removal, failed to infer shapes needed for object completion. The PICNet proposed in the last chapter produced multiple diverse results in which some shapes were correct but of limited quality. Our results are evaluated in higher resolution, with the short side at $512$ pixels and the long side at multiples of $2^5$, \eg 640. Our TFill model generated better visual results even under very challenging situations, \eg the heavily masked chicken in the second last row.

\begin{figure}[tb!]
    \centering
    \includegraphics[width=\linewidth]{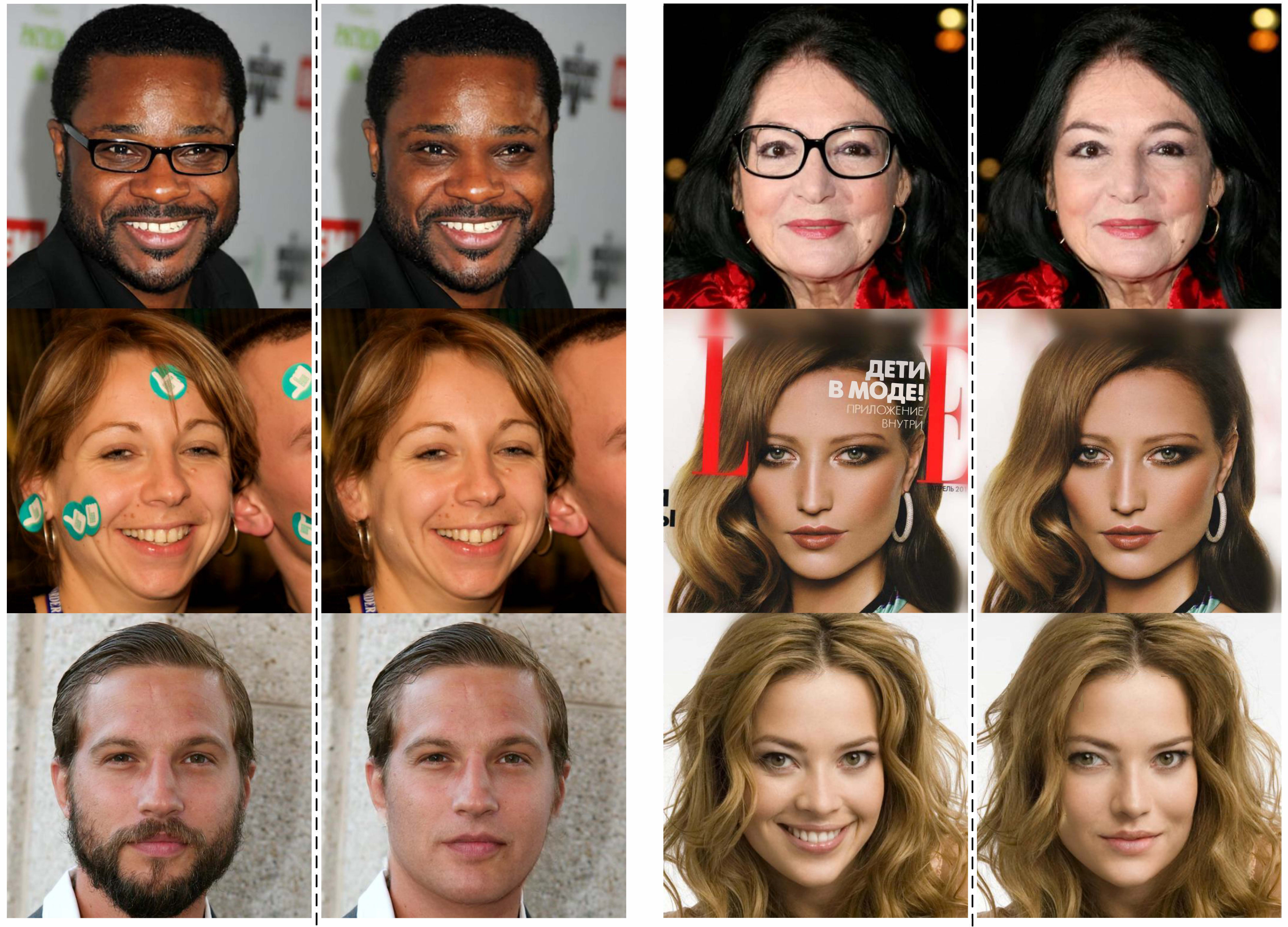}
    \begin{picture}(0,0)
    \put(-178,4){\footnotesize (a) Original}
    \put(-96,4){\footnotesize (b) Edited Output}
    \put(32,4){\footnotesize (a) Original}
    \put(118,4){\footnotesize (b) Edited Output}
    \end{picture}
    \caption[Qualitative results on CelebA-HQ and FFHQ testing set for free-form mask editing]{\textbf{Qualitative results on CelebA-HQ \cite{liu2015deep,karras2017progressive} and FFHQ \cite{karras2019style} testing set for free-form mask editing.} All results are reported at $512^2$ resolution.}
    \label{fig:ch5_sup-free-edit-face}
\end{figure}

\begin{table}[tb!]
    \centering
    \renewcommand{\arraystretch}{1.1}
    \setlength\tabcolsep{4pt}
    \begin{tabular}{@{}llccccc@{}}
         \hlineB{3.5}
         & \multirow{2}{*}{\textbf{Method}} & \multicolumn{2}{c}{\textbf{CelebA-HQ}}&& \multicolumn{2}{c}{\textbf{FFHQ}}\\
		\cline{3-4}\cline{6-7}
		& & LPIPS$\downarrow$& FID$\downarrow$ && LPIPS$\downarrow$ & FID$\downarrow$\\
		\hlineB{2}
		& CA~\cite{yu2018generative} & 0.104 & 9.53 &  & 0.127 & 8.78\\
		& PICNet~\cite{zheng2019pluralistic} & 0.061 & 6.43 & & 0.068 & 4.61\\
		& MEDFE~\cite{Liu2019MEDFE} & 0.067 & 7.01 & & - & - \\
		\cdashline{1-7}
		$\mathrm{A}$ & Traditional \emph{Conv} & 0.060 & 6.29 && 0.066 & 4.12 \\
		$\mathrm{B}$ & + Attention in G & 0.059 & 6.34 & & 0.064 & 4.01 \\
		$\mathrm{C}$ & + Restrictive \emph{Conv} &  0.056 & 4.68 && 0.060 & 3.87 \\
		$\mathrm{D}$ & + Transformer & 0.051 & 4.02 && 0.057 & 3.66\\
		$\mathrm{E}$ & + Masked Attention &  0.050 & 3.92 && 0.057 & 3.63\\
		$\mathrm{F}$ & + Refine Network & \textbf{0.048} & \textbf{3.86} & & \textbf{0.053} & \textbf{3.50} \\
        \hlineB{2}
    \end{tabular}
    \caption[Quantitative comparison of various completion networks on center masked images]{Learned Perceptual Image Patch Similarity (LPIPS) and Fr\'echet Inception Distance (FID) for various completion networks on center masked images. Here, we calculate the LPIPS and FID using all images in the corresponding test sets.}
    \label{tab:ch5_conv_vs_transform}
\end{table}

\subsection{Results and Analysis for Token Representation}\label{sec:ch5_analysis-trans}

\paragraph{Results} We first demonstrate experimentally that the transformer-based model outperforms previous CNN-based models. Table \ref{tab:ch5_conv_vs_transform} shows Learned Perceptual Image Patch Similarity (LPIPS)\footnote{While multi-modal generation tasks had previously been evaluated with LPIPS \cite{zhu2017toward,huang2018multimodal,zheng2019pluralistic} in Chapters \ref{ch:F(L)SeSim} and \ref{ch:PICNet}, it was used to measure diversity. Here, we apply it to measure the similarity between completed images and original ground-truth. A smaller value means the completed image is closer to the ground-truth image w.r.t.\ the learned perceptual similarity, rather than pixel-level reconstruction. We refer readers to \cite{zhang2018unreasonable} for details.} \cite{zhang2018unreasonable} and Fr\'echet Inception Distance (FID) \cite{heusel2017gans} for various image completion architectures on CelebA-HQ \cite{liu2015deep,karras2017progressive} and FFHQ \cite{karras2019style} datasets degraded by center masks. The traditional image quality results are given in Appendix \ref{app:sec:ch5_ana}. Here, we compared with three CNN-based models, for which CA \cite{yu2018generative} and PICNet \cite{zheng2019pluralistic} had the appropriate pretrained models available, while the latest MEDFE \cite{Liu2019MEDFE} was reproduced using their publicly available code. All scores are reported for $256^2$ resolution. Without bells and whistles, our TFill-\emph{Coarse} with configuration ($\mathrm{E}$) improved  LPIPS (18\% relative improvement) and FID (39\% relative improvement) quite significantly on CelebA-HQ, despite only using the transformer-based content inference network, without our refinement network.  

\begin{figure}[tb!]
    \centering
    \includegraphics[width=0.95\linewidth]{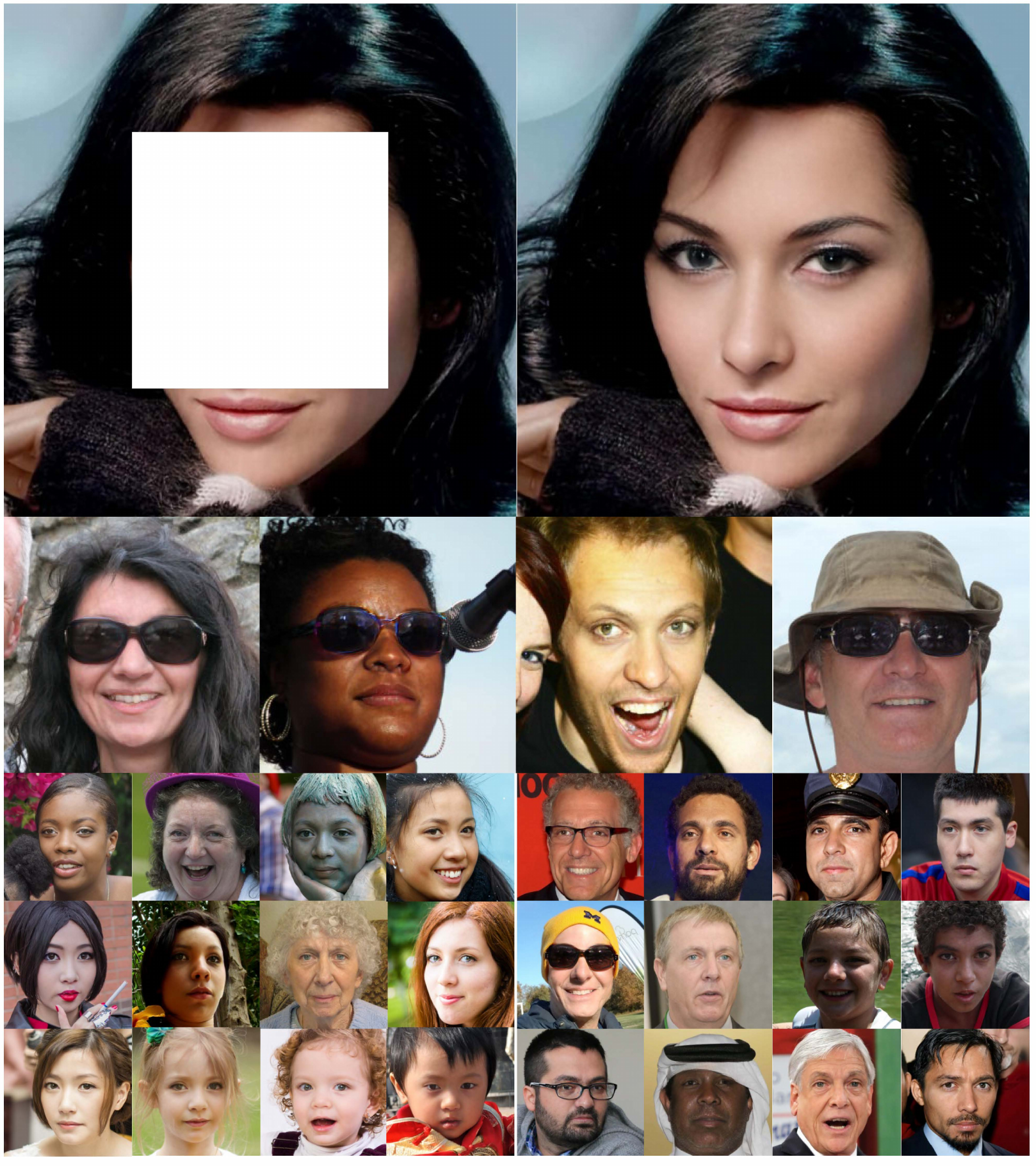}
    \caption[Example completion results of our method on face datasets]{\textbf{Example completion results of our method (config E) on face datasets.} Here, a center mask was used for all input images. The corresponding quantitative results are reported in Tables  \ref{tab:ch5_conv_vs_transform} and \ref{tab:ch5_ablation_transform}. One center masked example input is shown top-left.}
    \label{fig:ch5_center-face-examples}
\end{figure}

\begin{figure}[tb!]
    \centering
    \includegraphics[width=\linewidth,height=0.38\textheight]{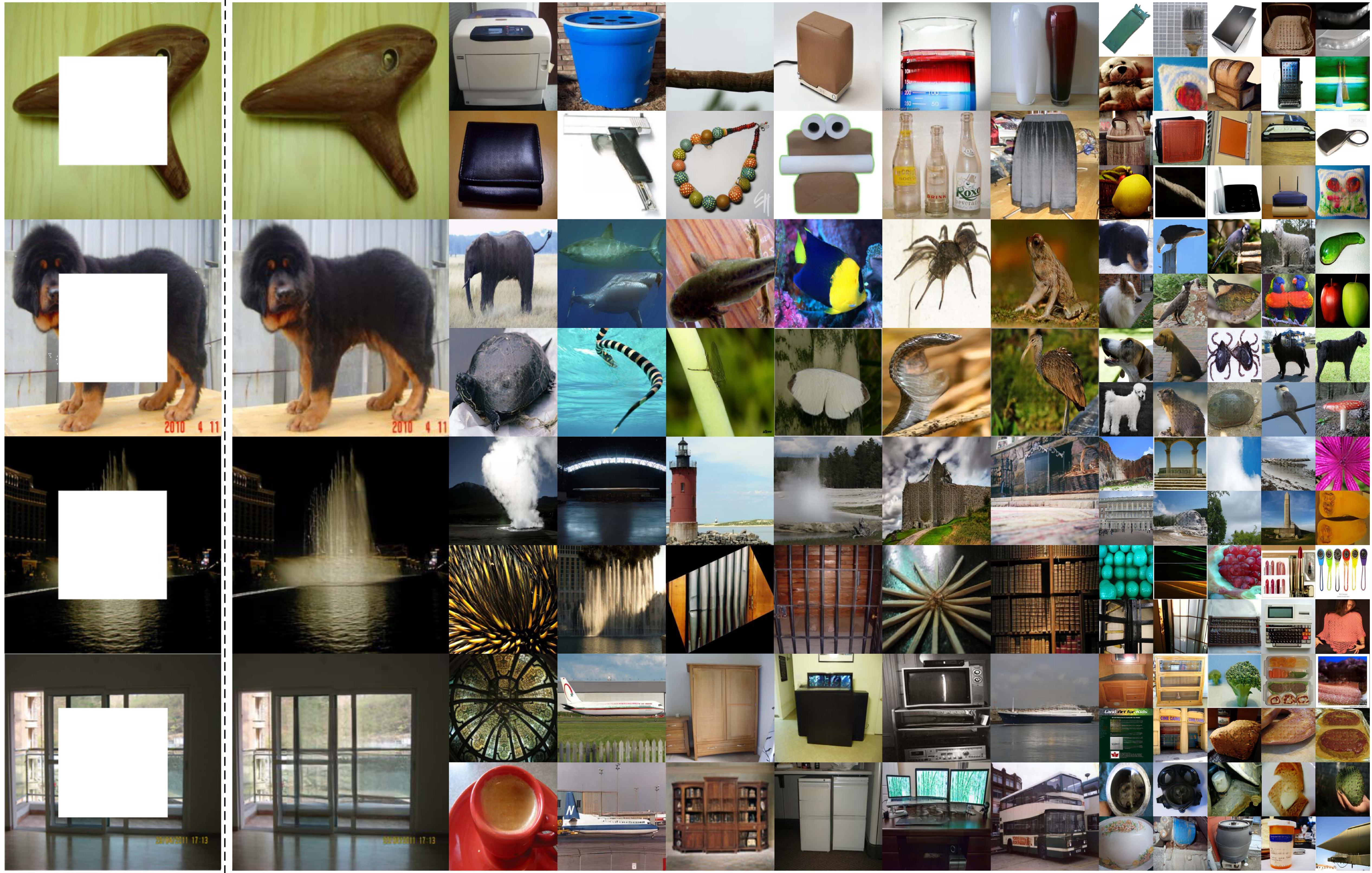}
    \caption[Completion results of our method on ImageNet datasets]{\textbf{Completion results of our method (config E) on ImageNet datasets \cite{russakovsky2015imagenet}.} All images come from the corresponding testing set that were \textbf{degraded by center masks}. Here, we show results for various categories, such as commodity, animal, plant, natural scene, building, food, furniture and so on.}
    \label{fig:ch5_sup-center-imgnet-examples}
\end{figure}

\begin{figure}[tb!]
    \centering
    \includegraphics[width=\linewidth,height=0.38\textheight]{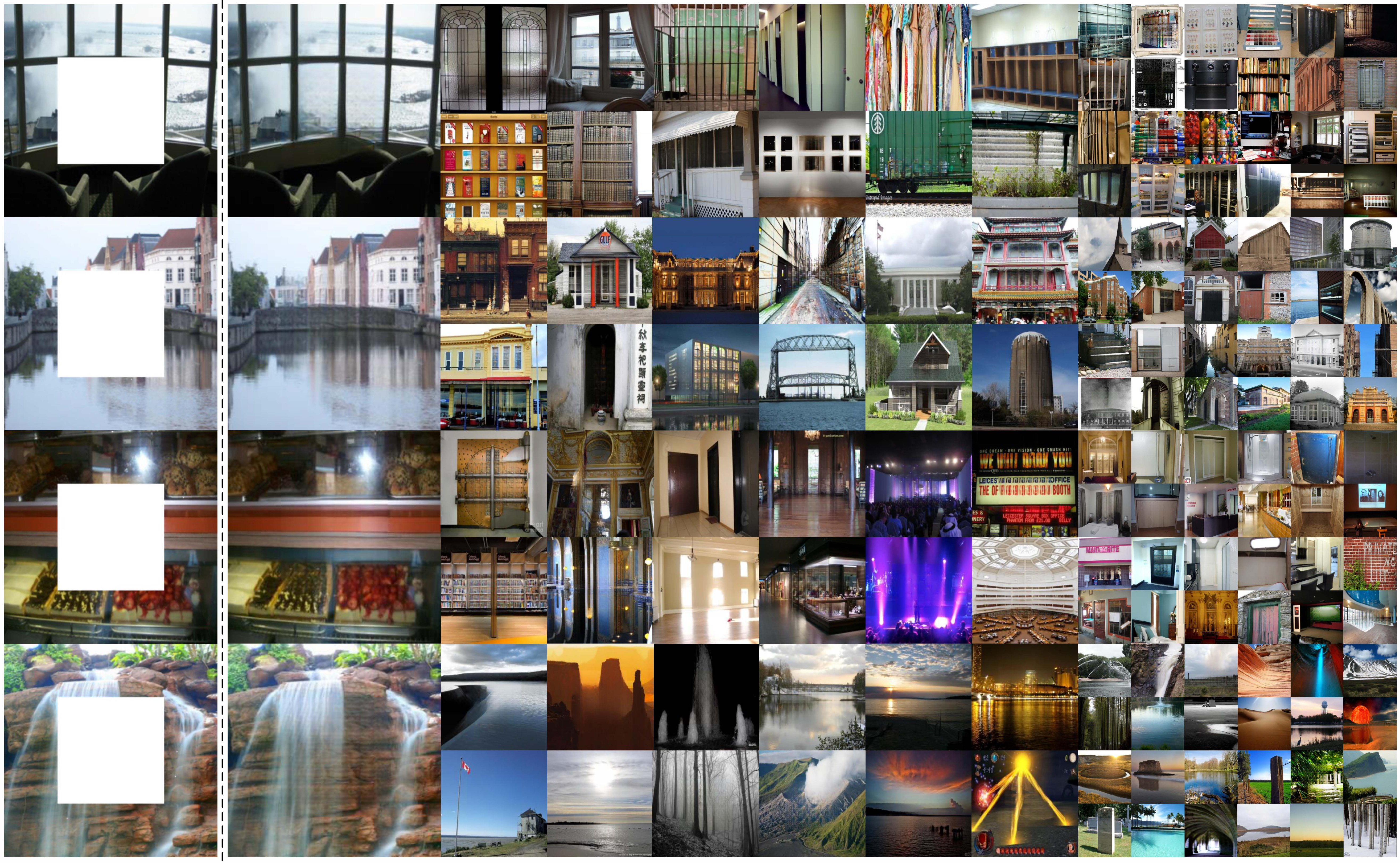}
    \caption[Completion results of our method on Places2 datasets]{\textbf{Completion results of our method (config E) on Places2 datasets \cite{zhang2018unreasonable}.} All images come from the corresponding testing set that were \textbf{degraded by center masks}.}
    \label{fig:ch5_sup-center-places2-examples}
\end{figure}

Figure~\ref{fig:ch5_center-face-examples} shows the visual results of our TFill on CelebA-HQ and FFHQ datasets. Here, all images are center masked in order to demonstrate its ability to go beyond object removal and to generate reasonable semantic content for large missing regions. As can be seen, the completed images are on average of high quality. Even for some challenging cases, such as when eyeglasses are center masked, our TFill can correctly repair the face \emph{with} eyeglasses. Furthermore, it generally works well for varied skin tones, poses, expressions, ages, and illumination.

In Figure~\ref{fig:ch5_sup-center-imgnet-examples}, we show more examples for object completion, such as the various items and animals on the top half. In Figure~\ref{fig:ch5_sup-center-places2-examples}, we display the completed images for various natural scenes. These examples are good evidence that our TFill model is suitable for both \emph{foreground} object completion and \emph{background} scene completion, where it can synthesize semantically consistent content with visually realistic appearance based on the presented visible pixels. 

\paragraph{Analysis} Our baseline configuration ($\mathrm{A}$) used the same encoder-decoder structure as VQGAN \cite{esser2020taming}, except here attention layers were removed for a pure CNN-based version. When combined with the powerful discriminator of StyleGANv2 \cite{karras2020analyzing}, the performance was comparable to PICNet \cite{zheng2019pluralistic}, in which the best results were selected from $50$ diverse samples. We first added the attention layer to the decoder (Generator, G) in ($\mathrm{B}$), but the performance remained similar to baseline ($\mathrm{A}$). In contrast, when we use our proposed \emph{restrictive CNN} in ($\mathrm{C}$), the performance improved substantially, especially for FID. This suggests that the input feature representation is significant for the attention layer to equally deliver all messages, as explained later. We then improved this new baseline by adding the transformer encoder ($\mathrm{D}$), which benefits from globally delivered messages at every layer. Finally, we introduced masked weights to each attention layer of the transformer ($\mathrm{E}$), improving results further.  

\begin{table}[tb!]
    \centering
     \renewcommand{\arraystretch}{1.1}
    \setlength\tabcolsep{5pt}
    \begin{tabular}{@{}llcccc@{}}
        \hlineB{3.5}
        & \textbf{Method} & LPIPS$\downarrow$& FID$\downarrow$  & Mem$\downarrow$ & Time$\downarrow$\\
        \hlineB{2}
        & IGPT~\cite{chen2020generative} (RF $1$) & 0.609 & 148.42 & 3.16 & 26.45\\
        & VIT~\cite{dosovitskiy2020image} (RF $16$) & 0.062 & 5.09 & 1.16 & 0.167\\
		& VQGAN~\cite{esser2020taming} & 0.226 & 11.92 & 2.36 & 4.29\\
		\cdashline{1-6}
		$\mathrm{B}$ & \emph{Conv} (RF $229$) & 0.064 & 4.01 & 0.99 & 0.162\\
        $\mathrm{C}$ & Ours \emph{R-Conv} (RF $16$) & 0.060 & 3.87 & \textbf{0.90} & \textbf{0.157}\\
		\cdashline{1-6}
        & T-based (RF $229$) & 0.062 & 3.92 & 1.25 & 0.188\\
        $\mathrm{E}$ & T-based (RF $16$) & \textbf{0.057} & \textbf{3.63} & 1.15 & 0.180\\
        \hlineB{2}
    \end{tabular}
    \caption[Ablation study on token representation]{\textbf{The effect of restrictive token embedding and transformer block} in our transformer-based completion network on FFHQ dataset. ``RF'' indicates the Receptive Field size. ``Mem'' denotes the memory (GB) cost during testing and ``Time'' is the testing time (s) for each center masked image. }
    \label{tab:ch5_ablation_transform}
\end{table}

To study the influence of the token representation, we conducted two experiments that compared with recent visual transformer works \cite{chen2020generative,dosovitskiy2020image,esser2020taming} and provided an ablation study by controlling the RF in Table \ref{tab:ch5_ablation_transform}. 

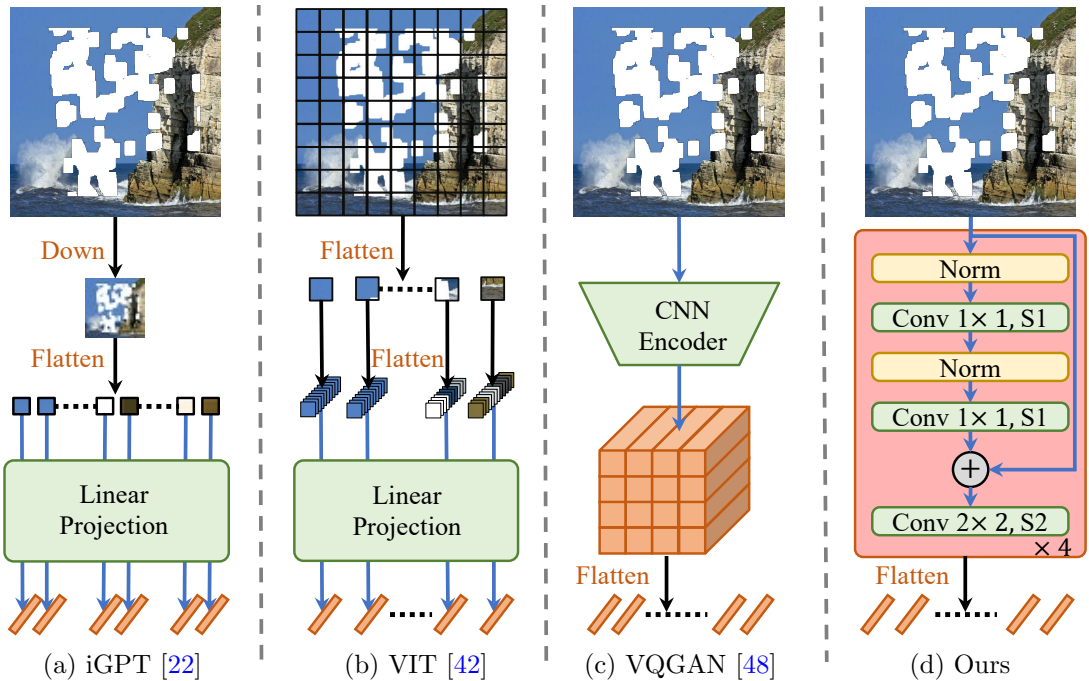
\begin{figure}[tb!]
    \centering
    \includegraphics[width=\linewidth]{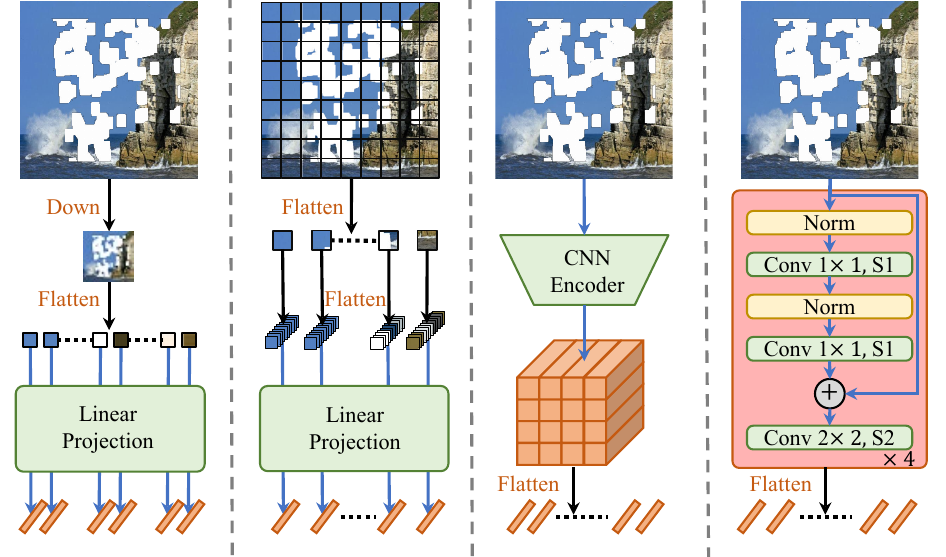}
    \begin{picture}(0,0)
    \put(-189, 10){\footnotesize (a) iGPT~\cite{chen2020generative}}
    \put(-76, 10){\footnotesize (b) VIT~\cite{dosovitskiy2020image}}
    \put(14, 10){\footnotesize (c) VQGAN~\cite{esser2020taming}}
    \put(136, 10){\footnotesize (d) Ours}
    \end{picture}
    \caption[Token representation]{\textbf{Token representation.} (a) Pixel to token. (b) Patch to token. (c) Feature to token. (d) Restrictive \textbf{R}eceptive \textbf{F}ield (RF) feature to token. Note our token has a small and non-overlapping RF like VIT \cite{dosovitskiy2020image}, but uses a complex CNN embedding. Each token represents locally isolated contexts, leaving the long-range relationship to be cleanly modeled in the transformer encoder. }
    \label{fig:ch5_tokens}
\end{figure}

\begin{figure}[tb!]
    \centering
    \includegraphics[width=\linewidth]{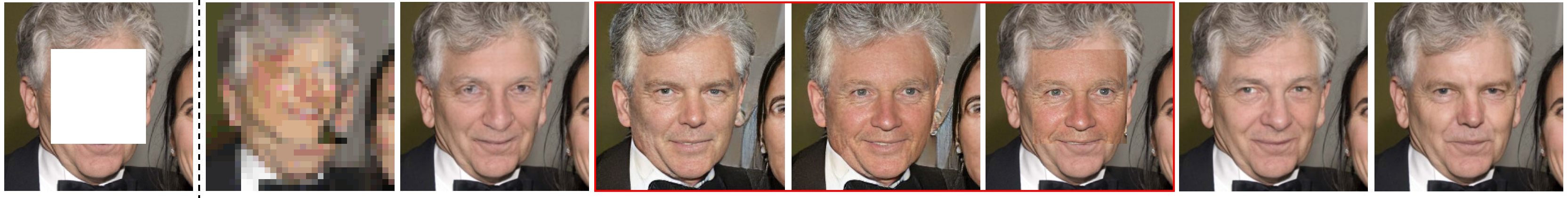}
    \begin{picture}(0,0)
    \put(-188,7){\tiny Input}
    \put(-145,7){\tiny iGPT~\cite{chen2020generative}}
    \put(-88, 7){\tiny VIT~\cite{dosovitskiy2020image}}
    \put(-46, 7){\tiny VQ~\cite{esser2020taming} Rec}
    \put(4, 7){\tiny VQ~\cite{esser2020taming} Comp}
    \put(56, 7){\tiny VQ~\cite{esser2020taming} Output}
    \put(114, 7){\tiny TFill-\emph{Coarse}}
    \put(162, 7){\tiny Ground Truth}
    \end{picture}
    \vspace{-0.3cm}
    \caption[Comparing results under different token representations]{\textbf{Comparing results under different token representations.} All transformers are based on the same transformer backbone \cite{vaswani2017attention}. For VQGAN \cite{esser2020taming}, we report reconstruction (Rec) image, completed (Comp) image and recomposed output image. TFill-\emph{Coarse} is our model with configuration E in Tables 1 and 2, \ie TFill without the refinement network. Please see main text for details.}
    \label{fig:ch5_center-face-comp}
\end{figure}
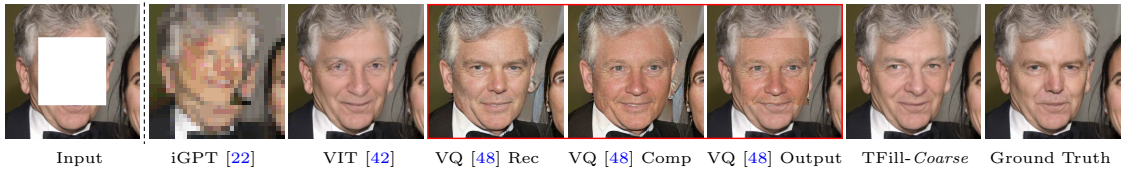

As illustrated in Figure~\ref{fig:ch5_tokens}, iGPT \cite{chen2020generative} downsamples the image to a fixed scale, \eg $32^2$ resolution, and embeds \emph{each pixel to a token}. While this may not impact the original classification task, which is robust to low resolutions \cite{torralba200880}, it has a large negative effect on generating high-quality images. Furthermore, the auto-regressive form resulted in the completed image being inconsistent with the bottom-right visible region (iGPT in Figure~\ref{fig:ch5_center-face-comp}), and each image runs an average of 26.45s during the testing. This is because the conditional sequence generation can only utilize the top-left visible pixels, generating new pixels one-by-one. In contrast, VIT \cite{dosovitskiy2020image} divides an image to a set of fixed patches and embeds \emph{each patch to a token}. As shown in Table \ref{tab:ch5_ablation_transform} and Figure~\ref{fig:ch5_center-face-comp}, it can achieve relatively good quantitative and qualitative results. However, some details are perceptually poor, \eg the strange eyes in Figure~\ref{fig:ch5_center-face-comp}, possibly due to the limited one-layer linear projection. Finally, VQGAN \cite{esser2020taming} employs a traditional CNN to encode an image to the feature domains and then \emph{quantizes each feature as a token} through a learned codebook \cite{van2017neural,razavi2019generating}. Figure~\ref{fig:ch5_center-face-comp} shows the generated images using tokens embedded from ground truth (VQ Rec), and tokens extracted from the center masked image (VQ Comp). While it generates the content of missing regions sequentially conditioned only on top-left visible tokens, we found the completed pixels to be consistent with the bottom-right region, even though these tokens were \emph{not} used to infer missing content in the transformer encoder. We believe this is due to the large RF in the CNN-based encoder causing each token to capture extended dependencies in a deep layer. However, this leads to two issues: \textbf{1)} even the original visible tokens are modified, resulting in different appearances for the visible regions \eg see VQ Rec \emph{vs} VQ Comp in Figure~\ref{fig:ch5_center-face-comp}; \textbf{2)} inferred tokens are unduly influenced by implicit CNN-based correlation to nearby tokens, and cannot establish ties cleanly to important but distant tokens. Thus it generates a visually realistic completion, but when pasted to the original masked input (VQ Output in Figure \ref{fig:ch5_center-face-comp}), there is an obvious gap between generated and visible pixels.

In contrast to \cite{chen2020generative,dosovitskiy2020image,esser2020taming}, our token representation is extracted using a \emph{restrictive CNN} (Figure~\ref{fig:ch5_tokens}(d)). In particular, the $1$$\times$$1$ filter and \texttt{layernorm} is applied for non-linear projection, followed by a partial convolution layer \cite{Liu_2018_ECCV} that uses a $2$$\times$$2$ filter with stride $2$ to extract visible information and reduce feature resolution simultaneously. For instance, if half of the pixels in a window are masked, we only embed the other $50\%$ comprising visible pixels as our token representation, and establish an initial weight of $0.5$ for the \emph{masked} self-attention layer. To do this, we ensure each token represents only the visible information in a small RF, \emph{leaving the long-range dependencies to be explicitly modeled by the transformer encoder in every layer}, without cross-contamination from implicit correlation due to larger CNN RF. To demonstrate the impact of RF, a thorough ablation study result is reported in Table \ref{tab:ch5_ablation_transform}, in which we find the small RF CNN improves both LPIPS and FID significantly, with the added benefit of low memory cost. Furthermore, our model runs at 180ms per image on an Nvidia GTX 1080Ti (+21ms CPU time for resizing input and storing output), due to predicting all output heads together, rather than auto-regressively as in existing work \cite{chen2020generative,esser2020taming}.

\subsection{Results and Analysis for AAL}\label{sec:ch5_attn-aware-ana}

\begin{figure}[tb!]
    \centering
    \includegraphics[width=\linewidth]{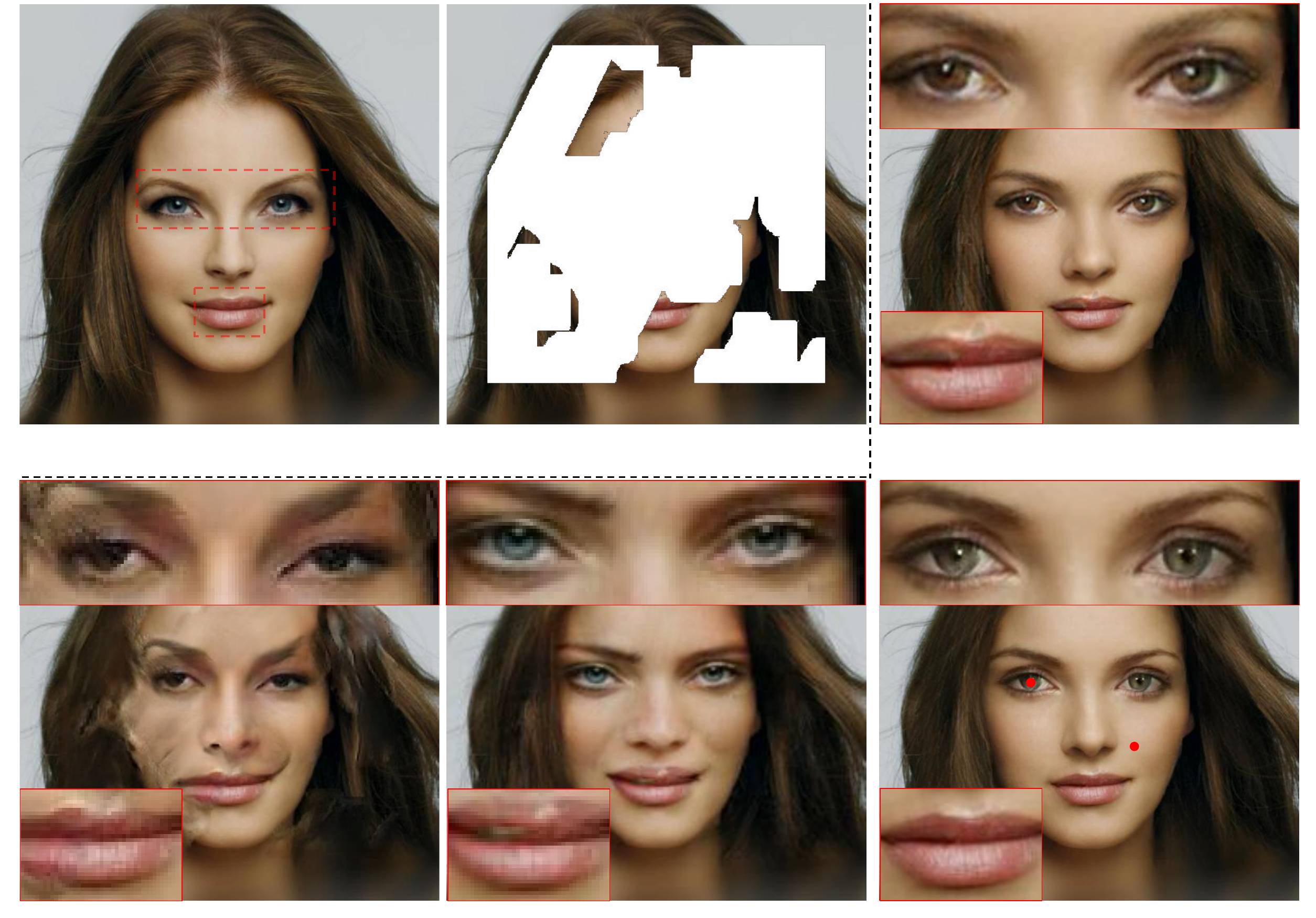}
    \begin{picture}(0,0)
    \put(-166, 162){\footnotesize Ground Truth}
    \put(-10, 162){\footnotesize Input}
    \put(124, 162){\footnotesize  TFill-\emph{SA}}
    \put(-152, 7){\footnotesize CA~\cite{yu2018generative}}
    \put(-24, 7){\footnotesize PICNet~\cite{zheng2019pluralistic}}
    \put(116, 7){\footnotesize TFill-\emph{Refined}}
    \end{picture}
    \vspace{-0.3cm}
    \caption[Results with different attention modules in various methods]{\textbf{Results with different attention modules} in various methods. Our attention-ware layer is able to adaptively select the features from both visible and generated content. In this example, the ratio for the two query points is ${\bf w}_v/{\bf w}_m$ = $0.77/0.23$ (skin) and ${\bf w}_v/{\bf w}_m$ = $0.08/0.92$ (eye), respectively.}
    \label{fig:ch5_attn-comp}
\end{figure}
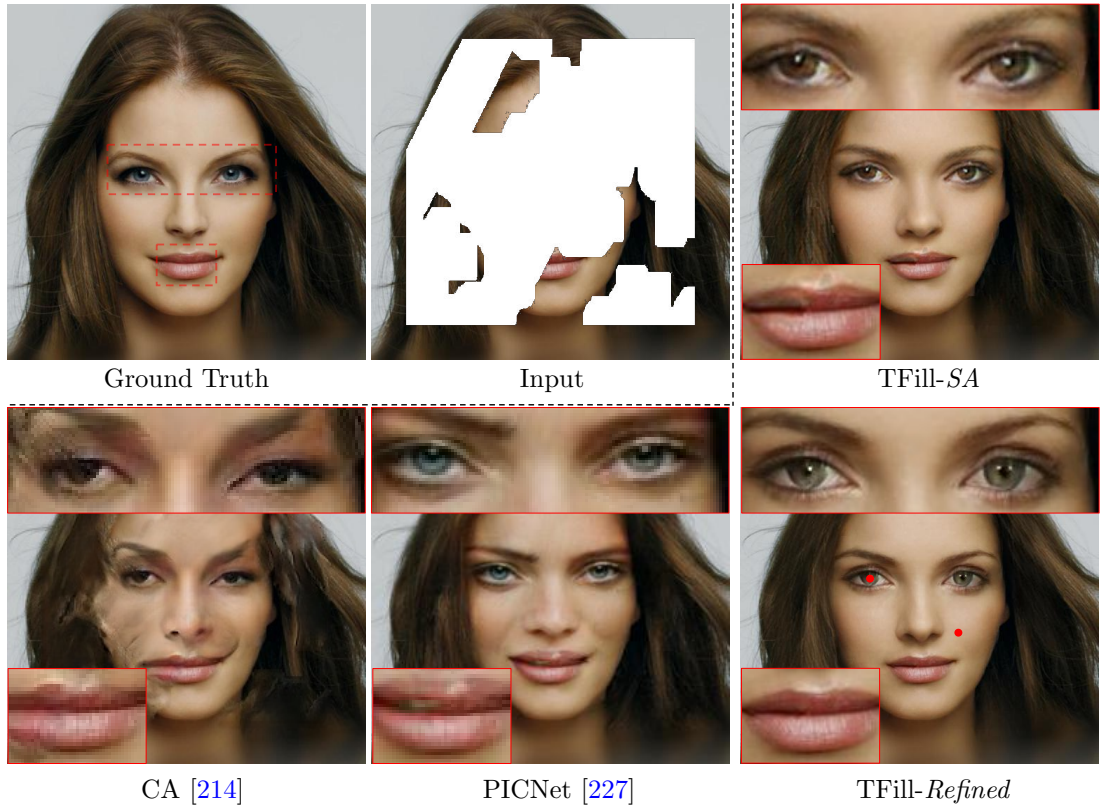

\begin{table}[htb!]
    \centering
    \renewcommand{\arraystretch}{1.0}
    \setlength\tabcolsep{4pt}
    \begin{tabular}{@{}l|l|cccc@{}}
        \hlineB{3.5}
         Mask Type & Metric & SA \cite{zhang2019self} & CA \cite{yu2018generative}  & SLTA \cite{zheng2019pluralistic}  & Ours-AAL\\
         \hlineB{2}
         \multirow{2}{*}{center} & LPIPS  & 0.058 & 0.061 & 0.056 & \bf{0.053} \\
         & FID  & 3.62 & 3.86 & 3.61 & \bf{3.50} \\
         \hline
          \multirow{2}{*}{random} & LPIPS & 0.047  & 0.044 & 0.045 &\bf{0.041} \\
         & FID  & 2.69 & 2.66 & 2.64 & \bf{2.57} \\
         \hlineB{2}
    \end{tabular}
    \caption[The effect of various attention layers on FFHQ dataset]{The effect of various attention layers on FFHQ dataset. ``center'' denotes the center mask, ``random'' denotes the random regular mask and ``SA'' is the basic self-attention layer. These attention layers were implemented within our TFill refinement framework. }
    \label{tab:ch5_attn_aware_ana}
\end{table}

We ran ablations to analyze our proposed AAL by replacing it with existing contextual attention models of SA \cite{vaswani2017attention,zhang2019self}, CA \cite{yu2018generative} and SLTA \cite{zheng2019pluralistic}. As shown in Table \ref{tab:ch5_attn_aware_ana}, SA showed similar performance to the coarse results in Table~\ref{tab:ch5_conv_vs_transform}, due to the insular attention problem mentioned earlier. CA \cite{yu2018generative} performed worse on large center masks than random regular masks (even worse than the coarse results of ($\mathrm{E}$) in Table \ref{tab:ch5_conv_vs_transform}), as it borrows context from visible regions only. When important context is not visible, \eg when both eyes are missing in Figure~\ref{fig:ch5_attn-comp}, it is unable to find the right context to copy. While our previous PICNet \cite{zheng2019pluralistic} focuses on both visible and invisible regions, selection was done by \emph{fixed} weights learned during training. This is also inferior, and in some cases we observed that it can have difficulty in selecting the best features for generation, especially on free-form masks. In contrast, our AAL selects features based on the largest attention scores, using weights \emph{dynamically mapped} during inference. For instance, in Figure~\ref{fig:ch5_attn-example}, only the left eye was masked, and it had a large attention score to the visible right eye, resulting in a ratio of ${\bf w}_v/{\bf w}_m$ = $0.91/0.09$. Conversely, when two eyes were masked in Figure~\ref{fig:ch5_attn-comp}, the attention score between the two eyes was still high, but the ratio was correctly flipped to ${\bf w}_v/{\bf w}_m$ = $0.08/0.92$ for the left eye. 

\subsection{Additional Results}\label{sec:ch5_add_result}

Following Chapter \ref{ch:PICNet}, I also show interesting applications of the proposed TFill model for free-form image editing on various higher resolution datasets.

As shown in Figure \ref{fig:ch5_sup-free-edit-imagenet0}, I edit the natural scene with object removal being the main task, as it is the main use case for image inpainting. Here, I enforce the input image size to be multiples of $32$, \eg $512\times384$ and provide the high-resolution results on the corresponding image size. As we can see, our TFill-\emph{Refined} model is able to handle high-resolution images for object removal in traditional image inpainting task. More results can be found in our online project\footnote{More results are available on \href{http://www.chuanxiaz.com/publication/tfill/}{http://www.chuanxiaz.com/publication/tfill/}}.

In Figure~\ref{fig:ch5_sup-free-edit-face}, more examples are shown for face editing at $512\times512$ resolution. For conventional object removal, \eg watermark removal, the proposed TFill addresses them easily. Furthermore, the TFill can handle more extensive face editing, such as removing substantial facial hair and changing mouth expressions.

\section{Limitations and Discussion}\label{ch5:conl}

Through the detailed analyses and experiments, I demonstrate that the transformer based architecture has exciting potential for image completion, due to its capacity for effectively modeling connections between distant image content. Unlike recent vision transformer models that either use shallow projections or large receptive fields for token representation, our \emph{restrictive CNN projection} provides the necessary separation between explicit attention modeling and implicit RF correlation that leads to substantial improvement in results. I also introduced a novel attention-aware layer that adaptively balances the attention for visible and masked regions, further improving the completed image quality.

While this \emph{TFill} model generates reasonable content as well as realistic appearance, it provides only one ``optimal'' result for this highly subjective task. It will be much more interesting if we can explore the transformer-based architecture for multiple and diverse results. The more recent work \cite{wan2021high} has made an initial step towards this goal. We would like to explore more in future work. 
\part{Modeling Shape and Appearance: \\ Completed Scene Decomposition} %

\chapter{Visiting the Invisible} 
\chaptermark{VIV}
\label{ch:VIV} 

The methods in Part \uppercase\expandafter{\romannumeral2} can produce plausible results given a masked image by filling into reasonable content as well as visually realistic appearance. However, these systems depend on manual masks as input, rather than automatically understanding the full scene. In this chapter, we present a higher-level structural scene decomposition and completion system, which has the ability to \emph{decompose} a scene into individual objects, \emph{infer} their underlying occlusion relationships and moreover \emph{imagine} what occluded objects may look like, while \emph{using only an image as input}.  In order to disentangle the occluded relationships of all objects in a complex scene, we use the fact that the front object, being free from occlusion, is easy to be identified, detected, and segmented. Our system interleaves the two tasks of instance segmentation and scene completion through multiple iterations, solving for objects layer-by-layer. We first provide a thorough experiment using a new realistically rendered dataset, where ground-truth is available for all invisible regions. To bridge the domain gap to real imagery, where ground-truth is not available, we then train another model with pseudo-ground-truths generated from our previously trained synthesis model. We demonstrate results on a wide variety of datasets and show significant improvement over the state-of-the-art. 

The rest of this chapter is organized as follows. We first describe our motivation in Section \ref{sec:ch6_intro}. Then, we discuss the related work in Section~\ref{sec:ch6_related}, and describe our rendered dataset in Section~\ref{sec:ch6_data}. In Section~\ref{sec:ch6_method} we present our layer-by-layer CSDNet method. We then show the experiment results on this synthetic dataset as well as the results on real-world images in Section~\ref{sec:ch6_experiment}, followed by a conclusion in Section~\ref{sec:ch6_conclusions}.

\section{Motivation}\label{sec:ch6_intro}

The vision community has made rapid advances in scene understanding tasks, such as object classification and localization~\cite{girshick2014rich,he2015spatial,ren2015faster}, scene parsing~\cite{long2015fully,chen2017deeplab,badrinarayanan2017segnet}, instance segmentation~\cite{pinheiro2015learning,he2017mask,chen2019hybrid}, and layered scene decomposition~\cite{gould2009decomposing,yang2010layered,zhang2015monocular}. Despite their impressive performance, these systems deal only with \emph{visible} parts of scenes without trying to exploit \emph{invisible} regions, which results in an uncompleted representation of real objects. 

In parallel, significantly progress for the generation task has been made with the emergence of deep generative networks, such as GAN-based models \cite{goodfellow2014generative,gulrajani2017improved,karras2019style}, VAE-based models \cite{kingma2013auto,van2017neural,vahdat2020NVAE}, and flow-based models \cite{dinh2014nice,dinh2016density,kingma2018glow}. Empowered by these techniques, image completion~\cite{,iizuka2017globally,yu2018generative,zheng2019pluralistic} and object completion~\cite{ehsani2018segan,zhan2020self,ling2020variational} have made it possible to create the plausible appearances for occluded objects and backgrounds, as shown in above chapters. However, these systems depend on manual masks or visible ground-truth masks as input, rather than automatically understand the full scene.

In this chapter, we will present a system that has the ability to \emph{decompose} a scene into individual objects, \emph{infer} their underlying occlusion relationships, and moreover \emph{imagine} what occluded objects may look like, \emph{while using only an image as input}. This novel task involves the classical recognition task of instance segmentation to predict the geometry and category of all objects in a scene, and the generation task of image completion to reconstruct invisible parts of objects and backgrounds. After a full decomposition of the given scene, users can freely edit the instances in the original 2D image, such as deleting object, moving object's positions, and further changing their occlusion relationships \ref{fig:ch6_results_app}.  

To decompose a scene into instances with completed appearances in one pass is extremely challenging. This is because realistic natural scenes often consist of a vast collection of physical objects, with complex scene structure and occlusion relationships, especially when one object is occluded by multiple objects, or when instances have deep hierarchical occlusion relationships.

\begin{figure}[tb!]
	\centering
	\includegraphics[width=\linewidth]{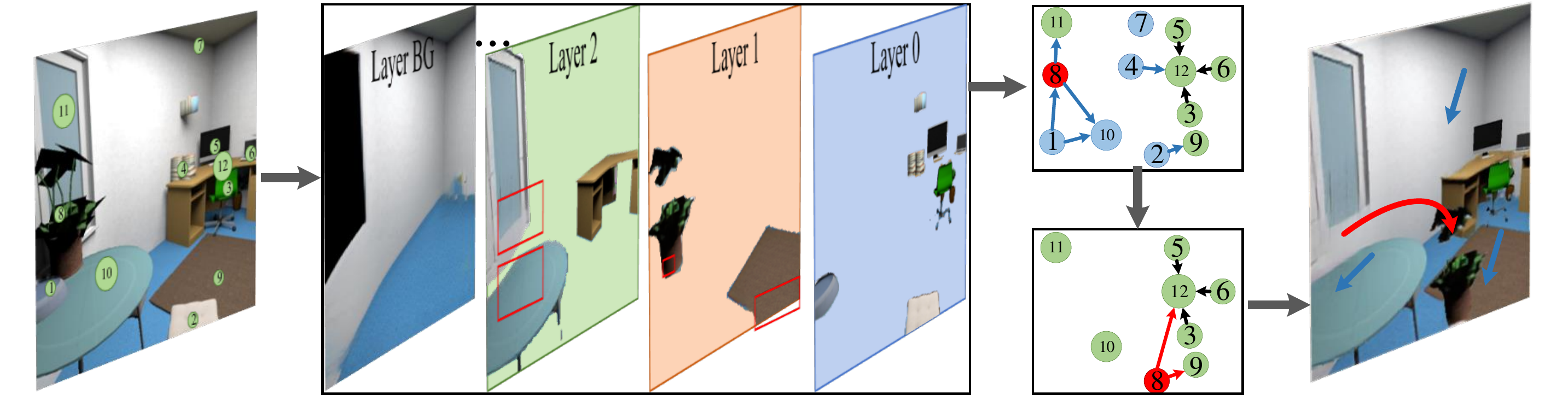}
	\begin{picture}(0,0)
		\put(-185,0){\scriptsize (a) Input}
		\put(-120,0){\scriptsize (b) Layered Completed Scene Decomposition}
		\put(54, 0) {\scriptsize (c) Pairwise Order}
		\put(124, 0){\scriptsize (d) Recomposed Image}
	\end{picture}
	\caption[Example results of scene decomposition and recomposition]{\textbf{Example results of scene decomposition and recomposition.} (a) Input. (b) Our model structurally decomposes a scene into individual completed objects. Red rectangles highlight the original \emph{invisible} parts. (c) The inferred pairwise order (top graph) and edited order (bottom graph) of the instances. Blue nodes indicate the deleted objects while the red node is the moved object. (d) The new recomposed scene.}
	\label{fig:ch6_example}
\end{figure}
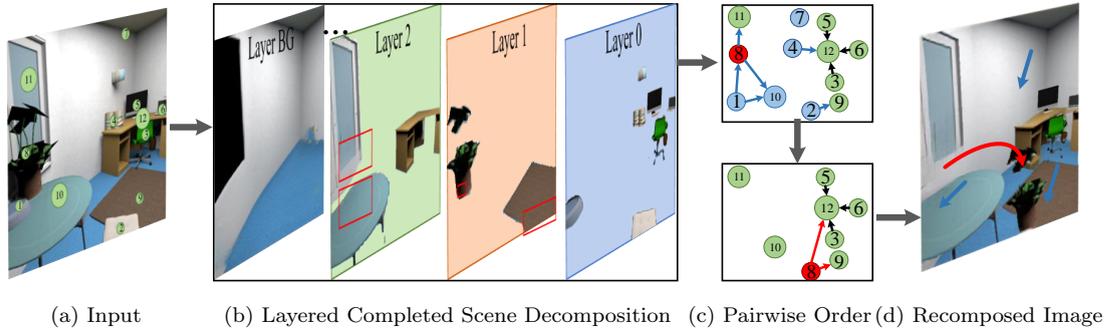

Our core idea is from the observation that \emph{it is much easier to identify, detect and segment foreground objects than occluded objects}. Motivated by this, we propose a {\bf C}ompleted {\bf S}cene {\bf D}ecomposition {\bf Net}work ({\bf CSDNet}) that learns to segment and complete each object in a scene layer-by-layer consecutively. As shown in Figure~\ref{fig:ch6_example}, our layered scene decomposition network only segments the fully visible objects out in each layer (Figure~\ref{fig:ch6_example}(b)). If the system is able to properly segment the foreground objects, it will automatically learn which parts of occluded objects are actually invisible that need to be filled in. The completed image is then passed back to the layered scene decomposition network, which can again focus purely on detecting and segmenting visible objects. As the interleaving proceeds, a structured instance depth order (Figure~\ref{fig:ch6_example}(c)) is progressively derived by using the inferred absolute layer order. The thorough decomposition of a scene along with spatial relationships allows the system to freely recompose a new scene (Figure~\ref{fig:ch6_example}(d)).

Another challenge in this novel task is the lack of data: there is no complex, realistic dataset that provides intact ground-truth appearance for originally occluded objects and backgrounds in a scene. While latest works~\cite{li2016amodal,zhan2020self} introduced a self-supervised way to tackle the amodal completion using only visible annotations, they can not do a fair quantitative comparison as no real ground-truths are available. To mitigate this issue, we constructed a high-quality rendered dataset, named \textbf{C}ompleted \textbf{S}cene \textbf{D}ecomposition (\textbf{CSD}), based on more than 2k indoor rooms. Unlike the datasets in \cite{ehsani2018segan,Dhamo2019iccv}, our dataset is designed to have more typical camera viewpoints, with near-realistic appearance.

As elaborated in Section~\ref{sec:ch6_csd_results}, the proposed system performs well on this rendered dataset, both qualitatively and quantitatively outperforming existing methods in completed scene decomposition, in terms of instance segmentation, depth ordering, and amodal mask and content completion. To further demonstrate the generalization of our system, we extend it to real datasets. As there are no ground truth annotations and appearance available for training, we created pseudo-ground-truths for real images using our model that is purely trained on \textbf{CSD}, and then fine-tuned this model accordingly. This model outperforms state-of-the-art methods~\cite{zhu2017semantic,qi2019amodal,zhan2020self} on amodal instance segmentation and depth ordering tasks, despite these methods being specialized to their respective tasks rather than our holistic completed scene decomposition task. While we are unable to quantitatively evaluate real-image scene completion without ground truth appearance for occluded objects, our method is able to create visually reasonable layer-by-layer decomposition results, and we further demonstrate its effectiveness in real scene recomposition. 

In summary, we propose a layer-by-layer scene decomposition network that jointly learns structural scene decomposition and completion, rather than treating them separately as the existing works \cite{ehsani2018segan,Dhamo2019iccv,zhan2020self}. To our knowledge, it is the first work that proposes to complete objects based on the global context, instead of tackling each object independently. To address this novel task, we render a high-quality rendered dataset with ground-truth for all instances. We then provide a thorough ablation study using this rendered dataset, in which we demonstrate that the method substantially outperforms existing methods that address the task in isolation. On real images, we improve the performance to the recent state-of-the-art methods by using pseudo-ground-truth as weakly-supervised labels. The experimental results show that our \textbf{CSDNet} is able to acquire a full decomposition of a scene, \emph{with only an image as input}, which conduces to a lot of applications, \eg object-level image editing.

\begin{table}[tb!]
    \centering
	\renewcommand{\arraystretch}{1.0}
	\setlength\tabcolsep{5pt}
	\begin{tabular}{@{}llll@{}}
		\hlineB{3.5}
		Paper & Outputs & Inputs & Data\\
		\hline
		 & SP, O & I & LabelMe, PASVOC, others \\
		Yang \etal \cite{yang2011layered} & In, O & I & PASVOC \\
		Tighe \etal \cite{tighe2014scene} & SP, O & I & LabelMe, SUN \\
		Zhang \etal \cite{zhang2015monocular} & In, O & I & KITTI \\
		\hline
		Guo \etal \cite{guo2012beyond} & AS & I & StreetScenes, SUN, others\\
		Kar \etal \cite{kar2015amodal} & AB & I & PASVOC, PAS3D \\
		Liu \etal \cite{liu2016layered} & AS, O & I, D & NTUv2-D\\
		\hline
		Li \etal \cite{li2016amodal} & A & I, In & PASVOC\\
		Zhu \etal \cite{zhu2017semantic} & A, O & I & COCOA (from COCO)\\
		Follmann \etal \cite{follmann2019learning} & A & I & COCOA, COCOA-cls, D2S\\
		Qi \etal \cite{qi2019amodal} & A & I & KINS (from KITTI) \\
		Hu \etal \cite{hu2019sail} & A & I & Synthesis video\\
		\hline
		Ehsani \etal \cite{ehsani2018segan} & A, O, IRGB & I, In & DYCE, PAS3D\\
		Zhan \etal \cite{zhan2020self} & A, O, IRGB & I, In & KINS, COCOA \\
		Ling \etal \cite{ling2020variational} & A, IRGB & I, In & KINS\\
		Yan \etal \cite{yan2019visualizing} & A, IRGB & I & Vehicle \\
		Burgess \etal \cite{burgess2019monet} & In, IRGB & I &  Toy\\
		Dhamo \etal \cite{Dhamo2019iccv} & A, D, IRGB& I & SUNCG, Stanford 2D-3D\\
		Ours & A, O, IRGB& I & KINS, COCOA, SUNCG\\
		\hlineB{2.5}
	\end{tabular}
	\caption[Comparison with related work based on three aspects: outputs, inputs and data]{\textbf{Comparison with related work based on three aspects: outputs, inputs and data}. I: image, In: inmodal segmentation, O: occlusion order, SP: scene parsing, AB: amodal bounding box, AS: amodal surface, A: amodal segmentation, D: depth, IRGB: intact RGB object.}
	\label{tab:ch6_related_work}
\end{table}

\section{Background}\label{sec:ch6_related}

A variety of scene understanding tasks have previously been proposed, including layered scene decomposition~\cite{yang2011layered}, instance segmentation~\cite{he2017mask}, amodal segmentation~\cite{li2016amodal}, and scene parsing~\cite{chen2017deeplab}. In order to clarify the relationships of our work to the relevant literature, Table~\ref{tab:ch6_related_work} gives a comparison based on three aspects: what the goals are, which information is used, and on which dataset is evaluated.

\subsection{Inmodal Perception}

The layered scene decomposition for visible regions has been extensively studied in the literature. Shade \etal \cite{shade1998layered} first proposed a representation called a layered depth image (LDI), which contains multiple layers for a complex scene. Based on this image representation that requires occlusion reasoning, the early works focused on ordering the semantic map as occluded and visible regions. Winn and Shotton \cite{winn2006layout} proposed a LayoutCRF to model several occlusions for segmenting partially occluded objects. Gould \etal \cite{gould2009decomposing} decomposed a scene into semantic regions together with their spatial relationships. Sun \etal \cite{sun2010layered} utilized an MRF to model the layered image motion with occlusion ordering. Yang \etal~\cite{yang2010layered,yang2011layered} formulated a layered object detection and segmentation model, in which occlusion ordering for all detected objects was derived. This inferred order for all objects has been used to improve scene parsing~\cite{tighe2014scene} through a CRF. Zhang \etal \cite{zhang2015monocular} combined CNN and MRF to predict instance segmentation with depth ordering. While these methods evaluate occlusion ordering, their main goal is to improve the inmodal perception accuracy for object detection, image parsing, or instance segmentation using the spatial occlusion information. In contrast to these methods, our method \emph{not} only focuses on visible regions with structural inmodal perception, but also tries to solve for amodal perception. \ie to learn \emph{what is behind the occlusion}.

\subsection{Amodal Image/Instance Perception}

Some initial steps have been taken toward amodal perception, exploring the invisible regions. Guo and Hoiem \cite{guo2012beyond} investigated background segmentation map completion by learning relationships between occluders and background. Subsequently, \cite{liu2016layered} introduced the Occlusion-CRF to handle occlusions and complete occluded surfaces. Kar \etal \cite{kar2015amodal} focused on amodal bounding box completion, where the goal is to predict the intact extent of the bounding box. The common attribute in these earlier amodal perception works is using piecemeal representations of a scene, rather than a full decomposition that infers the amodal shapes for all objects. 

The success of advanced deep networks trained on large-scale annotated datasets has recently led to the ability to get more comprehensive representations of a scene. Instance segmentation \cite{pinheiro2015learning,dai2016instance,pinheiro2016learning,li2017fully} deal with detecting, localizing and segmenting all objects of a scene into individual instances. This task combines the classical object detection \cite{girshick2014rich,he2015spatial,girshick2015fast,ren2015faster} and semantic segmentation \cite{long2015fully,chen2017deeplab,badrinarayanan2017segnet}. However, these notable methods typically segment the scene only into visible regions, and do \emph{not} have an explicit structural representation of a scene. We believe a key reason is the lack of large-scale datasets with corresponding annotations for amodal perception and occlusion ordering. The widely used datasets, such as Pascal VOC 2012~\cite{everingham2010pascal}, NYU Depth v2~\cite{silberman2012indoor}, COCO \cite{lin2014microsoft}, KITTI \cite{Geiger2012we}, and CityScapes \cite{Cordts2016Cityscapes}, contain only annotations for the visible instances, purely aiming for 2D inmodal perception.

To mitigate the lack of annotated datasets, Li \etal \cite{li2016amodal} presented a self-supervised approach by pasting occluders into an image. Although reasonable amodal segmentation results are shown, a quantitative comparison is unavailable due to the lack of ground-truth annotations for invisible parts. In more recent works, the completed masks for occluded parts are provided in COCOA~\cite{zhu2017semantic} and KINS~\cite{qi2019amodal}, which are respectively a subset of COCO~\cite{lin2014microsoft} and KITTI~\cite{Geiger2012we}. However, their annotations for invisible parts are manually labeled, which is highly subjective~\cite{ehsani2018segan,zhan2020self}. Furthermore, these datasets are mainly used for the task of inferring amodal semantic maps and are not suitable for the task of RGB appearance completion, since the ground truth RGB appearance for occluded parts are not available. In contrast, we jointly address these two amodal perception tasks using our constructed CSD dataset.

\subsection{Amodal Perception for both Mask and Appearance}

The problem of generating the amodal RGB appearance for the occluded parts is highly related to semantic image completion. The latest methods \cite{pathak2016context,yang2017high,iizuka2017globally,yu2018generative,zheng2019pluralistic,Nazeri_2019_ICCV} extend GANs \cite{goodfellow2014generative} and CGANs \citep{mirza2014conditional} to address this task, generating new imagery to fill in partially erased image regions. However, they mainly focus on object removal, needing users to interactively annotate the objects to be removed. 

SeGAN \cite{ehsani2018segan} involved an end-to-end network that sequentially infers amodal masks and generates complete RGB appearances for instances. The instance depth order is estimated by comparing the areas of the full and visible masks. PCNet~\cite{zhan2020self}  used a self-supervised learning approach to recover masks and content using only visible annotations. However, these works mainly present results in which the ground truth visible mask is used as input, and are sensitive to errors in this visible mask. As stated in~\cite{zhan2020self}, their focus is on amodal completion, rather than a scene understanding for amodal perception. While the recent work of Yan \etal \cite{yan2019visualizing} tried to visualize the invisible from a single input image, it only tackles the occluded ``vehicle'' category, for which there is much less variation in amodal shape and RGB appearance, and thus easier to model. 

There are two recent works that attempt to learn structural scene decomposition with amodal perception. MONet \cite{burgess2019monet} combined an attention network and a CVAE~\cite{kingma2013auto} for jointly modeling objects in a scene. While it is nominally able to do object appearance completion, this unsupervised method has only been shown to work on simple toy examples with minimal occlusions. Dhamo \etal \cite{Dhamo2019iccv} utilized Mask-RCNN \cite{he2017mask} to obtain visible masks, and conducted RGBA-D completion for each object. However, depth values are hard to accurately estimate from a single image, especially in real images without paired depth ground-truths. Besides, they still considered the decomposition and completion separately. In practice, even if we use domain transfer learning for depth estimation, the pixel-level depth value for all objects are unlikely to be consistent in a real scene. Therefore, our method uses an instance-level occlusion order, called the ``2.1D'' model \cite{yang2011layered}, to represent the structural information of a scene, which is easier to be inferred and manipulated.

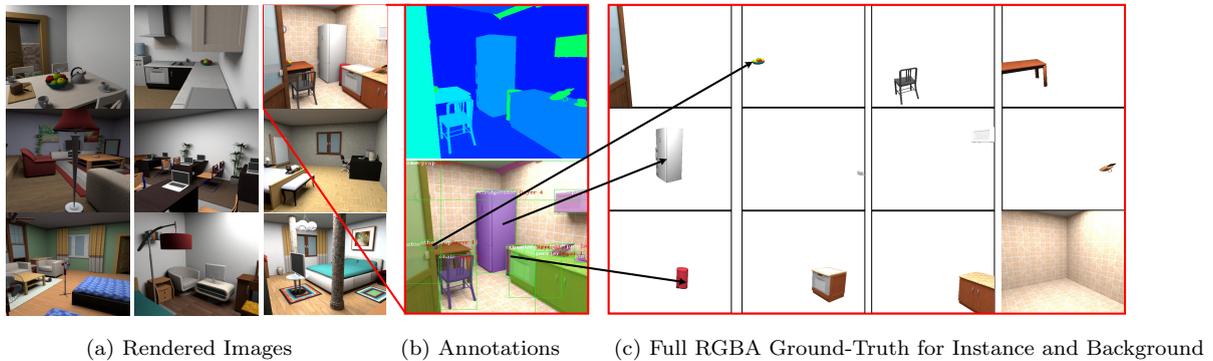
\begin{figure}[tb!]
	\centering
	\includegraphics[width=\linewidth]{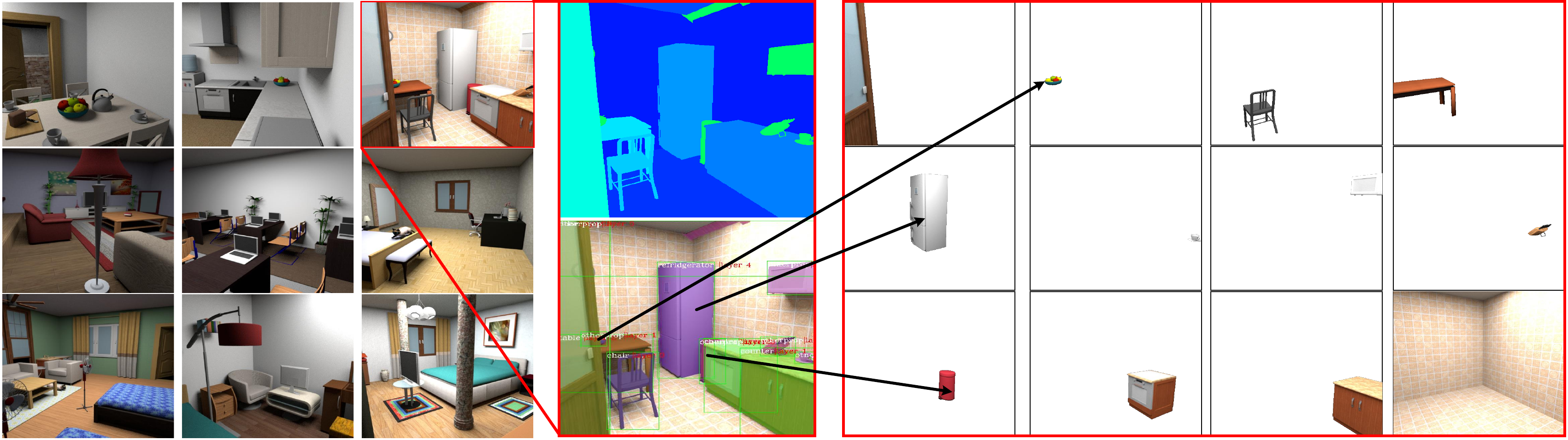}
	\begin{picture}(0,0)
	\put(-180,0){\scriptsize(a) Rendered Images}
	\put(-62,0){\scriptsize(b) Annotations}
	\put(18,0){\scriptsize(c) Full RGBA Ground-Truth for Instance and Background}
	\end{picture}
	\caption[Our rendered dataset]{\textbf{Our rendered dataset.} (a) High quality rendered RGB images. (b) Semantic map and instance annotations with bbox, category, ordering and segmentation map. (c) Intact RGBA ground-truth for instances and background.}
	\label{fig:ch6_datasets}
\end{figure}

\section{Data Collection}\label{sec:ch6_data}

Large datasets with complete ground-truth appearances for all objects are very limited. Burgess \etal \cite{burgess2019monet} created the \emph{Objects Room dataset}, but only with toy objects. Ehsani \etal \cite{ehsani2018segan} and Dhamo \etal \cite{Dhamo2019iccv} rendered synthetic datasets. However, the former only includes 11 rooms, with most viewpoints being atypical of real indoor images. The latter's OpenGL-rendered dataset appears to have more typical viewpoints with rich annotations, but the OpenGL-rendered images have low realism. Recently, Zhan \etal \cite{zhan2020self} explored the \emph{amodal completion} task through self-supervised learning without the need of amodal ground-truth. However, a fair quantitative comparison is not possible as no appearance ground-truth is available for invisible parts. 

To mitigate this issue, we rendered a realistic dataset with Maya \cite{Maya}, instead of the OpenGL-rendering used in Chapter \ref{ch:Synthestic2Real}. We can train the supervised model and test the unsupervised model on this synthetic data with masks and RGB appearance ground-truths for all occluded parts. 

\paragraph{Data Rendering} Our rendered data is based on a total of 10.2k views inside over 2k rooms (CAD models from SUNCG \cite{song2017semantic}) with various room types and lighting environments (see Figure~\ref{fig:ch6_datasets}(a)). To select the typical viewpoints, we first sampled many random positions, orientations and heights for the camera. Only when a view contains at least 5 objects will we render the image and the corresponding ground-truth of each instance. To avoid an excessive rendering workload, we separately rendered each isolated object, as well as the empty room, as shown in Figure~\ref{fig:ch6_datasets}(c). This allows us to then freely create the ground-truths of each layer by compositing these individual objects and background using the instance occlusion order. The rendering details and statistics of the dataset can be found in the Appendix \ref{ch6_appendix_data}. 

\paragraph{Data Annotation} Each rendered scene is accompanied by a global semantic map and dense annotations for all objects. As shown in Figure~\ref{fig:ch6_datasets}(b) and Figure~\ref{fig:ch6_datasets}(c), the intact RGB appearances are given, as well as categories (the 40 classes in NYUDv2 \cite{Silberman:ECCV12}), bounding boxes and masks for complete objects, as well as for only the visible regions. Furthermore, the absolute layer order and pairwise occlusion order shown in Figure~\ref{fig:ch6_layer_representation} are also defined in our rendered dataset. The detail examples are presented in Appendix \ref{ch6_appendix_data_annotation}.

\begin{figure}[tb!]
	\centering
	\includegraphics[width=\linewidth]{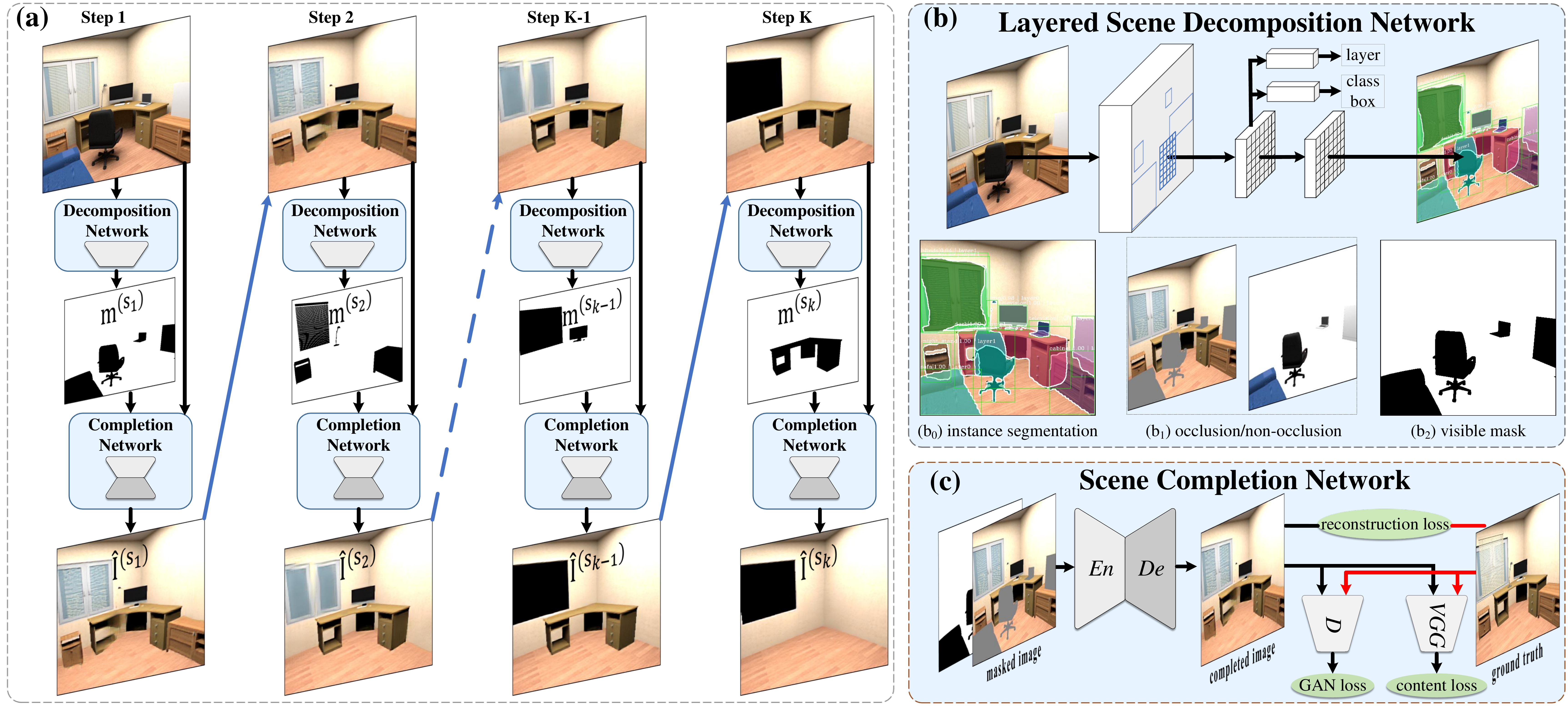}
	\caption[An illustration of the CSDNet framework]{\textbf{An illustration of the CSDNet framework.} (a) Overall layer-by-layer completed scene decomposition pipeline. In each step, the layered decomposition network selects out the fully visible objects. The completion network will complete the resultant holes with appropriate imagery. The next step starts again with the completed image. (b) The layered scene decomposition network estimates instance masks and binary occlusion relationships. (c) The completion network generates realistic content for the original \emph{invisible} regions.}
	\label{fig:ch6_framework}
\end{figure}

\section{Approach}\label{sec:ch6_method}

In this work, we aim to derive a higher-level structural decomposition of a scene. When given a single RGB image $\mathbf{I}$, our goal is to decompose all objects in it and infer their fully completed RGB appearances, together with their underlying occlusion relationships (As depicted in Figure~\ref{fig:ch6_example}). Our system is designed to carry out inmodal perception for \emph{visible} structured instance segmentation, and also solve the amodal perception task of completing shapes and appearances for originally \emph{invisible} parts.

Instead of directly predicting the invisible content and decoupling the occlusion relationships of all objects at one pass, we use the fact that foreground objects are more easily identified, detected and segmented without occlusion. Our CSDNet decomposes the scene layer-by-layer. As shown in Figure~\ref{fig:ch6_framework}(a), in each step $s_{k-1}$, given an image $\mathbf{I}^{(s_{k-1})}$, the layered segmentation network creates masks as well as occlusion labels for all detected objects. Those instances classified as fully visible are extracted out, and the scene completion network generates appropriate appearances for the invisible regions. The completed image $\mathbf{\hat{I}}^{(s_{k-1})}$ will then be resubmitted for layered instance segmentation in the next step $s_{k}$. This differs significantly from previous works~\cite{ehsani2018segan,Dhamo2019iccv,burgess2019monet,zhan2020self,ling2020variational}, which do not adapt the segmentation process based on completion results.

Our \emph{key novel insight} is that scene completion generates completed shapes for originally occluded objects by leveraging the global scene context, so that they are subsequently easier to be detected and segmented without occlusion. Conversely, better segmented masks are the cornerstones to complete individual objects by precisely predicting which regions are occluded. Furthermore, this interleaved process enables extensive \emph{information sharing between these two networks}, to holistically solve for multiple objects, and produces a structured representation for a scene. This contrasts with existing one-pass methods~\cite{ehsani2018segan,Dhamo2019iccv,burgess2019monet,zhan2020self,ling2020variational}, where the segmentation and completion are processed separately and instances are handled independently. Together with the benefit of occlusion reasoning, our system is able to explicitly learn \emph{which parts of the objects and background are occluded that need to be completed}, instead of freely extending to arbitrary shapes.

\subsection{Layered Scene Decomposition}\label{sec:ch6_layered_scene}

As shown in Figure~\ref{fig:ch6_framework}, our layered scene decomposition network comprehensively detect objects in a scene. For each candidate instance, it outputs a class label, a bounding-box offset and an instance mask. The system is an extension of Mask-RCNN \cite{he2017mask}, which consists of two main stages. In the first stage, the image is passed to a \emph{backbone network} (\eg ResNet-50-FPN \cite{lin2017feature}) and next to a \emph{region proposal network} (RPN \cite{ren2015faster}) to get object proposals. In the second stage, the network extracts features using \emph{RoIAlign} from each candidate box, for passing to object classification and mask prediction. We refer readers to \cite{he2017mask,chen2019hybrid} for details.

To determine if an object is fully visible \emph{or} partially occluded, a new parallel branch for this binary occlusion classification is added, as shown in Figure~\ref{fig:ch6_framework}(b). This decomposition is done consecutively layer-by-layer, where at each step it is applied to a single RGB derived from the counterpart scene completion step. While only binary decisions are made in each step, after a full set of iterations, a comprehensive layered occlusion ordering is obtained. The following parts describe how this is done, looking first at the instance depth order representation, followed by how the occlusion head is designed.

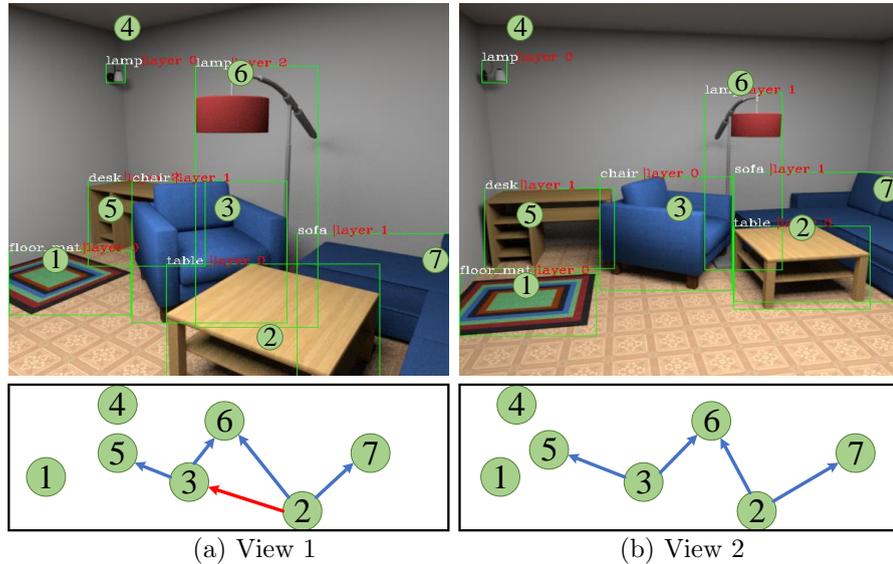
\begin{figure}[tb!]
	\centering
	\includegraphics[width=0.8\linewidth]{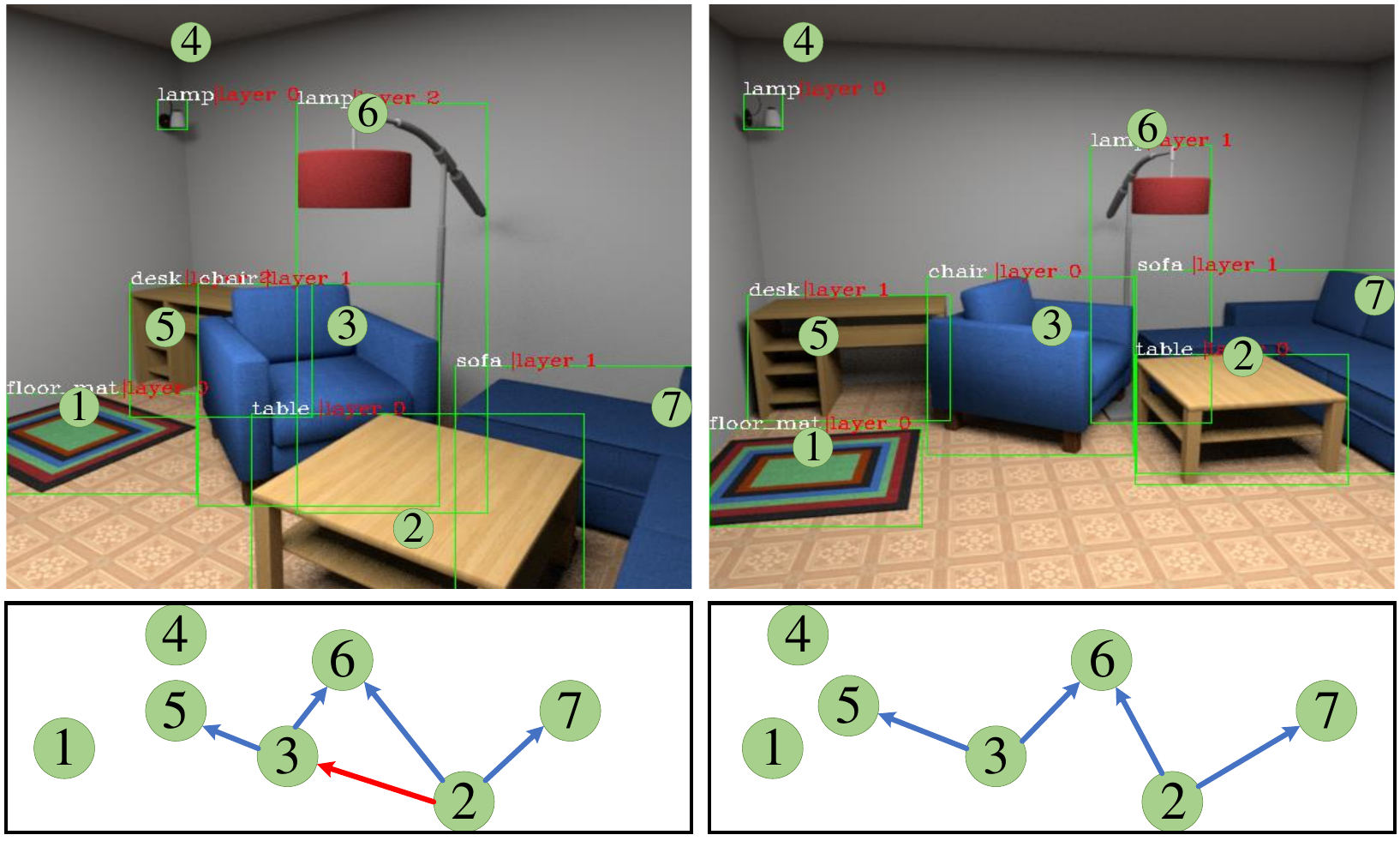}
	\begin{picture}(0,0)
	\put(-270,-8){\footnotesize (a) View 1}
	\put(-110,-8){\footnotesize (b) View 2}
	\end{picture}
	\caption[Instance depth order representation]{\textbf{Instance depth order representation.} Top images show absolute layer order~\cite{qi2019amodal} in different views. Bottom directed graphs give the pairwise order between objects. }
	\label{fig:ch6_layer_representation}
\end{figure}

\paragraph{Instance depth order representation} Absolute layer order and pairwise occlusion order are two standard representations for occlusion reasoning in a scene \cite{sun2010layered,tighe2014scene}. As shown in Figure~\ref{fig:ch6_layer_representation}, the definition for our \emph{absolute layer order} $\mathcal{L}$ follows~\cite{qi2019amodal}, where fully visible objects are labeled as 0, while other occluded objects have 1 order higher than the highest-order instance occluding them (see top images in Figure~\ref{fig:ch6_layer_representation}). We interpret the \emph{pairwise occlusion order matrix} as a directed graph $G=(\Omega,W)$ (see bottom graphs in Figure~\ref{fig:ch6_layer_representation}), where $\Omega$ is a discrete set of all instances with number $N$, and $W$ is a $N\times N$ matrix. $W_{i,j}$ is the occlusion relationship of instance $i$ to instance $j$. We use three numbers to encode the order --- $\{-1$: occluded, 0 : no relationship, 1: front$\}$. For example, the chair (instance \#3) is occluded by the table (instance \#2), so the pairwise order for the chair is $W_{3, 2} = -1$, while the pairwise order for the table is inversely labeled as $W_{2, 3} = 1$. 

\paragraph{Occlusion head} In practice, we find it hard to directly predict these instance depth orders. The absolute layer order index $l\in\mathcal{L}$ cannot be predicted purely from local features in a bounding box, since it depends on the global layout of all objects in a scene. Furthermore, this index is very sensitive to viewpoints, \eg in Figure~\ref{fig:ch6_layer_representation}, the desk (instance \#5) is occluded by only one object (instance \#3) in both views, but the absolute layer order indices of the desk are different: ``2'' \emph{vs} ``1''. In contrast, pairwise ordering $G$ captures the occlusion relationships between pairs of objects, but all pairs have to be analyzed, leading to scalability issues in the current instance segmentation network. As R-CNN-based system creates 2,000 candidate objects, this pairwise analysis requires building an unwieldy 2k$\times$2k features. Even if we were to restrict these to the 100 highest scoring detection boxes, it will still be very memory intensive.

We circumvent these problems as our occlusion classifier only predicts a \emph{binary occlusion label}: $\{0, 1\}$ in each step, where $0$ is fully visible, and $1$ is occluded, following the setting of absolute layer order. During training, each ground-truth binary occlusion label is determined from the pairwise order of the actual objects present in the scene (see details in the Appendix \ref{ch6_supp_experiment}). The occlusion head in our layered scene decomposition network is a \emph{fc} layer, which receives aligned features from each RoI and predicts the binary occlusion label. 

\paragraph{Decomposition Loss} The multi-task loss function for layered scene decomposition is defined as follows:
\begin{equation}
{L}_\text{decomp} = \sum_{t=1}^{T}\alpha_t({L}_\text{cls}^t + {L}_\text{bbox}^t + {L}_\text{mask}^t+ {L}_\text{occ}^t) + \beta{L}_\text{seg}
\label{eq:ch6_decomposition_loss}
\end{equation}
where classification loss ${L}_\text{cls}^t$, bounding-box loss ${L}_\text{bbox}^t$, mask loss ${L}_\text{mask}^t$ and semantic segmentation loss ${L}_\text{seg}$ are identical to those defined in HTC \cite{chen2019hybrid}, and ${L}_\text{occ}^t$ is the occlusion loss at the cascade refined stage $t$ (three cascade refined blocks in HTC~\cite{chen2019hybrid}), using binary cross-entropy loss~\cite{long2015fully} for each RoI.

\subsection{Visiting the Invisible by Exploring Global Context}\label{sec:ch6_viv_global}

In our solution, we treat visiting the invisible as a \emph{semantic image completion}~\cite{pathak2016context} problem. As illustrated in Figure~\ref{fig:ch6_framework}, in step $s_{k-1}$, after removing the front visible instances, the given image $\mathbf{I}^{(s_{k-1})}$ is degraded to become $\mathbf{I}_m^{(s_{k-1})}$. Our goal is to generate appropriate content to complete these previously \emph{invisible} regions (being occluded) for the next layer $\mathbf{I}^{(s_{k})}$. Unlike existing methods that complete each object independently~\cite{ehsani2018segan,Dhamo2019iccv,burgess2019monet,ling2020variational}, our model completes multiple objects in each step layer-by-layer, such that the information from earlier scene completions propagate to later ones. The global scene context is utilized in each step. 

To visit the invisible, it is critical to know which parts are invisible that need to be completed. The general image completion methods use manually interactive masks as input, which differs from our goal. Recent related works~\cite{ehsani2018segan,zhan2020self,ling2020variational} depend on the ground-truth visible masks as input to indicate which parts are occluded. In contrast, our system selects out fully visible objects and automatically learns which parts are occluded in each step. The holes left behind explicitly define the occluded regions for remaining objects, and thus the completed shapes for remaining objects must be deliberately \emph{restricted to these regions}, instead of being allowed to grow freely using only the predicted visible masks.

We use our previous PICNet~\cite{zheng2019pluralistic} framework to train our completion network. While our original PICNet was designed for diversity, here we only want to obtain the best result closest to the ground-truth. Therefore, we only use the encoder-decoder structure, and eschew the random sampling aspect.

\paragraph{Completion Loss} The overall scene completion loss function is given by
\begin{equation}
{L}_\text{comp} = \alpha_\text{rec}{L}_\text{rec} + \alpha_\text{ad}{L}_\text{ad} + \alpha_\text{per}{L}_\text{per}
\label{eq:ch6_completion_loss}
\end{equation}
where reconstruction loss ${L}_\text{rec}$ and adversarial loss ${L}_\text{ad}$ are identical to those in PICNet \citep{zheng2019pluralistic} proposed in Chapter \ref{ch:PICNet}. The perceptual loss ${L}_\text{per}=|\mathbf{F}^{(l)}(\mathbf{\hat{I}}^{(s_{k})}) - \mathbf{F}^{(l)}(\mathbf{I}^{(s_{k})})|$ \citep{johnson2016perceptual}, based on a pretrained VGG-19 \citep{vgg}, is the $l_1$ distance of features $\mathbf{F}$ in $l$-th layer between the generated image $\mathbf{\hat{I}}^{(s_{k})}$ and ground-truth $\mathbf{I}^{(s_{k})}$.

\subsection{Inferring Instance Pairwise Occlusion Order}\label{sec:ch6_infer_order}

As discussed in Section~\ref{sec:ch6_layered_scene}, absolute layer order $\mathcal{L}$ is sensitive to errors. If one object is incorrectly selected as a front object in an earlier step, objects behind it will have their absolute layer order incorrectly shifted. Hence in keeping with prior works~\cite{ehsani2018segan,zhan2020self}, we use the pairwise occlusion order $G=(\Omega,W)$ to represent our final instance occlusion relationships for evaluation. 

Given a single image $\mathbf{I}$, our model decomposes it into instances with completed RGB appearances $A_\Omega^{S_K}$. Here, $A$ denotes the amodal perception instance (inclusive of both mask and appearance), $\Omega$ specifies instances in the scene, and $S_K$ indicates which layers are the instances in (selected out in step $s_k$). When two segmented amodal masks $A_{\omega_i}^{s_i}$ and $A_{\omega_j}^{s_j}$ overlap, we infer their occlusion relationship based on the order of the object-removing process, formally:
\begin{equation}
W_{\omega_i,\omega_j} = \left\{\begin{array}{ll}
0 & \textrm{if $O(A_{\omega_i}^{s_i}, A_{\omega_j}^{s_j})=0$}\\
1 & \textrm{if $O(A_{\omega_i}^{s_i}, A_{\omega_j}^{s_j})>0$ and $s_i<s_j$}\\
-1 & \textrm{if $O(A_{\omega_i}^{s_i}, A_{\omega_j}^{s_j})>0$ and $s_i\geq s_j$}\\
\end{array} \right.
\end{equation}
where $O(A_{\omega_i}^{s_i}, A_{\omega_j}^{s_j})$ is the area of overlap between instances $\omega_i$ and $\omega_j$. If they do not overlap, they share no pairwise depth-order relationship in a scene. If there is an overlap and the instance $\omega_i$ is first selected out with a smaller layer order, the inferred pairwise order is $W_{\omega_i,\omega_j}$ = 1; otherwise it is labeled as $W_{\omega_i,\omega_j}$ = -1. Hence the instance occlusion order only depends on the order (selected out step) of removal between the two instances, and do not suffer from shift errors. 

\begin{figure}[tb!]
	\centering
	\includegraphics[width=\linewidth]{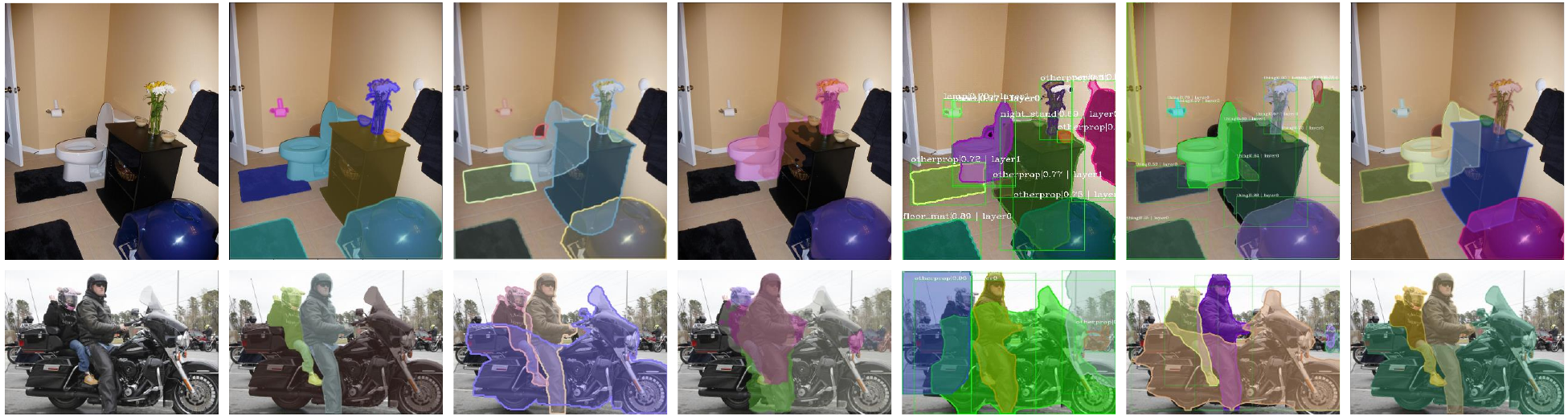}
	\begin{picture}(0,0)
	\put(-200,0){\footnotesize(a) Input}
	\put(-147,0){\footnotesize(b) Inmodal}
	\put(-84,0){\footnotesize(c) PCNet}
	\put(-32,0){\scriptsize(d) AmodalMask}
	\put(32,0){\footnotesize(e) Ours-Syn}
	\put(90,0){\footnotesize(f) Ours-Real}
	\put(164,0){\footnotesize(g) GT}
	\end{picture}
	\caption[Amodal instance segmentation results on the COCOA validation set]{\textbf{Amodal instance segmentation results on the COCOA validation set.} Our model trained on synthetic dataset (Ours-syn) achieves visually reasonable results in the similar real indoor scenes (example in top row), but it fails in some dissimilar real scenes (example in bottom row). After training on the real data with ``pseudo ground-truths'', the model (Ours-Real) performs much better. Note that, unlike the PCNet~\citep{zhan2020self} that need visible inmodal ground-truth masks as input, our system decomposes a scene using only an RGB image.}
	\label{fig:ch6_example_amodal_seg_real}
\end{figure}
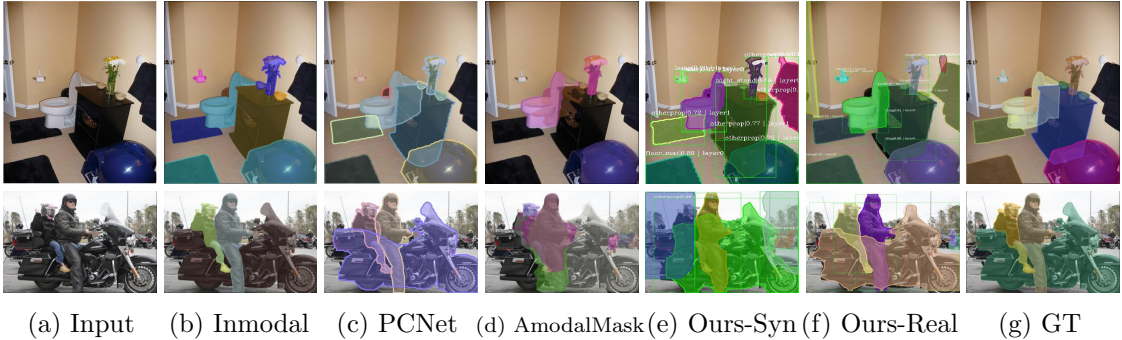

\subsection{Training on Real Data with Pseudo Ground-truth}

Real-world data appropriate for a completed scene decomposition task is difficult to acquire, because ground truth shapes and RGB appearance for occluded parts are hard to collect without very extensive manual interventions, \eg deliberate physical placement and removal of objects. Although our proposed model trained on the high-quality rendered data achieves visually reasonable results in some real scenes that share similarities to the rendered dataset (\eg indoor scene in top row of Figure~\ref{fig:ch6_example_amodal_seg_real}), it does not generalize well to dissimilar real scenes (\eg outdoor scene in bottom row of Figure~\ref{fig:ch6_example_amodal_seg_real}). These are caused by: 1) differences in labeled categories between synthetic and real datasets, especially between indoor and outdoor scenes; and 2) inconsistencies between synthetically trained image completion of masked regions and fully visible real pixels.

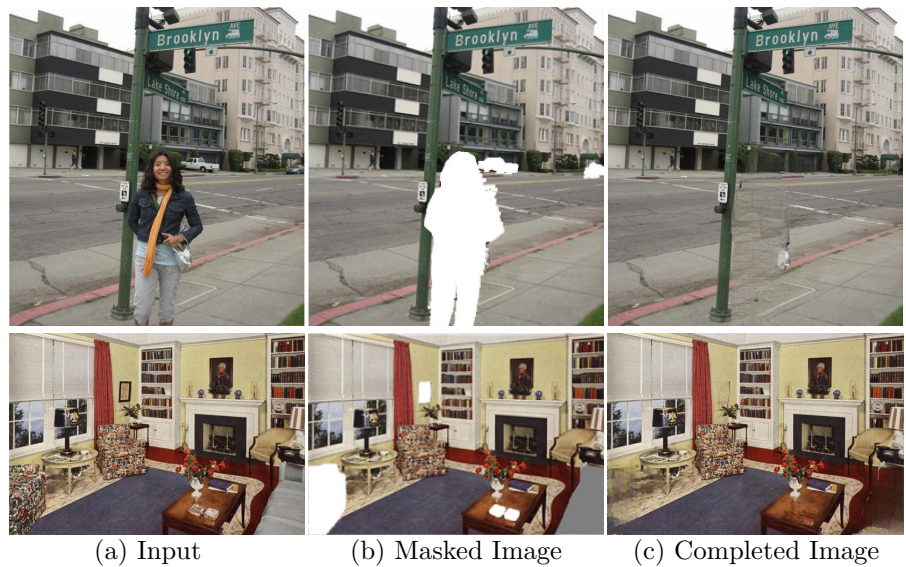
\begin{figure}[tb!]
    \centering
    \includegraphics[width=0.8\linewidth]{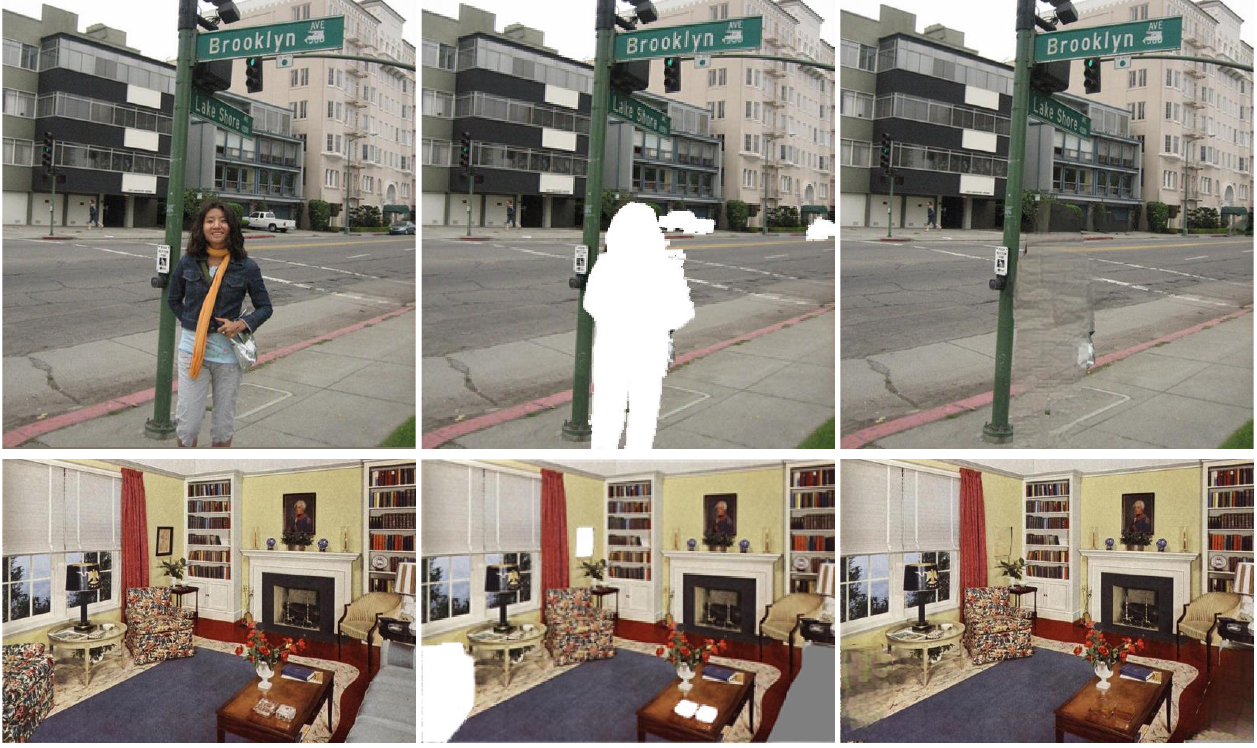}
        \begin{picture}(0,0)
			\put(-308,-8){\footnotesize (a) Input}
			\put(-212,-8){\footnotesize (b) Masked Image}
			\put(-106,-8){\footnotesize (c) Completed Image}
		\end{picture}
		\vspace{0.2cm}
		\caption[Pseudo RGB ground-truth]{\textbf{Pseudo RGB ground-truth.} (a) Input. (b) Masked image by selecting out the fully visible objects. (c) Pseudo ground-truth generated from our model trained on synthetic data.}
		\label{fig:ch6_syn_completed}
\end{figure}

One alternative is to simply use an image completion network trained only on real images. From our experience, this performs poorly in a scene decomposition task. The reason is that while real-trained image completion methods are able to create perceptually-pleasing regions and textures for a single completion, they do not appear to have the ability to adhere to consistent object geometry and boundaries when de-occluding, which is crucial for object shape completion. As a result, errors accumulate even more dramatically as the decomposition progresses.

Our \emph{key motivating insight} is this: instead of training the model entirely without ground-truth in the completion task, we train it in a semi-supervised learning manner, exploiting the scene structure and object shape knowledge that has been gained in our synthetically-trained CSDNet. As shown in Figure~\ref{fig:ch6_syn_completed}, this synthetic completion model is able to generate visually adequate appearance, but more importantly it is better able to retain appropriate geometry and shapes. We can use this to guide the main image completion process in real-word data, while allowing a GAN-based loss to increase the realism of the output.

\begin{figure}[tb!]
    \centering
    \includegraphics[width=0.8\linewidth]{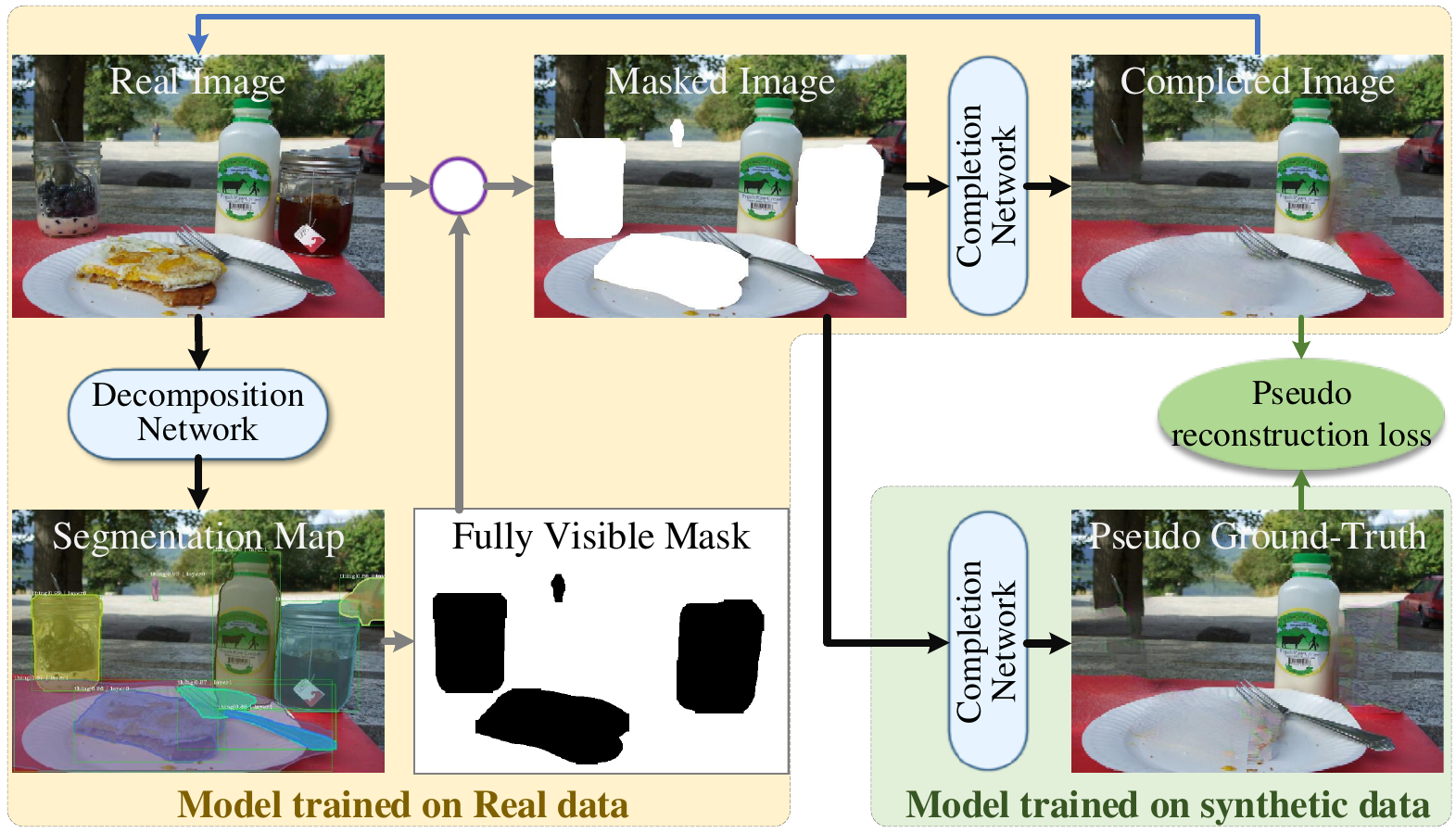}
    \caption[Training pipeline for real images]{\textbf{Training pipeline for real images.} We introduce a semi-supervised learning method for real data by providing \emph{pseudo} RGB ground-truth for originally invisible regions.}
    \label{fig:ch6_framework_real}
\end{figure}

Specifically, for a real image $\mathbf{I}$, we first train the layered decomposition network using the manual annotated amodal labels. In a step, after segmenting and selecting out the foreground objects, we obtain $\mathbf{\hat{I}}_{syn}^{(s_k)}=G(\mathbf{I}_m^{(s_k)};\theta_{\text{syn}})$ to serve as ``pseudo ground-truth'' (green box in Figure~\ref{fig:ch6_framework_real}) through the completion model trained on synthetic data. We then train the completion network $G(\mathbf{I}_m^{(s_k)};\theta_{\text{real}})$ using the loss function of equation (\ref{eq:ch6_completion_loss}) by comparing the output $\mathbf{\hat{I}}_{real}^{(s_k)}$ to ``pseudo ground-truth'' $\mathbf{\hat{I}}_{syn}^{(s_k)}$. Like \cite{sengupta2020background}, we also reduce the weights of reconstruction loss ${L}_\text{rec}$ and perceptual loss ${L}_\text{per}$ to encourage the output to be biased towards the real image distribution via the discriminator loss ${L}_\text{ad}$. It is worth noticing that the completed image is \emph{passed back} to the layered decomposition network in the next layer, where the decomposition loss ${L}_\text{decomp}$ in equation (\ref{eq:ch6_decomposition_loss}) will be backpropagated to the completion network. This connection allows the completion network to \emph{learn to complete real-world objects that might not be learned through the synthetic data}.

\section{Results and Applications}\label{sec:ch6_experiment}

\subsection{Setup}

\paragraph{Datasets} We evaluated our system on three datasets: \textbf{COCOA}~\cite{zhu2017semantic}, \textbf{KINS}~\cite{qi2019amodal} and the rendered \textbf{CSD}. \textbf{COCOA} is annotated from COCO2014~\cite{lin2014microsoft}, a large scale natural image datasets, in which 5,000 images are selected to manually label with pairwise occlusion orders and amodal masks. \textbf{KINS} is derived from the outdoor traffic dataset KITTI \cite{Geiger2012we}, in which 14,991 images were labeled with absolute layer orders and amodal masks. \textbf{CSD} is our rendered synthetic dataset, which contains 8,298 images, 95,030 instances for training and 1,012 images, 11,648 instances for testing. We conducted thorough experiments and ablation studies to assess the quality of the completed results for invisible appearance estimation (since the in-the-wild datasets lack ground-truth for the occluded parts).

\begin{figure}[tb!]
	\centering
	\includegraphics[width=\linewidth]{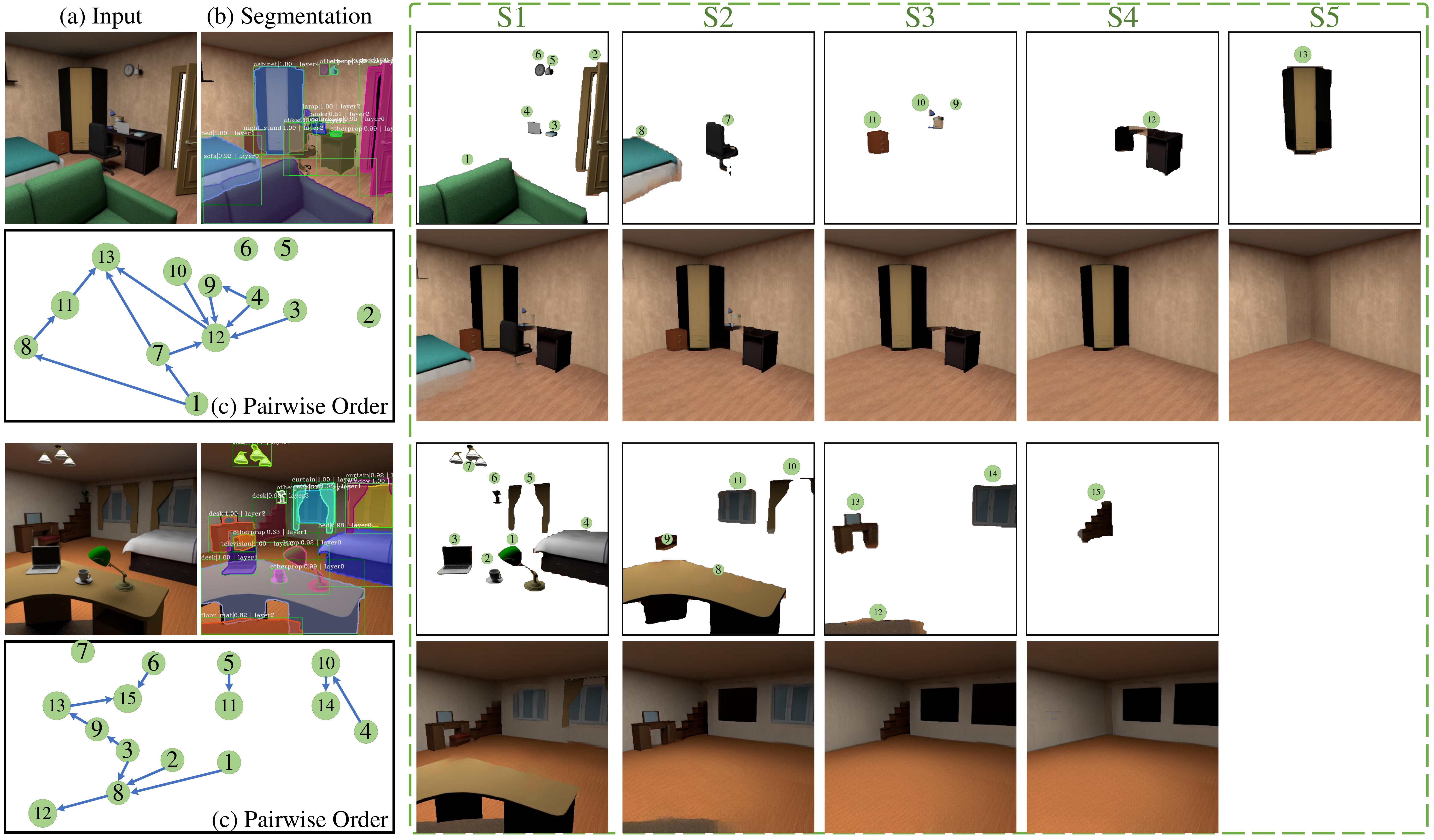}
	\caption[Layer-by-Layer Completed Scene Decomposition]{\textbf{Layer-by-Layer Completed Scene Decomposition} results on rendered CSD testing set. (a) Input RGB images. (b) Final amodal instance segmentation. (c) Inferred directed graph for pairwise order. (d) Columns labeled S1-5 show the decomposed instances (top) and completed scene (bottom) based on the predicted non-occluded masks. Note that the originally invisible parts are filled in with realistic appearance. }
	\label{fig:ch6_results_de_co}
\end{figure}

\paragraph{Metrics} For amodal instance segmentation, we report the standard COCO metrics \cite{lin2014microsoft}, including AP (average over IoU thresholds), $\text{AP}_{50}$, $\text{AP}_{75}$, and $\text{AP}_{S}$, $\text{AP}_{M}$ and $\text{AP}_{L}$ (AP at different scales). Unless otherwise stated, the AP is for mask IoU. For appearance completion, we used RMSE, SSIM and PSNR to evaluate the quality of generated images. All images were normalized to the range $\left[0, 1\right]$.

Since the occlusion order is related to the quality of instance segmentation, we defined a novel metric for evaluating the occlusion order that uses the previous benchmark criterion for instance segmentation. Specifically, given a pairwise occlusion order $G=(\Omega, W)$ predicted by the model, we only evaluate the order for these valid instances that have IoU with ground-truth masks over a given threshold. For instance, if we set the threshold as $0.5$, the predicted instance $\omega$ will be evaluated when we can identify a matched ground-truth mask with $\text{IoU}\geq 0.5$. Hence we can measure the \textbf{occlusion average precision} (OAP) as assessed with different thresholds. 

\subsection{Results on Synthetic CSD Dataset}\label{sec:ch6_csd_results}

We first present results that we obtained from our framework when experimenting on our synthetic CSD dataset.

\subsubsection{Main Results}

\paragraph{Completed scene decomposition} We show the qualitative results of CSDNet in Figure~\ref{fig:ch6_results_de_co}. Given a single RGB image, the system has learned to decompose it into semantically complete instances (\eg counter, table, window) and the background (wall, floor and ceiling), while completing RGB appearance for \emph{invisible} regions. Columns labeled S1-5 show the completed results layer-by-layer. In each layer, fully visible instances are segmented out, and after scene completion some previously occluded regions become fully visible in the next layer, \eg the table in the second example. The final amodal instance segmentation results shown in Figure~\ref{fig:ch6_results_de_co}(b) consist of the fully visible amodal masks in each layer. Note that unlike MONet~\cite{burgess2019monet}, our model does not need predefined slots. The process will stop when it is unable to detect any more objects.

\begin{table}[tb!]
	\centering
	\scriptsize
	\renewcommand{\arraystretch}{1.1}
	\setlength\tabcolsep{6pt}
	\begin{tabular}{@{}l|l|c |c c c |c c c@{}}
		\hlineB{3}
		& SegNet & box AP & mask AP & AP$_{50}$ & AP$_{75}$ & AP$_{S}$ & AP$_M$& AP$_L$ \\
		\hlineB{3}
		Mask-RCNN \citep{he2017mask}& Mask-RCNN & 51.3 & 46.8 & 67.2 & 50.6 & 14.5 & 43.0 & 49.9 \\
		MLC \citep{qi2019amodal}& Mask-RCNN & 52.3 & 47.2 & 67.5 & 50.9 & 14.7 & 43.8 & 50.2 \\
		PCNet \citep{zhan2020self} & Mask-RCNN & - & 43.6 & 59.1 & 43.4 & 11.5 & 40.4 & 46.0 \\
		HTC \citep{chen2019hybrid} & HTC &52.9 & 47.3 & 65.9 & 51.6 & 12.2 & 41.3 & 51.0 \\
		MLC \citep{qi2019amodal}& HTC & 53.6 & 47.9 & 66.1 & 52.3 & 13.1 & 41.9 & 51.7 \\
		PCNet \citep{zhan2020self} & HTC & - & 45.7 & 60.6 & 49.2 & 10.2 & 39.3 & 48.4 \\
		\hline
		\textbf{CSDNet}& Mask-RCNN & 52.6  & 48.7  &  66.2 &  53.1 &  15.7 & 42.8 & 52.2 \\
		\textbf{CSDNet} & HTC &\bf{56.3} & \bf{50.3} & \bf{67.7} & \bf{53.4} & \bf{17.4} & \bf{44.2} & \bf{53.1} \\
		\hline
		\textbf{CSDNet}-gt& Mask-RCNN &  54.9 &  53.1 &  66.5 &  56.9 &  21.4$^*$ &  49.9 &  57.0 \\
		\textbf{CSDNet}-gt & HTC & 60.3$^*$ & 56.0$^*$ & 67.9$^*$ & 59.3$^*$ & 19.6 & 53.4$^*$ & 59.5$^*$ \\
		\hlineB{2.5}
	\end{tabular}
	\caption[Amodal Instance Segmentation on CSD testing sets]{\textbf{Amodal Instance Segmentation on CSD testing sets}. Mask-RCNN~\citep{he2017mask} and HTC~\citep{chen2019hybrid} are the state-of-the-arts of the COCO segmentation challenges. MLC~\citep{qi2019amodal} is the latest amodal instance segmentation work for outdoor scene. PCNet~\citep{zhan2020self} is the self-supervised amodal completion work. The \textbf{CSDNet}-gt holds same training environment as \textbf{CSDNet}, but is tested with completed ground-truths images $\mathbf{I}^{s_*}$ in each step. Best results used ground-truth annotations are marked with $^*$, while best results only used RGB images are in bold.}
	\label{table:ch6_syn_seg}
\end{table}

\begin{table}[tb!]
	\centering
	\scriptsize
	\renewcommand{\arraystretch}{1.1}
	\setlength\tabcolsep{2pt}
	\begin{tabular}{@{}l| l|l | l |c c c c|c c c@{}}
		\hlineB{3}
		& \multicolumn{2}{c|}{Inputs} & \multicolumn{1}{c|}{\multirow{2}{*}{\makecell[c]{Ordering \\ Algorithm}}} & \multirow{2}{*}{OAP} & \multirow{2}{*}{OAP$_{50}$} & \multirow{2}{*}{OAP$_{75}$} & \multirow{2}{*}{OAP$_{85}$} & \multirow{2}{*}{OAP$_{S}$} & \multirow{2}{*}{OAP$_M$}& \multirow{2}{*}{OAP$_L$} \\
		\cline{2-3}
		& \makecell[c]{Amodal} & \makecell[c]{Ordering} & & & & & & & & \\
		\hlineB{3}
		SeGAN \citep{ehsani2018segan} & I + $\text{V}_{gt}$ & $\text{V}_{gt}$ + $\hat{\text{F}}_{pre}$ & IoU Area & 68.4 & - & - & - & - & - & -\\
		SeGAN \citep{ehsani2018segan} & I + $\hat{\text{V}}_{pred}$ & $\text{V}_{gt}$ + $\hat{\text{F}}_{pre}$ & IoU Area & 66.1 & 50.2 & 65.6 & 70.4 & 10.6 & 65.0 & 63.8 \\
		HTC + MLC \citep{qi2019amodal} & I & $\text{V}_{gt}$ + $\hat{\text{F}}_{pre}$ & IoU Area &  76.5 &  70.3 &  77.1 &  79.8 &  11.6 &  69.8 &  78.2 \\
		HTC + MLC \citep{qi2019amodal} & I & $\hat{\text{F}}_{pre}$ + layer &layer order$^1$ & 51.9 & 44.3 & 50.8 & 54.6 & 11.7 & 60.8 & 46.2 \\
		HTC + PCNet \citep{zhan2020self} & I + $\hat{\text{V}}_{pred}$ & $\text{V}_{gt}$ + $\hat{\text{F}}_{pre}$ & IoU Area &  70.8 & 56.9 & 71.3 & 76.0 & 11.3 & 67.1 & 68.6 \\
		\hline
		\textbf{CSDNet} & I & $\hat{\text{F}}_{pre}$ & Area & 44.7 & 45.3 & 45.7 & 45.1 & 17.4 & 34.5 & 41.5 \\
		\textbf{CSDNet} & I & $\hat{\text{F}}_{pre}$ & Y-axis & 62.0 & 60.1 & 61.2 & 62.7 & \bf{63.4} & 58.6 & 66.1 \\
		\textbf{CSDNet} & I & $\text{V}_{gt}$ + $\hat{\text{F}}_{pre}$ & IoU Area & 80.7 & \bf{77.2} & \bf{81.0} & 82.9 & 61.1 & 73.7 & 80.5\\
		\textbf{CSDNet} & I & $\hat{\text{F}}_{pre}$ + layer & layer order & \bf{81.7} & 76.6 & 80.9 & \bf{84.6} & 15.7 & \bf{75.9} & \bf{82.6}\\
		\hline
		\textbf{CSDNet}-gt & I$^{s*}$ & $\hat{\text{F}}_{pre}$ + layer & layer order & 88.9$^*$ & 85.2$^*$ & 88.5$^*$ & 90.1$^*$ & 49.6 & 84.3$^*$ & 89.9$^*$ \\
		\hlineB{2.5}
	\end{tabular}
	\caption[Instance depth ordering on CSD testing sets]{\textbf{Instance depth ordering on CSD testing sets}. We report the pairwise depth ordering on occluded instance pairs $\text{OAP}_{occ}$. I$^{s*}=$ ground-truth completed image in each step ${s*}$, $\text{V}_{gt}=$ visible ground-truth mask, $\hat{\text{V}}_{pred}=$ visible predicated mask, and $\hat{\text{F}}_{pre}=$ full (amodal) predicated mask. $\text{layer order}^1$ only predicts the occlusion / non-occlusion labels in the original image (the first step in our model).}
	\label{table:ch6_syn_depth}
\end{table}

\paragraph{Amodal instance segmentation} We compare CSDNet to the state-of-the-art methods in amodal instance segmentation in  Table~\ref{table:ch6_syn_seg}. As the existing works Mask-RCNN~\cite{he2017mask} and HTC~\cite{chen2019hybrid} are aimed at inmodal perception for visible parts, we retrained their models for amodal perception task, by providing amodal ground-truths. We also retrained MLC~\cite{qi2019amodal} on our rendered dataset, which are the latest work for amodal perception. For PCNet~\cite{zhan2020self}, we used the predicted visible mask as input, rather than the original visible ground-truth annotations. While the HTC~\cite{chen2019hybrid} improves on Mask-RCNN's~\cite{he2017mask}  bounding box AP by about 1.6 points by refining the bounding box offsets in three cascade steps, the improvement for amodal mask segmentation is quite minor at 0.5 points. We believe this is an inherent limitation of methods that attempt amodal segmentation of occluded objects directly without first reasoning about occluding objects and masking their image features, as such the front objects' features will distract the network. In contrast, our CSDNet is able to improve the amodal mask segmentation accuracy by a relative $6.3\%$ with the same \emph{backbone segmentation network} (HTC), by jointing segmentation and completion with layer-by-layer decomposition.

To further demonstrate that better completed images improve amodal segmentation, we consider a scenario with a completion oracle, by using ground-truth appearances to repair the occluded parts in each step. This is denoted as the \textbf{CSDNet}-gt, for which amodal instance segmentation accuracy increases from $47.3\%$ to $56.0\%$ (relative $18.4\%$ improvement). We also note that, while the \textbf{CSDNet}-gt using Mask-RCNN achieves lower bounding box score than our HTC \textbf{CSDNet} (``54.9'' \emph{vs} ``56.3''), the mask accuracy is much higher (``53.1'' \emph{vs} ``50.3''). This suggests that amodal segmentation benefits from better completed images. 

\paragraph{Instance depth ordering} Following~\cite{zhu2017semantic}, we report the pairwise instance depth ordering for correctly detected instances in Table~\ref{table:ch6_syn_depth}. The original SeGAN and PCNet used ground-truth visible masks $\text{V}_{gt}$ as input. For a fair comparison, we first retrained them on our synthetic data using the same segmentation network (HTC~\cite{chen2019hybrid}) for all models. After predicting amodal masks, we assessed various instance depth ordering algorithms: two baselines proposed in AmodalMask~\cite{zhu2017semantic} of ordering by \emph{area}\footnote{We used the heuristic in PCNet~\cite{zhan2020self} --- larger masks are ordered in front for KINS, and behind for COCOA and CSD.} and by \emph{y-axis} (amodal masks closest to image bottom in front), ordering by \emph{incremental area} defined as the \emph{IoU area} between visible and amodal masks\footnote{See details in \cite{zhan2020self}, where the visible ground-truth masks $\text{V}_{gt}$ are used for ordering.}, and our ordering by absolute \emph{layer order} (Section~\ref{sec:ch6_infer_order}).

As can be seen in Table~\ref{table:ch6_syn_depth}, all instantiations of our model outperformed baselines as well as previous models. Unlike SeGAN~\cite{ehsani2018segan} and PCNet \cite{zhan2020self}, our final model explicitly predicts the occlusion labels of instances, which improved the OAP substantially. While MLC \cite{qi2019amodal} predicts the instance occlusion order in a network, it only contains one layer for binary occlusion / non-occlusion labeling. In contrast, our method provides a fully structural decomposition of a scene in multiple steps. Additionally, we observed that our model achieves better performance with a higher IoU threshold for selecting the segmented mask (closer match to the ground-truth masks). We further observed that the occlusion relationships of small objects are difficult to infer in our method. However, the \emph{Y-axis} ordering method had similar performance under various metrics as it only depends on the locations of objects. Note that our depth ordering does \emph{not} rely on any ground-truth that is used in~\cite{ehsani2018segan,zhan2020self}.

\begin{table}[tb!]
	\centering
	\renewcommand{\arraystretch}{1.1}
	\setlength\tabcolsep{4pt}
	\begin{tabular}{@{}l |c c c | c |c c c@{}}
		\hlineB{3}
		& \multicolumn{3}{c|}{C1-$\text{F}_{gt}$} & & \multicolumn{3}{c}{C2} \\
		\cline{2-4}\cline{6-8}
		& RMSE & SSIM & PSNR & & RMSE & SSIM & PSNR \\
		\hlineB{3}
		SeGAN \cite{ehsani2018segan} & 0.1246 & 0.8140 & 21.42 & \multirow{2}{*}{C2a-$\text{V}_{gt}$} & 0.2390 & 0.6045 & 16.03 \\
		PCNet \cite{zhan2020self} & 0.1129 & 0.8267 & 23.16 & & 0.2483 & 0.5931 & 15.54 \\
		\cline{1-8}
		DNT \cite{Dhamo2019iccv} & 0.1548 & 0.7642 & 20.32 & & 0.2519 & 0.5721 & 15.10 \\
		PICNet \cite{zheng2019pluralistic} & 0.0927 & 0.8355 & 28.81 & C2b & 0.1401 & 0.7730 &  24.71 \\
		\textbf{CSDNet} & \bf{0.0614} & \bf{0.9179} & \bf{35.24} & & \bf{0.0914} & \bf{0.8768} & \bf{30.45} \\
		\hlineB{2.5}
	\end{tabular}	
	\caption[Object Completion]{\textbf{Object Completion}. $\text{F}_{gt}$ = full ground-truth mask, $\text{V}_{gt}$ = visible ground-truth mask.  For methods provided with $\text{F}_{gt}$, we only evaluate the completion networks. }
	\label{table:ch6_completion}
\end{table}

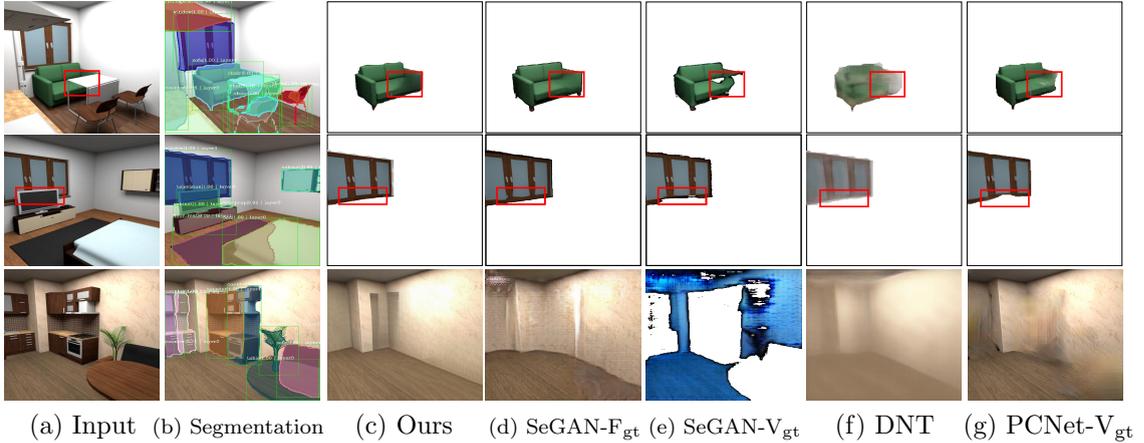
\begin{figure}[tb!]
	\centering
	\includegraphics[width=\linewidth]{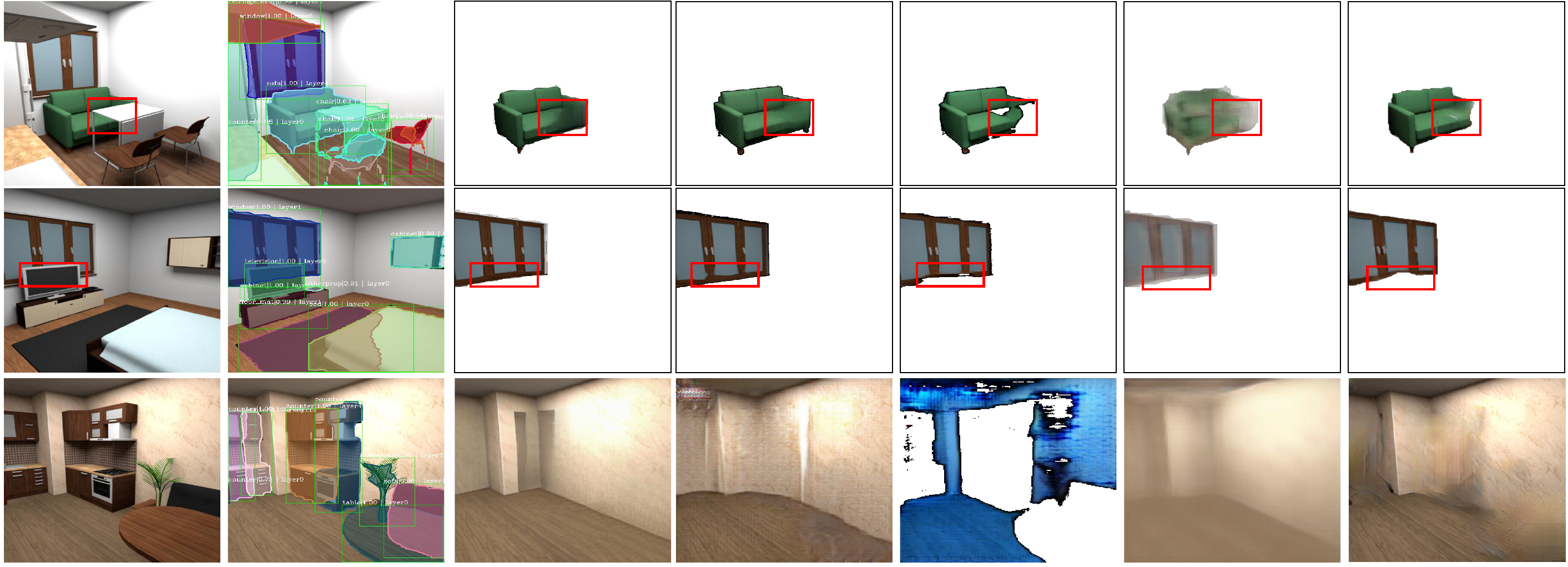}
	\begin{picture}(0,0)
	\put(-200,3){\footnotesize(a) Input}
	\put(-154,3){\scriptsize(b) Segmentation}
	\put(-78,3){\footnotesize(c) Ours}
	\put(-28,3){\scriptsize(d) SeGAN-F$_\text{gt}$}
	\put(32,3){\scriptsize(e) SeGAN-V$_\text{gt}$}
	\put(102,3){\footnotesize(f) DNT}
	\put(150,3){\footnotesize(g) PCNet-V$_\text{gt}$}
	\end{picture}
	\caption[Results for Visiting the Invisible]{Results for \textbf{Visiting the Invisible}. We show the input image, our amodal instance segmentation results, and the objects and background we try to visit. The red rectangles highlight the previously \emph{invisible} regions of occluded objects. }
	\label{fig:ch6_results_invisible}
\end{figure} 

\paragraph{Object completion.} We finally evaluated the quality of generated appearances. We compared our results to those from SeGAN \cite{ehsani2018segan}, Dhamo \etal \cite{Dhamo2019iccv} (abbrev.\ as DNT), PCNet \cite{zhan2020self} and our previous PICNet \cite{zheng2019pluralistic} (original point-attention) in Table \ref{table:ch6_completion}. We evaluated different conditions of: C1) when the ground-truth full mask $\text{F}_{gt}$ is provided to all methods, C2a) when the ground-truth visible mask $\text{V}_{gt}$ is the input to SeGAN and PCNet, and C2b) when an RGB image is the only input to other methods. C2a-$\text{V}_{gt}$ is considered because SeGAN and PCNet assumes that a predefined mask is provided as input. 

In C1-$\text{F}_{gt}$, CSDNet substantially outperformed the other methods. In C2, even when given only RGB images \emph{without} ground-truth masks, our method worked better than SeGAN and PCNet with $\text{V}_{gt}$. One important reason for the strong performance of CSDNet is the \emph{occlusion reasoning} component, which constraints the completed shapes of partly occluded objects based on the global scene context and other objects \emph{during testing}. 

Qualitative results are visualized in Figure~\ref{fig:ch6_results_invisible}. We noted that SeGAN worked well only when ground-truth amodal masks $\text{F}_{gt}$ were available to accurately label which parts were \emph{invisible} that needed filling in, while DNT generated blurry results from simultaneously predicting RGB appearance and depth maps in one network, which is not an ideal approach~\cite{zamir2018taskonomy}. The PCNet \cite{zhan2020self} can not correctly repair the object shape as it trained without ground-truth object shape and appearance. Our CSDNet performed much better on background completion, as it only masked fully visible objects in each step instead of all objects at a go, so that \emph{earlier completed information propagates to later steps}.

\begin{table}[tb!]
	\begin{minipage}[t!]{0.52\textwidth}
		\begin{center}
		    \scriptsize
			\renewcommand{\arraystretch}{1.1}
			\setlength\tabcolsep{2pt}
			\scalebox{1.0}{\begin{tabular}{@{}l | c |c c c| c c c@{}}
					\hlineB{3}
					& train & AP & AP$_{50}$ & AP$_{75}$ & AP$_{S}$ & AP$_M$ & AP$_L$ \\
					\hlineB{3}
					gt & - & 56.0$^*$ & 67.9$^*$ & 59.3$^*$ & 19.6$^*$ & 53.4$^*$ & 59.5$^*$ \\
					w/o & - &  36.8 & 52.6 & 38.3 & 10.8 & 31.6 & 38.2\\
					\hline
					PICNet-point & sep & 40.8 & 63.0 & 43.5 & 12.6 & 37.2 & 43.7 \\
					PICNet-patch & sep & 43.8 & 60.7 & 46.6 & 11.0 & 36.3 & 45.5\\
					\hline
					PICNet-point & end & 47.7 & 63.2 & 50.6 & 14.9 & 41.7  & 51.3 \\
					PICNet-patch & end & \bf{50.3} & \bf{67.7} & \bf{53.4} & \bf{17.4} & \bf{44.2} & \bf{53.1}\\
					\hlineB{2.5}
			\end{tabular}}
		\end{center}
	\end{minipage}	
	\hfill
	\begin{minipage}[t!]{0.48\textwidth}
		\begin{center}
			\scriptsize
			\renewcommand{\arraystretch}{1.2}
			\setlength\tabcolsep{4pt}
			\scalebox{1.0}{\begin{tabular}{@{}l|c|c c c@{}}
					\hlineB{3}
					& train & RMSE & SSIM & PSNR \\
					\hlineB{3}
					gt & - & 0.0614$^*$ & 0.9179$^*$ & 35.24$^*$\\
					\hline 
					M-RCNN\citep{he2017mask} & sep & 0.1520 & 0.7781 & 22.34 \\
					HTC \citep{chen2019hybrid} & sep & 0.1496 & 0.7637 & 26.75\\
					\hline
					M-RCNN\citep{he2017mask}  & end & 0.1345 & 0.7824& 27.31\\
					HTC \citep{chen2019hybrid} & end & \bf{0.0914} & \bf{0.8768} & \bf{30.45} \\
					\hlineB{2.5}
			\end{tabular}}
		\end{center}
	\end{minipage}
	\begin{minipage}[t!]{0.52\textwidth}
	    \scriptsize
		(a) \textbf{Effect of different completion methods on instance segmentation (HTC-based decomposition)}. ``sep'' = separate training of the 2 networks, ``w/o'' = without any completion, and ``end'' = joint training.
	\end{minipage}
	\begin{minipage}[t!]{0.10\textwidth}
	\end{minipage}
	\begin{minipage}[t!]{0.44\textwidth}
	    \scriptsize
		(b) \textbf{Effect of different decomposition methods on scene completion (Patch-Attention PICNet)}. Better scene decomposition improved scene completion.
	\end{minipage}
	\caption[Ablations for joint optimization]{\textbf{Ablations} for joint optimization. In each table, we fixed one model for one subtask and trained different models for the other subtask. Better performance in one task can improve the performance in the other, which demonstrates the joint training of two tasks with layer-by-layer decomposition contributes to each other.}
	\label{table:ch6_ablation}
\end{table}

\subsubsection{Ablation Studies}

To demonstrate the two tasks can contribute to a better scene understanding system, instead of solving them isolated, we ran a number of ablations. 

\paragraph{Does better \emph{completion} help decomposition?} We show quantitative results for a fixed decomposition network (layered HTC \cite{chen2019hybrid} with two completion methods in Table~\ref{table:ch6_ablation}(a). Without any completion (``w/o''), segmented results were naturally bad (``36.8'' \emph{vs} ``50.3'') as it had to handle empty regions. More interestingly, even if advanced methods were used to generate visual completion, the isolated training of the decomposition and completion networks led to degraded performance. This suggests that even when generated imagery looks good visually, there is still a domain or semantic gap to the original visible pixels, and thus flaws and artifacts will affect the next segmentation step. Our original PICNet with patch attention provides better completed results than the original point attention PICNet \cite{zheng2019pluralistic}, resulting in a large improvement (``50.3'' \emph{vs} ``47.7'') of amodal instance segmentation. 

\paragraph{Does better \emph{decomposition} help completion?} To answer this, we report the results of using different scene segmentation networks with a same completion network (Patch-attention PICNet~\cite{zheng2019pluralistic}) in Table~\ref{table:ch6_ablation}(b). We also first considered the ideal situation that ground-truth segmentation masks were provided in each decomposition step. As shown in Table~\ref{table:ch6_ablation}(b), the completion quality significantly improved (RMSE: ``0.0614'', SSIM: ``0.9179'' and PSNR: ``35.24'') as occluded parts were correctly pointed out and the completion network precisely knows which parts need to be completed. HTC~\cite{chen2019hybrid} provided better instance masks than Mask-RCNN~\cite{he2017mask}, which resulted in more accurately completed scene imagery. The best results were with end-to-end jointly training.

\begin{figure}[tb!]
	\centering
	\includegraphics[width=\linewidth]{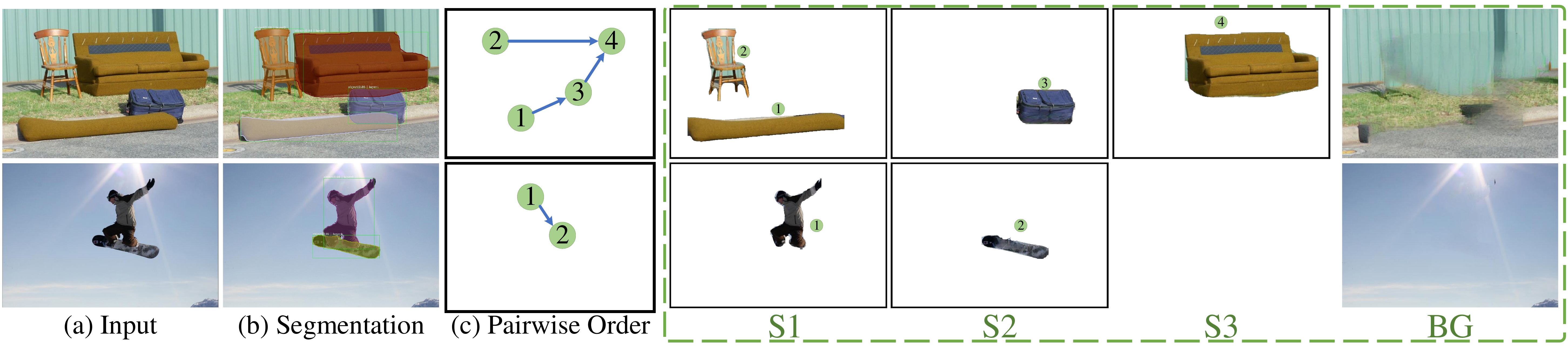}
	\caption[Layer-by-layer completed scene decomposition on natural images]{\textbf{Layer-by-layer completed scene decomposition on natural images}. (a) Inputs. (b) Final amodal instance segmentation. (c) Inferred directed graph for pairwise occlusion order. (d) Columns labeled S1-3 show the decomposed instances with completed appearance in each step.}
	\label{fig:ch6_results_real_csd}	
\end{figure}

\subsection{Results on Real Datasets}

We now assess our model on real images. Since the ground-truth appearances are unavailable, we only provide the visual \emph{scene manipulation} results in Section~\ref{sec:ch6_results_application}, instead of quantitative results for \emph{invisible completion}.

\paragraph{Completed scene decomposition} In Figure~\ref{fig:ch6_results_real_csd}, we visualize the layer-by-layer completed scene decomposition results on real images. Our CSDNet is able to decompose a scene into completed instances with correct ordering. The originally occluded invisible parts of ``suitcase'', for instance, is completed with full shape and realistic appearance. Note that, our system is a fully scene understanding method that only takes an image as input, without requiring the other manual annotations as ~\cite{ehsani2018segan,zhan2020self}.

\begin{table}[tb!]
	\centering
	\scriptsize
	\renewcommand{\arraystretch}{1.2}
	\setlength\tabcolsep{8pt}
	\begin{tabular}{@{}l|l|l|c|c@{}}
		\hlineB{3.5}
		&  Inputs & SegNet & \makecell[c]{COCOA \\ ($\%$mAP)} & \makecell[c]{KINS\\($\%$mAP)} \\
		\hline 
		Amodel~\cite{zhu2017semantic} & I & Sharp~\cite{pinheiro2016learning} & \cellcolor[rgb]{0.7,0.7,0.7}7.7 & - \\
		Mask-RCNN~\cite{he2017mask} & I & Mask-RCNN~\cite{he2017mask} & \cellcolor[rgb]{0.7,0.7,0.7}31.8 & \cellcolor[rgb]{0.7,0.7,0.7}29.3 \\
		ORCNN~\cite{follmann2019learning} & I & Mask-RCNN~\cite{he2017mask} & \cellcolor[rgb]{0.7,0.7,0.7}33.2 & \cellcolor[rgb]{0.7,0.7,0.7}29.0 \\
		MLC~\cite{qi2019amodal} & I& Mask-RCNN~\cite{he2017mask} & 34.0 & \cellcolor[rgb]{0.7,0.7,0.7}31.1\\
		MLC~\cite{qi2019amodal} & I& HTC~\cite{chen2019hybrid} & 34.4 & 31.6 \\
		PCNet~\cite{zhan2020self} & I+$\hat{\text{V}}_{pred}$ & Mask-RCNN~\cite{he2017mask} & 30.3 & 28.6 \\
		PCNet~\cite{zhan2020self} & I+$\hat{\text{V}}_{pred}$ & HTC~\cite{chen2019hybrid} & 32.6 & 30.1\\
		\hline
		\textbf{CSDNet} & I & Mask-RCNN~\cite{he2017mask} & 34.1 & 31.5 \\
		\textbf{CSDNet} & I & HTC~\cite{chen2019hybrid} & \textbf{34.8} & \textbf{32.2} \\
		\hlineB{2.5}
	\end{tabular}
	\caption[Amodal Instance Segmentation on COCOA and KINS sets]{\textbf{Amodal Instance Segmentation on COCOA and KINS sets}. The gray color shows results reported in existing works and the others are our reported results by using the released codes and our CSDNet. }
	\label{table:ch6_real_seg}
\end{table}

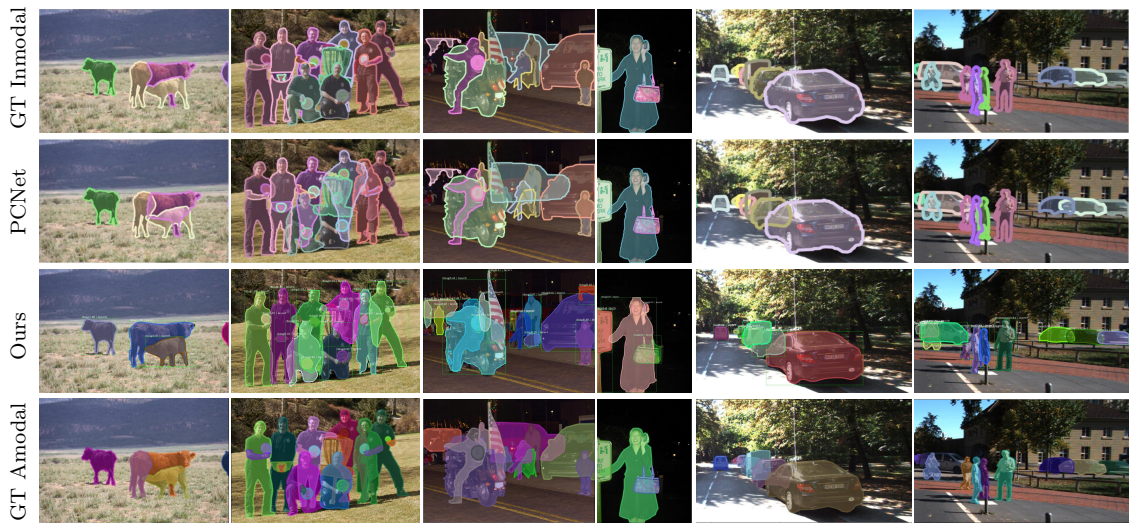
\begin{figure}[tb!]
	\centering
	\includegraphics[width=\linewidth]{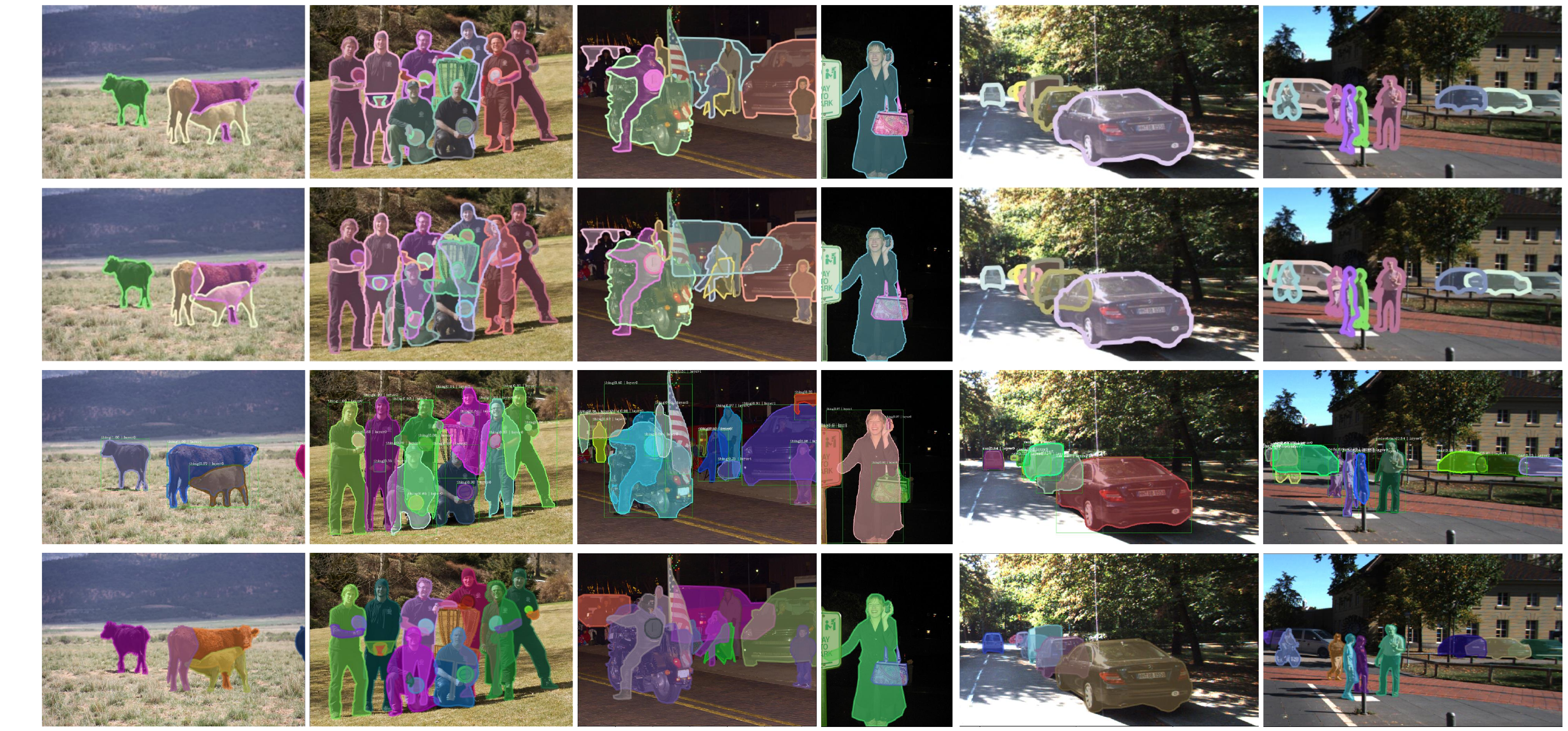}
	\caption[Amodal instance segmentation results on natural images]{\textbf{Amodal instance segmentation results on natural images}. Our CSDNet learns to predict the intact mask for the occluded objects (\eg animals and human). Note that, unlike PCNet~\cite{zhan2020self}, our model does \emph{not} depend on the visible mask (first row) as input. Hence it can handle some objects without ground-truth annotation, such as two `humans' in the third column and the `smartphone' in the fourth column. }
	\begin{picture}(0,0)
	\put(-208,238){\rotatebox{90}{\scriptsize GT Inmodal}}
	\put(-208,198){\rotatebox{90}{\scriptsize PCNet}}
	\put(-208,151){\rotatebox{90}{\scriptsize Ours}}
	\put(-208,90){\rotatebox{90}{\scriptsize GT Amodal}}
	\end{picture}
	\vspace{-0.3cm}
	\label{fig:ch6_results_real_mask}	
\end{figure}

\paragraph{Amodal instance segmentation} Next, we compare with state-of-the-art methods on amodal instance segmentation. Among these, AmodalMask~\cite{zhu2017semantic} and ORCNN \cite{follmann2019learning} were trained for the COCOA dataset, MLC~\cite{qi2019amodal} works for the KINS dataset, and PCNet~\cite{zhan2020self} is focused on amodal completion (mask completion) rather than amodal instance segmentation (requiring precise visible masks). For a fair comparison, when these methods do not provide results on a given dataset, we trained their models using publicly released code. For COCOA, we only report the results for ``thing'' category (\eg car, person, chair), because the ``stuff'' category (\eg glass, cloud, water) does not have specific shapes.

Table~\ref{table:ch6_real_seg} shows that our results (34.8 mAP and 32.2 mAP) are 0.4 points and 0.6 points higher than the recent MLC using the same segmentation structure (HTC) in COCOA and KINS, respectively. PCNet~\cite{zhan2020self} considers amodal perception in two steps and assumes that visible masks are available. We note that their mAP scores were very high when the visible ground-truth masks were provided. This is because all initial masks were matched to the annotations (without detection and segmentation errors for instances, as shown in Figure~\ref{fig:ch6_results_real_mask}). However, when we used a segmentation network to obtain visible masks $\hat{\text{V}}_{pred}$ for PCNet, the amodal instance segmentation results became lower than other methods, suggesting that it is much harder to segment a visible mask and then complete it.

In Figure~\ref{fig:ch6_results_real_mask}, we compare our CSDNet and PCNet~\cite{zhan2020self}. PCNet only completes the given visible annotated objects which had visible masks. In contrast, our CSDNet produces more contiguous amodal instance segmentation maps even for some unlabeled objects, for instance, the two ``humans'' in the third column. Furthermore, our model can directly create a deep hierarchical representation of a scene, producing a layer order for each instance. 

\begin{table}[tb!]
    \centering
    \scriptsize
	\renewcommand{\arraystretch}{1.2}
	\setlength\tabcolsep{8pt}
	\begin{tabular}{@{}l|l|l|c|c@{}}
		\hlineB{3}
		& \makecell[c]{Ordering \\ Inputs}& \makecell[c]{Ordering \\Algorithm}& \makecell[c]{COCOA \\ (OAP) }& \makecell[c]{KINS \\ (OAP)}\\
		\hlineB{3}
		OrderNet~\cite{zhu2017semantic} & I+$\text{F}_{gt}$ & Network & \cellcolor[rgb]{0.6,0.8,1.0}88.3 & \cellcolor[rgb]{0.6,0.8,1.0}94.1 \\
		PCNet~\cite{zhan2020self}  & $\text{V}_{gt}$ +$\hat{\text{F}}_{pre}$ & IoU Area & \cellcolor[rgb]{0.6,0.8,1.0}84.6 & \cellcolor[rgb]{0.6,0.8,1.0}86.0 \\
		MLC~\cite{qi2019amodal} & $\text{V}_{gt}$ + $\hat{\text{F}}_{pre}$ & IoU Area & \cellcolor[rgb]{0.6,0.8,1.0}80.3 & \cellcolor[rgb]{0.6,0.8,1.0}82.3 \\
		\textbf{CSDNet} & $\text{V}_{gt}$ + $\hat{\text{F}}_{pre}$ & IoU Area & \cellcolor[rgb]{0.6,0.8,1.0}84.7 & \cellcolor[rgb]{0.6,0.8,1.0}86.4 \\
		\hline
		MLC~\cite{qi2019amodal} & $\hat{\text{V}}_{pred}$ + $\hat{\text{F}}_{pre}$ & IoU Area & 74.2 & 80.2 \\
		MLC~\cite{qi2019amodal} & $\hat{\text{F}}_{pre}$ + layer & layer order$^1$ &  66.5 & 71.8 \\
		PCNet~\cite{zhan2020self} & $\hat{\text{V}}_{pred}$ + $\hat{\text{F}}_{pre}$ & IoU Area &  72.4 & 79.8\\
		\hline
		\textbf{CSDNet} & $\hat{\text{V}}_{pred}$ + $\hat{\text{F}}_{pre}$ & IoU Area & 75.4 & 81.6\\
		\textbf{CSDNet} & $\hat{\text{F}}_{pre}$ + layer & layer order & 80.9 & 82.2\\
		\hlineB{2.5}
	\end{tabular}
	\caption[Instance depth ordering on COCOA and KINS sets]{\textbf{Instance depth ordering on COCOA and KINS sets}. The blue rows show the results that uses ground-truth annotations as inputs. }
	\label{table:ch6_real_order}
\end{table}

\paragraph{Instance depth ordering.} Finally, we report the instance depth ordering results in Table~\ref{table:ch6_real_order}. In order to compare with existing work, we considered two settings: ground-truths provided (blue rows in Table~\ref{table:ch6_real_order}), and only RGB images given. The OrderNet obtained the best results as the ground-truth full masks $\text{F}_{gt}$ were given. We note that PCNet and our model achieved comparable performance when the visible ground-truth masks were used. Note that, we only used $\text{V}_{gt}$ for depth ordering, while PCNet utilized the visible mask as input for both mask prediction and depth ordering. Furthermore, when no ground-truth annotation was provided as input, our model performed better than MLCand PCNet. 

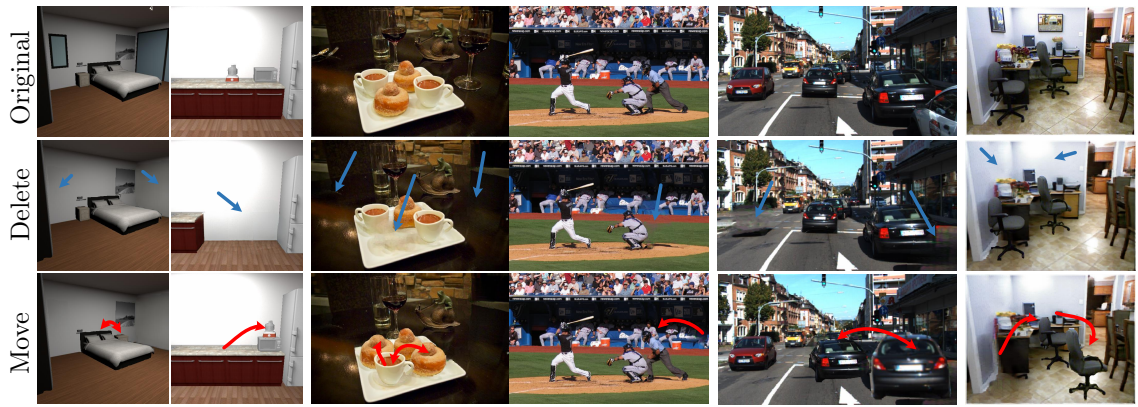
\begin{figure*}[tb!]
	\centering
	\includegraphics[width=\linewidth]{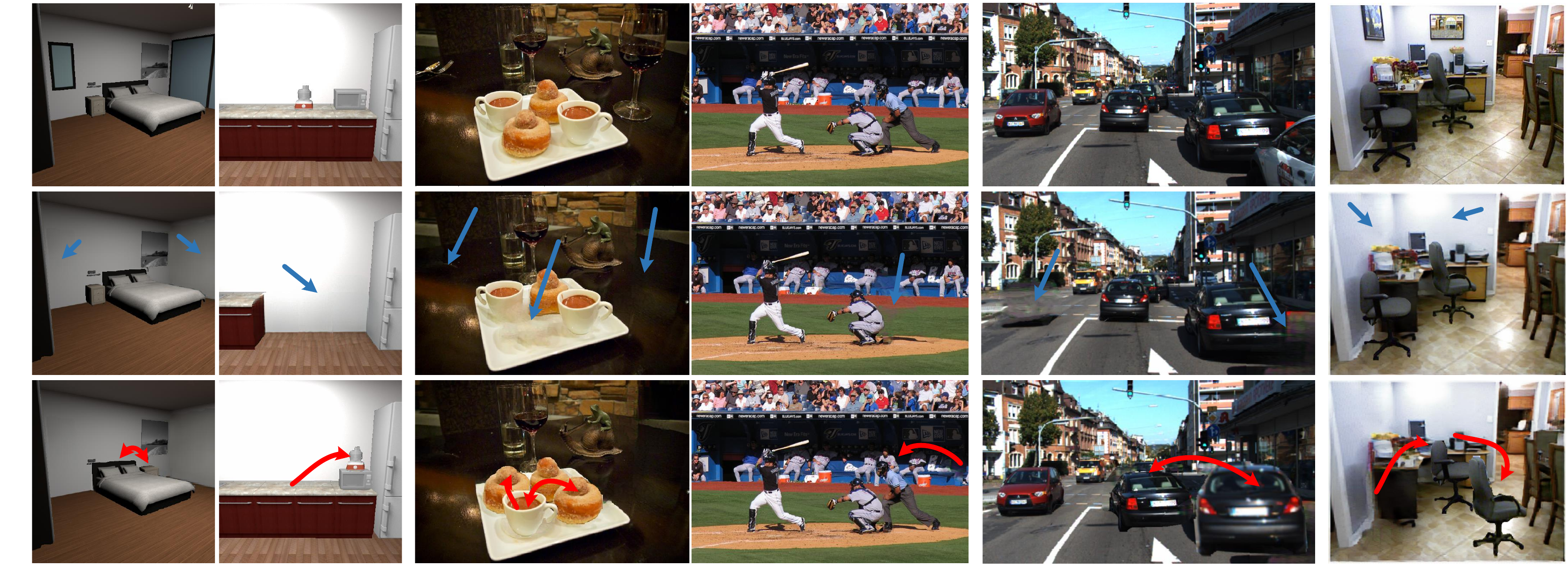}
	\caption[Free editing based on the results of our system]{\textbf{Free editing based on the results of our system} on images from various datasets. Note that our method is able to automatically detect, segment and complete the objects in the scene, \emph{without the need for manual interactive masking}, with interactive operations limited to only ``delete'' and ``drag-and-drop''. The blue arrows show object removal, while red arrows show object moving operations. We can observe that the originally \emph{invisible} regions are fully visible after editing.}
	\begin{picture}(0,0)
	\put(-210,208){\rotatebox{90}{\footnotesize Original}}
	\put(-210,164){\rotatebox{90}{\footnotesize Delete}}
	\put(-210,114){\rotatebox{90}{\footnotesize Move}}
	\end{picture}
	\label{fig:ch6_results_app}	
\end{figure*}

\subsection{Applications}\label{sec:ch6_results_application}
We illustrate some image editing / re-composition applications of this novel task, after the system has learned to decompose a scene into isolated completed objects together with their spatial occlusion relationships. In Figure~\ref{fig:ch6_results_app}, we visualize some recomposed scenes on various datasets, including our CSD, real COCOA \cite{zhu2017semantic}, KITTI \cite{Geiger2012we} and NYU-v2 \cite{Silberman:ECCV12}. 

In these cases, we directly modified the positions and occlusion ordering of individual objects. For instance, in the first bedroom example, we \emph{deleted} the ``window'', and \emph{moved} the ``bed'' and the ``counter'', which resulted in also \emph{modifying} their occlusion order. Note that all original \emph{invisible} regions were filled in with reasonable appearance. We also tested our model on real NYU-v2~\cite{Silberman:ECCV12} images which do \emph{not} belong to any of the training sets used. As shown in the last column of Figure~\ref{fig:ch6_results_app}, our model was able to detect and segment the object and complete the scene. The ``picture'', for instance, is deleted and filled in with background appearance.

\section{Limirations and Discussion}\label{sec:ch6_conclusions}

Building on previous inmodal and amodal instance perception work, we explored a higher-level structure scene understanding task that aims to decompose a scene into semantic instances, with completed RGB appearance and spatial occlusion relationships. We presented a layer-by-layer CSDNet, an iterative method to address this novel task. The main motivation behind our method is that fully visible objects, at each step, can be relatively easily detected and selected out without concern about occlusion. To do this, we simplified this complex task to two subtasks: instance segmentation and scene completion. We analyzed CSDNet and compared it with recent works on various datasets. Experimental results show that our model can handle an arbitrary number of objects and is able to generate the appearance of occluded parts. Our model outperformed current state-of-the-art methods that address this problem in one pass. The thorough ablation studies on synthetic data demonstrate that the two subtasks can contribute to each other through the layer-by-layer processing.

Although we have achieved good results for visiting the invisible, there are some limitations to the proposed method. First, if there are too many objects in a complex scene, the progressively introduced artifacts in image completion will have an increasing impact on subsequent steps. Second, limited by GPU memory, the completion network currently operates at a lower resolution than the scene decomposition network. Besides, we freely remove and move objects in a natural scene, but it is still an operation in a 2D image. It will be much more interesting to do the free editing in 3D space, just like the real-world object interaction \cite{yang2021objectnerf}.


\chapter{Conclusion and Future Directions} 
\chaptermark{Discussion}
\label{ch:Discussion} 

In previous chapters, a few novel learning-based methods have been presented for visual synthesis and generation, including changing visual appearance for I2I translation (Chapters \ref{ch:Synthestic2Real} and \ref{ch:F(L)SeSim}), generating semantic content for image completion (Chapters \ref{ch:PICNet} and \ref{ch:TFill}), and simultaneously modeling shapes and appearance for scene decomposition and completion (Chapter \ref{ch:VIV}). In each chapter, a new model is introduced to advance the state-of-the-art in the corresponding task and I hope to bring some new perspectives for each task. The extensive experiments have demonstrated that the proposed approaches can generate reasonable content as well as visually realistic appearance results compared to previous methods.   

For changing visual appearance in \textbf{Part \uppercase\expandafter{\romannumeral1}}, we found that the learned model mainly focused on modifying local patch textures, regardless of the global semantic information. In particular, when we aimed to generate multiple and diverse results in an I2I translation task, repeated noises tend to be added at different image or feature positions, such as in BicycleGAN \cite{zhu2017toward} and MUNIT \cite{huang2018multimodal}. This resulted in unwanted texture modification, \eg the zebra-stripe in background in \emph{horse$\rightarrow$zebra}. 
To generate semantic content in \textbf{Part \uppercase\expandafter{\romannumeral2}}, we needed to correctly model long-range dependencies, instead of focusing on local texture information. Therefore, the general network architecture involved downsampling the image to lower resolutions to extract global information, \eg 5 times downsampling (Chapter \ref{ch:PICNet}) compared to 2 times in I2I translation (Chapters \ref{ch:Synthestic2Real} and \ref{ch:F(L)SeSim}). The patch discriminator \cite{zhu2017semantic} is also replaced by the global discriminator (Chapters \ref{ch:PICNet} and \ref{ch:TFill}) that can model long-range relationships. Furthermore, transformer-based architectures (Chapter \ref{ch:TFill}, \cite{wan2021high}) have rapidly improved the image completion results for both single and multiple solutions, which further demonstrated that directly modeling the long-range visible information is quite important for semantic content generation. 

As for simultaneously modeling shapes and appearance in \textbf{Part \uppercase\expandafter{\romannumeral3}}, it remains quite a challenging problem, which requires global perception of a scene to be able to decompose all instances as well as infer their underlying occlusion relationships, with local texture modeling needed to generate visually realistic appearance for occluded regions. Furthermore, although we rendered a high-quality synthetic dataset with RGB ground truth for all instances and background in this thesis, the dataset still has a gap to real images. Building a publicly available dataset for this higher-level scene understanding task is still some distance away.

Next, I discuss several possible future research directions, building on our current visual synthesis and generation algorithms.

\paragraph{Visual Word Representation in Generator} As mentioned above, while visually plausible results have been achieved in image translation and completion, there are a number of failure cases in all methods. A possible factor is the distribution of the training features and the testing features being different. Recently, vector quantization (VQ) has been re-used in the computer vision community and contributed to excellent performance in image generation \cite{oord2018representation,razavi2019generating,peng2021generating}. Due to the quantization and online learned dictionary, training features and testing features will belong more closely to the same domain, which is naturally suitable for image generation, resulting in lower risk of mode ``collapse''. Furthermore, the quantized visual words can be processed with frameworks used in NLP, in which the transformer \cite{esser2020taming} has shown excellent performance. 

\paragraph{Image Editing with Interactive Inputs} Existing learning-based image editing approaches have achieved rapid improvement over a short period of time, but most of the results are not manual editable. While the latest EdgeConnect approach \cite{Nazeri_2019_ICCV} provides edge input during the completion, it is difficult to train networks to recognize arbitrary edges, due to the gap between manual input edges and ground truth edges. However, is it enough to only provide edges? On the other hand, some recent works \cite{radford2021learning,patashnik2021styleclip} have succeeded in text-guided image generation and manipulation, where the content and style in the generative image is controllable using language. These breakthroughs may enable high-level controllable image editing applications. In particular, it may be possible to learn the joint distribution of visual words and language words in the future.    

\paragraph{3D View Synthesis} While working on 2D scene synthesis and generation, I realized that in addition to content generation in a 2D plane, the GAN-based method is a potentially powerful method to create different views from a single image or limited numbers of images. In particular, we can rebuild the 3D shapes and try to hallucinate unobserved parts based on prior knowledge and limited visible information, similar to the 2D image completion in \textbf{Part \uppercase\expandafter{\romannumeral2}}. However, the existing methods \cite{choy20163d,girdhar2016learning,wang2018pixel2mesh,gkioxari2019mesh} that learns to rebuild the 3D scene in 3D format, such as point cloud \cite{lin2018learning}, voxel \cite{girdhar2016learning} and mesh \cite{wang2018pixel2mesh}, we would like to represent the 3D structure as a 3D feature in latent space, where a corresponding generator can be applied as a render simulator to generate visually realistic images from arbitrary views.


\addtocontents{toc}{\vspace{0.8em}} 

\appendix 



\chapter{Proofs for Chapter \ref{ch:PICNet}} 

\label{AppendixA} 

\section{Mathematical Derivation and Analysis}\label{ch4:math}
\subsection{Difficulties with Using the Classical CVAE for Image Completion}

Here we elaborate on the difficulties encountered when using the classical CVAE formulation for pluralistic image completion, expanding on the shorter description in Section~\ref{ch4_probabilistic}.

\subsubsection{Background: Derivation of the Conditional Variational Auto-Encoder
	(CVAE)}

The broad CVAE framework of Sohn \etal\cite{sohn2015learning} is a straightforward conditioning of the classical VAE. Using the notation in Chapter \ref{ch:PICNet}, a latent variable $\mathbf{z}_{c}$ is assumed to stochastically generate the hidden partial image $\mathbf{I}_{c}$. When conditioned on the visible partial image $\mathbf{I}_{m}$, we get the conditional probability:
\begin{equation}
	p(\mathbf{I}_{c}|\mathbf{I}_{m})=\int p_{\phi}(\mathbf{z}_{c}|\mathbf{I}_{m})p_{\theta}(\mathbf{I}_{c}|\mathbf{z}_{c},\mathbf{I}_{m})d\mathbf{z}_{c}
\end{equation}
The variance of the Monte Carlo estimate can be reduced by importance sampling:
\begin{align}
	p(\mathbf{I}_{c}|\mathbf{I}_{m}) & =\int q_{\psi}(\mathbf{z}_{c}|\mathbf{I}_{c},\mathbf{I}_{m})\frac{p_{\phi}(\mathbf{z}_{c}|\mathbf{I}_{m})}{q_{\psi}(\mathbf{z}_{c}|\mathbf{I}_{c},\mathbf{I}_{m})}p_{\theta}(\mathbf{I}_{c}|\mathbf{z}_{c},\mathbf{I}_{m})d\mathbf{z}_{c}\nonumber\\
	& =\mathbb{E}_{\mathbf{z}_{c}\sim q_{\psi}(\mathbf{z}_{c}|\mathbf{I}_{c},\mathbf{I}_{m})}\left[\frac{p_{\phi}(\mathbf{z}_{c}|\mathbf{I}_{m})}{q_{\psi}(\mathbf{z}_{c}|\mathbf{I}_{c},\mathbf{I}_{m})}p_{\theta}(\mathbf{I}_{c}|\mathbf{z}_{c},\mathbf{I}_{m})\right]
\end{align}
Taking logs and apply Jensen's inequality leads to
\begin{align}
	\log p(\mathbf{I}_{c}|\mathbf{I}_{m}) & \geq\mathbb{E}_{\mathbf{z}_{c}\sim q_{\psi}(\mathbf{z}_{c}|\mathbf{I}_{c},\mathbf{I}_{m})}\left[\log p_{\theta}(\mathbf{I}_{c}|\mathbf{z}_{c},\mathbf{I}_{m})-\log\frac{q_{\psi}(\mathbf{z}_{c}|\mathbf{I}_{c},\mathbf{I}_{m})}{p_{\phi}(\mathbf{z}_{c}|\mathbf{I}_{m})}\right]\nonumber\\
	\mathcal{V} & =\mathbb{E}_{\mathbf{z}_{c}\sim q_{\psi}(\mathbf{z}_{c}|\mathbf{I}_{c},\mathbf{I}_{m})}\left[\log p_{\theta}(\mathbf{I}_{c}|\mathbf{z}_{c},\mathbf{I}_{m})\right]-\mathrm{KL}\left(q_{\psi}(\mathbf{z}_{c}|\mathbf{I}_{c},\mathbf{I}_{m})||p_{\phi}(\mathbf{z}_{c}|\mathbf{I}_{m})\right)
	\label{eq:ch4_appendix_cvae}
\end{align}
The variational lower bound $\mathcal{V}$ totaled over all training data is jointly maximized \wrt the network parameters $\theta$, $\phi$ and $\psi$ in attempting to maximize the total log likelihood of the observed training instances.

\subsubsection{Single Instance Per Conditioning Label}

As is typically the case for image completion, there is only one training instance of $\mathbf{I}_{c}$ for each unique $\mathbf{I}_{m}$. This means that for the function $q_{\psi}(\mathbf{z}_{c}|\mathbf{I}_{c},\mathbf{I}_{m})$, $\mathbf{I}_{c}$ can be learned into the network as a hard-coded dependency of the input $\mathbf{I}_{m}$, so $q_{\psi}(\mathbf{z}_{c}|\mathbf{I}_{c},\mathbf{I}_{m})\cong\hat{q}_{\psi}(\mathbf{z}_{c}|\mathbf{I}_{m})$. Assuming that the network for $p_{\phi}(\mathbf{z}_{c}|\mathbf{I}_{m})$ has similar or higher modeling power and there are no other explicit constraints imposed on it, then in training $p_{\phi}(\mathbf{z}_{c}|\mathbf{I}_{m})\rightarrow\hat{q}_{\psi}(\mathbf{z}_{c}|\mathbf{I}_{m})$, and the KL divergence in (\ref{eq:ch4_appendix_cvae}) goes to zero.

In this situation of zero KL divergence, we can rewrite the variational lower bound and replace $\hat{q}_{\psi}(\mathbf{z}_{c}|\mathbf{I}_{m})$ with $p_{\phi}(\mathbf{z}_{c}|\mathbf{I}_{m})$ without loss of generality, as
\begin{equation}
	\mathcal{V}=\mathbb{E}_{\mathbf{z}_{c}\sim p_{\phi}(\mathbf{z}_{c}|\mathbf{I}_{m})}\left[\log p_{\theta}(\mathbf{I}_{c}|\mathbf{z}_{c},\mathbf{I}_{m})\right]
\end{equation}

\subsubsection{Unconstrained Learning of the Conditional Prior}

We can analyze how $\mathcal{V}$ can be maximized, by using Jensen's inequality again (reversing earlier use)
\begin{align}
	\mathcal{V} & \leq\log\mathbb{E}_{\mathbf{z}_{c}\sim p_{\phi}(\mathbf{z}_{c}|\mathbf{I}_{m})}\left[p_{\theta}(\mathbf{I}_{c}|\mathbf{z}_{c},\mathbf{I}_{m})\right]\nonumber\\
	& =\log\int p_{\phi}(\mathbf{z}_{c}|\mathbf{I}_{m})p_{\theta}(\mathbf{I}_{c}|\mathbf{z}_{c},\mathbf{I}_{m})d\mathbf{z}_{c}
\end{align}
By further applying Hölder's inequality (\ie $\left\Vert fg\right\Vert _{1}\leq\left\Vert f\right\Vert _{p}\left\Vert g\right\Vert _{q}$
for $\frac{1}{p}+\frac{1}{q}=1$), we get
\begin{align}
	\mathcal{V} & \leq\log\left[\left|\int\left|p_{\phi}(\mathbf{z}_{c}|\mathbf{I}_{m})\right|d\mathbf{z}_{c}\right|\left|\int\left|p_{\theta}(\mathbf{I}_{c}|\mathbf{z}_{c},\mathbf{I}_{m})\right|^{\infty}d\mathbf{z}_{c}\right|^{\frac{1}{\infty}}\right]\quad(\text{by setting }p=1,q=\infty)\nonumber\\
	& =\log\left[1\cdot\max_{\mathbf{z}_{c}}p_{\theta}(\mathbf{I}_{c}|\mathbf{z}_{c},\mathbf{I}_{m})\right]=\max_{\mathbf{z}_{c}}\log p_{\theta}(\mathbf{I}_{c}|\mathbf{z}_{c},\mathbf{I}_{m})
\end{align}
Assuming that there is a unique global maximum for $\log p_{\phi}(\mathbf{z}_{c}|\mathbf{I}_{m})$, the bound achieves equality when the conditional prior becomes a Dirac delta function centered at the maximum latent likelihood point
\begin{equation}
	p_{\phi}(\mathbf{z}_{c}|\mathbf{I}_{m})\rightarrow\delta(\mathbf{z}_{c}-\mathbf{z}_{c}^{*})\quad\textrm{where }\mathbf{z}_{c}^{*}=\arg\max_{\mathbf{z}_{c}}\log p_{\theta}(\mathbf{I}_{c}|\mathbf{z}_{c},\mathbf{I}_{m})
\end{equation}
Intuitively, subject to the vagaries of stochastic gradient descent, the network for $p_{\phi}(\mathbf{z}_{c}|\mathbf{I}_{m})$ without further constraints will learn a narrow delta-like function that sifts out maximum latent likelihood value of $\log p_{\theta}(\mathbf{I}_{c}|\mathbf{z}_{c},\mathbf{I}_{m})$. 

As mentioned in Section~\ref{ch4_probabilistic}, although this narrow conditional prior may be helpful in estimating a single solution for $\mathbf{I}_{c}$ given $\mathbf{I}_{m}$ during testing, this is poor for sampling a diversity of solutions. In our framework, the (unconditional) latent priors are imposed for the partial images themselves, which prevent this delta function degeneracy.

\subsubsection{CVAE with Fixed Prior}

An alternative CVAE variant~\cite{walker2016} assumes that conditional prior is independent of the $\mathbf{I}_m$ and fixed, so $p(\mathbf{z}_c|\mathbf{I}_m)\cong p(\mathbf{z}_c)$, where $p(\mathbf{z}_c)$ is a fixed distribution (\eg standard normal). This means
\begin{equation}
	p(\mathbf{I}_c|\mathbf{I}_m) = \int p(\mathbf{I}_c|\mathbf{z}_c, \mathbf{I}_m) p(\mathbf{z}_c) d\mathbf{z}_c
	\label{eq:ch4_cvae_fixed_prior}
\end{equation}

Now we can consider the case for a fixed $\mathbf{I}_m=\mathbf{I}^*_m$, and rewrite (\ref{eq:ch4_cvae_fixed_prior}) as
\begin{equation}
	p_{\mathbf{I}^*_m}(\mathbf{I}_c) = \int p_{\mathbf{I}^*_m}(\mathbf{I}_c|\mathbf{z}_c) p(\mathbf{z}_c) d\mathbf{z}_c
\end{equation}
Doing so makes it obvious we can then derive the standard (unconditional) VAE formulation from here. Thus, an appropriate interpretation of this CVAE variant is that it uses $\mathbf{I}_m$ as a ``switch'' parameter to choose between different VAE models that are trained for the specific conditions.

Once again, this is fine if there are multiple training instances per conditional label. However, in the image completion problem, there is only one $\mathbf{I}_c$ per unique $\mathbf{I}_m$, so the condition-specific VAE model will simply ignore the sampling ``noise'' and learn to predict the single instance of $\mathbf{I}_c$ from $\mathbf{I}_m$ directly, \ie $p(\mathbf{I}_c|\mathbf{z}_c, \mathbf{I}_m) \approx p(\mathbf{I}_c|\mathbf{I}_m)$, which incidentally achieves equality for the variational lower bound. This results in negligible variation of output despite now sampling from $p(\mathbf{z}_c)=\mathcal{N}(0,1)$.

Our framework resolves this in part by defining all (unconditional) partial images of $\mathbf{I}_c$ as sharing a common latent space with adaptive priors, with the likelihood parameters learned as an unconditional VAE, and further coupling on the conditional portion (\ie the generative path) to get a more distinct but regularized estimate for $p(\mathbf{z}_c|\mathbf{I}_m)$.

\subsection{Joint Maximization of Unconditional and Conditional Variational Lower
	Bounds}

The overall training loss function (\ref{eq:ch4_total_loss}) used in our framework has a direct link to jointly maximizing the unconditional and unconditional variational lower bounds, respectively expressed by (\ref{eq:ch4_VAE}) and (\ref{eq:ch4_mixed_models}). Using simplified notation, we rewrite these bounds respectively as:
\begin{align}
	\mathcal{B}_{1} & =\mathbb{E}_{q_{\psi}}\log p_{\theta}^{r}-\mathrm{KL}(q_{\psi}||p_{z_{c}})\nonumber\\
	\mathcal{B}_{2} & =\lambda\left(\mathbb{E}_{q_{\psi}}\log p_{\theta}^{r}-\mathrm{KL}(q_{\psi}||p_{z_{c}})\right)+\mathbb{E}_{p_{\phi}}\log p_{\theta}^{g}
\end{align}
To clarify, $\mathcal{B}_{1}$ is the lower bound related to the unconditional log likelihood of observing $\mathbf{I}_{c}$, while $\mathcal{B}_{2}$ relates to the log likelihood of observing $\mathbf{I}_{c}$ conditioned on $\mathbf{I}_{m}$. The expression of $\mathcal{B}_{2}$ reflects a blend of conditional likelihood formulations with and without the use of importance sampling, which are matched to different likelihood models, as explained in Section~\ref{ch4_probabilistic}. Note that the $(1-\lambda)$ coefficient from (\ref{eq:ch4_mixed_models}) is left out here for simplicity, but there is no loss of generality since we can ignore a constant factor of the true lower bound if we are simply maximizing it. We can then define a combined objective function as our maximization goal
\begin{align}
	\mathcal{B}&= \beta \, \mathcal{B}_{1}+\mathcal{B}_{2}\nonumber\\
	&= (\beta+\lambda)\mathbb{E}_{q_{\psi}}\log p_{\theta}^{r}+\mathbb{E}_{p_{\phi}}\log p_{\theta}^{g}-\left[\beta\mathrm{KL}(q_{\psi}||p_{z_{c}})+\lambda\mathrm{KL}(q_{\psi}||p_{\phi})\right]
	\label{eq:ch4_appendix_total_bound}
\end{align}
with $\beta\geq0$.

To understand the relation between $\mathcal{B}$ in (\ref{eq:ch4_appendix_total_bound}) and $\mathcal{L}$ in (\ref{eq:ch4_total_loss}), we consider the equivalence of:
\begin{equation}
	-\mathcal{B}\cong\mathcal{L}=\alpha_{\mathrm{KL}}(\mathcal{L}_{\mathrm{KL}}^{r}+\mathcal{L}_{\mathrm{KL}}^{g})+\alpha_{\mathrm{app}}(\mathcal{L}_{\mathrm{app}}^{r}+\mathcal{L}_{\mathrm{app}}^{g})+\alpha_{\mathrm{ad}}(\mathcal{L}_{\mathrm{ad}}^{r}+\mathcal{L}_{\mathrm{ad}}^{g})
\end{equation}
Comparing terms
\begin{equation}
	\mathcal{L}_{\mathrm{KL}}^{r}\cong\mathrm{KL}(q_{\psi}||p_{z_{c}}),\quad\mathcal{L}_{\mathrm{KL}}^{g}\cong\mathrm{KL}(q_{\psi}||p_{\phi})\quad\Rightarrow\beta=\lambda=\alpha_{\mathrm{KL}}
\end{equation}

For the reconstructive path that involves sampling from the (posterior) importance function $q_{\psi}(\mathbf{z}_{c}|\mathbf{I}_{c})$ of (\ref{eq:ch4_CVAE_with_prior}), we can substitute $(\beta+\lambda)=2\alpha_{\mathrm{KL}}$ and get the reconstructive log likelihood formulation as
\begin{equation}
	-\mathbb{E}_{q_{\psi}}\log p_{\theta}^{r}\cong\frac{\alpha_{\mathrm{app}}}{2\alpha_{\mathrm{KL}}}\mathcal{L}_{\mathrm{app}}^{r}+\frac{\alpha_{\mathrm{ad}}}{2\alpha_{\mathrm{KL}}}\mathcal{L}_{\mathrm{ad}}^{r}
\end{equation}
Here, $\mathbf{I}_{c}$ is available, with $\mathcal{L}_{\mathrm{app}}^{r}$ reconstructing both $\mathbf{I}_{c}$ and $\mathbf{I}_{m}$ as in (\ref{eq:ch4_app_loss_rec}), while $\mathcal{L}_{\mathrm{ad}}^{r}$ involves GAN-based pairwise feature matching (\ref{eq:ch4_ad_loss_rec}).

For the generative path that involves sampling from the conditional prior $p_{\phi}(\mathbf{z}_{c}|\mathbf{I}_{m})$, we have the generative log likelihood formulation as
\begin{equation}
	-\mathbb{E}_{p_{\phi}}\log p_{\theta}^{g}\cong\alpha_{\mathrm{app}}\mathcal{L}_{\mathrm{app}}^{g}+\alpha_{\mathrm{ad}}\mathcal{L}_{\mathrm{ad}}^{g}
\end{equation}

As explained in Sections~\ref{ch4_probabilistic} and \ref{sec:ch4_training}, the generative path does not have direct access to $\mathbf{I}_{c}$, and this is reflected in the likelihood $p_{\theta}^{g}$ in which the instances of $\mathbf{I}_{c}$ are ignored. Thus $\mathcal{L}_{\mathrm{app}}^{g}$ is only for reconstructing $\mathbf{I}_{m}$ in a deterministic auto-encoder fashion as per (\ref{eq:ch4_app_loss_gen}), while $\mathcal{L}_{\mathrm{ad}}^{g}$ in (\ref{eq:ch4_ad_loss_gen}) only tries to enforce that the generated distribution be consistent with the training set distribution (hence without per-instance knowledge), as implemented in the form of a GAN.


\chapter{Supplementary Material for Chapter \ref{ch:TFill}} 

\label{AppendixB} 

\section{Additional Quantitative Results}\label{app:sec:ch5_ana}

We further report quantitative results using traditional pixel-level and patch-level image quality evaluation metrics. 

\begin{table}[htb!]
    \centering
    \scriptsize
    \renewcommand{\arraystretch}{1.2}
    \setlength\tabcolsep{8pt}
    \begin{tabular}{@{}llccccccc@{}}
         \hlineB{3.5}
         & \multirow{2}{*}{\textbf{Method}} & \multicolumn{3}{c}{\textbf{CelebA-HQ}}&& \multicolumn{3}{c}{\textbf{FFHQ}}\\
		\cline{3-5}\cline{7-9}
		& & $\ell_1 \text{loss}\downarrow$& SSIM$\uparrow$ & PSNR$\uparrow$ && $\ell_1 \text{loss}\downarrow$ & SSIM$\uparrow$ & PSNR$\uparrow$ \\
		\hlineB{2}
		& CA~\cite{yu2018generative} & 0.0310 & 0.8201 & 23.5667 && 0.0337 & 0.8099 & 22.7745 \\
		& PICNet~\cite{zheng2019pluralistic} & 0.0209 & 0.8668 & 24.6860 & & 0.0241 & 0.8547 & 24.3430\\
		& MEDFE~\cite{Liu2019MEDFE} & 0.0208 & 0.8691 & 24.4733 & & -& - & - \\
		\cdashline{1-7}
		$\mathrm{A}$ & Traditional \emph{Conv} & 0.0199 & 0.8693 & 24.5800 && 0.0241 & 0.8559 & 24.2271 \\
		$\mathrm{B}$ & + Attention in G & 0.0196 & 0.8717 & 24.6512 && 0.0236 & 0.8607 & 24.4384 \\
		$\mathrm{C}$ & + Restrictive \emph{Conv} & 0.0191 & 0.8738 & 24.8067 && 0.0220 & 0.8681 & 24.9280 \\
		$\mathrm{D}$ & + Transformer & 0.0189 & 0.8749 & 24.9467 && 0.0197 & 0.8751 & 25.1002\\
		$\mathrm{E}$ & + Masked Attention & 0.0183 & 0.8802 & 25.2510 && 0.0188 & 0.8765 & 25.1204 \\
		$\mathrm{F}$ & + Refine Network & \textbf{0.0180} & \textbf{0.8821} & \textbf{25.4220} && \textbf{0.0184} & \textbf{0.8778}& \textbf{25.2061} \\
        \hlineB{2}
    \end{tabular}
    \caption[Quantitative results for traditional metrics on center masked images]{Quantitative results for traditional pixel-level and patch-level metrics on center masked images. }
    \label{tab:ch5_sup_conv_vs_transform}
\end{table}

Table \ref{tab:ch5_sup_conv_vs_transform} provides a comparison of our results to state-of-the-art CNN-based models, as well as various alternative configurations for our design, on the center masked face testing set. This is an extension of Table \ref{tab:ch5_conv_vs_transform} in the main text. All images were normalized to the range [0,1] for quantitative evaluation. While there is no necessity to strongly encourage the completed images to be the same as the original ground-truth images, our TFill model nonetheless achieved better performance on these metrics too, including $\ell_1$ loss, structure similarity index (SSIM) and peak signal-to-noise ratio (PSNR), suggesting that our TFill model is more capable of generating closer content to the original unmasked images. 

\begin{table}[tb!]
    \centering
    \scriptsize
    \renewcommand{\arraystretch}{1.2}
    \setlength\tabcolsep{8pt}
    \begin{tabular}{@{}l|c|ccccc@{}}
        \hlineB{3.5}
         &  \textbf{Size} &  {\bf GL} \cite{iizuka2017globally} & {\bf CA} \cite{yu2018generative}  & {\bf PICNet} \cite{zheng2019pluralistic} & {\bf HiFill} \cite{yi2020contextual} & \textbf{TFill}\\
         \hlineB{2}
        \multirow{6}{1em}{\rotatebox[origin=c]{90}{$\ell_1 \text{loss}^{\dagger}$}} & [0.01, 0.1] & 0.0233 & 0.0241 & 0.0097 & 0.0195 & \textbf{0.0093} \\
        & (0.1, 0.2] & 0.0346 & 0.0338 & 0.0164 & 0.0282 & \textbf{0.0153}\\
        & (0.2, 0.3] & 0.0500 & 0.0471 & 0.0249 & 0.0390 & \textbf{0.0231}\\
        & (0.3, 0.4] & 0.0659 & 0.0612 & 0.0348 & 0.0513 & \textbf{0.0322}\\
        & (0.4, 0.5] & 0.0808 & 0.0753 & 0.0456 & 0.0657 & \textbf{0.0422}\\
        & (0.5, 0.6] & 0.0945 & 0.0925 & 0.0641 & 0.0885 & \textbf{0.0591}\\
        \hline
        \multirow{6}{1em}{\rotatebox[origin=c]{90}{SSIM$^{\star}$}} & [0.01, 0.1] & 0.9150 & 0.9079 & 0.9634 & 0.9245 & \textbf{0.9695}\\
        & (0.1, 0.2] & 0.8526 & 0.8447 & 0.9137 & 0.8603 & \textbf{0.9253} \\
        & (0.2, 0.3] & 0.7672 & 0.7652 & 0.8520 & 0.7838 & \textbf{0.8686}\\
        & (0.3, 0.4] & 0.6823 & 0.6906 & 0.7850 & 0.7057 & \textbf{0.8063}\\
        & (0.4, 0.5] & 0.5987 & 0.6133 & 0.7119 & 0.6193 & \textbf{0.7391}\\
        & (0.5, 0.6] & 0.5185 & 0.5322 & 0.6077 & 0.5137 & \textbf{0.6428}\\
         \hline
        \multirow{6}{1em}{\rotatebox[origin=c]{90}{PSNR$^{\star}$}} & [0.01, 0.1] & 28.4151 & 26.8452 & 32.2579 & 28.3955 & \textbf{33.0585}\\
        & (0.1, 0.2] & 24.4074 & 23.1766 & 27.3320 & 24.5495 & \textbf{28.0670}\\
        & (0.2, 0.3] & 21.3296 & 20.4427 & 24.4423 & 22.0604 & \textbf{25.0951}\\
        & (0.3, 0.4] & 19.1118 & 18.6337 & 22.3238 & 20.1451 &  \textbf{22.8942}\\
        & (0.4, 0.5] & 17.5594 & 17.2978 & 20.7146 & 18.4715 & \textbf{21.2200}\\
        & (0.5, 0.6] & 16.4831 & 16.0824 & 18.7234 & 16.4998 & \textbf{19.1040}\\
        \hlineB{2}
    \end{tabular}
    \caption[Quantitative comparisons on Places2 with free-form masks]{Quantitative comparisons on Places2~\cite{zhou2018places} with free-form masks \cite{Liu_2018_ECCV}.  $^{\dagger}$Lower is better. $^{\star}$Higher is better. Without bells and whistles, TFill outperformed all traditional CNN-based models.}
    \label{tab:ch5_sup_SOTA_comp}
\end{table}

Table \ref{tab:ch5_sup_SOTA_comp} provides a comparison of our results to state-of-the-art methods on the Places2~\cite{zhou2018places} testing set with free-form masks \cite{Liu_2018_ECCV}. This is an extension of Table \ref{tab:ch5_SOTA_comp} in the main text. As we can see in Figure \ref{fig:ch5_sup-center-places2-examples}, while our TFill model does \emph{not} generate the same content as the original unmasked images, it filled the masked holes with semantically appropriate content of consistent realistic appearance. There were no obvious artifacts when the completed pixels were recomposed with the original visible pixels, resulting in quite a significant improvement in image quality. 

\section{Experiment Details}\label{app:sec:ch5_experiment}

Here we first present the novel layers and loss functions used to train our model, followed by the training details. 

\subsection{Multihead \emph{Masked} Self-Attention}\label{app:sec:ch5_msa}

Our transformer encoder is built on the standard \textbf{qkv} \texttt{self-attention} (SA) \cite{vaswani2017attention} with a learned position embedding in each layer. Given an input sequence $\textbf{z}\in\mathbb{R}^{N\times C}$, we first calculate the pairwise similarity $\textbf{A}$ between each two elements as follows:
\begin{align}
    [\textbf{q}, \textbf{k}, \textbf{v}] & = \textbf{W}_{qkv}\textbf{z} \\
    \textbf{A} & = \texttt{softmax}(\textbf{q}\textbf{k}^{\top}/\sqrt{C_h})
\end{align}
where $\textbf{W}_{qkv}\in\mathbb{R}^{C\times3C_{h}}$ is the learned parameter to refine the features \textbf{z} for the query $\textbf{q}$, the key $\textbf{k}$ and the value $\textbf{v}$. $\textbf{A}\in\mathbb{R}^{N\times N}$ is the dot similarity of N tokens, which is scaled by the square root of feature dimension $C_h$. Then, we compute a weighted sum over all values $\textbf{v}$ via:
\begin{equation}
    \text{SA}(\textbf{z}) = \textbf{A}\textbf{v}
\end{equation}
where the value $z$ in the sequence is connected through their learned similarity $A$, rather than purely depending on a fixed learned weight $w$. 

The multihead \texttt{self-attention} (MSA) is an extension of SA, in which $H$ heads are run in parallel to get multiple attention scores and the corresponding projected results. Then we get the following function:
\begin{equation}
    \text{MSA}(\textbf{z})=[\text{SA}_1(\textbf{z});\text{SA}_2(\textbf{z});\dots;\text{SA}_h(\textbf{z})]
\end{equation}

To encourage the model to \emph{bias} to the important visible values, we further modify the MSA with a \emph{masked} self-attention layer, in which a masked weight is applied to scale the attention score $\textbf{A}$. Given a feature $\textbf{x}$ and the corresponding mask $\textbf{m}$ (1 denotes visible pixel and 0 is masked pixel). The original partial convolution operation is operated as:
\begin{align}
    x^\prime & = 
    \begin{cases}
    \textbf{W}_p(\textbf{x}_p\bigodot\textbf{m}_p)\frac{1}{\sum(\textbf{m}_p)} + b, & \mbox{if} \sum(\textbf{m}_p) > 0 \\
    0, & \mbox{otherwise}
    \end{cases} \\
    m^\prime & = 
    \begin{cases}
    1, & \mbox{if} \sum(\textbf{m}_p) > 0 \\
    0, & \mbox{otherwise}
    \end{cases}
\end{align}
where $\textbf{W}_p$ contain the convolution filter weights, $b$ is the corresponding bias, while $\textbf{x}_p$ and $\textbf{m}_p$ are the feature values and mask values in the current convolution window (\eg $2\times2$ in our \emph{restrictive CNN}), respectively. Here, we replace the $m^\prime$ as a float value:
\begin{equation}
    m^\prime = \frac{\sum(\textbf{m}_p)}{S}
\end{equation}
where $S$ is the size of each convolution filter, $2\times2$ used in our \emph{restrictive CNN}. To do this, each token only extracts the visible information. What's more, the final $m$ for each token denotes the percentage of valid values in each token under a small RF. Then, for each sequence $\textbf{z}\in\mathbb{R}^{N\times C}$, we obtain a corresponding masked weight $\textbf{m}\in\mathbb{R}^{N\times 1}$ by flattening the updated mask. Finally, we update the original attention score by multiplying with the repeated masked weight $\textbf{m}\in\mathbb{R}^{N\times 1}$:
\begin{equation}
    \textbf{A}_m = \textbf{A}\bigodot\textbf{m}_r
\end{equation}
where $\textbf{m}_r\in\mathbb{R}^{N\times N}$ is the extension of masked weight $\textbf{m}\in\mathbb{R}^{N\times 1}$ in the final dimension. 

\chapter{Supplementary Material for Chapter \ref{ch:VIV}} 

\label{AppendixC} 


\section{Experimental Details} \label{ch6_supp_experiment}

\paragraph{Training} We trained our model on the synthetic data into three phases: 1) the layered scene decomposition network (Figure~\ref{fig:ch6_framework}(b)) is trained with loss ${L}_{decomp}$ for 24 epochs, where at each layer, re-composited layered ground-truths are used as input. 2) Separately, the completion network (Figure~\ref{fig:ch6_framework}(c)) is trained with loss ${L}_{comp}$ for 50 epochs, wherein the ground-truth layer orders and segmented masks are used to designate the \emph{invisible} regions for completion. 3) Both decomposition and completion networks were trained jointly for 12 epochs, \emph{without relying on ground-truths as input} at any layer (Figure~\ref{fig:ch6_framework}(a)). Doing so allows the scene decomposition network to \emph{learn to cope with flaws} (\eg texture artifacts) in the scene completion network, and vice versa. For each scene, the iteration ends when no more objects are detected, or a maximum 10 iterations is reached.

The training on real data only involved phases 1 and 3, as no ground-truth appearances are available for the invisible parts. The layered decomposition network is trained only for one layer (original image) in phase 1 due to \emph{no} re-composed ground-truth images. Since phase 3 does not rely on ground-truths as input, we trained it layer-by-layer on real images by providing the ``pseudo ground truth'' appearances to calculate the reconstruction loss. To reduce the effect of progressively introduced artifacts in image completion, we used bounding boxes detected in the first layer as proposals for remaining decomposition steps. 

\paragraph{Inference} During testing, fully visible instances were selected out and assigned an absolute layer order corresponding to the step index $s_k$. In each layer, the decomposition network selects the highest scoring 100 detected boxes for mask segmentation and non-occlusion predication. As we observed that higher object classification scores provided more accurately segmented boundaries, we only selected non-occluded objects with high object classification scores and non-occlusion scores (thresholds of $0.5$ for synthetic images and $0.3$ for real images) among these 100 candidates. We further observed that in some cases, we detected multiple objects with high object classification confidences, yet none were classified as fully visible due to low non-occlusion scores, especially in complex scenes with steps larger than 5. We will then choose the instance with the highest non-occlusion score so that \emph{at least one object is selected at each layer}. When no objects are detected, the iteration stops. 

\begin{figure}[tb!]
	\centering
	\includegraphics[width=0.8\linewidth]{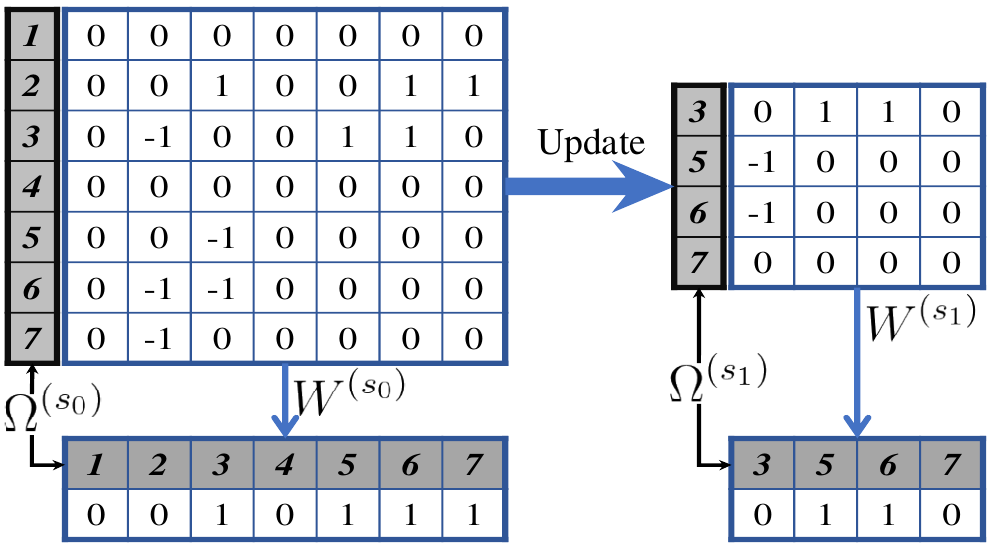}
	\begin{picture}(0,0)
	\put(-296,-6){\footnotesize Binary Occlusion Labels}
	\put(-106,-6){\footnotesize Binary Occlusion Labels}
	\end{picture}
	\vspace{0.2cm}
	\caption[Update for binary occlusion labels]{\textbf{An illustration of obtaining the ground-truth binary occlusion labels from pairwise order graph $G=(\Omega,W)$ in each step $s_k$.} If the indegree of a vertex is $0$, it will be labeled as $0$, a fully visible instance. Otherwise, the instance will be labeled as $1$, being occluded. When some objects are detected and selected out in the previous step, the object indexes and the corresponding occlusions will be eliminated. }
	\label{fig:ch6_layer_update}
\end{figure}
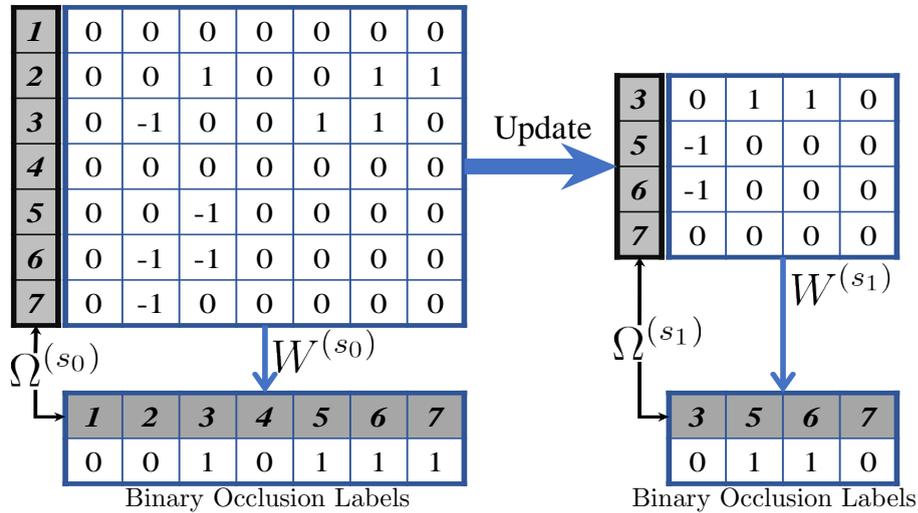 

\begin{figure}[tb!]
	\centering
	\includegraphics[width=\linewidth]{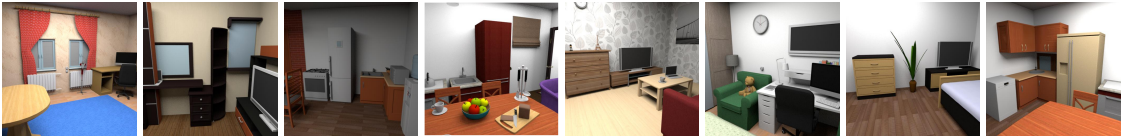}
	\caption[Realistic rendered images in the CSD dataset]{\textbf{Realistic rendered images in the CSD dataset} with various environment and lighting.}
	\label{fig:ch6_decom_appendix_dataset_rendering}	
\end{figure}

\paragraph{Instance depth ordering update} As illustrated in Figure~\ref{fig:ch6_layer_update}, we calculate the indegree $deg^{-}(\omega)$ (counts of $-1$) of each instance in the matrix. If $deg^{-}(\omega)=0$, meaning no objects are in front of it, its binary occlusion label will be 0. Otherwise, the object is occluded, labeled as 1. At each step, the fully visible objects will be eliminated from the directed graph $G$, and the ground-truth binary occlusion labels will be updated in each step. So if the table (instance \#2) was selected in the previous step, the vertex index $\Omega$ will be updated after the corresponding object $\omega_2$ is deleted from the occlusion matrix. 

\section{Rendering Dataset}\label{ch6_appendix_data}

\subsection{Data Rendering}\label{ch6_appendix_data_rendering}

Our \textbf{completed scene decomposition (CSD) dataset} was created using Maya \cite{Maya}, based on the SUNCG CAD models \cite{song2017semantic}. The original SUNCG dataset contains 45,622 different houses with realistically modeled rooms. As realistically rendering needs a lot of time (average 1 hour for each house), we only selected 2,456 houses in current work. The set of camera views was based on the original OpenGL-rendering method in SUNCG, but further filtered such that a camera view was only be picked when at least 5 objects appeared in that view. We then realistically rendered RGB images for the selected views. Eight examples are shown in Figure~\ref{fig:ch6_decom_appendix_dataset_rendering} for various room types and lighting environments. Notice that our rendered images are much more realistic than the OpenGL rendered versions from the original SUNCG and likewise in \cite{Dhamo2019iccv}. 

To \emph{visit the invisible}, the supervised method needs ground truth for the original occluded regions of each object. One possible way is to remove the fully visible objects in one layer and re-render the updated scene for the next layer, repeating this for all layers. However, during the training, the fully visible objects are not always correctly detected by the models. Thus, for more robust learning, we need to consider all different combinations of objects and the background. Given $N$ objects, we would need to render $2^N$ images for each view. As we can see from the data statistics presented in Figure \ref{fig:ch6_decom_appendix_dataset_statistic}, an average of 11 objects are visible in each view. Due to slow rendering, we do not have the capacity to render all such scenes (average $2^{11}=2048$ images per view). Instead, we separately rendered each isolated object with the full RGB appearance, as well as the empty room. 

During training, the image of a scene is created by using a combination of the rendered images of these individual objects and the background to create a composed image, based on the remaining objects left after applying the scene decomposition network at each step. Since the room environment is empty for each individual objects during the rendering, the re-composited scenes have lower realism than the original scenes, due to missing shadows and lack of indirect illumination from other objects. In this project, we do not consider the challenges of working with shadows and indirect illumination, leaving those for future research.

\begin{figure}[tb!]
	\centering
	\includegraphics[width=\linewidth]{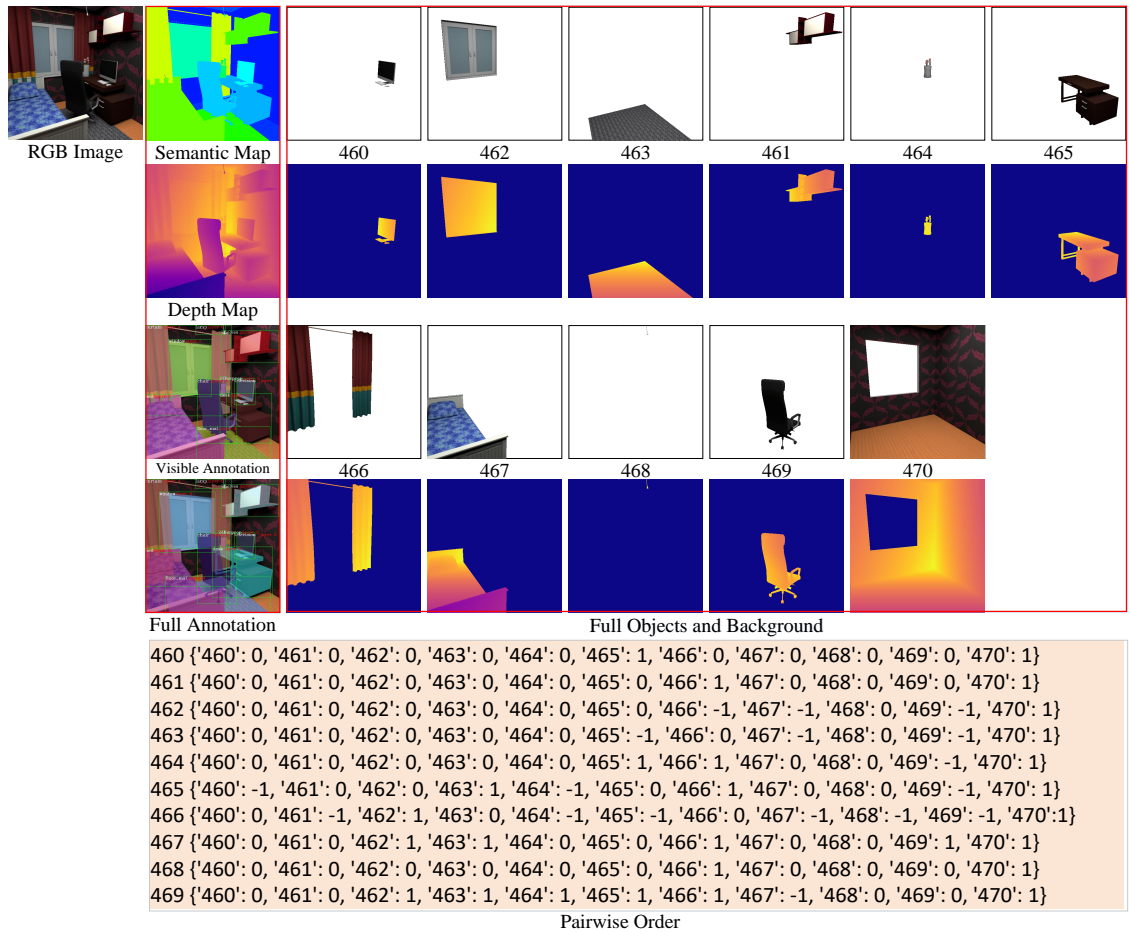}
	\caption[Illustration of Data Annotation]{\textbf{Illustration of Data Annotation.} For each rendered image, we have a corresponding semantic map, a depth map, and dense annotation, including class category, bounding box, instance mask, absolute layer order and pairwise order. In addition, for each object, we have a full RGBA image and depth map.}
	\label{fig:ch6_decom_appendix_dataset_annotation}
\end{figure}

\subsection{Data Annotation}\label{ch6_appendix_data_annotation}

In Figure~\ref{fig:ch6_decom_appendix_dataset_annotation}, we show one example of a rendered image with rich annotations, consisting of a semantic map, a depth map, visible annotations and full (amodal) annotations. For the semantic maps, we transferred the SUNCG class categories to NYUD-V2 40 categories so that this rendered dataset can be tested on real-world images. The depth map is stored in 16-bit format, with the largest indoor depth value at 20m. The class category and layer order (both absolute layer order and pairwise occlusion order) are included for visible annotations and full annotations. The visible annotations also contain the visible bounding-box offset and visible binary mask for each instance. Additionally, we also have the full (amodal) bounding-box offset and completed mask for each individual object.

\paragraph{Pairwise Occlusion Order} The pairwise order for each object is a vector storing the occlusion relationship between itself and all other objects. We use three numbers $\{-1, 0, 1\}$ to encode the occlusion relationship between two objects --- -1: occluded, 0 : no relationship, 1: front (\ie occluding). As can be seen in Figure~\ref{fig:ch6_decom_appendix_dataset_annotation}, the computer (object number: \#460) does not overlap the shelves (object number: \#461), so the pairwise order is ``0'', indicating these two objects have no occlusion relationship. The computer is however on top of the desk (object number: \#465), hence the pairwise order for $W_{460, 465}$ is ``1'', and conversely the pairwise order for $W_{465, 460}$ is ``-1'', representing that the desk is occluded by the computer. 

\subsection{Data Statistics}\label{ch6_appendix_data_statistics}

\begin{figure}[tb!]
	\centering
	\includegraphics[width=\linewidth]{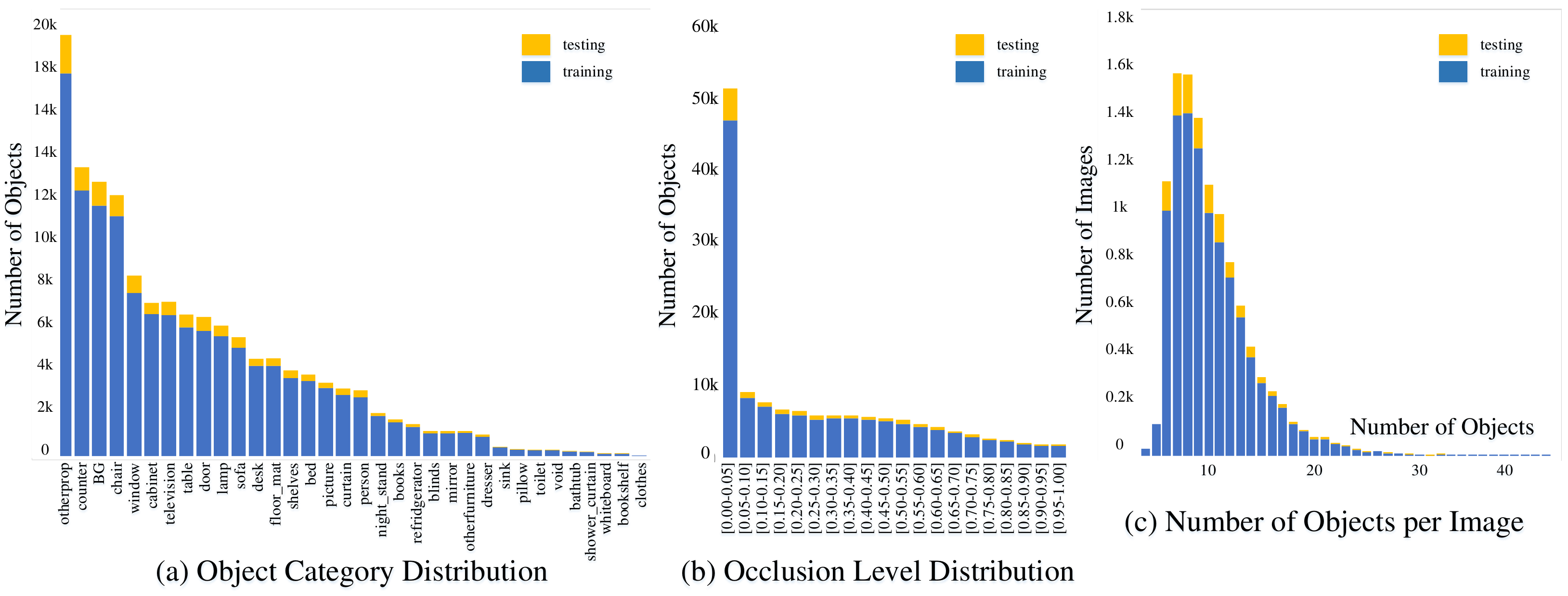}
	\caption[Data Statistics]{\textbf{Data Statistics.} Left: the object category distribution. Middle: the occlusion level distribution. Right: distribution of number of objects per image. On average there are 11 objects in each room. }
	\label{fig:ch6_decom_appendix_dataset_statistic}	
\end{figure}

In total, there are 11,434 views encompassing 129,336 labeled object instances in our rendered dataset. On average, there are 11 individual objects per view. Among these, 63.58\% objects are partially occluded by other objects and the average occlusion ratio (average IoU between two objects) is 26.27\%.

\paragraph{Object Category Statistics} Figure~\ref{fig:ch6_decom_appendix_dataset_statistic}(a) shows the overall object category distribution in our CSD dataset. Overall, the distribution is highly similar to the object distribution of NYUD-V2 dataset \cite{Silberman:ECCV12}, containing a diverse set of common furniture and objects in indoor rooms. ``Other props'' and ``Other furniture'' are atypical objects that do not belong in a common category. In particular, "Other props" are small objects that can be easily removed, while "Other furniture" are large objects with more permanent locations. Additionally, we merge floors, ceilings, and walls as ``BG'' in this work. If the user wants to obtain the separated semantic maps for these structures, these are also available. 

\paragraph{Occlusion Statistics} The occlusion level is defined as the fraction of overlapping regions between two objects (Intersection over Union, or IOU). We divide the occlusion into 20 levels from highly visible (denoted as [0.00-0.05] fraction of occlusion) to highly invisible (denoted as (0.95-1.00] fraction of occlusion), with 0.05 increment in the fraction of occlusion for each level. Figure~\ref{fig:ch6_decom_appendix_dataset_statistic}(b) shows the occlusion level in our dataset. In general, the distribution of occlusion levels is similar to the distribution in \cite{zhu2017semantic}, where a vast number of the instances are slightly occluded, while only a small number of instances are heavily occluded.

\paragraph{Object Count Distribution} Figure~\ref{fig:ch6_decom_appendix_dataset_statistic}(c) shows the distribution of the number of objects present per view. On average, there are more than 11 objects in each view. This supports the learning of rich scene contextual information for a completed scene decomposition task, instead of processing each object in isolation. 

\subsection{Data Encoding}\label{ch6_appendix_data_encoding}

After we get the views and corresponding dense annotations, we encode the data annotation to COCO format\footnote{http://cocodataset.org}. The annotations are stored using JSON, and the CSD API will be made available for visualizing and utilizing the rendered dataset. The JSON file contains a series of fields, including ``categories'', ``images'' and ``annotations''.

\cleardoublepage    


\authorpublications{

\section*{Conference Proceedings}

\begin{itemize}
  \item \textbf{Chuanxia Zheng}, Tat-Jen Cham, Jianfei Cai, ``T$^2$net: Synthetic-to-realistic translation for solving single-image depth estimation tasks,'' in \emph{Proceedings of the European Conference on Computer Vision (\textbf{ECCV}), 2018}.
  \item \textbf{Chuanxia Zheng}, Tat-Jen Cham, Jianfei Cai, ``Pluralistic image completion,'' in \emph{Proceedings of the IEEE/CVF Conference on Computer Vision and Pattern Recognition (\textbf{CVPR}), 2019}.
  \item \textbf{Chuanxia Zheng}, Tat-Jen Cham, Jianfei Cai, ``The Spatially-Correlative Loss for Various Image Translation Tasks,'' in \emph{Proceedings of the IEEE/CVF Conference on Computer Vision and Pattern Recognition (\textbf{CVPR}), 2021}.
  \item Guoxian Song, Linjie Luo, Jing Liu, Chunpong Lai, \textbf{Chuanxia Zheng}, and Tat-Jen Cham, ``Agilegan: Stylizing portraits by inversion-consistent transfer learning,'' in \emph{ACM Transactions on Graphics (Proceedings of ACM \textbf{SIGGRAPH}), 2021}.
  \item Tianyi Zhang, Jingyi Yang, \textbf{Chuanxia Zheng}, Guosheng Lin, Jianfei Cai, Alex C Kot, ``Task-in-all domain adaptation for semantic segmentation,'' in \emph{IEEE Visual Communications and Image Processing (\textbf{VCIP}), 2019}.
\end{itemize}

\section*{Journals}

\begin{itemize}
    \item \textbf{Chuanxia Zheng}, Tat-Jen Cham, Jianfei Cai, ``Pluralistic free-form image completion,'' \emph{International Journal of Computer Vision (\textbf{IJCV}), 2021}.
    \item \textbf{Chuanxia Zheng}, Duy-Son Dao, Guoxian Song, Tat-Jen Cham, Jianfei Cai, ``Visiting the invisible: layer-by-layer completed scene decomposition,'' \emph{International Journal of Computer Vision (\textbf{IJCV}), 2021}.
\end{itemize}

\section*{Preprints}

\begin{itemize}
    \item \textbf{Chuanxia Zheng}, Tat-Jen Cham, Jianfei Cai, ``Tfill: Image completion via a transformer-based architecture,'' in \emph{(arXiv), 2021}.
\end{itemize}

}

\backmatter


\label{Bibliography}
\bibliographystyle{ieee_fullname}
\bibliography{References/egbib}

\begin{thebibliography}{100}\itemsep=-1pt

\bibitem{alami2018unsupervised}
Youssef Alami~Mejjati, Christian Richardt, James Tompkin, Darren Cosker, and
  Kwang~In Kim.
\newblock Unsupervised attention-guided image-to-image translation.
\newblock In {\em Proceedings of the International Conference on Neural
  Information Processing Systems}, volume~31, pages 3693--3703, 2018.

\bibitem{amodio2019travelgan}
Matthew Amodio and Smita Krishnaswamy.
\newblock Travelgan: Image-to-image translation by transformation vector
  learning.
\newblock In {\em Proceedings of the IEEE Conference on Computer Vision and
  Pattern Recognition}, pages 8983--8992, 2019.

\bibitem{Maya}
{Autodesk Maya}, 2019.
\newblock \url{https://www.autodesk.com/products/maya/overview}.

\bibitem{bachman2019learning}
Philip Bachman, R~Devon Hjelm, and William Buchwalter.
\newblock Learning representations by maximizing mutual information across
  views.
\newblock In {\em Proceedings of the International Conference on Neural
  Information Processing Systems}, pages 15535--15545, 2019.

\bibitem{badrinarayanan2017segnet}
Vijay Badrinarayanan, Alex Kendall, and Roberto Cipolla.
\newblock Segnet: A deep convolutional encoder-decoder architecture for image
  segmentation.
\newblock {\em IEEE Transactions on Pattern Analysis and Machine Intelligence},
  39(12):2481--2495, 2017.

\bibitem{baek2020rethinking}
Kyungjune Baek, Yunjey Choi, Youngjung Uh, Jaejun Yoo, and Hyunjung Shim.
\newblock Rethinking the truly unsupervised image-to-image translation.
\newblock {\em arXiv preprint arXiv:2006.06500}, 2020.

\bibitem{bahdanau2014neural}
Dzmitry Bahdanau, Kyung~Hyun Cho, and Yoshua Bengio.
\newblock Neural machine translation by jointly learning to align and
  translate.
\newblock In {\em Processing of the 3rd International Conference on Learning
  Representations, ICLR 2015}, 2015.

\bibitem{ballester2001filling}
Coloma Ballester, Marcelo Bertalmio, Vicent Caselles, Guillermo Sapiro, and
  Joan Verdera.
\newblock Filling-in by joint interpolation of vector fields and gray levels.
\newblock {\em IEEE Transactions on Image Processing}, 10(8):1200--1211, 2001.

\bibitem{bao2017cvae}
Jianmin Bao, Dong Chen, Fang Wen, Houqiang Li, and Gang Hua.
\newblock Cvae-gan: Fine-grained image generation through asymmetric training.
\newblock In {\em Proceedings of the IEEE International Conference on Computer
  Vision (ICCV)}, pages 2764--2773. IEEE, 2017.

\bibitem{barnes2009patchmatch}
Connelly Barnes, Eli Shechtman, Adam Finkelstein, and Dan~B Goldman.
\newblock Patchmatch: A randomized correspondence algorithm for structural
  image editing.
\newblock {\em ACM Transactions on Graphics (ToG)}, 28:24, 2009.

\bibitem{benaim2017one}
Sagie Benaim and Lior Wolf.
\newblock One-sided unsupervised domain mapping.
\newblock In {\em Proceedings of the International Conference on Neural
  Information Processing Systems}, pages 752--762, 2017.

\bibitem{bengio2013representation}
Yoshua Bengio, Aaron Courville, and Pascal Vincent.
\newblock Representation learning: A review and new perspectives.
\newblock {\em IEEE Transactions on Pattern Analysis and Machine Intelligence},
  35(8):1798--1828, 2013.

\bibitem{bertalmio2000image}
Marcelo Bertalmio, Guillermo Sapiro, Vincent Caselles, and Coloma Ballester.
\newblock Image inpainting.
\newblock In {\em Proceedings of the 27th Annual Conference on Computer
  Graphics and Interactive Techniques}, pages 417--424. ACM
  Press/Addison-Wesley Publishing Co., 2000.

\bibitem{bertalmio2003simultaneous}
Marcelo Bertalmio, Luminita Vese, Guillermo Sapiro, and Stanley Osher.
\newblock Simultaneous structure and texture image inpainting.
\newblock {\em IEEE Transactions on Image Processing}, 12(8):882--889, 2003.

\bibitem{bourlard1988auto}
Herv{\'e} Bourlard and Yves Kamp.
\newblock Auto-association by multilayer perceptrons and singular value
  decomposition.
\newblock {\em Biological Cybernetics}, 59(4):291--294, 1988.

\bibitem{burgess2019monet}
Christopher~P Burgess, Loic Matthey, Nicholas Watters, Rishabh Kabra, Irina
  Higgins, Matt Botvinick, and Alexander Lerchner.
\newblock {MONet}: Unsupervised scene decomposition and representation.
\newblock {\em arXiv preprint arXiv:1901.11390}, 2019.

\bibitem{cao2017estimating}
Yuanzhouhan Cao, Zifeng Wu, and Chunhua Shen.
\newblock Estimating depth from monocular images as classification using deep
  fully convolutional residual networks.
\newblock {\em IEEE Transactions on Circuits and Systems for Video Technology},
  2017.

\bibitem{carion2020end}
Nicolas Carion, Francisco Massa, Gabriel Synnaeve, Nicolas Usunier, Alexander
  Kirillov, and Sergey Zagoruyko.
\newblock End-to-end object detection with transformers.
\newblock In {\em Proceedings of the European Conference on Computer Vision
  (ECCV)}, pages 213--229. Springer, 2020.

\bibitem{chen2020pre}
Hanting Chen, Yunhe Wang, Tianyu Guo, Chang Xu, Yiping Deng, Zhenhua Liu, Siwei
  Ma, Chunjing Xu, Chao Xu, and Wen Gao.
\newblock Pre-trained image processing transformer.
\newblock {\em arXiv preprint arXiv:2012.00364}, 2020.

\bibitem{chen2019hybrid}
Kai Chen, Jiangmiao Pang, Jiaqi Wang, Yu Xiong, Xiaoxiao Li, Shuyang Sun,
  Wansen Feng, Ziwei Liu, Jianping Shi, Wanli Ouyang, et~al.
\newblock Hybrid task cascade for instance segmentation.
\newblock In {\em Proceedings of the IEEE Conference on Computer Vision and
  Pattern Recognition}, pages 4974--4983, 2019.

\bibitem{chen2017deeplab}
Liang-Chieh Chen, George Papandreou, Iasonas Kokkinos, Kevin Murphy, and Alan~L
  Yuille.
\newblock Deeplab: Semantic image segmentation with deep convolutional nets,
  atrous convolution, and fully connected crfs.
\newblock {\em IEEE Transactions on Pattern Analysis and Machine Intelligence},
  40(4):834--848, 2017.

\bibitem{chen2020generative}
Mark Chen, Alec Radford, Rewon Child, Jeffrey Wu, Heewoo Jun, David Luan, and
  Ilya Sutskever.
\newblock Generative pretraining from pixels.
\newblock In {\em Proceedings of the International Conference on Machine
  Learning}, pages 1691--1703. PMLR, 2020.

\bibitem{chen2017photographic}
Qifeng Chen and Vladlen Koltun.
\newblock Photographic image synthesis with cascaded refinement networks.
\newblock In {\em Proceedings of the IEEE International Conference on Computer
  Vision}, pages 1511--1520, 2017.

\bibitem{chen2020simple}
Ting Chen, Simon Kornblith, Mohammad Norouzi, and Geoffrey Hinton.
\newblock A simple framework for contrastive learning of visual
  representations.
\newblock {\em arXiv preprint arXiv:2002.05709}, 2020.

\bibitem{chen2021s2r}
Xiaotian Chen, Yuwang Wang, Xuejin Chen, and Wenjun Zeng.
\newblock S2r-depthnet: Learning a generalizable depth-specific structural
  representation.
\newblock In {\em Proceedings of the IEEE/CVF Conference on Computer Vision and
  Pattern Recognition}, pages 3034--3043, 2021.

\bibitem{chen2018attention}
Xinyuan Chen, Chang Xu, Xiaokang Yang, and Dacheng Tao.
\newblock Attention-gan for object transfiguration in wild images.
\newblock In {\em Proceedings of the European Conference on Computer Vision
  (ECCV)}, pages 164--180, 2018.

\bibitem{chen2018high}
Zeyuan Chen, Shaoliang Nie, Tianfu Wu, and Christopher~G Healey.
\newblock High resolution face completion with multiple controllable attributes
  via fully end-to-end progressive generative adversarial networks.
\newblock {\em arXiv preprint arXiv:1801.07632}, 2018.

\bibitem{choi2018stargan}
Yunjey Choi, Minje Choi, Munyoung Kim, Jung-Woo Ha, Sunghun Kim, and Jaegul
  Choo.
\newblock Stargan: Unified generative adversarial networks for multi-domain
  image-to-image translation.
\newblock In {\em Proceedings of the IEEE/CVF Conference on Computer Vision and
  Pattern Recognition}, pages 8789--8797, 2018.

\bibitem{choi2020stargan}
Yunjey Choi, Youngjung Uh, Jaejun Yoo, and Jung-Woo Ha.
\newblock Stargan v2: Diverse image synthesis for multiple domains.
\newblock In {\em Proceedings of the IEEE/CVF Conference on Computer Vision and
  Pattern Recognition}, pages 8188--8197, 2020.

\bibitem{choy20163d}
Christopher~B Choy, Danfei Xu, JunYoung Gwak, Kevin Chen, and Silvio Savarese.
\newblock 3d-r2n2: A unified approach for single and multi-view 3d object
  reconstruction.
\newblock In {\em Proceedings of the European conference on computer vision
  (ECCV)}, pages 628--644. Springer, 2016.

\bibitem{Cordts2016Cityscapes}
Marius Cordts, Mohamed Omran, Sebastian Ramos, Timo Rehfeld, Markus Enzweiler,
  Rodrigo Benenson, Uwe Franke, Stefan Roth, and Bernt Schiele.
\newblock The cityscapes dataset for semantic urban scene understanding.
\newblock In {\em Processing of the IEEE Conference on Computer Vision and
  Pattern Recognition (CVPR)}, 2016.

\bibitem{criminisi2003object}
Antonio Criminisi, Patrick Perez, and Kentaro Toyama.
\newblock Object removal by exemplar-based inpainting.
\newblock In {\em Proceedings of the IEEE Computer Society Conference on
  Computer Vision and Pattern Recognition}, volume~2, pages II--II. IEEE, 2003.

\bibitem{criminisi2004region}
Antonio Criminisi, Patrick P{\'e}rez, and Kentaro Toyama.
\newblock Region filling and object removal by exemplar-based image inpainting.
\newblock {\em IEEE Transactions on Image Processing}, 13(9):1200--1212, 2004.

\bibitem{dai2016instance}
Jifeng Dai, Kaiming He, and Jian Sun.
\newblock Instance-aware semantic segmentation via multi-task network cascades.
\newblock In {\em Proceedings of the IEEE Conference on Computer Vision and
  Pattern Recognition}, pages 3150--3158, 2016.

\bibitem{deng2009imagenet}
Jia Deng, Wei Dong, Richard Socher, Li-Jia Li, Kai Li, and Li Fei-Fei.
\newblock Imagenet: A large-scale hierarchical image database.
\newblock In {\em Proceedings of the IEEE Conference on Computer Vision and
  Pattern Recognition}, pages 248--255. Ieee, 2009.

\bibitem{deng2020image}
Ye Deng and Jinjun Wang.
\newblock Image inpainting using parallel network.
\newblock In {\em Proceedings of the IEEE International Conference on Image
  Processing (ICIP)}, pages 1088--1092. IEEE, 2020.

\bibitem{devlin2018bert}
Jacob Devlin, Ming-Wei Chang, Kenton Lee, and Kristina Toutanova.
\newblock Bert: Pre-training of deep bidirectional transformers for language
  understanding.
\newblock {\em arXiv preprint arXiv:1810.04805}, 2018.

\bibitem{Dhamo2019iccv}
Helisa Dhamo, Nassir Navab, and Federico Tombari.
\newblock Object-driven multi-layer scene decomposition from a single image.
\newblock In {\em Proceedings of the IEEE International Conference on Computer
  Vision (ICCV)}, 2019.

\bibitem{dinh2014nice}
Laurent Dinh, David Krueger, and Yoshua Bengio.
\newblock Nice: Non-linear independent components estimation.
\newblock {\em arXiv preprint arXiv:1410.8516}, 2014.

\bibitem{dinh2016density}
Laurent Dinh, Jascha Sohl-Dickstein, and Samy Bengio.
\newblock Density estimation using real nvp.
\newblock In {\em Proceedings of the International Conference on Learning
  Representations}, 2017.

\bibitem{doersch2012makes}
Carl Doersch, Saurabh Singh, Abhinav Gupta, Josef Sivic, and Alexei Efros.
\newblock What makes paris look like paris?
\newblock {\em ACM Transactions on Graphics}, 31(4), 2012.

\bibitem{dosovitskiy2020image}
Alexey Dosovitskiy, Lucas Beyer, Alexander Kolesnikov, Dirk Weissenborn,
  Xiaohua Zhai, Thomas Unterthiner, Mostafa Dehghani, Matthias Minderer, Georg
  Heigold, Sylvain Gelly, Uszkoreit Jakob, and Houlsby Neil.
\newblock An image is worth 16x16 words: Transformers for image recognition at
  scale.
\newblock In {\em Proceedings of the International Conference on Learning
  Representations}, 2020.

\bibitem{dosovitskiy2016generating}
Alexey Dosovitskiy and Thomas Brox.
\newblock Generating images with perceptual similarity metrics based on deep
  networks.
\newblock In {\em Proceedings of the International Conference on Neural
  Information Processing Systems}, pages 658--666, 2016.

\bibitem{ehsani2018segan}
Kiana Ehsani, Roozbeh Mottaghi, and Ali Farhadi.
\newblock {SeGAN}: Segmenting and generating the invisible.
\newblock In {\em Proceedings of the IEEE Conference on Computer Vision and
  Pattern Recognition}, pages 6144--6153, 2018.

\bibitem{eigen2015predicting}
David Eigen and Rob Fergus.
\newblock Predicting depth, surface normals and semantic labels with a common
  multi-scale convolutional architecture.
\newblock In {\em Proceedings of the IEEE International Conference on Computer
  Vision (ICCV)}, pages 2650--2658, 2015.

\bibitem{eigen2014depth}
David Eigen, Christian Puhrsch, and Rob Fergus.
\newblock Depth map prediction from a single image using a multi-scale deep
  network.
\newblock In {\em Proceedings of the International Conference on Neural
  Information Processing Systems}, pages 2366--2374, 2014.

\bibitem{Eslami2018}
S.~M.~Ali Eslami, Danilo Jimenez~Rezende, Frederic Besse, Fabio Viola, Ari~S.
  Morcos, Marta Garnelo, Avraham Ruderman, Andrei~A. Rusu, Ivo Danihelka, Karol
  Gregor, David~P. Reichert, Lars Buesing, Theophane Weber, Oriol Vinyals, Dan
  Rosenbaum, Neil Rabinowitz, Helen King, Chloe Hillier, Matt Botvinick, Daan
  Wierstra, Koray Kavukcuoglu, and Demis Hassabis.
\newblock Neural scene representation and rendering.
\newblock {\em Science}, 360(6394):1204--1210, 2018.

\bibitem{esser2020taming}
Patrick Esser, Robin Rombach, and Björn Ommer.
\newblock Taming transformers for high-resolution image synthesis.
\newblock In {\em Proceedings of the IEEE/CVF Conference on Computer Vision and
  Pattern Recognition}, 2021.

\bibitem{everingham2010pascal}
Mark Everingham, Luc Van~Gool, Christopher~KI Williams, John Winn, and Andrew
  Zisserman.
\newblock The pascal visual object classes (voc) challenge.
\newblock {\em International Journal of Computer Vision}, 88(2):303--338, 2010.

\bibitem{follmann2019learning}
Patrick Follmann, Rebecca~K{\"o} Nig, Philipp~H{\"a} Rtinger, Michael
  Klostermann, and Tobias~B{\"o} Ttger.
\newblock Learning to see the invisible: End-to-end trainable amodal instance
  segmentation.
\newblock In {\em Proceedings of the IEEE Winter Conference on Applications of
  Computer Vision (WACV)}, pages 1328--1336. IEEE, 2019.

\bibitem{fu20203d}
Huan Fu, Bowen Cai, Lin Gao, Lingxiao Zhang, Cao Li, Zengqi Xun, Chengyue Sun,
  Yiyun Fei, Yu Zheng, Ying Li, et~al.
\newblock 3d-front: 3d furnished rooms with layouts and semantics.
\newblock {\em arXiv preprint arXiv:2011.09127}, 2020.

\bibitem{fu2019geometry}
Huan Fu, Mingming Gong, Chaohui Wang, Kayhan Batmanghelich, Kun Zhang, and
  Dacheng Tao.
\newblock Geometry-consistent generative adversarial networks for one-sided
  unsupervised domain mapping.
\newblock In {\em Proceedings of the IEEE Conference on Computer Vision and
  Pattern Recognition}, pages 2427--2436, 2019.

\bibitem{gaidon2016virtualworlds}
Adrien Gaidon, Qiao Wang, Yohann Cabon, and Eleonora Vig.
\newblock Virtualworlds as proxy for multi-object tracking analysis.
\newblock In {\em Processing of the IEEE Computer Vision and Pattern
  Recognition (CVPR), 2016 IEEE Conference on}, pages 4340--4349. IEEE, 2016.

\bibitem{ganin2015unsupervised}
Yaroslav Ganin and Victor Lempitsky.
\newblock Unsupervised domain adaptation by backpropagation.
\newblock In {\em Processing of the International Conference on Machine
  Learning (ICML)}, pages 1180--1189, 2015.

\bibitem{garg2016unsupervised}
Ravi Garg, Vijay~Kumar BG, Gustavo Carneiro, and Ian Reid.
\newblock Unsupervised cnn for single view depth estimation: Geometry to the
  rescue.
\newblock In {\em Proceedings of the European Conference on Computer Vision
  (ECCV)}, pages 740--756. Springer, 2016.

\bibitem{gatys2016image}
Leon~A Gatys, Alexander~S Ecker, and Matthias Bethge.
\newblock Image style transfer using convolutional neural networks.
\newblock In {\em Proceedings of the Computer Vision and Pattern Recognition
  (CVPR)}, pages 2414--2423. IEEE, 2016.

\bibitem{Geiger2012we}
Andreas Geiger, Philip Lenz, and Raquel Urtasun.
\newblock Are we ready for autonomous driving? the kitti vision benchmark
  suite.
\newblock In {\em Processing of the IEEE Computer Vision and Pattern
  Recognition (CVPR), 2012 IEEE Conference on}, pages 3354--3361. IEEE, 2012.

\bibitem{geirhos2018imagenet}
Robert Geirhos, Patricia Rubisch, Claudio Michaelis, Matthias Bethge, Felix~A
  Wichmann, and Wieland Brendel.
\newblock Imagenet-trained cnns are biased towards texture; increasing shape
  bias improves accuracy and robustness.
\newblock {\em arXiv preprint arXiv:1811.12231}, 2018.

\bibitem{girdhar2016learning}
Rohit Girdhar, David~F Fouhey, Mikel Rodriguez, and Abhinav Gupta.
\newblock Learning a predictable and generative vector representation for
  objects.
\newblock In {\em Proceedings of the European Conference on Computer Vision},
  pages 484--499. Springer, 2016.

\bibitem{girshick2015fast}
Ross Girshick.
\newblock Fast {R-CNN}.
\newblock In {\em Proceedings of the IEEE International Conference on Computer
  Vision}, pages 1440--1448, 2015.

\bibitem{girshick2014rich}
Ross Girshick, Jeff Donahue, Trevor Darrell, and Jitendra Malik.
\newblock Rich feature hierarchies for accurate object detection and semantic
  segmentation.
\newblock In {\em Proceedings of the IEEE Conference on Computer Vision and
  Pattern Recognition}, pages 580--587, 2014.

\bibitem{gkioxari2019mesh}
Georgia Gkioxari, Jitendra Malik, and Justin Johnson.
\newblock Mesh r-cnn.
\newblock In {\em Proceedings of the IEEE/CVF International Conference on
  Computer Vision}, pages 9785--9795, 2019.

\bibitem{godard2017unsupervised}
Cl{\'e}ment Godard, Oisin Mac~Aodha, and Gabriel~J Brostow.
\newblock Unsupervised monocular depth estimation with left-right consistency.
\newblock In {\em Proceedings of the IEEE Conference on Computer Vision and
  Pattern Recongition (CVPR)}, 2017.

\bibitem{goodfellow2016deep}
Ian Goodfellow, Yoshua Bengio, Aaron Courville, and Yoshua Bengio.
\newblock {\em Deep learning}, volume~1.
\newblock MIT press Cambridge, 2016.

\bibitem{goodfellow2014generative}
Ian Goodfellow, Jean Pouget-Abadie, Mehdi Mirza, Bing Xu, David Warde-Farley,
  Sherjil Ozair, Aaron Courville, and Yoshua Bengio.
\newblock Generative adversarial nets.
\newblock In {\em Proceedings of the International Conference on Neural
  Information Processing Systems}, volume~27, 2014.

\bibitem{gould2009decomposing}
Stephen Gould, Richard Fulton, and Daphne Koller.
\newblock Decomposing a scene into geometric and semantically consistent
  regions.
\newblock In {\em Proceedings of the IEEE International Conference on Computer
  Vision}, pages 1--8, 2009.

\bibitem{gulrajani2017improved}
Ishaan Gulrajani, Faruk Ahmed, Martin Arjovsky, Vincent Dumoulin, and Aaron~C
  Courville.
\newblock Improved training of wasserstein gans.
\newblock In {\em Proceedings of the International Conference on Neural
  Information Processing Systems}, pages 5767--5777, 2017.

\bibitem{guo2012beyond}
Ruiqi Guo and Derek Hoiem.
\newblock Beyond the line of sight: labeling the underlying surfaces.
\newblock In {\em Proceedings of the European Conference on Computer Vision},
  pages 761--774. Springer, 2012.

\bibitem{hara2020spherical}
Takayuki Hara and Tatsuya Harada.
\newblock Spherical image generation from a single normal field of view image
  by considering scene symmetry.
\newblock {\em arXiv preprint arXiv:2001.02993}, 2020.

\bibitem{hays2007scene}
James Hays and Alexei~A Efros.
\newblock Scene completion using millions of photographs.
\newblock In {\em ACM Transactions on Graphics (TOG)}, volume~26, page~4. ACM,
  2007.

\bibitem{he2020momentum}
Kaiming He, Haoqi Fan, Yuxin Wu, Saining Xie, and Ross Girshick.
\newblock Momentum contrast for unsupervised visual representation learning.
\newblock In {\em Proceedings of the IEEE/CVF Conference on Computer Vision and
  Pattern Recognition}, pages 9729--9738, 2020.

\bibitem{he2017mask}
Kaiming He, Georgia Gkioxari, Piotr Doll{\'a}r, and Ross Girshick.
\newblock Mask {R-CNN}.
\newblock In {\em Proceedings of the IEEE International Conference on Computer
  Vision}, pages 2961--2969, 2017.

\bibitem{he2015spatial}
Kaiming He, Xiangyu Zhang, Shaoqing Ren, and Jian Sun.
\newblock Spatial pyramid pooling in deep convolutional networks for visual
  recognition.
\newblock {\em IEEE Transactions on Pattern Analysis and Machine Intelligence},
  37(9):1904--1916, 2015.

\bibitem{he2016deep}
Kaiming He, Xiangyu Zhang, Shaoqing Ren, and Jian Sun.
\newblock Deep residual learning for image recognition.
\newblock In {\em Proceedings of the IEEE Conference on Computer Vision and
  Pattern Recognition}, pages 770--778, 2016.

\bibitem{heise2013pm}
Philipp Heise, Sebastian Klose, Brian Jensen, and Alois Knoll.
\newblock Pm-huber: Patchmatch with huber regularization for stereo matching.
\newblock In {\em Processing of the IEEE International Conference on Computer
  Vision}, pages 2360--2367. IEEE, 2013.

\bibitem{henaff2019data}
Olivier~J H{\'e}naff, Aravind Srinivas, Jeffrey De~Fauw, Ali Razavi, Carl
  Doersch, SM Eslami, and Aaron van~den Oord.
\newblock Data-efficient image recognition with contrastive predictive coding.
\newblock In {\em Proceedings of the IEEE/CVF Conference on Computer Vision and
  Pattern Recognition}, 2019.

\bibitem{heusel2017gans}
Martin Heusel, Hubert Ramsauer, Thomas Unterthiner, Bernhard Nessler, and Sepp
  Hochreiter.
\newblock Gans trained by a two time-scale update rule converge to a local nash
  equilibrium.
\newblock In {\em Proceedings of the International Conference on Neural
  Information Processing Systems}, pages 6626--6637, 2017.

\bibitem{hjelm2018learning}
R~Devon Hjelm, Alex Fedorov, Samuel Lavoie-Marchildon, Karan Grewal, Phil
  Bachman, Adam Trischler, and Yoshua Bengio.
\newblock Learning deep representations by mutual information estimation and
  maximization.
\newblock In {\em Proceedings of the International Conference on Learning
  Representations}, 2018.

\bibitem{hoffman2018cycada}
Judy Hoffman, Eric Tzeng, Taesung Park, Jun-Yan Zhu, Phillip Isola, Kate
  Saenko, Alexei Efros, and Trevor Darrell.
\newblock Cycada: Cycle-consistent adversarial domain adaptation.
\newblock In {\em Proceedings of the International Conference on Machine
  Learning}, pages 1989--1998, 2018.

\bibitem{Hoiem2005}
Derek Hoiem, Alexei~A. Efros, and Martial Hebert.
\newblock Automatic photo pop-up.
\newblock 24(3):577--584, 2005.

\bibitem{hu2019sail}
Yuan-Ting Hu, Hong-Shuo Chen, Kexin Hui, Jia-Bin Huang, and Alexander~G
  Schwing.
\newblock Sail-vos: Semantic amodal instance level video object segmentation-a
  synthetic dataset and baselines.
\newblock In {\em Proceedings of the IEEE Conference on Computer Vision and
  Pattern Recognition}, pages 3105--3115, 2019.

\bibitem{huang2017arbitrary}
Xun Huang and Serge Belongie.
\newblock Arbitrary style transfer in real-time with adaptive instance
  normalization.
\newblock In {\em Proceedings of the IEEE International Conference on Computer
  Vision}, pages 1501--1510, 2017.

\bibitem{huang2018multimodal}
Xun Huang, Ming-Yu Liu, Serge Belongie, and Jan Kautz.
\newblock Multimodal unsupervised image-to-image translation.
\newblock In {\em Proceedings of the European Conference on Computer Vision
  (ECCV)}, pages 172--189, 2018.

\bibitem{hudson2021gansformer}
Drew~A Hudson and C.~Lawrence Zitnick.
\newblock Generative adversarial transformers.
\newblock {\em arXiv preprint}, 2021.

\bibitem{iizuka2017globally}
Satoshi Iizuka, Edgar Simo-Serra, and Hiroshi Ishikawa.
\newblock Globally and locally consistent image completion.
\newblock {\em ACM Transactions on Graphics (TOG)}, 36(4):107, 2017.

\bibitem{isola2017image}
Phillip Isola, Jun-Yan Zhu, Tinghui Zhou, and Alexei~A Efros.
\newblock Image-to-image translation with conditional adversarial networks.
\newblock In {\em Proceedings of the IEEE Conference on Computer Vision and
  Pattern Recognition}, pages 1125--1134, 2017.

\bibitem{jaderberg2015spatial}
Max Jaderberg, Karen Simonyan, Andrew Zisserman, et~al.
\newblock Spatial transformer networks.
\newblock In {\em Proceedings of the International Conference on Neural
  Information Processing Systems}, pages 2017--2025, 2015.

\bibitem{jia2004inference}
Jiaya Jia and Chi-Keung Tang.
\newblock Inference of segmented color and texture description by tensor
  voting.
\newblock {\em IEEE Transactions on Pattern Analysis and Machine Intelligence},
  26(6):771--786, 2004.

\bibitem{jiang2020tsit}
Liming Jiang, Changxu Zhang, Mingyang Huang, Chunxiao Liu, Jianping Shi, and
  Chen~Change Loy.
\newblock Tsit: A simple and versatile framework for image-to-image
  translation.
\newblock In {\em Proceedings of the European Conference on Computer Vision
  (ECCV)}, 2020.

\bibitem{jiang2021transgan}
Yifan Jiang, Shiyu Chang, and Zhangyang Wang.
\newblock Transgan: Two transformers can make one strong gan.
\newblock {\em arXiv preprint arXiv:2102.07074}, 2021.

\bibitem{jo2019sc}
Youngjoo Jo and Jongyoul Park.
\newblock Sc-fegan: Face editing generative adversarial network with user's
  sketch and color.
\newblock {\em arXiv preprint arXiv:1902.06838}, 2019.

\bibitem{johnson2016perceptual}
Justin Johnson, Alexandre Alahi, and Li Fei-Fei.
\newblock Perceptual losses for real-time style transfer and super-resolution.
\newblock In {\em Proceedings of the European Conference on Computer Vision
  (ECCV)}, pages 694--711. Springer, 2016.

\bibitem{kar2015amodal}
Abhishek Kar, Shubham Tulsiani, Joao Carreira, and Jitendra Malik.
\newblock Amodal completion and size constancy in natural scenes.
\newblock In {\em Proceedings of the IEEE International Conference on Computer
  Vision}, pages 127--135, 2015.

\bibitem{karras2017progressive}
Tero Karras, Timo Aila, Samuli Laine, and Jaakko Lehtinen.
\newblock Progressive growing of gans for improved quality, stability, and
  variation.
\newblock {\em arXiv preprint arXiv:1710.10196}, 2017.

\bibitem{karras2019style}
Tero Karras, Samuli Laine, and Timo Aila.
\newblock A style-based generator architecture for generative adversarial
  networks.
\newblock In {\em Proceedings of the IEEE Conference on Computer Vision and
  Pattern Recognition}, pages 4401--4410, 2019.

\bibitem{karras2020analyzing}
Tero Karras, Samuli Laine, Miika Aittala, Janne Hellsten, Jaakko Lehtinen, and
  Timo Aila.
\newblock Analyzing and improving the image quality of stylegan.
\newblock In {\em Proceedings of the IEEE/CVF Conference on Computer Vision and
  Pattern Recognition}, pages 8110--8119, 2020.

\bibitem{karsch2012depth}
Kevin Karsch, Ce Liu, and Sing~Bing Kang.
\newblock Depth extraction from video using non-parametric sampling.
\newblock In {\em Proceedings of the European Conference on Computer Vision
  (ECCV)}, pages 775--788. Springer, 2012.

\bibitem{kim2018learning}
Dahun Kim, Donghyeon Cho, Donggeun Yoo, and In~So Kweon.
\newblock Learning image representations by completing damaged jigsaw puzzles.
\newblock In {\em Proceedings of the IEEE Winter Conference on Applications of
  Computer Vision (WACV)}, pages 793--802. IEEE, 2018.

\bibitem{Kim20DST}
Sunnie S.~Y. Kim, Nicholas Kolkin, Jason Salavon, and Gregory Shakhnarovich.
\newblock Deformable style transfer.
\newblock In {\em Proceedings of the European Conference on Computer Vision
  (ECCV)}, 2020.

\bibitem{kim2017learning}
Taeksoo Kim, Moonsu Cha, Hyunsoo Kim, Jung~Kwon Lee, and Jiwon Kim.
\newblock Learning to discover cross-domain relations with generative
  adversarial networks.
\newblock In {\em Proceedings of the International Conference on Machine
  Learning (ICML)}, pages 1857--1865, 2017.

\bibitem{kingma2018glow}
Durk~P Kingma and Prafulla Dhariwal.
\newblock Glow: Generative flow with invertible 1x1 convolutions.
\newblock In {\em Proceedings of the International Conference on Neural
  Information Processing Systems}, pages 10215--10224, 2018.

\bibitem{kingma2013auto}
Diederik~P Kingma and Max Welling.
\newblock Auto-encoding variational bayes.
\newblock {\em arXiv preprint arXiv:1312.6114}, 2013.

\bibitem{kohler2014mask}
Rolf K{\"o}hler, Christian Schuler, Bernhard Sch{\"o}lkopf, and Stefan
  Harmeling.
\newblock Mask-specific inpainting with deep neural networks.
\newblock In {\em Proceedings of the German Conference on Pattern Recognition},
  pages 523--534. Springer, 2014.

\bibitem{kolkin2019style}
Nicholas Kolkin, Jason Salavon, and Gregory Shakhnarovich.
\newblock Style transfer by relaxed optimal transport and self-similarity.
\newblock In {\em Proceedings of the IEEE Conference on Computer Vision and
  Pattern Recognition}, pages 10051--10060, 2019.

\bibitem{kuznietsov2017semi}
Yevhen Kuznietsov, J{\"o}rg St{\"u}ckler, and Bastian Leibe.
\newblock Semi-supervised deep learning for monocular depth map prediction.
\newblock In {\em Processing of the IEEE Conference on Computer Vision and
  Pattern Recognition (CVPR)}, pages 6647--6655, 2017.

\bibitem{ladicky2014pulling}
L'ubor Ladick\'{y}, Jianbo Shi, and Marc Pollefeys.
\newblock Pulling things out of perspective.
\newblock In {\em Proceedings of the IEEE Conference on Computer Vision and
  Pattern Recognition (CVPR)}, pages 89--96, 2014.

\bibitem{laina2016deeper}
Iro Laina, Christian Rupprecht, Vasileios Belagiannis, Federico Tombari, and
  Nassir Navab.
\newblock Deeper depth prediction with fully convolutional residual networks.
\newblock In {\em Proceedings of the Fourth International Conference on 3D
  Vision (3DV)}, pages 239--248. IEEE, 2016.

\bibitem{landau1988importance}
Barbara Landau, Linda~B Smith, and Susan~S Jones.
\newblock The importance of shape in early lexical learning.
\newblock {\em Cognitive Development}, 3(3):299--321, 1988.

\bibitem{lecun2015deep}
Yann LeCun, Yoshua Bengio, and Geoffrey Hinton.
\newblock Deep learning.
\newblock {\em Nature}, 521(7553):436--444, 2015.

\bibitem{lee2018diverse}
Hsin-Ying Lee, Hung-Yu Tseng, Jia-Bin Huang, Maneesh Singh, and Ming-Hsuan
  Yang.
\newblock Diverse image-to-image translation via disentangled representations.
\newblock In {\em Proceedings of the European Conference on Computer Vision
  (ECCV)}, pages 35--51, 2018.

\bibitem{lee2020drit++}
Hsin-Ying Lee, Hung-Yu Tseng, Qi Mao, Jia-Bin Huang, Yu-Ding Lu, Maneesh Singh,
  and Ming-Hsuan Yang.
\newblock Drit++: Diverse image-to-image translation via disentangled
  representations.
\newblock {\em International Journal of Computer Vision}, pages 1--16, 2020.

\bibitem{levin2003learning}
Anat Levin, Assaf Zomet, and Yair Weiss.
\newblock Learning how to inpaint from global image statistics.
\newblock In {\em Proceedings of the IEEE International Conference on Computer
  Vision (ICCV)}, page 305. IEEE, 2003.

\bibitem{li2016amodal}
Ke Li and Jitendra Malik.
\newblock Amodal instance segmentation.
\newblock In {\em Proceedings of the European Conference on Computer Vision},
  pages 677--693. Springer, 2016.

\bibitem{li2017generative}
Yijun Li, Sifei Liu, Jimei Yang, and Ming-Hsuan Yang.
\newblock Generative face completion.
\newblock In {\em Proceedings of the IEEE Computer Vision and Pattern
  Recognition (CVPR)}, pages 5892--5900. IEEE, 2017.

\bibitem{li2017fully}
Yi Li, Haozhi Qi, Jifeng Dai, Xiangyang Ji, and Yichen Wei.
\newblock Fully convolutional instance-aware semantic segmentation.
\newblock In {\em Proceedings of the IEEE Conference on Computer Vision and
  Pattern Recognition}, pages 2359--2367, 2017.

\bibitem{lin2018learning}
Chen-Hsuan Lin, Chen Kong, and Simon Lucey.
\newblock Learning efficient point cloud generation for dense 3d object
  reconstruction.
\newblock In {\em Proceedings of the AAAI Conference on Artificial
  Intelligence}, volume~32, 2018.

\bibitem{lin2017feature}
Tsung-Yi Lin, Piotr Doll{\'a}r, Ross Girshick, Kaiming He, Bharath Hariharan,
  and Serge Belongie.
\newblock Feature pyramid networks for object detection.
\newblock In {\em Proceedings of the IEEE Conference on Computer Vision and
  Pattern Recognition}, pages 2117--2125, 2017.

\bibitem{lin2014microsoft}
Tsung-Yi Lin, Michael Maire, Serge Belongie, James Hays, Pietro Perona, Deva
  Ramanan, Piotr Doll{\'a}r, and C~Lawrence Zitnick.
\newblock Microsoft coco: Common objects in context.
\newblock In {\em Proceedings of the European Conference on Computer Vision},
  pages 740--755. Springer, 2014.

\bibitem{ling2020variational}
Huan Ling, David Acuna, Karsten Kreis, Seung~Wook Kim, and Sanja Fidler.
\newblock Variational amodal object completion.
\newblock {\em Proceedings of the International Conference on Neural
  Information Processing Systems}, 33, 2020.

\bibitem{liu2016layered}
Chen Liu, Pushmeet Kohli, and Yasutaka Furukawa.
\newblock Layered scene decomposition via the {Occlusion-CRF}.
\newblock In {\em Proceedings of the IEEE Conference on Computer Vision and
  Pattern Recognition}, pages 165--173, 2016.

\bibitem{liu2016learning}
Fayao Liu, Chunhua Shen, Guosheng Lin, and Ian Reid.
\newblock Learning depth from single monocular images using deep convolutional
  neural fields.
\newblock {\em IEEE Transactions on Pattern Analysis and Machine Intelligence
  (TPAMI)}, 38(10):2024--2039, 2016.

\bibitem{Liu_2018_ECCV}
Guilin Liu, Fitsum~A. Reda, Kevin~J. Shih, Ting-Chun Wang, Andrew Tao, and
  Bryan Catanzaro.
\newblock Image inpainting for irregular holes using partial convolutions.
\newblock In {\em Proceedings of the European Conference on Computer Vision
  (ECCV)}, September 2018.

\bibitem{Liu2019MEDFE}
Hongyu Liu, Bin Jiang, Yibing Song, and Wei~Huang andand Chao~Yang.
\newblock Rethinking image inpainting via a mutual encoder-decoder with feature
  equalizations.
\newblock In {\em Proceedings of the European Conference on Computer Vision},
  2020.

\bibitem{liu2017unsupervised}
Ming-Yu Liu, Thomas Breuel, and Jan Kautz.
\newblock Unsupervised image-to-image translation networks.
\newblock In {\em Proceedings of the International Conference on Neural
  Information Processing Systems (NIPS)}, pages 700--708, 2017.

\bibitem{liu2016coupled}
Ming-Yu Liu and Oncel Tuzel.
\newblock Coupled generative adversarial networks.
\newblock In {\em Proceedings of the International Conference on Neural
  Information Processing Systems}, pages 469--477, 2016.

\bibitem{liu2015deep}
Ziwei Liu, Ping Luo, Xiaogang Wang, and Xiaoou Tang.
\newblock Deep learning face attributes in the wild.
\newblock In {\em Proceedings of the IEEE International Conference on Computer
  Vision}, pages 3730--3738, 2015.

\bibitem{long2015fully}
Jonathan Long, Evan Shelhamer, and Trevor Darrell.
\newblock Fully convolutional networks for semantic segmentation.
\newblock In {\em Proceedings of the IEEE Conference on Computer Vision and
  Pattern Recognition}, pages 3431--3440, 2015.

\bibitem{luan2017deep}
Fujun Luan, Sylvain Paris, Eli Shechtman, and Kavita Bala.
\newblock Deep photo style transfer.
\newblock In {\em Proceedings of the IEEE Conference on Computer Vision and
  Pattern Recognition}, pages 4990--4998, 2017.

\bibitem{mao2016multi}
Xudong Mao, Qing Li, Haoran Xie, Raymond~YK Lau, and Zhen Wang.
\newblock Multi-class generative adversarial networks with the l2 loss
  function.
\newblock {\em CoRR, abs/1611.04076}, 2, 2016.

\bibitem{mao2017least}
Xudong Mao, Qing Li, Haoran Xie, Raymond~YK Lau, Zhen Wang, and Stephen~Paul
  Smolley.
\newblock Least squares generative adversarial networks.
\newblock In {\em Proceedings of the IEEE International Conference on Computer
  Vision (ICCV)}, pages 2813--2821. IEEE, 2017.

\bibitem{mathieu2015deep}
Michael Mathieu, Camille Couprie, and Yann LeCun.
\newblock Deep multi-scale video prediction beyond mean square error.
\newblock {\em arXiv preprint arXiv:1511.05440}, 2015.

\bibitem{mechrez2018contextual}
Roey Mechrez, Itamar Talmi, and Lihi Zelnik-Manor.
\newblock The contextual loss for image transformation with non-aligned data.
\newblock In {\em Proceedings of the European Conference on Computer Vision
  (ECCV)}, pages 768--783, 2018.

\bibitem{Menze2015CVPR}
Moritz Menze and Andreas Geiger.
\newblock Object scene flow for autonomous vehicles.
\newblock In {\em Processing of the IEEE Conference on Computer Vision and
  Pattern Recognition (CVPR)}, 2015.

\bibitem{mescheder2018training}
Lars Mescheder, Andreas Geiger, and Sebastian Nowozin.
\newblock Which training methods for gans do actually converge?
\newblock In {\em Proceedings of the International Conference on Machine
  learning (ICML)}, pages 3481--3490. PMLR, 2018.

\bibitem{mirza2014conditional}
Mehdi Mirza and Simon Osindero.
\newblock Conditional generative adversarial nets.
\newblock {\em arXiv preprint arXiv:1411.1784}, 2014.

\bibitem{misra2020self}
Ishan Misra and Laurens van~der Maaten.
\newblock Self-supervised learning of pretext-invariant representations.
\newblock In {\em Proceedings of the IEEE/CVF Conference on Computer Vision and
  Pattern Recognition}, pages 6707--6717, 2020.

\bibitem{ferjad2020icml}
Muhammad~Ferjad Naeem, Seong~Joon Oh, Youngjung Uh, Yunjey Choi, and Jaejun
  Yoo.
\newblock Reliable fidelity and diversity metrics for generative models.
\newblock In {\em Proceedings of the International Conference on Machine
  Learning}, 2020.

\bibitem{Silberman:ECCV12}
Pushmeet~Kohli Nathan~Silberman, Derek~Hoiem and Rob Fergus.
\newblock Indoor segmentation and support inference from {RGBD} images.
\newblock In {\em Proceedings of the European Conference on Computer Vision},
  2012.

\bibitem{Nazeri_2019_ICCV}
Kamyar Nazeri, Eric Ng, Tony Joseph, Faisal Qureshi, and Mehran Ebrahimi.
\newblock Edgeconnect: Structure guided image inpainting using edge prediction.
\newblock In {\em Processing of the IEEE International Conference on Computer
  Vision (ICCV) Workshops}, Oct 2019.

\bibitem{nizan2020breaking}
Ori Nizan and Ayellet Tal.
\newblock Breaking the cycle-colleagues are all you need.
\newblock In {\em Proceedings of the IEEE/CVF Conference on Computer Vision and
  Pattern Recognition}, pages 7860--7869, 2020.

\bibitem{oord2018representation}
Aaron van~den Oord, Yazhe Li, and Oriol Vinyals.
\newblock Representation learning with contrastive predictive coding.
\newblock {\em arXiv preprint arXiv:1807.03748}, 2018.

\bibitem{park2017transformation}
Eunbyung Park, Jimei Yang, Ersin Yumer, Duygu Ceylan, and Alexander~C Berg.
\newblock Transformation-grounded image generation network for novel 3d view
  synthesis.
\newblock In {\em Processing of the IEEE Conference on Computer Vision and
  Pattern Recognition (CVPR)}, pages 702--711. IEEE, 2017.

\bibitem{park2020cut}
Taesung Park, Alexei~A. Efros, Richard Zhang, and Jun-Yan Zhu.
\newblock Contrastive learning for unpaired image-to-image translation.
\newblock In {\em Proceedings of the European Conference on Computer Vision},
  2020.

\bibitem{park2019semantic}
Taesung Park, Ming-Yu Liu, Ting-Chun Wang, and Jun-Yan Zhu.
\newblock Semantic image synthesis with spatially-adaptive normalization.
\newblock In {\em Proceedings of the IEEE Conference on Computer Vision and
  Pattern Recognition}, pages 2337--2346, 2019.

\bibitem{patashnik2021styleclip}
Or Patashnik, Zongze Wu, Eli Shechtman, Daniel Cohen-Or, and Dani Lischinski.
\newblock Styleclip: Text-driven manipulation of stylegan imagery.
\newblock {\em arXiv preprint arXiv:2103.17249}, 2021.

\bibitem{pathak2016context}
Deepak Pathak, Philipp Krahenbuhl, Jeff Donahue, Trevor Darrell, and Alexei~A
  Efros.
\newblock Context encoders: Feature learning by inpainting.
\newblock In {\em Proceedings of the IEEE Conference on Computer Vision and
  Pattern Recognition}, pages 2536--2544, 2016.

\bibitem{peng2021generating}
Jialun Peng, Dong Liu, Songcen Xu, and Houqiang Li.
\newblock Generating diverse structure for image inpainting with hierarchical
  vq-vae.
\newblock {\em arXiv preprint arXiv:2103.10022}, 2021.

\bibitem{pinheiro2015learning}
Pedro~O Pinheiro, Ronan Collobert, and Piotr Doll{\'a}r.
\newblock Learning to segment object candidates.
\newblock In {\em Proceedings of the International Conference on Neural
  Information Processing Systems}, pages 1990--1998, 2015.

\bibitem{pinheiro2016learning}
Pedro~O Pinheiro, Tsung-Yi Lin, Ronan Collobert, and Piotr Doll{\'a}r.
\newblock Learning to refine object segments.
\newblock In {\em Proceedings of the European Conference on Computer Vision},
  pages 75--91. Springer, 2016.

\bibitem{portenier2018faceshop}
Tiziano Portenier, Qiyang Hu, Attila Szabo, Siavash~Arjomand Bigdeli, Paolo
  Favaro, and Matthias Zwicker.
\newblock Faceshop: Deep sketch-based face image editing.
\newblock {\em ACM Transactions on Graphics (TOG)}, 37(4):99, 2018.

\bibitem{qi2019amodal}
Lu Qi, Li Jiang, Shu Liu, Xiaoyong Shen, and Jiaya Jia.
\newblock Amodal instance segmentation with kins dataset.
\newblock In {\em Proceedings of the IEEE Conference on Computer Vision and
  Pattern Recognition}, pages 3014--3023, 2019.

\bibitem{qiu2016unrealcv}
Weichao Qiu and Alan Yuille.
\newblock Unrealcv: Connecting computer vision to unreal engine.
\newblock In {\em Processing of the European Conference on Computer Vision},
  pages 909--916. Springer, 2016.

\bibitem{radford2021learning}
Alec Radford, Jong~Wook Kim, Chris Hallacy, Aditya Ramesh, Gabriel Goh,
  Sandhini Agarwal, Girish Sastry, Amanda Askell, Pamela Mishkin, Jack Clark,
  et~al.
\newblock Learning transferable visual models from natural language
  supervision.
\newblock {\em arXiv preprint arXiv:2103.00020}, 2021.

\bibitem{radford2018improving}
Alec Radford, Karthik Narasimhan, Tim Salimans, and Ilya Sutskever.
\newblock Improving language understanding by generative pre-training.
\newblock Technical report, OpenAI, 2018.

\bibitem{radford2019language}
Alec Radford, Jeffrey Wu, Rewon Child, David Luan, Dario Amodei, and Ilya
  Sutskever.
\newblock Language models are unsupervised multitask learners.
\newblock {\em OpenAI blog}, 1(8):9, 2019.

\bibitem{ranftl2020towards}
Ren{\'e} Ranftl, Katrin Lasinger, D Hafner, Konrad Schindler, and Vladlen
  Koltun.
\newblock Towards robust monocular depth estimation: Mixing datasets for
  zero-shot cross-dataset transfer.
\newblock {\em IEEE Transactions on Pattern Analysis and Machine Intelligence},
  2020.

\bibitem{razavi2019generating}
Ali Razavi, Aaron van~den Oord, and Oriol Vinyals.
\newblock Generating diverse high-fidelity images with vq-vae-2.
\newblock {\em arXiv preprint arXiv:1906.00446}, 2019.

\bibitem{ren2015shepard}
Jimmy~SJ Ren, Li Xu, Qiong Yan, and Wenxiu Sun.
\newblock Shepard convolutional neural networks.
\newblock In {\em Proceedings of the International Conference on Neural
  Information Processing Systems}, pages 901--909, 2015.

\bibitem{ren2015faster}
Shaoqing Ren, Kaiming He, Ross Girshick, and Jian Sun.
\newblock Faster {R-CNN}: Towards real-time object detection with region
  proposal networks.
\newblock In {\em Proceedings of the International Conference on Neural
  Information Processing Systems}, pages 91--99, 2015.

\bibitem{ronneberger2015u}
Olaf Ronneberger, Philipp Fischer, and Thomas Brox.
\newblock U-net: Convolutional networks for biomedical image segmentation.
\newblock In {\em Processing of the International Conference on Medical Image
  Computing and Computer-Assisted Intervention}, pages 234--241. Springer,
  2015.

\bibitem{rosca2017variational}
Mihaela Rosca, Balaji Lakshminarayanan, David Warde-Farley, and Shakir Mohamed.
\newblock Variational approaches for auto-encoding generative adversarial
  networks.
\newblock {\em arXiv preprint arXiv:1706.04987}, 2017.

\bibitem{russakovsky2015imagenet}
Olga Russakovsky, Jia Deng, Hao Su, Jonathan Krause, Sanjeev Satheesh, Sean Ma,
  Zhiheng Huang, Andrej Karpathy, Aditya Khosla, Michael Bernstein, et~al.
\newblock Imagenet large scale visual recognition challenge.
\newblock {\em International Journal of Computer Vision}, 115(3):211--252,
  2015.

\bibitem{salimans2016improved}
Tim Salimans, Ian Goodfellow, Wojciech Zaremba, Vicki Cheung, Alec Radford, and
  Xi Chen.
\newblock Improved techniques for training gans.
\newblock In {\em Proceedings of the International Conference on Neural
  Information Processing Systems}, pages 2234--2242, 2016.

\bibitem{Saxena2008}
Ashutosh Saxena, Sung~H. Chung, and Andrew~Y. Ng.
\newblock 3-d depth reconstruction from a single still image.
\newblock {\em International Journal of Computer Vision}, 76(1):53--69, 2008.

\bibitem{saxena2009make3d}
Ashutosh Saxena, Min Sun, and Andrew~Y Ng.
\newblock Make3d: Learning 3d scene structure from a single still image.
\newblock {\em IEEE Transactions on Pattern Analysis and Machine Intelligence},
  31(5):824--840, 2009.

\bibitem{sengupta2020background}
Soumyadip Sengupta, Vivek Jayaram, Brian Curless, Steven~M Seitz, and Ira
  Kemelmacher-Shlizerman.
\newblock Background matting: The world is your green screen.
\newblock In {\em Proceedings of the IEEE/CVF Conference on Computer Vision and
  Pattern Recognition}, pages 2291--2300, 2020.

\bibitem{shade1998layered}
Jonathan Shade, Steven Gortler, Li-wei He, and Richard Szeliski.
\newblock Layered depth images.
\newblock In {\em Proceedings of the 25th Annual Conference on Computer
  Graphics and Interactive Techniques}, pages 231--242, 1998.

\bibitem{shaham2019singan}
Tamar~Rott Shaham, Tali Dekel, and Tomer Michaeli.
\newblock Singan: Learning a generative model from a single natural image.
\newblock In {\em Proceedings of the IEEE International Conference on Computer
  Vision}, pages 4570--4580, 2019.

\bibitem{shechtman2007matching}
Eli Shechtman and Michal Irani.
\newblock Matching local self-similarities across images and videos.
\newblock In {\em Proceedings of the IEEE Conference on Computer Vision and
  Pattern Recognition}, pages 1--8. IEEE, 2007.

\bibitem{shi2000normcuts}
Jianbo Shi and Jitendra Malik.
\newblock Normalized cuts and image segmentation.
\newblock {\em IEEE Transactions on Pattern Analysis and Machine Intelligence},
  22(8):888--905, 2000.

\bibitem{InGAN}
Assaf Shocher, Shai Bagon, Phillip Isola, and Michal Irani.
\newblock Ingan: Capturing and retargeting the "dna" of a natural image.
\newblock In {\em Proceedings of the IEEE International Conference on Computer
  Vision (ICCV)}, 2019.

\bibitem{shrivastava2017learning}
Ashish Shrivastava, Tomas Pfister, Oncel Tuzel, Josh Susskind, Wenda Wang, and
  Russ Webb.
\newblock Learning from simulated and unsupervised images through adversarial
  training.
\newblock In {\em Proceedings of the IEEE Conference on Computer Vision and
  Pattern Recognition (CVPR)}, 2017.

\bibitem{silberman2012indoor}
Nathan Silberman, Derek Hoiem, Pushmeet Kohli, and Rob Fergus.
\newblock Indoor segmentation and support inference from rgbd images.
\newblock In {\em Processing of the European Conference on Computer Vision
  (ECCV)}, pages 746--760. Springer, 2012.

\bibitem{vgg}
K. Simonyan and A. Zisserman.
\newblock Very deep convolutional networks for large-scale image recognition.
\newblock In {\em Proceedings of the International Conference on Learning
  Representations}, May 2015.

\bibitem{smolensky1986information}
Paul Smolensky.
\newblock Information processing in dynamical systems: Foundations of harmony
  theory.
\newblock Technical report, Colorado Univ at Boulder Dept of Computer Science,
  1986.

\bibitem{sohn2015learning}
Kihyuk Sohn, Honglak Lee, and Xinchen Yan.
\newblock Learning structured output representation using deep conditional
  generative models.
\newblock In {\em Proceedings of the International Conference on Neural
  Information Processing Systems}, pages 3483--3491, 2015.

\bibitem{song2017semantic}
Shuran Song, Fisher Yu, Andy Zeng, Angel~X Chang, Manolis Savva, and Thomas
  Funkhouser.
\newblock Semantic scene completion from a single depth image.
\newblock In {\em Proceedings of the IEEE Conference on Computer Vision and
  Pattern Recognition}, pages 1746--1754, 2017.

\bibitem{song2018contextual}
Yuhang Song, Chao Yang, Zhe Lin, Xiaofeng Liu, Qin Huang, Hao Li, and CC Jay.
\newblock Contextual-based image inpainting: Infer, match, and translate.
\newblock In {\em Proceedings of the European Conference on Computer Vision
  (ECCV)}, pages 3--19, 2018.

\bibitem{song2018spg}
Yuhang Song, Chao Yang, Yeji Shen, Peng Wang, Qin Huang, and C-C~Jay Kuo.
\newblock Spg-net: Segmentation prediction and guidance network for image
  inpainting.
\newblock {\em arXiv preprint arXiv:1805.03356}, 2018.

\bibitem{su2015render}
Hao Su, Charles~R Qi, Yangyan Li, and Leonidas~J Guibas.
\newblock Render for cnn: Viewpoint estimation in images using cnns trained
  with rendered 3d model views.
\newblock In {\em Proceedings of the IEEE International Conference on Computer
  Vision}, pages 2686--2694, 2015.

\bibitem{sun2010layered}
Deqing Sun, Erik~B Sudderth, and Michael~J Black.
\newblock Layered image motion with explicit occlusions, temporal consistency,
  and depth ordering.
\newblock In {\em Proceedings of the International Conference on Neural
  Information Processing Systems}, pages 2226--2234, 2010.

\bibitem{tighe2014scene}
Joseph Tighe, Marc Niethammer, and Svetlana Lazebnik.
\newblock Scene parsing with object instances and occlusion ordering.
\newblock In {\em Proceedings of the IEEE Conference on Computer Vision and
  Pattern Recognition}, pages 3748--3755, 2014.

\bibitem{tonioni2017unsupervised}
Alessio Tonioni, Matteo Poggi, Stefano Mattoccia, and Luigi Di~Stefano.
\newblock Unsupervised adaptation for deep stereo.
\newblock In {\em Proceedings of the IEEE International Conference on Computer
  Vision (ICCV)}, pages 1605--1613, 2017.

\bibitem{torralba200880}
Antonio Torralba, Rob Fergus, and William~T Freeman.
\newblock 80 million tiny images: A large data set for nonparametric object and
  scene recognition.
\newblock {\em IEEE Transactions on Pattern Analysis and Machine Intelligence},
  30(11):1958--1970, 2008.

\bibitem{ulyanov2017improved}
Dmitry Ulyanov, Andrea Vedaldi, and Victor Lempitsky.
\newblock Improved texture networks: Maximizing quality and diversity in
  feed-forward stylization and texture synthesis.
\newblock In {\em Proceedings of the IEEE Conference on Computer Vision and
  Pattern Recognition}, pages 6924--6932, 2017.

\bibitem{vahdat2020NVAE}
Arash Vahdat and Jan Kautz.
\newblock {NVAE}: A deep hierarchical variational autoencoder.
\newblock In {\em Proceedings of the International Conference on Neural
  Information Processing Systems}, 2020.

\bibitem{van2017neural}
Aaron van~den Oord, Oriol Vinyals, and Koray Kavukcuoglu.
\newblock Neural discrete representation learning.
\newblock In {\em Proceedings of the 31st International Conference on Neural
  Information Processing Systems}, pages 6309--6318, 2017.

\bibitem{vaswani2017attention}
Ashish Vaswani, Noam Shazeer, Niki Parmar, Jakob Uszkoreit, Llion Jones,
  Aidan~N Gomez, {\L}ukasz Kaiser, and Illia Polosukhin.
\newblock Attention is all you need.
\newblock In {\em Proceedings of the International Conference on Neural
  Information Processing Systems}, pages 5998--6008, 2017.

\bibitem{walker2016}
Jacob Walker, Carl Doersch, Abhinav Gupta, and Martial Hebert.
\newblock An uncertain future: Forecasting from static images using variational
  autoencoders.
\newblock In {\em Proceedings of the European Conference on Computer Vision
  (ECCV)}, 2016.

\bibitem{wan2021high}
Ziyu Wan, Jingbo Zhang, Dongdong Chen, and Jing Liao.
\newblock High-fidelity pluralistic image completion with transformers.
\newblock {\em arXiv preprint arXiv:2103.14031}, 2021.

\bibitem{wang2018pixel2mesh}
Nanyang Wang, Yinda Zhang, Zhuwen Li, Yanwei Fu, Wei Liu, and Yu-Gang Jiang.
\newblock Pixel2mesh: Generating 3d mesh models from single rgb images.
\newblock In {\em Proceedings of the European Conference on Computer Vision
  (ECCV)}, pages 52--67, 2018.

\bibitem{wang2015towards}
Peng Wang, Xiaohui Shen, Zhe Lin, Scott Cohen, Brian Price, and Alan~L Yuille.
\newblock Towards unified depth and semantic prediction from a single image.
\newblock In {\em Proceedings of the IEEE Conference on Computer Vision and
  Pattern Recognition}, pages 2800--2809, 2015.

\bibitem{wang2018high}
Ting-Chun Wang, Ming-Yu Liu, Jun-Yan Zhu, Andrew Tao, Jan Kautz, and Bryan
  Catanzaro.
\newblock High-resolution image synthesis and semantic manipulation with
  conditional gans.
\newblock In {\em Proceedings of the IEEE Conference on Computer Vision and
  Pattern Recognition}, pages 8798--8807, 2018.

\bibitem{wang2018non}
Xiaolong Wang, Ross Girshick, Abhinav Gupta, and Kaiming He.
\newblock Non-local neural networks.
\newblock In {\em Proceedings of the IEEE Conference on Computer Vision and
  Pattern recognition}, pages 7794--7803, 2018.

\bibitem{wang2018image}
Yi Wang, Xin Tao, Xiaojuan Qi, Xiaoyong Shen, and Jiaya Jia.
\newblock Image inpainting via generative multi-column convolutional neural
  networks.
\newblock In {\em Proceedings of the International Conference on Neural
  Information Processing Systems}, pages 331--340, 2018.

\bibitem{wang2004image}
Zhou Wang, Alan~C Bovik, Hamid~R Sheikh, and Eero~P Simoncelli.
\newblock Image quality assessment: from error visibility to structural
  similarity.
\newblock {\em IEEE Transactions on Image Processing}, 13(4):600--612, 2004.

\bibitem{winn2006layout}
John Winn and Jamie Shotton.
\newblock The layout consistent random field for recognizing and segmenting
  partially occluded objects.
\newblock In {\em Proceedings of the IEEE Computer Society Conference on
  Computer Vision and Pattern Recognition (CVPR'06)}, volume~1, pages 37--44.
  IEEE, 2006.

\bibitem{wu2020visual}
Bichen Wu, Chenfeng Xu, Xiaoliang Dai, Alvin Wan, Peizhao Zhang, Masayoshi
  Tomizuka, Kurt Keutzer, and Peter Vajda.
\newblock Visual transformers: Token-based image representation and processing
  for computer vision.
\newblock {\em arXiv preprint arXiv:2006.03677}, 2020.

\bibitem{wu2018unsupervised}
Zhirong Wu, Yuanjun Xiong, Stella~X Yu, and Dahua Lin.
\newblock Unsupervised feature learning via non-parametric instance
  discrimination.
\newblock In {\em Proceedings of the IEEE Conference on Computer Vision and
  Pattern Recognition}, pages 3733--3742, 2018.

\bibitem{xu2015show}
Kelvin Xu, Jimmy Ba, Ryan Kiros, Kyunghyun Cho, Aaron Courville, Ruslan
  Salakhudinov, Rich Zemel, and Yoshua Bengio.
\newblock Show, attend and tell: Neural image caption generation with visual
  attention.
\newblock In {\em Proceedings of the International Conference on Machine
  Learning}, pages 2048--2057, 2015.

\bibitem{yan2019visualizing}
Xiaosheng Yan, Feigege Wang, Wenxi Liu, Yuanlong Yu, Shengfeng He, and Jia Pan.
\newblock Visualizing the invisible: Occluded vehicle segmentation and
  recovery.
\newblock In {\em Proceedings of the IEEE International Conference on Computer
  Vision}, pages 7618--7627, 2019.

\bibitem{Yan_2018_ECCV}
Zhaoyi Yan, Xiaoming Li, Mu Li, Wangmeng Zuo, and Shiguang Shan.
\newblock Shift-net: Image inpainting via deep feature rearrangement.
\newblock In {\em Proceedings of the European Conference on Computer Vision
  (ECCV)}, September 2018.

\bibitem{yang2021objectnerf}
Bangbang Yang, Yinda Zhang, Yinghao Xu, Yijin Li, Han Zhou, Hujun Bao, Guofeng
  Zhang, and Zhaopeng Cui.
\newblock Learning object-compositional neural radiance field for editable
  scene rendering.
\newblock In {\em International Conference on Computer Vision ({ICCV})},
  October 2021.

\bibitem{yang2017high}
Chao Yang, Xin Lu, Zhe Lin, Eli Shechtman, Oliver Wang, and Hao Li.
\newblock High-resolution image inpainting using multi-scale neural patch
  synthesis.
\newblock In {\em Proceedings of the IEEE Conference on Computer Vision and
  Pattern Recognition (CVPR)}, volume~1, page~3, 2017.

\bibitem{yang2010layered}
Yi Yang, Sam Hallman, Deva Ramanan, and Charless Fowlkes.
\newblock Layered object detection for multi-class segmentation.
\newblock In {\em Proceedings of the IEEE Conference on Computer Vision and
  Pattern Recognition}, pages 3113--3120, 2010.

\bibitem{yang2011layered}
Yi Yang, Sam Hallman, Deva Ramanan, and Charless~C Fowlkes.
\newblock Layered object models for image segmentation.
\newblock {\em IEEE Transactions on Pattern Analysis and Machine Intelligence},
  34(9):1731--1743, 2011.

\bibitem{yeh2017semantic}
Raymond~A Yeh, Chen Chen, Teck~Yian Lim, Alexander~G Schwing, Mark
  Hasegawa-Johnson, and Minh~N Do.
\newblock Semantic image inpainting with deep generative models.
\newblock In {\em Proceedings of the Computer Vision and Pattern Recognition
  (CVPR)}, pages 6882--6890. IEEE, 2017.

\bibitem{yi2020contextual}
Zili Yi, Qiang Tang, Shekoofeh Azizi, Daesik Jang, and Zhan Xu.
\newblock Contextual residual aggregation for ultra high-resolution image
  inpainting.
\newblock In {\em Proceedings of the IEEE/CVF Conference on Computer Vision and
  Pattern Recognition}, pages 7508--7517, 2020.

\bibitem{yi2017dualgan}
Zili Yi, Hao Zhang, Ping Tan, and Minglun Gong.
\newblock Dualgan: Unsupervised dual learning for image-to-image translation.
\newblock In {\em Proceedings of the IEEE International Conference on Computer
  Vision (ICCV)}, pages 2849--2857, 2017.

\bibitem{yoo2019photorealistic}
Jaejun Yoo, Youngjung Uh, Sanghyuk Chun, Byeongkyu Kang, and Jung-Woo Ha.
\newblock Photorealistic style transfer via wavelet transforms.
\newblock In {\em Proceedings of the IEEE International Conference on Computer
  Vision}, pages 9036--9045, 2019.

\bibitem{you2016image}
Quanzeng You, Hailin Jin, Zhaowen Wang, Chen Fang, and Jiebo Luo.
\newblock Image captioning with semantic attention.
\newblock In {\em Proceedings of the IEEE Conference on Computer Vision and
  Pattern Recognition}, pages 4651--4659, 2016.

\bibitem{YuKoltun2016}
Fisher Yu and Vladlen Koltun.
\newblock Multi-scale context aggregation by dilated convolutions.
\newblock In {\em Processing of the International Conference on Learning
  Representations(ICLR)}, 2016.

\bibitem{yu2017dilated}
Fisher Yu, Vladlen Koltun, and Thomas Funkhouser.
\newblock Dilated residual networks.
\newblock In {\em Proceedings of the IEEE Conference on Computer Vision and
  Pattern Recognition}, pages 472--480, 2017.

\bibitem{yu2018generative}
Jiahui Yu, Zhe Lin, Jimei Yang, Xiaohui Shen, Xin Lu, and Thomas~S Huang.
\newblock Generative image inpainting with contextual attention.
\newblock In {\em Proceedings of the IEEE Conference on Computer Vision and
  Pattern Recognition}, pages 5505--5514, 2018.

\bibitem{yu2019free}
Jiahui Yu, Zhe Lin, Jimei Yang, Xiaohui Shen, Xin Lu, and Thomas~S Huang.
\newblock Free-form image inpainting with gated convolution.
\newblock In {\em Proceedings of the IEEE International Conference on Computer
  Vision}, pages 4471--4480, 2019.

\bibitem{zamir2018taskonomy}
Amir~R Zamir, Alexander Sax, William Shen, Leonidas~J Guibas, Jitendra Malik,
  and Silvio Savarese.
\newblock Taskonomy: Disentangling task transfer learning.
\newblock In {\em Proceedings of the IEEE Conference on Computer Vision and
  Pattern Recognition}, pages 3712--3722, 2018.

\bibitem{zeng2020image}
Yu Zeng, Zhe Lin, Huchuan Lu, and Vishal~M Patel.
\newblock Image inpainting with contextual reconstruction loss.
\newblock {\em arXiv preprint arXiv:2011.12836}, 2020.

\bibitem{zeng2020high}
Yu Zeng, Zhe Lin, Jimei Yang, Jianming Zhang, Eli Shechtman, and Huchuan Lu.
\newblock High-resolution image inpainting with iterative confidence feedback
  and guided upsampling.
\newblock In {\em Proceedings of the European Conference on Computer Vision},
  pages 1--17. Springer, 2020.

\bibitem{zhan2020self}
Xiaohang Zhan, Xingang Pan, Bo Dai, Ziwei Liu, Dahua Lin, and Chen~Change Loy.
\newblock Self-supervised scene de-occlusion.
\newblock In {\em Proceedings of the IEEE/CVF Conference on Computer Vision and
  Pattern Recognition}, pages 3784--3792, 2020.

\bibitem{zhang2019self}
Han Zhang, Ian Goodfellow, Dimitris Metaxas, and Augustus Odena.
\newblock Self-attention generative adversarial networks.
\newblock In {\em Proceedings of the International Conference on Machine
  Learning}, pages 7354--7363. PMLR, 2019.

\bibitem{zhang2016colorful}
Richard Zhang, Phillip Isola, and Alexei~A Efros.
\newblock Colorful image colorization.
\newblock In {\em Proceedings of the European Conference on Computer Vision},
  pages 649--666. Springer, 2016.

\bibitem{zhang2018unreasonable}
Richard Zhang, Phillip Isola, Alexei~A Efros, Eli Shechtman, and Oliver Wang.
\newblock The unreasonable effectiveness of deep features as a perceptual
  metric.
\newblock In {\em Proceedings of the IEEE Conference on Computer Vision and
  Pattern Recognition}, pages 586--595, 2018.

\bibitem{zhang2015monocular}
Ziyu Zhang, Alexander~G Schwing, Sanja Fidler, and Raquel Urtasun.
\newblock Monocular object instance segmentation and depth ordering with cnns.
\newblock In {\em Proceedings of the IEEE International Conference on Computer
  Vision}, pages 2614--2622, 2015.

\bibitem{zhao2020uctgan}
Lei Zhao, Qihang Mo, Sihuan Lin, Zhizhong Wang, Zhiwen Zuo, Haibo Chen, Wei
  Xing, and Dongming Lu.
\newblock Uctgan: Diverse image inpainting based on unsupervised cross-space
  translation.
\newblock In {\em Proceedings of the IEEE/CVF Conference on Computer Vision and
  Pattern Recognition}, pages 5741--5750, 2020.

\bibitem{zhao2019geometry}
Shanshan Zhao, Huan Fu, Mingming Gong, and Dacheng Tao.
\newblock Geometry-aware symmetric domain adaptation for monocular depth
  estimation.
\newblock In {\em Proceedings of the IEEE/CVF Conference on Computer Vision and
  Pattern Recognition}, pages 9788--9798, 2019.

\bibitem{zheng2018t2net}
Chuanxia Zheng, Tat-Jen Cham, and Jianfei Cai.
\newblock T2net: Synthetic-to-realistic translation for solving single-image
  depth estimation tasks.
\newblock In {\em Proceedings of the European Conference on Computer Vision
  (ECCV)}, pages 767--783, 2018.

\bibitem{zheng2019pluralistic}
Chuanxia Zheng, Tat-Jen Cham, and Jianfei Cai.
\newblock Pluralistic image completion.
\newblock In {\em Proceedings of the IEEE Conference on Computer Vision and
  Pattern Recognition}, pages 1438--1447, 2019.

\bibitem{zheng2021spatiallycorrelative}
Chuanxia Zheng, Tat-Jen Cham, and Jianfei Cai.
\newblock The spatially-correlative loss for various image translation tasks.
\newblock In {\em Proceedings of the IEEE/CVF Conference on Computer Vision and
  Pattern Recognition (CVPR)}, June 2021.

\bibitem{zheng2021tfill}
Chuanxia Zheng, Tat-Jen Cham, and Jianfei Cai.
\newblock Tfill: Image completion via a transformer-based architecture, 2021.

\bibitem{zheng2021visiting}
Chuanxia Zheng, Duy-Son Dao, Guoxian Song, Tat-Jen Cham, and Jianfei Cai.
\newblock Visiting the invisible: Layer-by-layer completed scene decomposition,
  2021.

\bibitem{SETR}
Sixiao Zheng, Jiachen Lu, Hengshuang Zhao, Xiatian Zhu, Zekun Luo, Yabiao Wang,
  Yanwei Fu, Jianfeng Feng, Tao Xiang, Philip~H.S. Torr, and Li Zhang.
\newblock Rethinking semantic segmentation from a sequence-to-sequence
  perspective with transformers, 2020.

\bibitem{zhou2018places}
Bolei Zhou, Agata Lapedriza, Aditya Khosla, Aude Oliva, and Antonio Torralba.
\newblock Places: A 10 million image database for scene recognition.
\newblock {\em IEEE Transactions on Pattern Analysis and Machine Intelligence},
  40(6):1452--1464, 2018.

\bibitem{zhou2016view}
Tinghui Zhou, Shubham Tulsiani, Weilun Sun, Jitendra Malik, and Alexei~A Efros.
\newblock View synthesis by appearance flow.
\newblock In {\em Proceedings of the European Conference on Computer Vision},
  pages 286--301. Springer, 2016.

\bibitem{zhu2017unpaired}
Jun-Yan Zhu, Taesung Park, Phillip Isola, and Alexei~A Efros.
\newblock Unpaired image-to-image translation using cycle-consistent
  adversarial networks.
\newblock In {\em Proceedings of the IEEE International Conference on Computer
  Vision (ICCV)}, pages 2223--2232, 2017.

\bibitem{zhu2017toward}
Jun-Yan Zhu, Richard Zhang, Deepak Pathak, Trevor Darrell, Alexei~A Efros,
  Oliver Wang, and Eli Shechtman.
\newblock Toward multimodal image-to-image translation.
\newblock In {\em Proceedings of the International Conference on Neural
  Information Processing Systems}, pages 465--476, 2017.

\bibitem{zhu2020deformable}
Xizhou Zhu, Weijie Su, Lewei Lu, Bin Li, Xiaogang Wang, and Jifeng Dai.
\newblock Deformable detr: Deformable transformers for end-to-end object
  detection.
\newblock {\em arXiv preprint arXiv:2010.04159}, 2020.

\bibitem{zhu2017semantic}
Yan Zhu, Yuandong Tian, Dimitris Metaxas, and Piotr Doll{\'a}r.
\newblock Semantic amodal segmentation.
\newblock In {\em Proceedings of the IEEE Conference on Computer Vision and
  Pattern Recognition}, pages 1464--1472, 2017.

\end{thebibliography}

\end{document}